\documentclass[11pt]{article}

\usepackage[letterpaper,margin=1in]{geometry}
\makeatletter
\renewcommand{\section}{\@startsection{section}{1}{\z@}{-0.24in}{0.10in}%
  {\large\bf\raggedright}}
\renewcommand{\subsection}{\@startsection{subsection}{2}{\z@}{-0.20in}{0.08in}%
  {\normalsize\bf\raggedright}}
\renewcommand{\subsubsection}{\@startsection{subsubsection}{3}{\z@}{-0.18in}{0.08in}%
  {\normalsize\sc\raggedright}}
\renewcommand{\paragraph}{\@startsection{paragraph}{4}{\z@}%
  {1.5ex \@plus 0.5ex \@minus .2ex}{-1em}{\normalsize\bf}}
\def\@seccntformat#1{\csname @secprefix@#1\endcsname\csname the#1\endcsname\csname @secsep@#1\endcsname}
\def\@secsep@section{.\hskip 0.5em}
\def\@secsep@subsection{\hskip 0.5em}
\def\@secsep@subsubsection{\hskip 0.5em}
\g@addto@macro\appendix{\def\@secprefix@section{Appendix\ }}

\newlength\aftertitskip      
\newlength\beforetitskip     
\newlength\interauthorskip   
\newlength\aftermaketitskip  
\renewcommand{\maketitle}{\par
  \begingroup
    \def\thefootnote{\fnsymbol{footnote}}%
    \@maketitle \@thanks
  \endgroup
  \thispagestyle{plain}%
  \setcounter{footnote}{0}%
  \let\maketitle\relax \let\@maketitle\relax
  \gdef\@thanks{}\let\thanks\relax}
\def\@maketitle{\vbox{\hsize\textwidth
  \linewidth\hsize \vskip \beforetitskip
  {\begin{center} \Large\bf \@title \par \end{center}} \vskip \aftertitskip
  {\def\name{\normalsize\bf}%
   \def\addr{\normalsize\it}%
   \def\AND{\par\vskip \interauthorskip\noindent}%
   \noindent \@author \par}
  \vskip \aftermaketitskip
}}

\renewenvironment{abstract}
  {\centerline{\large\bf Abstract}\vspace{0.7ex}%
   \bgroup\leftskip 20pt\rightskip 20pt\noindent}%
  {\par\egroup\vskip 0.25ex}
\newcommand{\keywords}[1]{\par\vspace{0.5ex}%
  {\leftskip 20pt\rightskip 20pt\noindent{\bf Keywords:} #1\par}\vskip 0.25ex}
\makeatother

\usepackage{tocloft} % compact table of contents

\usepackage[american]{babel}

\usepackage[numbers]{natbib} % numeric mode, so alphanumeric labels are used
\renewcommand{\bibsection}{\section*{References}\addcontentsline{toc}{section}{References}}
\usepackage{mathtools} % amsmath with fixes and additions
\usepackage{booktabs} % commands to create good-looking tables
\usepackage{tikz} % nice language for creating drawings and diagrams

\usepackage{amssymb,amsmath,amsfonts,commath}

\usepackage{amsthm} % should be loaded after amsmath

\usepackage{float}

\newtheorem{example}{Example} 
\newtheorem{theorem}{Theorem}
\newtheorem{lemma}{Lemma} 
\newtheorem{proposition}{Proposition} 
\newtheorem{remark}{Remark}
\newtheorem{corollary}{Corollary}
\newtheorem{definition}{Definition}

\usepackage{graphicx} 
\usepackage{hyperref}
\usepackage{multirow}
\usepackage{enumitem}
\usepackage{wrapfig}
\usepackage{colortbl}

\DeclareMathOperator*{\argmin}{arg\,min}

\newcommand{\C}[1]{{\mathcal{#1}}} % math calligraphy
\newcommand{\BF}[1]{{\mathbf{#1}}} % math bold

\renewcommand{\t}{\text}
\newcommand{\op}[1]{\operatorname{#1}}

\newcommand{\w}{\BF{w}}
\newcommand{\x}{\BF{x}}
\renewcommand{\a}{\BF{a}}

\renewcommand{\v}{\BF{v}}
\newcommand{\y}{\BF{y}}
\newcommand{\z}{\BF{z}}
\newcommand{\vv}{\BF{v}}
\newcommand{\uu}{\BF{u}}
\newcommand{\dd}{\BF{d}}
\newcommand{\g}{\BF{g}}

\usepackage{tikz}
\usetikzlibrary{positioning,arrows.meta}

\usepackage{algorithm}
\usepackage{algorithmicx}
\usepackage{algpseudocode}
\usepackage{caption,enumitem}
\usepackage{placeins} % \FloatBarrier
\providecommand{\R}{\mathbb R}

\providecommand{\ip}[2]{\langle #1,#2\rangle}
\providecommand{\norm}[1]{\lVert#1\rVert}

\providecommand{\Reg}{\operatorname{Reg}}
\providecommand{\SReg}{\operatorname{SReg}}

\providecommand{\NP}{\mathbf{NP}}
\providecommand{\RP}{\mathbf{RP}}

\providecommand{\one}{\mathbf 1}
\usepackage{tabularx,ragged2e} % landscape table

\title{Tilted Convexity and an Exact Offline--Online Separation}
\author{%
  \name Mohammad Pedramfar \\ \addr Mila -- Quebec AI Institute / McGill University
  \AND
  \name Rajiv Khanna \\ \addr Purdue University
  \AND
  \name Christopher John Quinn \\ \addr Iowa State University
  \AND
  \name Vaneet Aggarwal \\ \addr Purdue University
}
\date{}

\begin{document}

% ---------------- Front matter (roman page numbers) ----------------
\pagenumbering{roman}

% Cover page: title, authors, abstract, keywords
\maketitle
\thispagestyle{plain}

\begin{abstract}
Convexity has three equivalent descriptions: support by affine functions, monotonicity of the subdifferential, and increasing secant slopes. We weaken all three by one distortion with parameter $\gamma\in(0,1]$, which multiplies positive changes by $\gamma$ and divides negative ones by $\gamma$. The resulting classes are the convex functions at $\gamma=1$ and, for $\gamma<1$, are distinct and nested, with the parameter dropping by at most a square along the chain. For proper lower semicontinuous functions in these classes, the effective domain is convex, local minimizers are global, and the function is locally Lipschitz on the relative interior of its domain. 

The classes separate offline from online optimization over compact convex sets with efficient projection, with access to function values and bounded support vectors. A smooth member of any class can be minimized to additive accuracy $\varepsilon$ with $\mathrm{poly}(1/\varepsilon)$ first-order queries and projections. Online, with nonpositive losses and $\alpha$--regret measured against $\alpha$ times the best fixed point, a polynomial-time algorithm achieves $\gamma$--regret $O(\sqrt T)$ on the largest class. For every fixed, effectively specified $\gamma\in(0,1)$ and every $\alpha\in(\gamma,1]$, no randomized polynomial-time algorithm has sublinear expected $\alpha$--regret unless $\NP=\RP$, already for smooth concave polynomial losses in the smallest class, each minimized by one linear optimization. Thus, unless $\NP=\RP$, $\gamma$ is the exact online coefficient of each class, while the offline coefficient is one.
\end{abstract}


\clearpage

% Table of contents
\setcounter{tocdepth}{2} % sections only
\setlength{\cftbeforesecskip}{3pt}
{\setlength{\parskip}{0pt}\tableofcontents}

\clearpage

% ---------------- Main paper (arabic page numbers from 1) ----------------
\pagenumbering{arabic}
\setcounter{page}{1}

\section{Introduction}
\label{sec:intro}

\newtheorem{introthm}{Theorem}
\renewcommand{\theintrothm}{\Alph{introthm}}

Convexity makes offline and online optimization efficient for many problems, but many objectives in learning, statistics, and control are not convex \citep{ghadimi2013stochastic,zhang2019deep,bedi2021escaping}. 
Since convexity has numerous equivalent characterizations, it can be relaxed in several ways, and extensive work has sought relaxations that preserve properties enabling efficient optimization, such as a unique minimizer or monotone slopes~\citep{cambini2009generalized,hadjisavvas2006handbook}.  

A convex function can be described in three equivalent ways: by the gradient inequality $f(\y)-f(\x)\ge\langle\v,\y-\x\rangle$ for a gradient $\v$ at $\x$; by monotonicity of the gradient, $\langle\uu-\v,\y-\x\rangle\ge0$ for gradients $\v$ at $\x$ and $\uu$ at $\y$, which uses only gradients; and by Jensen's inequality $f(\lambda\y+(1-\lambda)\z)\le\lambda f(\y)+(1-\lambda)f(\z)$, or equivalently by the increase of the slopes of secants, which uses only function values and is equivalent to $f$ being a pointwise supremum of affine functions. The classical relaxations of convexity keep the sign in these inequalities but not their size. Quasiconvexity replaces the right side of Jensen's inequality by $\max\{f(\y),f(\z)\}$, so that every sublevel set is convex, and pseudoconvexity requires only that $\langle\nabla f(\x),\y-\x\rangle\ge0$ implies $f(\y)\ge f(\x)$ \citep{mangasarian1965pseudo}; each of the two also has an equivalent form in terms of gradients alone \citep{hadjisavvas2006handbook}. 
Both relaxations suffice for offline tractability: a quasiconvex function can be minimized by bisection on its value \citep{boyd2004convex} and a pseudoconvex function by cutting planes \citep{ben2001lectures}.

We study a common multiplicative distortion of the three convexity conditions.% 
    \footnote{While we view all three as tilted forms of convexity, we call \eqref{eq:intro-mwc} as ``$\gamma$-tilted convexity'' to be consistent with \cite{martinezrubio2022global}.}
For $\gamma\in(0,1]$, put $\psi_\gamma(s)=\gamma s$ when $s\ge0$ and $\psi_\gamma(s)=s/\gamma$ when $s<0$. At points where the values are finite, the three conditions are
\begin{alignat}{2}
&\text{$\gamma$--tilted convexity:}\qquad&f(\y)-f(\x)&\ge\psi_\gamma\bigl(\langle\v,\y-\x\rangle\bigr),\label{eq:intro-mwc}\\
&\text{$\gamma$--tilted monotonicity:}\qquad&\langle\uu,\y-\x\rangle&\ge\psi_\gamma\bigl(\langle\v,\y-\x\rangle\bigr),\label{eq:intro-mono}\\
&\text{$\gamma$--tilted secant convexity:}\qquad&\frac{f(\y)-f(\x)}{\norm{\y-\x}}&\ge\psi_\gamma\Bigl(\frac{f(\x)-f(\z)}{\norm{\x-\z}}\Bigr),\label{eq:intro-jensen}
\end{alignat}
where in~\eqref{eq:intro-mwc} and~\eqref{eq:intro-mono} $\overline\partial f$ is the Clarke subdifferential, $\v\in\overline\partial f(\x)$, and $\uu\in\overline\partial f(\y)$, and in~\eqref{eq:intro-jensen} $\y\ne\z$ lie in $\op{dom}(f)$, $\lambda\in(0,1)$, and $\x=\lambda\y+(1-\lambda)\z$ is required to lie in $\op{dom}(f)$ as well, so that the two sides compare the slopes of the secants on the two sides of $\x$. Condition~\eqref{eq:intro-jensen} is equivalent to the Jensen-type inequality $f(\lambda\y+(1-\lambda)\z)\le M_{\gamma,\lambda}\bigl(f(\y),f(\z)\bigr)$ (Proposition~\ref{prop:secant-jensen}), where the mean $M_{\gamma,\lambda}(a,b)=\bigl(\lambda a+\gamma(1-\lambda)b\bigr)/\bigl(\lambda+\gamma(1-\lambda)\bigr)$ for $a\ge b$, and symmetrically for $a<b$, reduces the weight of the smaller value by a factor $\gamma$;
it is the upper expectile of the two-point distribution at level $1/(1+\gamma)$~\citep{newey1987asymmetric,bellini2014generalized}. 

For differentiable functions,  \eqref{eq:intro-mwc} is exactly the equal-parameter case $\gamma_n=\gamma_p=\gamma$ of tilted convexity introduced by Mart\'inez-Rubio~\cite[Eq.~(3)]{martinezrubio2022global}. 
We write $\mathcal F^\gamma$, $\mathcal F^\gamma_{\op{mono}}$, and $\mathcal F^\gamma_{\op{sec}}$ for tilted convex functions, functions with tilted monotone subdifferentials, and $\gamma$--tilted secant convex functions, respectively (Definitions~\ref{def:gamma-MWC}, \ref{def:gamma-MW-mono}, and~\ref{def:gamma-MW-Jensen-convex}). 
To the best of our knowledge, 
$\mathcal F^\gamma_{\op{mono}}$ and $\mathcal F^\gamma_{\op{sec}}$ have not been previously proposed.
Our contribution is the proper lower semicontinuous theory, the sharp relations among these formulations, and the exact coefficient that polynomial-time online algorithms can achieve on these classes.

At $\gamma=1$ all three classes coincide with the proper lower semicontinuous convex functions, and their unions over positive parameters agree and form a strict subclass of the pseudoconvex functions. The support and operator formulations have distinct $0^+$ and $0$ endpoints, corresponding to pseudoconvexity and quasiconvexity for radially continuous functions, that is, functions whose restriction to every closed segment between points of their domain is continuous. The secant formulation reaches quasiconvexity directly at $\gamma=0$. Theorem~\ref{introthm:interp} and Figure~\ref{fig:inclusions} give the precise inclusions.

In \eqref{eq:intro-mwc}, a predicted increase $\langle\v,\y-\x\rangle>0$ must be realized up to a factor $\gamma$, and an actual decrease may exceed the predicted one by at most a factor $1/\gamma$. Since $\psi_\gamma$ preserves signs, a point at which $\BF0$ is a generalized gradient is still a global minimizer. Additive relaxations lose this property: $-t^2$ on $(-1,1)$ becomes convex after $t^2$ is added, which makes it weakly convex in the sense of Davis and Drusvyatskiy~\cite{davis2019stochastic}, yet its stationary point $\BF0$ is a maximizer.   The multiplicative condition is also unchanged when $f$ is multiplied by a positive constant.

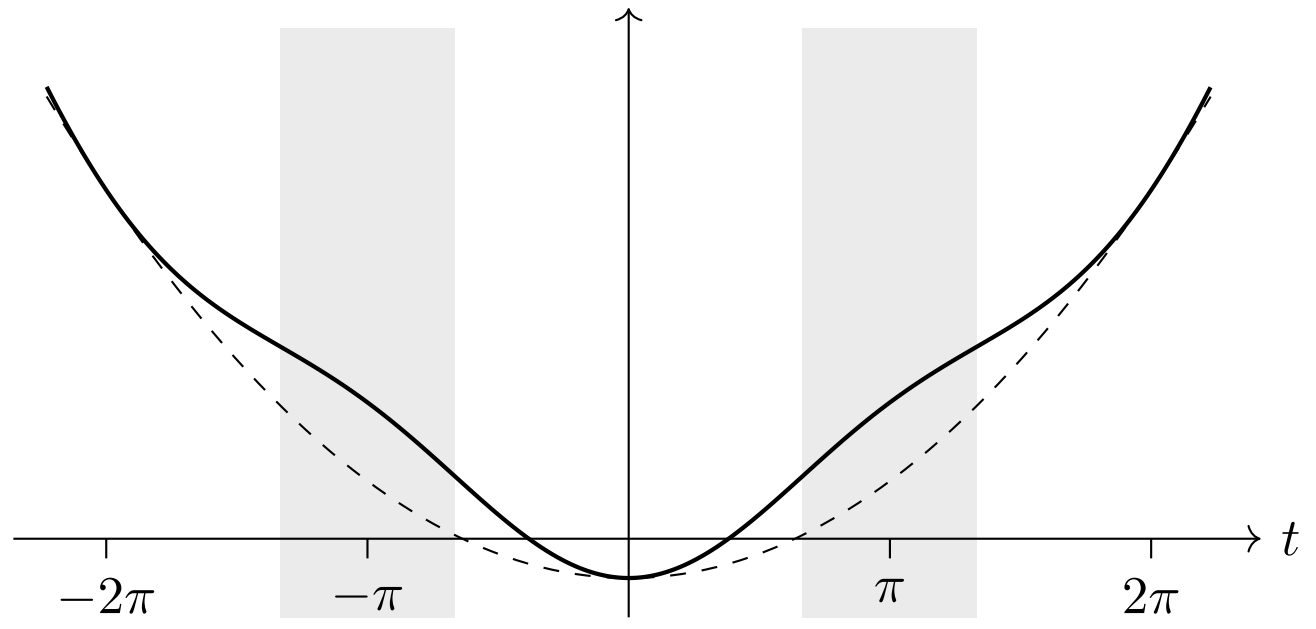
\begin{figure}[t!]
\centering
\begin{tikzpicture}[x=0.55cm,y=0.13cm,font=\small]
\fill[black!8] (2.0944,-4) rectangle (4.1888,26);
\fill[black!8] (-4.1888,-4) rectangle (-2.0944,26);
\draw[->] (-7.4,0) -- (7.6,0) node[right] {$t$};
\draw[->] (0,-4) -- (0,27);
\foreach \p/\l in {3.1416/\pi,-3.1416/-\pi,6.2832/2\pi,-6.2832/-2\pi}
  \draw (\p,0) -- (\p,-1) node[below] {$\l$};
\draw[dashed, domain=-7:7, samples=100] plot (\x,{0.5*(\x)*(\x)-2});
\draw[thick, domain=-7:7, samples=300] plot (\x,{0.5*(\x)*(\x)-2*cos(\x r)});
\end{tikzpicture}
\caption{%
The function $f(t)=\frac12t^2-2\cos t$ (solid) and the convex function $\frac12t^2-2$ (dashed). On the shaded intervals $f''(t)=1+2\cos t<0$, so $f$ is not convex. Its only stationary point is the minimizer $t=0$, and its subdifferential is $\gamma$--tilted monotone, so it lies in all three classes, with $\gamma=(4\pi-3\sqrt3)/(2\pi+3\sqrt3)$ (Example~\ref{ex:bowl-cosine}).
}
\label{fig:example}
\end{figure}

Each class contains objectives from applications (Section~\ref{sec:examples}). Concave polynomials of one coordinate with a large enough linear term, radial functions, robust regression losses such as the Cauchy and Geman--McClure losses \citep{black1996unification,barron2019general}, and distorted convex functions, which include worst-margin costs on a norm ball \citep{cortes1995support}, satisfy the monotonicity condition \eqref{eq:intro-mono} and therefore lie in all three classes. 
Ratios of a cost to a positive output, as in fractional programming \citep{charnes1962programming,dinkelbach1967nonlinear,schaible1976fractional}, including risk per unit of expected excess return in portfolio selection \citep{sharpe1966mutual} and energy per transmitted bit in wireless networks \citep{zappone2015energy}, satisfy \eqref{eq:intro-jensen} with a parameter at which they need not satisfy \eqref{eq:intro-mono}, and so do concave quadratics such as $-t^2-ct$ on $[0,1]$ with $c>0$. 
For example, $f(t)=\frac12t^2-2\cos t$ on $\R$ (Figure~\ref{fig:example}) is not convex, since $f''(t)=1+2\cos t<0$ on $(2\pi/3,4\pi/3)$, but its only stationary point is its minimizer. 
It is radial and satisfies \eqref{eq:intro-mono} with $\gamma=(4\pi-3\sqrt3)/(2\pi+3\sqrt3)$, so it lies in all three classes with this parameter (Example~\ref{ex:bowl-cosine}).

\paragraph{Overview of Main Results:}
Our first main result is that the three classes are nested within a squared parameter of each other, with $\mathcal F^\gamma_{\op{sec}}$ equal to the class of proper pointwise suprema of the functions $\y\mapsto c+\psi_{\sqrt\gamma}(\langle\z,\y-\w\rangle)$, 
and that they keep the main structural properties of convex functions, without any differentiability assumption (Theorems~\ref{introthm:interp} and~\ref{introthm:structure}). 
Our second main result separates the (offline) minimization of one function from the (online) minimization of a sequence of functions revealed one at a time: one smooth function from any of the classes can be minimized to any accuracy with polynomially many queries, while the largest factor $\alpha$ for which a polynomial-time online algorithm can guarantee sublinear regret against $\alpha$ times the best fixed point is exactly $\gamma$, unless $\NP=\RP$ (Theorem~\ref{introthm:separation}).

\subsection{Main Result 1: The Three Classes: Relations and Structure}
\label{sec:intro:structural}

\paragraph{How the classes relate.}
The three characterizations are equivalent for convex functions ($\gamma=1$) but not after relaxation ($\gamma<1$); there $\mathcal F^\gamma_{\op{mono}}\subseteq\mathcal F^\gamma_{\op{sec}}\subseteq\mathcal F^\gamma$, and passing from $\mathcal F^\gamma$ to $\mathcal F^\gamma_{\op{mono}}$ costs at most a square of the parameter. 
The class $\mathcal F^\gamma_{\op{sec}}$ also has a description by envelopes: it consists exactly of the proper pointwise suprema of the functions $\y\mapsto c+\psi_{\sqrt\gamma}(\langle\z,\y-\w\rangle)$, which we call \emph{tilted affine functions}, just as lower semicontinuous convex functions are the suprema of affine ones; it is therefore a class of abstract convex functions in the sense of Rubinov~\cite{rubinov2013abstract}. 
The positive-parameter classes form a strict hierarchy below the pseudoconvex class; 
the separately defined $0^+$ and $0$ endpoints give pseudoconvexity and quasiconvexity under radial continuity. 
Pointwise suprema stay in $\mathcal F^\gamma$ and in $\mathcal F^\gamma_{\op{sec}}$, 
while the maximum of two members of $\mathcal F^\gamma_{\op{mono}}$ can leave that class. 
Sums leave all three classes, and sums of their members plus one convex function reach every smooth function whose curvature is bounded below, 
so minimizing each function individually, as offline methods do, does not by itself solve the online comparator problem (Section~\ref{sec:intro:hardness}).

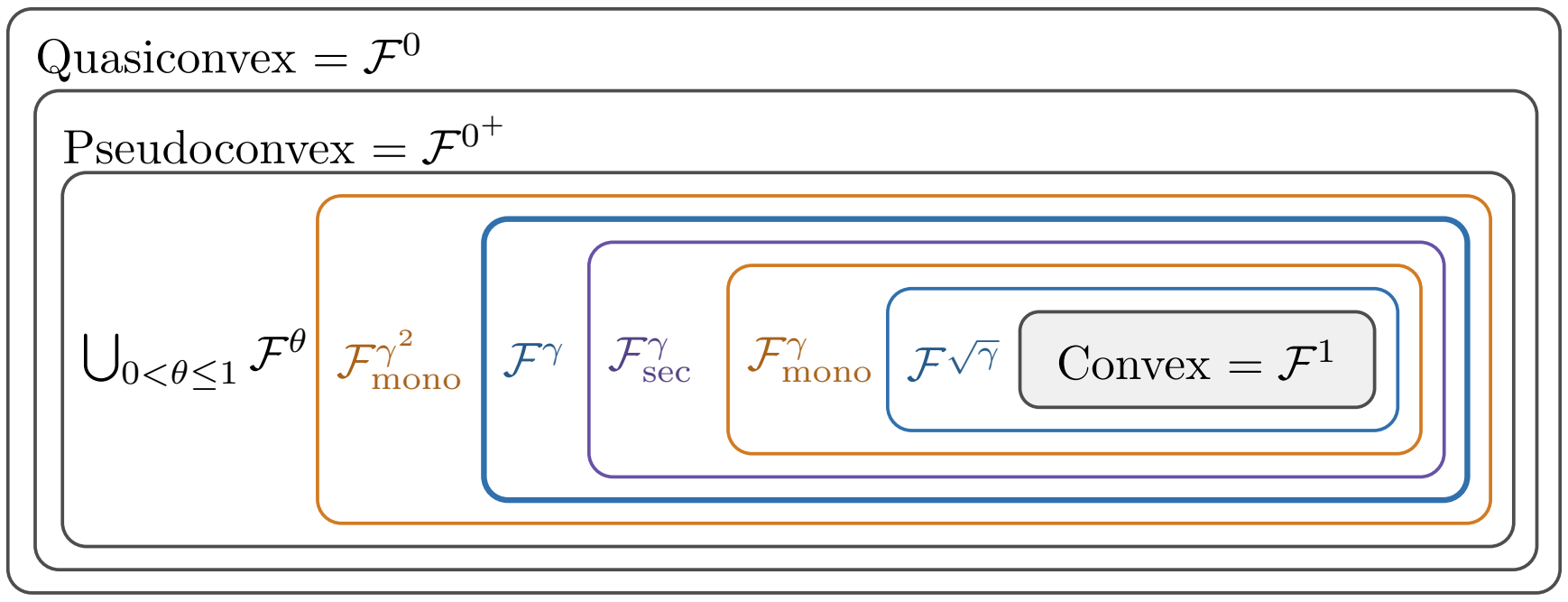
\begin{figure}[ht]
\centering
\definecolor{clsF}{RGB}{48,112,173}
\definecolor{clsM}{RGB}{104,82,166}
\definecolor{clsMono}{RGB}{210,123,34}
\begin{tikzpicture}[font=\small,line width=0.7pt]
\draw[rounded corners=4pt,fill=black!6,draw=black!70] (0.00,0.00) rectangle (2.60,0.70);
\node at (1.30,0.35) {Convex $=\mathcal F^1$};
\draw[rounded corners=5pt,draw=clsF,line width=0.7pt] (-0.97,-0.17) rectangle (2.77,0.87);
\node[anchor=west,text=clsF!80!black,inner sep=1pt] at (-0.88,0.35) {$\mathcal F^{\sqrt\gamma}$};
\draw[rounded corners=5pt,draw=clsMono,line width=0.7pt] (-2.14,-0.34) rectangle (2.94,1.04);
\node[anchor=west,text=clsMono!80!black,inner sep=1pt] at (-2.05,0.35) {$\mathcal F^{\gamma}_{\op{mono}}$};
\draw[rounded corners=5pt,draw=clsM,line width=0.7pt] (-3.16,-0.51) rectangle (3.11,1.21);
\node[anchor=west,text=clsM!80!black,inner sep=1pt] at (-3.07,0.35) {$\mathcal F^{\gamma}_{\op{sec}}$};
\draw[rounded corners=5pt,draw=clsF,line width=1.2pt] (-3.93,-0.68) rectangle (3.28,1.38);
\node[anchor=west,text=clsF!80!black,inner sep=1pt] at (-3.84,0.35) {$\mathcal F^{\gamma}$};
\draw[rounded corners=5pt,draw=clsMono,line width=0.7pt] (-5.15,-0.85) rectangle (3.45,1.55);
\node[anchor=west,text=clsMono!80!black,inner sep=1pt] at (-5.06,0.35) {$\mathcal F^{\gamma^2}_{\op{mono}}$};
\draw[rounded corners=5pt,draw=black!70,line width=0.7pt] (-7.02,-1.02) rectangle (3.62,1.72);
\node[anchor=west,inner sep=1pt] at (-6.93,0.35) {$\bigcup_{0<\theta\le1}\mathcal F^\theta$};
\draw[rounded corners=5pt,draw=black!70,line width=0.7pt] (-7.22,-1.19) rectangle (3.79,2.32);
\node[anchor=north west] at (-7.16,2.28) {Pseudoconvex $=\mathcal F^{0^+}$};
\draw[rounded corners=5pt,draw=black!70,line width=0.7pt] (-7.42,-1.36) rectangle (3.96,2.92);
\node[anchor=north west] at (-7.36,2.88) {Quasiconvex $=\mathcal F^0$};
\end{tikzpicture}
\caption{%
Nested classes for $0<\gamma<1$; each class contains the classes drawn inside it. The three families interleave as $\mathcal F^{\sqrt\gamma}\subseteq\mathcal F^\gamma_{\op{mono}}\subseteq\mathcal F^\gamma_{\op{sec}}\subseteq\mathcal F^\gamma\subseteq\mathcal F^{\gamma^2}_{\op{mono}}$ (Corollary~\ref{cor:relations}), and the pattern repeats at each squared parameter. Every inclusion is strict (Examples~\ref{ex:sep-1}--\ref{ex:sep-3} and Theorem~\ref{thm:hierarchy}). The identifications with the pseudoconvex and the quasiconvex functions hold for radially continuous functions.
}
\label{fig:inclusions}
\end{figure}

\begin{introthm}[Relations and representation; informal]\label{introthm:interp}
Let $0<\gamma'<\gamma<1$.
\begin{enumerate}[label=(\roman*)]
\item The classes are nested,
\[
\mathcal F^{\sqrt\gamma}\subseteq\mathcal F^\gamma_{\op{mono}}\subseteq\mathcal F^\gamma_{\op{sec}}\subseteq\mathcal F^\gamma,
\]
each inclusion is strict, and some functions lose the full square of the parameter between $\mathcal F^\gamma$ and $\mathcal F^\gamma_{\op{mono}}$ (Corollary~\ref{cor:relations}, Examples~\ref{ex:sep-1}--\ref{ex:sep-3}, and Figure~\ref{fig:inclusions}). The concave quadratic $-t^2-ct$ on $[0,1]$, with $c>0$, has parameter $c/(c+1)$ in $\mathcal F^\gamma_{\op{sec}}$ and $\mathcal F^\gamma$ but only $c/(c+2)$ in $\mathcal F^\gamma_{\op{mono}}$ (Example~\ref{ex:quadratic}).
\item A proper lower semicontinuous function is $\gamma$--tilted secant convex exactly when it is a nonempty pointwise supremum of tilted affine functions $\y\mapsto c+\psi_{\sqrt\gamma}(\langle\z,\y-\w\rangle)$ and exactly when its values satisfy an inequality at every finite set of points (Theorems~\ref{thm:H-convex-equivalent-def} and~\ref{thm:gamma-MW-Jensen-equivalent-def}).
\item Let $\mathcal F^{0^+}$ be defined by \eqref{eq:intro-mwc} with $\psi_\gamma$ replaced by its pointwise limit as $\gamma\downarrow0$, and let $\mathcal F^0$ be defined in the same way except that the inequality is not required when $\langle\v,\y-\x\rangle=0$ (Section~\ref{subsec:end-point}). If $\mathcal C^\theta$ stands for $\mathcal F^\theta_{\op{mono}}$, $\mathcal F^\theta_{\op{sec}}$, or $\mathcal F^\theta$ for every $\theta$, then
\[
\mathcal C^1\subsetneq\mathcal C^\gamma\subsetneq\mathcal C^{\gamma'}\subsetneq\bigcup_{0<\theta\le1}\mathcal C^\theta=\bigcup_{0<\theta\le1}\mathcal F^\theta\subsetneq\mathcal F^{0^+}\subsetneq\mathcal F^0,
\] 
where $\mathcal C^1$ consists of the convex functions and, among radially continuous functions, $\mathcal F^{0^+}$ and $\mathcal F^0$ consist of the pseudoconvex and the quasiconvex functions (Theorem~\ref{thm:hierarchy}).
\item Sums of $\gamma$--tilted convex functions need not be $\gamma$--tilted convex (Example~\ref{ex:sum-not-closed}), and every twice continuously differentiable function on $[0,1]^n$ whose Hessian is bounded below is the sum of a convex function and $n$ concave quadratics that lie in all three classes with parameter $\gamma$ (Theorem~\ref{thm:decomposition}).
\item A proper pointwise supremum of members of $\mathcal F^\gamma$ lies in $\mathcal F^\gamma$, and the same holds for $\mathcal F^\gamma_{\op{sec}}$. A maximum of two members of $\mathcal F^\gamma_{\op{mono}}$ can lie outside $\mathcal F^\gamma_{\op{mono}}$, but it always lies in $\mathcal F^{\gamma^2}_{\op{mono}}$, and the square is sharp (Theorems~\ref{thm:sup-closure} and~\ref{thm:gamma-MW-Jensen-basics}, Example~\ref{ex:mono-max}, and Section~\ref{subsec:mono-max}).
\end{enumerate}
\end{introthm}

\paragraph{A convex-like theory.}
Convex functions have convex sublevel sets and local minima that are global, they are locally Lipschitz on the relative interiors of their effective domains, and convexity can be checked along lines and is preserved under pointwise limits. 
Theorem~\ref{introthm:structure} establishes the corresponding properties for all three classes, including nonsmooth extended-valued functions; 
maxima of tilted affine functions and distorted worst-margin costs are nonsmooth examples (Section~\ref{sec:examples}).

\begin{introthm}[Structure; informal]\label{introthm:structure}
Let $\gamma\in(0,1]$ and let $f$ belong to $\mathcal F^\gamma_{\op{sec}}$, $\mathcal F^\gamma$, or $\mathcal F^\gamma_{\op{mono}}$.

\begin{enumerate}[label=(\roman*)]

\item The set where $f$ is finite is convex, and $f$ is quasiconvex. 
Every local minimizer is a global minimizer, and $\x$ is a minimizer exactly when $\BF0$ is a generalized gradient at $\x$ (Theorems~\ref{thm:mwc-domain-is-convex} and~\ref{thm:local-min-is-global-min}, 
Corollary~\ref{cor:gamma-MW-Jensen-structural-properties}).

\item The function $f$ is locally Lipschitz on the relative interior of its effective domain, and its values there determine $f$ on the whole domain (Theorems~\ref{thm:gamma-MW-convex-is-locally-Lipschitz} and~\ref{thm:mwc-lsc-envelope}).

\item Membership in each class can be checked on the restrictions of $f$ to lines and is preserved under restriction to closed convex sets, and if members of a class converge pointwise to finite values on a dense subset of a convex set, the lower semicontinuous envelope of the limit lies in the class (Theorems~\ref{thm:gamma-MW-convex-on-lines}, \ref{thm:gamma-MW-convex-pointwise-limit}, \ref{thm:gamma-MW-monotone-on-lines}, \ref{thm:gamma-MW-monotone-pointwise-limit}, \ref{thm:gamma-MW-Jensen-basics}, and~\ref{thm:gamma-MW-Jensen-pointwise-limit}, and Corollaries~\ref{cor:gamma-MW-convex-restrictions} and~\ref{cor:gamma-MW-monotone-restrictions}). 
Membership in $\mathcal F^\gamma$ does not depend on which standard notion of generalized gradient is used in \eqref{eq:intro-mwc} (Theorem~\ref{thm:mwc-alternative-characterizations}).
\end{enumerate}
\end{introthm}

\noindent
We prove (i) and (ii) 
for $\mathcal F^\gamma$ only, and they hold for the other two classes by inclusion 
(Theorem~\ref{introthm:interp}).

\paragraph{Techniques for the structural results.}
The proofs rest on a mean-value inequality for nonsmooth functions (Appendix~\ref{apx:useful}), which turns the pointwise condition \eqref{eq:intro-mwc} into statements about whole segments, as Jensen's inequality does for convex functions. The weights in the resulting segment inequality depend on the sign of a directional derivative, and the argument along segments absorbs this dependence. This argument yields the convexity of the domain and both conversions between \eqref{eq:intro-mwc} and \eqref{eq:intro-mono}: \eqref{eq:intro-mono} implies \eqref{eq:intro-mwc} with the same parameter, and \eqref{eq:intro-mwc} implies \eqref{eq:intro-mono} with the sharp loss of $\gamma^2$.
Together with Rademacher's theorem, which gives gradients on a dense subset of the relative interior, and a bound of the horizon subdifferential at boundary points by the normal cone of the domain (Appendix~\ref{apx:mwc-alternative-characterizations}), 
it also shows that all standard notions of generalized gradient define the same class $\mathcal F^\gamma$. Regularity and stability under limits come from a weaker notion, approximate $\gamma$--tilted convexity (Appendix~\ref{apx:approx}), in which the vector in \eqref{eq:intro-mwc} may depend on the desired accuracy and is required only on a dense set of points; such approximate supports force local Lipschitz continuity and become exact on the relative interior. 
Because these Lipschitz bounds are uniform along a locally bounded sequence, members that converge on a dense set converge uniformly on compact subsets of the relative interior, and the lower semicontinuous envelope of the limit is again in the class. 
For $\mathcal F^\gamma_{\op{sec}}$, an equivalent form of \eqref{eq:intro-jensen} applies a distortion with parameter $\sqrt\gamma$ to differences of values and extends to any finite number of points, which gives a membership test that uses function values alone (Theorem~\ref{thm:gamma-MW-Jensen-equivalent-def}).

\subsection{Main Result 2: Offline--Online Separation}
\label{sec:intro:hardness}

\paragraph{One function versus a sequence.}
Convexity makes 
optimization tasks tractable: offline, gradient or cutting-plane methods minimize a single convex function to any accuracy;
online, where a learner picks $\x_t$ in a convex set $\mathcal K$ in each round $t=1,\dots,T$ and an adversary then reveals a convex loss $f_t$, 
gradient descent stays within $O(\sqrt T)$ regret with respect to the best fixed point in hindsight \citep{zinkevich2003online,pedramfar2026generalized}. For nonconvex losses the two tasks can differ, because the online benchmark minimizes the sum $\sum_tf_t$ 
and by Theorem~\ref{introthm:interp}
(iv) such sums can encode general smooth nonconvex minimization even when each $f_t$ is easy to minimize. 
After $f_t$ is revealed, our learner may evaluate $f_t$ and its generalized gradients at points of its choice, 
and we measure its performance by approximate regret. 
Losses are nonpositive, so multiplying the benchmark by $\alpha\le1$ weakens it, and for $\alpha\in(0,1]$ the \emph{$\alpha$--regret} is
\[
\operatorname{Reg}_\alpha(T)=\sum_{t=1}^T f_t(\x_t)-\alpha\min_{\uu\in\mathcal K}\sum_{t=1}^T f_t(\uu),
\]
which is the usual regret when $\alpha=1$. A coefficient $\alpha$ is \emph{achievable} for a class of losses if a possibly randomized algorithm that is given the horizon $T$ and runs in time polynomial in the input size and $T$ keeps the expected $\operatorname{Reg}_\alpha(T)$ below $P\,T^{1-\rho}$ for every horizon and every loss sequence from the class, with $\rho>0$ a constant and $P$ polynomial in the input size, which bounds the descriptions of $\mathcal K$ and of a single loss but not the horizon (Section~\ref{subsec:online}). The \emph{online coefficient} is the supremum of the achievable coefficients, with value zero if none is positive; we show that it is attained at $\gamma$. The \emph{offline coefficient} of a class is $1$ when a single member can be minimized to additive accuracy $\varepsilon$ with a number of oracle and projection calls polynomial in $1/\varepsilon$ and the numerical parameters. We write $\mathbf{RP}$ for the class of decision problems solved by randomized polynomial-time algorithms with one-sided error~\citep{arora2009computational}.  

For $\alpha<1$ the $\alpha$--regret is not invariant under adding constants to the losses: replacing $f_t$ by $f_t-U_t$ subtracts $(1-\alpha)\sum_{t=1}^TU_t$ from it, so the sign convention is part of the model. Losses with known upper bounds $U_t\ge\sup_{\mathcal K}f_t$ can be shifted to nonpositive losses in the same class, and the choice $U_t=\sup_{\mathcal K}f_t$ gives the most demanding version. Both the algorithm (Section~\ref{subsec:online}) and the lower bound (Section~\ref{sec:hardness}) use nonpositive losses, and the hard losses take values in $[-1,0]$. Offline minimization to additive accuracy does not depend on this convention.

Because the classes are nested, an algorithm for the largest class $\mathcal F^\gamma$ runs on all three, a lower bound for the smallest class $\mathcal F^\gamma_{\op{mono}}$ holds for all three, and Theorem~\ref{introthm:separation} provides both.

\begin{introthm}[Offline--online separation; informal]\label{introthm:separation}
Fix an effectively specified $0<\gamma<1$, and let $\mathcal C$ be one of $\mathcal F^\gamma_{\op{mono}}$, $\mathcal F^\gamma_{\op{sec}}$, and $\mathcal F^\gamma$. Use the value-and-support and projection model of Section~\ref{sec:offline-opt} on a nonempty compact convex set of diameter at most $D$, with $G$--Lipschitz functions and, at each query point $\x$, a selected support vector, that is, a vector $\v$ satisfying~\eqref{eq:intro-mwc} for every feasible $\y$, of norm at most $G$.
\begin{enumerate}[label=(\roman*)]
\item \emph{Offline.} An $L$--smooth member of $\mathcal C$ can be minimized to additive accuracy $\varepsilon$ with $O(1+LD^2/(\gamma^2\varepsilon))$ gradient evaluations and projections by projected gradient descent, and with $\widetilde O(\gamma^{-1}\sqrt{LD^2/\varepsilon})$ first-order queries by an accelerated method, where $\widetilde O$ hides logarithmic factors (Corollary~\ref{cor:tilted-pgd}, based on~\cite{martinezrubio2025smooth}).
\item \emph{Online, achievability.} For a continuous nonpositive loss $f_t$ in the largest class $\mathcal F^\gamma$ and a point $\x_t\in\mathcal K$, some vector $g_t$ with $\norm{g_t}\le G$ satisfies
\[
f_t(\x_t)-\gamma f_t(\y)\le\langle g_t,\x_t-\y\rangle\qquad(\y\in\mathcal K)
\]
(Theorem~\ref{thm:linearization}). A polynomial-time procedure computes such a vector up to an additive error in this inequality that can be made inverse-polynomially small, and passing the computed vectors to a linear online learner gives $O(DG\sqrt T)$ static $\gamma$--regret. The algorithm makes polynomially many value, support, and projection queries to each revealed loss, and its running time is polynomial under the conventions of Section~\ref{subsec:online} (Theorem~\ref{thm:query}, Corollary~\ref{cor:ombq}). It applies to all three classes.
\item \emph{Online, hardness.} For every fixed $\alpha\in(\gamma,1]$, a randomized polynomial-time learner with expected static $\alpha$--regret at most $P(N)T^{1-\rho}$, for a fixed polynomial $P$ and constant $\rho>0$, on all sequences from $\mathcal F^\gamma_{\op{mono}}$ would imply $\NP=\RP$. This holds for explicit $[-1,0]$--valued, $1$--Lipschitz, $1$--smooth concave polynomials on products of simplices. Their degree and number of relevant coordinates depend only on $(\gamma,\alpha)$; their Hessians have rank at most one, and one linear optimization minimizes each loss exactly. Independent uniform samples from a polynomial-size family suffice. For $\mathcal F^\gamma$ alone, concave quadratics suffice (Theorem~\ref{thm:hardness}).
\end{enumerate}
\end{introthm}

Each class therefore has online coefficient exactly $\gamma$, unless $\NP=\RP$, under the stated model and normalization, 
and this already holds for its smooth members, whose offline coefficient is one (Corollary~\ref{cor:endpoints},  
Table~\ref{tab:landscape}). For the hard losses of Theorem~\ref{introthm:separation}(iii), minimizing a single loss is also polynomial in running time,  
since one linear optimization minimizes each loss exactly, so the lower bound concerns their combination. Convex optimization has a (trivial) approximation coefficient $\alpha=1$ in both settings;  
matching $\alpha<1$ offline and online coefficients are also known in several submodular settings (Section~\ref{sec:related}). 
We determine the exact nontrivial gap for three nested classes.
To the best of our knowledge, the classes studied here are the first with an online coefficient $\alpha$ that is determined exactly ($\alpha=\gamma$) and lies strictly between $0$ and the offline coefficient (i.e.~$1$), and this coefficient can be any $\gamma\in(0,1)$.

\paragraph{Techniques for the separation.}
Offline, smooth members are tilted convex and therefore quasar-convex~\citep{martinezrubio2022global,martinezrubio2025smooth}, 
so known first-order methods apply (Section~\ref{sec:offline-opt}). Online, the comparison must hold against every point, because the best fixed point in hindsight rarely minimizes any single loss, and the vector of Theorem~\ref{introthm:separation}(ii) provides it. Losses with such vectors are called lower-linearizable, and the reduction of Pedramfar and Aggarwal~\cite{pedramfar2024linear} passes the vectors to any online algorithm for linear losses. We obtain the vector by rescaling a generalized gradient $\v$ at $\x$ to $t\v$ with $t\in[\gamma,1]$. Each comparator $\y$ forces $t$ above or below a threshold according to the sign of $\langle\v,\y-\x\rangle$. The thresholds are compatible because two comparators of opposite signs can be paired through the point between them at which $\langle\v,\cdot-\x\rangle$ vanishes, and an interpolation inequality for $-f$ applies on their segment (Lemma~\ref{lem:interpolation}). Since the vector is defined by infinitely many constraints, we find $t$ by bisection and test each candidate with a projected method that searches for a violated comparator, which takes polynomially many value, support, and projection queries (Theorem~\ref{thm:query}). The number of calls has no inverse-power dependence on $\gamma$.

For the lower bound, a class has the \emph{online gap property} if SAT reduces to finite families of losses whose average equals $-Q$ up to scale, for a score $Q$ that reaches $1$ on satisfiable instances and stays below some $\tau<\alpha$ on unsatisfiable ones (Definition~\ref{def:online-gap-property}). Running a learner with coefficient $\alpha$ on random members of such a family decides SAT with one-sided error, so no class with this property admits a polynomial-time algorithm with coefficient $\alpha$ unless $\mathbf{NP}=\mathbf{RP}$; this step uses no form of convexity (Theorem~\ref{thm:online-gap-hardness-criterion}). We build the families from Label Cover, a constraint satisfaction problem for which the PCP theorem makes it hard to tell instances in which all constraints can be satisfied from instances in which only a small fraction can \citep{moshkovitz2010two}. Each right vertex $v$ and label $b$ give a loss $-H(z_{v,b})$, where $z_{v,b}$ is the mass that a fractional labeling of the left vertices sends to $b$ along the edges at $v$, and $H$ is an increasing convex function with $H(0)=0$ and $H(1)=1$. The linear part of $H$ pays for labels whether or not they agree across edges, so on unsatisfiable instances the score is at most $H'(0)$ plus a term that Label Cover makes small (Theorem~\ref{thm:direct-hard-profile-criterion}). Membership of $-H$ in $\mathcal F^\gamma_{\op{mono}}$ forces $H'(0)\ge\gamma H'(1)\ge\gamma$, and polynomial choices of $H$ in the class bring $H'(0)$ arbitrarily close to $\gamma$ (Lemma~\ref{lem:hardness-tight-monotone-profile}), so the threshold of the reduction equals the parameter of the class.

\paragraph{Contributions.}
\begin{itemize}

\item \emph{Three classes from one distortion.}
We apply one multiplicative distortion to the three standard characterizations of convexity (the support inequality; monotonicity of the subdifferential; and increasing secant slopes or, equivalently, Jensen's inequality) 
and study the resulting classes for proper lower semicontinuous functions. 
The support version extends tilted convexity~\citep{martinezrubio2022global} from differentiable to nonsmooth extended-valued functions; 
the monotonicity and secant versions are new.
\item \emph{Convex-like structure without smoothness.} For proper lower semicontinuous functions in each class, the domain is convex, local minimizers are global, 
the function is locally Lipschitz on the relative interior of its domain, and membership can be checked on lines and passes to limits (Theorem~\ref{introthm:structure}).

\item \emph{Exact relations among the classes.} The classes are nested, $\mathcal F^{\sqrt\gamma}\subseteq\mathcal F^\gamma_{\op{mono}}\subseteq\mathcal F^\gamma_{\op{sec}}\subseteq\mathcal F^\gamma$, every inclusion is strict, 
and some functions lose the full square of the parameter. 
The secant class $\mathcal F^\gamma_{\op{sec}}$ consists exactly of the proper suprema of tilted affine functions, 
as the lower semicontinuous convex functions are the suprema of affine ones. 
None of the classes is closed under addition (Theorem~\ref{introthm:interp}).
\item \emph{An exact offline--online separation.} A smooth function from any of the classes can be minimized with polynomially many queries by known methods for quasar-convex functions. Online, a polynomial-time algorithm achieves coefficient $\gamma$ on the largest class, and unless $\NP=\RP$ no polynomial-time algorithm achieves a larger coefficient on the smallest class (Theorem~\ref{introthm:separation} and Table~\ref{tab:landscape}).
\end{itemize}

\begin{table}[ht]
\centering
\renewcommand{\arraystretch}{1.2}
\caption{Offline--online separation. Offline coefficient $1$ means that a single objective can be minimized to arbitrary additive accuracy with polynomially many queries. Online coefficients refer to $\alpha$--regret bounded by a polynomial times $T^{1-\rho}$, achieved in polynomial time, for losses with values in $[-B,0]$ and $B$, $D$, and $G$ polynomial in the input size. The convex and positive-parameter rows use Section~\ref{sec:offline-opt}. The endpoint offline rows assume the stronger feasibility or cut access specified in their cells, with the usual geometric and encoding conditions. Online lower bounds use explicit smooth individual losses and hold unless $\NP=\RP$.}
\label{tab:landscape}
\begin{tabularx}{\textwidth}{@{}>{\RaggedRight\arraybackslash}p{0.17\textwidth}>{\RaggedRight\arraybackslash}X>{\RaggedRight\arraybackslash}X@{}}
\toprule
Class & Offline coefficient and individual minimization & Online coefficient\\
\midrule
Convex & $1$; classical cutting-plane or projected methods \citep{ben2001lectures} & $1$, with $O(DG\sqrt T)$ regret \citep{zinkevich2003online}\\
$\mathcal F^\gamma_{\op{mono}}$, $\mathcal F^\gamma_{\op{sec}}$, $\mathcal F^\gamma$, $0<\gamma<1$ & $1$ for $L$--smooth members; projected gradient descent uses $O(1+LD^2/(\gamma^2\varepsilon))$ calls, and acceleration uses~\eqref{eq:tilted-accelerated-rate} queries (Section~\ref{sec:offline-opt}). & $\gamma$ is achievable on $\mathcal F^\gamma$; every fixed $\alpha>\gamma$ implies $\NP=\RP$ already on $\mathcal F^\gamma_{\op{mono}}$ (Corollary~\ref{cor:ombq}, Theorem~\ref{thm:hardness})\\
Pseudoconvex & $1$ under an efficient valid-cut oracle and the usual cutting-plane assumptions~\citep{ben2001lectures} & Every fixed $\alpha>0$ implies $\NP=\RP$ (Corollary~\ref{cor:endpoints})\\
Quasiconvex & $1$ under efficient weak feasibility for convex sublevel sets and the usual geometric and encoding assumptions~\citep{boyd2004convex} & Every fixed $\alpha>0$ implies $\NP=\RP$ (Corollary~\ref{cor:endpoints})\\
\bottomrule
\end{tabularx}
\end{table}

\section{Related Work}
\label{sec:related}

\paragraph{Generalized convexity and generalized monotonicity.}
Quasiconvexity, pseudoconvexity, and invexity each keep part of the global geometry of convex functions \citep{mangasarian1965pseudo,mangasarian1975pseudo,craven1981invex,cambini2009generalized}. Their nonsmooth versions are characterized through generalized monotonicity of subdifferentials, with quasiconvexity corresponding to quasimonotonicity \citep{aussel1994subdifferential} and pseudoconvexity to pseudomonotonicity \citep{penot1997generalized}; see also~\cite{hadjisavvas2006handbook,hadjisavvas2024generalized,soleimanidamaneh07_chara_nonsm,rockafellar1998variational}. These conditions record the sign of a change in value but not its size. Our three classes give a quantitative, sign-dependent version of each standard characterization of convexity, they reduce to convexity at $\gamma=1$, and $\mathcal F^\gamma$ reaches pseudoconvexity and quasiconvexity at its endpoints $0^+$ and $0$. The size bound is what yields local Lipschitz continuity without differentiability, a hierarchy that is strict at every parameter, and sharp conversions between the classes with a loss of at most a square in the parameter.

\paragraph{Convexity with respect to means, and abstract convexity.}
Replacing the arithmetic mean in Jensen's inequality by another mean is a classical way to define convexity classes: $f$ is convex with respect to means $P$ and $Q$ if $f(P(x,y))\le Q(f(x),f(y))$ \citep{niculescu2003convexity,anderson2007generalized,maksa2015convexity}. In this language, the Jensen form of $\gamma$--tilted secant convexity uses the weighted arithmetic mean on the arguments and the mean $M_{\gamma,\lambda}$ on the values, where $M_{\gamma,\lambda}$ is an upper expectile~\citep{newey1987asymmetric,bellini2014generalized}. We characterize the resulting function class, including its finite Jensen inequalities and its envelope representation. Abstract convexity represents a function as the upper envelope of a chosen family of elementary functions, generalizing the description of a proper lower semicontinuous convex function as a supremum of affine functions, and closure of the resulting classes under suprema is standard \citep{rubinov2013abstract,demyanov1990foundations,penot2013calculus}. The two viewpoints meet in $\mathcal F^\gamma_{\op{sec}}$: the upper envelopes of the tilted affine functions $\y\mapsto c+\psi_\gamma(\langle\z,\y-\w\rangle)$ are exactly the $\gamma^2$--tilted secant convex functions and are characterized by an inequality on finitely many function values, 
just as the proper upper envelopes of affine functions are exactly the proper lower semicontinuous functions that satisfy Jensen's inequality.

\paragraph{Additive and multiplicative relaxations.}
Weak convexity and semiconvexity bound negative curvature by adding a quadratic; 
prox-regularity gives a local quadratic support property \citep{hiriart2013convex,davis2019stochastic,patrascu2015efficient}. 
Our distortion rescales slopes instead, so the classes are unchanged when $f$ is multiplied by a positive constant and stationary points remain minimizers, whereas weak convexity with a large modulus contains every twice differentiable function whose Hessian is bounded below. The tilted condition is distinct from multiplicative, or geometric, convexity, which compares positive function values through geometric means~\citep{niculescu2000convexity}.

\paragraph{Tilted convexity and quasar-convexity.}
For differentiable functions, the support condition~\eqref{eq:intro-mwc} is tilted convexity with $\gamma_n=\gamma_p=\gamma$, introduced by Mart\'inez-Rubio~\citep{martinezrubio2022global} in the analysis of Riemannian acceleration: a geodesically convex function on a subset of hyperbolic or spherical space, mapped to Euclidean space by a geodesic map, satisfies it with constants that depend on the radius of the subset. That work notes that the condition is stronger than quasar-convexity and uses it for accelerated first-order methods on convex sets. 
We develop the nonsmooth extended-valued theory, sharp relationships with the secant and operator formulations, 
and an exact polynomial-time online coefficient. 
Tilted-convex functions are $\gamma$--quasar-convex~\citep{martinezrubio2022global,martinezrubio2025smooth}, 
a condition introduced by Hardt, Ma, and Recht and by Hinder, Sidford, and Sohoni~\citep{hardt2018gradient,hinder2020near} and studied online by Pun et al.~\citep{pun2024online}. 
Since quasar-convex functions need not be quasiconvex, they form a strictly larger class; the rates for its constrained form in~\citep{martinezrubio2025smooth} give our offline bounds.

\paragraph{Offline and online approximation coefficients.}
Several established settings have matching offline and online coefficients under their respective access and constraint models. Convex losses have coefficient $1$ in both settings \citep{zinkevich2003online}. For monotone submodular maximization, the greedy coefficient $1-1/e$ is attained online \citep{streeter08_onlin_algor_maxim_submod_funct}, and offline greedy algorithms of a general form convert into online algorithms with the same coefficient \citep{niazadeh21_onlin_learn_offlin_greed_algor}. For monotone continuous DR-submodular maximization, the coefficient is $1-1/e$ both offline and online \citep{chen18_projec_free_onlin_optim_stoch_gradien,zhang19_onlin_contin_submod_maxim,pedramfar2024unified}, and Mualem and Feldman~\cite{mualem2023resolving} showed that the two coefficients also coincide for non-monotone maximization over general convex sets. Linearizable formulations transfer offline guarantees of this kind to the online setting across a broad range of up-concave objectives \citep{pedramfar2024linear,pedramfar2026gamma}. For the three classes of this paper the coefficients differ, $1$ offline and exactly $\gamma$ online for every fixed, effectively specified $\gamma\in(0,1)$, unless $\NP=\RP$.

\paragraph{Online optimization beyond convexity.}
For convex losses, projected online gradient descent achieves $O(\sqrt T)$ static regret with first-order full-information feedback~\citep{zinkevich2003online,hazan2016introduction}. The classical one-point bandit method of Flaxman, Kalai, and McMahan has a $T^{3/4}$ rate, whereas two-point feedback permits square-root-type regret~\citep{flaxman2005online,agarwal2010optimal}. Separate guarantees concern moving comparators or every time interval~\citep{zhang18_adapt_onlin_learn_dynam_envir,hazan09_effic,daniely2015strongly}. For nonconvex losses, one line of work replaces the global benchmark by a local stationarity criterion \citep{hazan2017efficient} or by per-round minimizers \citep{gao2018online}. Another keeps the global benchmark but assumes an oracle that minimizes the sum of past losses plus a linear perturbation \citep{agarwal2019learning,suggala2020online}. We keep the global benchmark and allow access only to individual losses, through their values and generalized gradients. Under this access, Theorem~\ref{introthm:separation} shows that the benchmark can be approached up to the factor $\gamma$ in polynomial time and, unless $\NP=\RP$, not beyond it, for each of the three classes.

\paragraph{Linearizable optimization.}
Pedramfar and Aggarwal~\cite{pedramfar2024linear} defined linearizable and quadratizable optimization, in which a comparison inequality that holds for all comparators is passed to an online linear learner. Later work added uniform wrappers, decentralized and projection-free variants, and weakly up-concave classes \citep{pedramfar2025uniform,lu2025decentralized,pedramfar2026gamma}. We use this reduction without change; what is new is the existence, for every continuous nonpositive $\gamma$--tilted convex loss, of a comparison vector with coefficient exactly $\gamma$, and its computation with polynomially many queries.

\paragraph{Computational limits of online learning.}
Hazan and Koren~\cite{hazan2016computational} established computational limits on online learning with access to an offline optimization oracle. Our lower bound has a different source: each hard loss is minimized by one linear optimization, and the obstruction lies in their sum, which encodes a Label Cover instance whose gap follows from the PCP theorem \citep{moshkovitz2010two,moshkovitz2015projection}. The hard losses lie in the smallest class $\mathcal F^\gamma_{\op{mono}}$, and the threshold of the reduction equals its parameter.

\section{Background and notation}
\label{sec:background}

We use $\overline{\mathbb{R}}$ to denote the set of extended real numbers $\mathbb{R} \cup \{-\infty, +\infty\}$.
For $f : \mathbb{R}^d \to \overline{\mathbb{R}}$, we define the \emph{(effective) domain} of $f$ as the set $\op{dom}(f) := \{\x \in \mathbb{R}^d \mid f(\x) \in \mathbb{R} \}$.
We say $f$ is \emph{proper} if $\op{dom}(f) \neq \emptyset$ and $f > -\infty$.
\footnote{This convention suits minimization; for maximization one instead requires that $-f$ be proper.}
Given a non-empty set $\mathcal X \subseteq \mathbb{R}^d$ and a function $f : \mathcal X \to \overline{\mathbb{R}}$, its \emph{proper extension} $\widetilde f:\mathbb R^d\to\overline{\mathbb R}$ equals $f$ on $\mathcal X$ and $+\infty$ on $\mathbb{R}^d \setminus \mathcal X$, and it is proper if and only if $f$ is proper.
With the indicator $\delta_{\mathcal X}$, which is $0$ on $\mathcal X$ and $+\infty$ elsewhere, we also write $\widetilde f=f+\delta_{\mathcal X}$.
Subdifferentials of $f$, and membership of $f$ in the function classes of Section~\ref{sec:classes}, always refer to $\widetilde f$.
For a convex set $\mathcal K$, we write $\op{aff}(\mathcal K)$ for its affine hull and $\op{ri}(\mathcal K)$ for its relative interior, the interior of $\mathcal K$ relative to $\op{aff}(\mathcal K)$. If $f$ is differentiable at $\x\in\op{ri}(\mathcal K)$ as a function on $\op{aff}(\mathcal K)$, its gradient \emph{with respect to $\op{aff}(\mathcal K)$} is the vector in the linear subspace parallel to $\op{aff}(\mathcal K)$ that represents this derivative.

We say $f : \mathcal X \to \overline{\mathbb{R}}$ is \emph{lower semi-continuous (lsc)} at $\x \in \mathcal X$ if
\[
f(\x) \leq \liminf_{\y \in \C{X}, \y \to \x} f(\y).
\]
The function $f$ is \emph{locally lsc} at $\x$ if it is lsc on an open neighborhood of $\x$.
The function $f$ is lsc at $\x \in \C{X}$ if and only if its proper extension is lsc at $\x$.
\footnote{
The extension may fail to be lsc outside $\C{X}$.
For example, the function $f : (0, \infty) \to \mathbb{R}$ defined by $f(x) = 0$ has $\widetilde{f}(x) = 0$ for $x > 0$ and $\widetilde{f}(x) = \infty$ for $x \leq 0$, so both $f$ and $\widetilde{f}$ are lsc on $(0,\infty)$, but $\widetilde{f}$ is not lsc at $0$.
}
If $f$ is \emph{locally} lsc at $\x$, its proper extension need not be locally lsc at $\x$.
\footnote{
For example, $f : \{(0, 0)\} \cup (0, 1] \times [0, 1] \to \mathbb{R}$ defined identically zero on its domain is locally lsc on its domain, but its proper extension is not locally lsc at $(0,0)$.
}

Given $f : \mathcal X \to \overline{\mathbb R}$, let $\widetilde f$ equal $f$ on $\mathcal X$ and $+\infty$ outside $\mathcal X$. Its \emph{lower semicontinuous envelope} is
\[
\op{lsc}(f)(\x)
:=\liminf_{\y\to\x}\widetilde f(\y)
=\sup_{r>0}\inf\{\widetilde f(\y):\|\y-\x\|<r\},
\]
where the lower limit is nonpunctured and $\inf\varnothing=+\infty$.
It is the largest lower semicontinuous function below $\widetilde f$, or equivalently the largest lower semicontinuous function $g$ satisfying $g\le f$ on $\mathcal X$, so $\op{lsc}(f)=\widetilde f$ whenever $\widetilde f$ is lower semicontinuous.
For a proper $f$, the envelope equals $+\infty$ outside $\overline{\op{dom}(f)}$, but it need not be proper.
For example, $f(x)=1/x$ for $x\ne0$ and $f(0)=0$ is proper, whereas $\op{lsc}(f)(0)=-\infty$.

\begin{definition}
\label{def:radially-continuous}
A function $f : \mathbb R^d \to \overline{\mathbb{R}}$ is \emph{radially continuous}, or \emph{continuous on lines}, if, for all $\x, \y \in \op{dom}(f)$, the restriction of $f$ to the closed line segment $[\x,\y]$ is continuous.
\end{definition}

Given a sequence $\x_k$ and a point $\y$ in $\C{X}$, we say $\x_k$ \emph{converges $f$-attentively} to $\y$, denoted by $\x_k \xrightarrow{f} \y$, when $\x_k \to \y$ and $f(\x_k) \to f(\y)$.

For sets $A$ and $B$, a multi-valued operator $F : A \rightrightarrows B$ is a function from $A$ to the power set of $B$, so that $F(a) \subseteq B$ for each $a \in A$.
The domain of $F$ is
\[
\op{dom}(F) := \{a \in A \mid F(a) \neq \emptyset \}.
\]

We use the following generalized derivatives \cite{rockafellar1998variational}.

\begin{definition}
Let $f : \mathbb{R}^d \to \overline{\mathbb{R}}$. 
The \emph{Fr\'echet subdifferential} of $f$ at $\x \in \op{dom}(f)$ is
\[
\hat{\partial} f(\x)
:=
\left\{
\v \in \mathbb{R}^d \;\middle|\;
\liminf_{\substack{\y \to \x \\ \y \in \op{dom}(f), \y \neq \x}}
\frac{f(\y) - f(\x) - \langle \v, \y - \x \rangle}{\|\y - \x\|}
\ge 0
\right\}.
\]
The \emph{limiting (Mordukhovich) subdifferential} of $f$ at $\x \in \op{dom}(f)$ is
\[
\partial f(\x)
:=
\left\{
\v \in \mathbb{R}^d \;\middle|\;
\exists\, \x_k \in \op{dom}(f),\ \x_k \xrightarrow{f} \x,\ 
\exists\, \v_k \in \hat{\partial} f(\x_k)
\text{ with } \v_k \to \v
\right\},
\]
and the \emph{horizon limiting (Mordukhovich) subdifferential} of $f$ at $\x \in \op{dom}(f)$ is
\[
\partial^\infty f(\x)
:=
\left\{
\vv \in \mathbb{R}^d \;\middle|\;
\begin{array}{l}
\exists\, \x_k \in \operatorname{dom} f,\ 
\exists\, \lambda_k > 0,\ 
\x_k \xrightarrow{f} \x,\ 
\lambda_k \to 0,\\[1mm]
\exists\, \vv_k \in \widehat{\partial} f(\x_k)
\text{ with } \lambda_k \vv_k \to \vv
\end{array}
\right\}.
\]
Finally, if $f$ is proper and locally lsc at $\x \in \op{dom}(f)$, then the \emph{Clarke--Rockafellar subdifferential} (or simply \emph{Clarke subdifferential}) of $f$ at $\x$ is
\footnote{
When $f$ is proper and locally lsc at $\x \in \op{dom}(f)$, this formula agrees with the classical definition of the Clarke--Rockafellar subdifferential; see Theorem 3.57 of~\cite{mordukhovich2006variational1}.
}
\[
\overline{\partial} f(\x)
:= 
\overline{\op{conv}(\partial f(\x)) + \op{conv}(\partial^\infty f(\x))}.
\]
\end{definition}

\begin{proposition}
\label{prop:subdiff-basics}
Let $f,g : \mathbb{R}^d \to \overline{\mathbb{R}}$ be proper and $\x \in \op{dom}(f) \cap \op{dom}(g)$.
Then the following are true.
\begin{itemize}
\item[(i)] If $f$ is locally lsc at $\x$, then 
\[
\overline{\partial} f(\x) \supseteq \partial f(\x) \supseteq \hat{\partial} f(\x).
\]
Moreover, there is a sequence $\x_i \xrightarrow{f} \x$ where $\hat{\partial} f(\x_i)$ is non-empty.
\item[(ii)] If $f$ is locally Lipschitz at $\x$, then $\partial f(\x)$ is non-empty.
Specifically, if
\[
\overline{\nabla} f(\x)
:=
\left\{
\v \in \mathbb{R}^d \;\middle|\;
\exists\, \x_k \in \op{dom}(f),\ \x_k \to \x,\ 
\nabla f(\x_k) \text{ exists },\ 
\nabla f(\x_k) \to \v
\right\},
\]
then $\overline{\nabla} f(\x)$ is a non-empty compact set and 
\[
\overline{\nabla} f(\x) 
\subseteq 
\partial f(\x)
\subseteq
\overline{\partial} f(\x)
= \op{conv}(\overline{\nabla} f(\x)).
\]
\item[(iii)] If $f$ is differentiable at $\x$, then $\partial f(\x) \supseteq \hat{\partial} f(\x) = \{ \nabla f(\x) \}$.
\item[(iv)] If $f$ is continuously differentiable on a neighborhood of $\x$, then $\overline{\partial} f(\x) = \partial f(\x) = \hat{\partial} f(\x) = \{ \nabla f(\x) \}$.
\item[(v)] If $g$ is continuously differentiable on a neighborhood of $\x$, then
\[
\hat{\partial} (f+g)(\x) 
= \hat{\partial} f(\x) + \nabla g(\x).
\]
The same identity holds for the limiting subdifferential. If $f$ is also locally lsc at $\x$, it holds for the Clarke--Rockafellar subdifferential as well.
In particular, let $\mathcal K\subseteq\R^d$ be nonempty, closed, and convex, let $\x\in\mathcal K$, and let $N_{\mathcal K}(\x)=\{\BF n\in\R^d:\ip{\BF n}{\y-\x}\le0\text{ for all }\y\in\mathcal K\}$ be the normal cone of $\mathcal K$ at $\x$.
If $f$ is the indicator $\delta_{\mathcal K}$, then $f+g$ is the proper extension $\widetilde g$ of $g|_{\mathcal K}$, and
\[
\overline{\partial}\widetilde g(\x)=\nabla g(\x)+N_{\mathcal K}(\x).
\]
Since $\BF0\in N_{\mathcal K}(\x)$, this gives $\nabla g(\x)\in\overline{\partial}\widetilde g(\x)$.
\end{itemize}
\end{proposition}

The inclusions in (i) follow from the definitions: for $\v\in\hat{\partial} f(\x)$, the constant sequences $\x_k:=\x$ and $\v_k:=\v$ show that $\v\in\partial f(\x)$.
If $\partial f(\x)\neq\emptyset$, then there are $\x_k\xrightarrow{f}\x$ and $\v_k\in\hat{\partial} f(\x_k)$ with $\v_k\to\v$ for some $\v$, and $\lambda_k:=1/k$ gives $\lambda_k\v_k\to\BF{0}$, so $\BF{0}\in\partial^\infty f(\x)$ and $\partial f(\x)\subseteq\partial f(\x)+\BF{0}\subseteq\overline{\partial} f(\x)$.
The second claim in (i) is Corollary~8.10 in~\cite{rockafellar1998variational}, and parts (ii)--(v) are Theorems~9.13 and~9.61 and Exercise~8.8 in~\cite{rockafellar1998variational}.
For the indicator in (v), the limiting subdifferential $\partial\delta_{\mathcal K}(\x)$ and the horizon subdifferential $\partial^\infty\delta_{\mathcal K}(\x)$ both equal $N_{\mathcal K}(\x)$ \citep[Chapters~6 and~8]{rockafellar1998variational}.
Since $N_{\mathcal K}(\x)$ is a closed convex cone, the definition of the Clarke subdifferential gives $\overline{\partial}\delta_{\mathcal K}(\x)=\overline{N_{\mathcal K}(\x)+N_{\mathcal K}(\x)}=N_{\mathcal K}(\x)$, and the sum rule in (v) for $\overline\partial$ gives the formula for $\overline{\partial}\widetilde g(\x)$.

\paragraph{The tilt function $\psi_\gamma$.}
\label{subsec:psi}

Classical convexity is characterized by the affine support inequality
\[
f(\y)-f(\x)
\ge
\langle \y-\x,\v\rangle,
\qquad
\v \in \overline{\partial} f(\x).
\]
Scaling the left side of a scalar inequality $a\le b$ by a factor $\gamma$ relaxes it for only one sign of $a$: for $\gamma\in(0,1]$ and $a\ge0$, the inequality $\gamma a\le b$ is weaker than $a\le b$, but for $a<0$ it is stronger.
To relax the inequality for both signs, for $\gamma \in (0, 1]$ we define the functions $\psi_\gamma$ and $\overline{\psi}_\gamma$ over extended real numbers with
\begin{align*}
\psi_{\gamma}(z) 
:= \gamma^{\op{sign}(z)} \, z
= \min\{\gamma z, \gamma^{-1} z\}
,\qquad
\overline{\psi}_{\gamma}(z) 
:= \gamma^{-\op{sign}(z)} \, z
= \max\{\gamma z, \gamma^{-1} z\}.
\end{align*}
For $\gamma = 0$ we set
\begin{align*}
\psi_{0}(z) 
:= \begin{cases}
0                 &\t{ if } z > 0 \\
\text{undefined}  &\t{ if } z = 0 \\
-\infty           &\t{ if } z < 0
\end{cases}
,\qquad
\overline{\psi}_{0}(z) 
:= \begin{cases}
\infty            &\t{ if } z > 0 \\
\text{undefined}  &\t{ if } z = 0 \\
0                 &\t{ if } z < 0
\end{cases}
\end{align*}

\begin{lemma}\label{lem:psi_basics}
Let $a, b \in \mathbb R$, $\gamma, \gamma' \in [0,1]$.
Then the following are true.
\begin{enumerate}
\item If $\gamma > 0$, then both $\psi_\gamma$ and $\overline{\psi}_\gamma$ are continuous, monotone increasing, bijective, and positively homogeneous.
\item Whenever $\psi_\gamma(a)$ and $\overline{\psi}_\gamma(a)$ are defined, $\psi_\gamma(a) \leq a \leq \overline{\psi}_\gamma(a)$.
\item $\overline{\psi}_{\gamma}(a) = - \psi_{\gamma} (-a)$ whenever both sides are defined.
\item $\psi_{\gamma}(\psi_{\gamma'}(a)) = \psi_{\gamma \gamma'} (a)$ whenever both sides are defined.
\item If $\gamma > 0$, then 
$\psi_\gamma(a) + \psi_\gamma(b) \leq \psi_\gamma(a + b)$ and $\overline{\psi}_\gamma(a) + \overline{\psi}_\gamma(b) \geq \overline{\psi}_\gamma(a + b)$.
\item If $\gamma > 0$, then $\overline{\psi}_{\gamma}(\psi_{\gamma}(a)) = \psi_{\gamma}(\overline{\psi}_{\gamma}(a)) = a$.
\item If $\gamma' < \gamma$ and $\psi_{\gamma'}(a)$ is defined, then $\psi_{\gamma'}(a) \leq \psi_\gamma(a)$.
\end{enumerate}
\end{lemma}

\begin{proof}
For $\gamma\in(0,1]$ the definitions read
\begin{equation}\label{eq:m1-psi-pieces}
\psi_\gamma(z)=
\begin{cases}
\gamma z, & z\ge0,\\
z/\gamma, & z<0,
\end{cases}
\qquad
\overline{\psi}_\gamma(z)=
\begin{cases}
z/\gamma, & z\ge0,\\
\gamma z, & z<0,
\end{cases}
\end{equation}
and $\psi_\gamma(\pm\infty)=\overline{\psi}_\gamma(\pm\infty)=\pm\infty$.
In particular, for $\gamma>0$ the numbers $\psi_\gamma(z)$, $\overline{\psi}_\gamma(z)$, and $z$ have the same sign.
For $\gamma=0$, both scalar expressions are defined exactly when their argument is nonzero. The calculations below use finite arguments; the same identities extend to infinite arguments whenever all expressions and extended-real sums are defined.

(1) By \eqref{eq:m1-psi-pieces}, $\psi_\gamma$ consists of two linear pieces with positive slopes $\gamma$ and $1/\gamma$ that both vanish at $0$.
Hence $\psi_\gamma$ is continuous and strictly increasing, and it maps $[0,\infty)$ onto $[0,\infty)$ and $(-\infty,0)$ onto $(-\infty,0)$, so it is bijective.
For $t>0$ we have $\op{sign}(tz)=\op{sign}(z)$, so $\psi_\gamma(tz)=\gamma^{\op{sign}(z)}tz=t\,\psi_\gamma(z)$.
The same arguments apply to $\overline{\psi}_\gamma$.

(2) Let $\gamma>0$.
If $a\ge0$, then \eqref{eq:m1-psi-pieces} gives $\psi_\gamma(a)=\gamma a\le a\le a/\gamma=\overline{\psi}_\gamma(a)$.
If $a<0$, then it gives $\psi_\gamma(a)=a/\gamma\le a\le\gamma a=\overline{\psi}_\gamma(a)$.
Let $\gamma=0$, so that $a\neq0$.
If $a>0$, then $\psi_0(a)=0<a<\infty=\overline{\psi}_0(a)$, and if $a<0$, then $\psi_0(a)=-\infty<a<0=\overline{\psi}_0(a)$.

(3) Let $\gamma>0$.
Since $\op{sign}(-a)=-\op{sign}(a)$, we get $-\psi_\gamma(-a)=-\gamma^{-\op{sign}(a)}(-a)=\gamma^{-\op{sign}(a)}a=\overline{\psi}_\gamma(a)$.
Let $\gamma=0$.
If $a>0$, then $-\psi_0(-a)=-(-\infty)=\infty=\overline{\psi}_0(a)$, and if $a<0$, then $-\psi_0(-a)=0=\overline{\psi}_0(a)$.

(4) Let $\gamma,\gamma'>0$.
Since $\psi_{\gamma'}(a)$ has the sign of $a$, we get $\psi_\gamma(\psi_{\gamma'}(a))=\gamma^{\op{sign}(a)}\gamma'^{\op{sign}(a)}a=(\gamma\gamma')^{\op{sign}(a)}a=\psi_{\gamma\gamma'}(a)$.
Now let $\gamma\gamma'=0$.
Then $\psi_{\gamma\gamma'}=\psi_0$, and both sides being defined forces $a\neq0$.
Let $a<0$, so that $\psi_0(a)=-\infty$.
If $\gamma'=0$, then $\psi_{\gamma'}(a)=-\infty$ and $\psi_\gamma(-\infty)=-\infty$.
If $\gamma=0$, then $\psi_{\gamma'}(a)<0$ and hence $\psi_0(\psi_{\gamma'}(a))=-\infty$.
If $a>0$ and $\gamma'=0$, then $\psi_{\gamma'}(a)=0$, at which $\psi_\gamma$ is defined only when $\gamma>0$, and then $\psi_\gamma(0)=0=\psi_0(a)$.
If $a>0$ and $\gamma=0<\gamma'$, then $\psi_0(\gamma'a)=0=\psi_0(a)$.

(5) Let $\gamma>0$.
By definition $\psi_\gamma(z)=\min\{\gamma z,z/\gamma\}$.
For real numbers $p,q,r,s$ we have $\min\{p,q\}+\min\{r,s\}\le\min\{p+r,q+s\}$, since the left-hand side is at most $p+r$ and at most $q+s$.
Applying this with $p=\gamma a$, $q=a/\gamma$, $r=\gamma b$, $s=b/\gamma$ gives
\[
\psi_\gamma(a)+\psi_\gamma(b)
\le\min\{\gamma(a+b),(a+b)/\gamma\}
=\psi_\gamma(a+b).
\]
The inequality for $\overline{\psi}_\gamma(z)=\max\{\gamma z,z/\gamma\}$ follows in the same way from $\max\{p,q\}+\max\{r,s\}\ge\max\{p+r,q+s\}$.

(6) Let $\gamma>0$.
Since $\psi_\gamma(a)$ has the sign of $a$, we get $\overline{\psi}_\gamma(\psi_\gamma(a))=\gamma^{-\op{sign}(a)}\gamma^{\op{sign}(a)}a=a$.
The identity $\psi_\gamma(\overline{\psi}_\gamma(a))=a$ follows in the same way.

(7) Let $\gamma'<\gamma$, so that $\gamma>0$.
If $\gamma'>0$ and $a\ge0$, then $\psi_{\gamma'}(a)=\gamma'a\le\gamma a=\psi_\gamma(a)$.
If $\gamma'>0$ and $a<0$, then $1/\gamma'>1/\gamma$ gives $\psi_{\gamma'}(a)=a/\gamma'<a/\gamma=\psi_\gamma(a)$.
If $\gamma'=0$, then $a\neq0$; for $a>0$ we have $\psi_0(a)=0<\gamma a=\psi_\gamma(a)$, and for $a<0$ we have $\psi_0(a)=-\infty<\psi_\gamma(a)$.
\end{proof}

\section{The Function Classes}
\label{sec:classes}

This section applies the distortion $\psi_\gamma$ of Section~\ref{sec:background} to the gradient inequality, to monotonicity of the subdifferential, and to the increase of secant slopes, which is equivalent to Jensen's inequality (Sections~\ref{subsec:def-mwc}--\ref{subsec:def-jensen}). Section~\ref{subsec:abstract} describes the secant class as the envelopes of tilted affine functions, and Section~\ref{subsec:end-point} describes the endpoints $\gamma=1$, $0^+$, and $0$. Throughout, $\mathcal K\subseteq\R^d$ denotes a convex set.

\subsection{The gradient inequality: \texorpdfstring{$\gamma$}{gamma}--tilted convexity}
\label{subsec:def-mwc}

\begin{definition}[$\gamma$--tilted convexity]
\label{def:gamma-MWC}
Let $f : \mathbb{R}^d \to \overline{\mathbb{R}}$ be a proper lsc function, and let $F : \mathbb R^d \rightrightarrows \mathbb R^d$ be a multi-valued operator.
For $\gamma \in [0,1]$, we say $f$ is
\emph{$\gamma$--tilted convex with respect to $F$} if, for all $\x, \y \in \op{dom}(f)$ and $\v \in F(\x)$,
\begin{align}
f(\y) - f(\x)
\;\ge\;
\psi_\gamma(\langle \y - \x, \v \rangle)
\label{eq:gamma-mwc}
\end{align}
whenever the right-hand side is well-defined.
We say $f$ is $\gamma$--tilted convex if it is $\gamma$--tilted convex with respect to $\overline{\partial} f$.
\footnote{
Similarly, we say a usc function $f : \C{K} \to \mathbb{R}$ is \emph{$\gamma$--tilted concave with respect to $F$} if $-f$ is $\gamma$--tilted convex with respect to $-F$. 
Equivalently, for all $\x, \y \in \op{dom}(f)$ and $\v \in F(\x)$,
\begin{align*}
f(\y) - f(\x)
\;\le\;
\overline{\psi}_\gamma(\langle \y - \x, \v \rangle),
\end{align*}
whenever the right-hand side is well-defined.
We say $f$ is $\gamma$--tilted concave if $-f$ is $\gamma$--tilted convex, that is, if $f$ is $\gamma$--tilted concave with respect to $- \overline{\partial} (-f)$, which in general differs from $\overline{\partial} f$.
}
For any such function we define $\gamma^*(f):=\sup\{\gamma\in[0,1]: f\text{ is }\gamma\text{--tilted convex}\}$, with $\sup\emptyset:=0$.
\end{definition}

For a differentiable function and $\v=\nabla f(\x)$, the support inequality is tilted convexity with constants $\gamma_n=\gamma_p=\gamma$~\citep[Eq.~(3)]{martinezrubio2022global}. 
On a closed convex feasible set, the Clarke subdifferential of a continuously differentiable ambient function adds normal vectors, whose inner products with feasible directions are nonpositive, so the identification persists. The definition above extends the condition to proper lsc extended-valued functions; convexity of the effective domain and equivalence of the subdifferential formulations are proved in Section~\ref{sec:structure}.

If $F(\x) \subseteq G(\x)$ for every $\x$ and $f$ is $\gamma$--tilted convex with respect to $G$, then $f$ is $\gamma$--tilted convex with respect to $F$, so by Proposition~\ref{prop:subdiff-basics}(i) a $\gamma$--tilted convex function is also $\gamma$--tilted convex with respect to $\partial f$ and $\hat{\partial} f$.

For $\gamma\in[0,1]$ we write $\mathcal F^\gamma$ for the class of proper lsc $\gamma$--tilted convex functions $f:\R^d\to\overline{\R}$. By the convention of Section~\ref{sec:background}, a function on a set $\mathcal K$ lies in $\mathcal F^\gamma$ when its proper extension does.

\subsection{Monotone generalized gradients: \texorpdfstring{$\gamma$}{gamma}--tilted monotonicity}
\label{subsec:def-mono}

\begin{definition}[$\gamma$--tilted monotone operator]
\label{def:gamma-MW-mono}
For $\gamma \in [0,1]$, we say the multi-valued operator $F : \mathbb R^d \rightrightarrows \mathbb{R}^d$ is
\emph{$\gamma$--tilted monotone} if for all
$\x, \y \in \op{dom}{F}$, $\x^* \in F(\x)$, and $\y^* \in F(\y)$,
\begin{align}
\langle \y^*, \y - \x \rangle
\geq \psi_{\gamma}(\langle \x^*, \y - \x \rangle)
= 
\gamma^{\op{sign}\!\left(
\langle \x^*, \y - \x \rangle
\right)}
\,
\langle \x^*, \y - \x \rangle,
\label{eq:gamma-mw-mono}
\end{align}
whenever both sides are well-defined.
\end{definition}

\begin{definition}
For $\gamma\in[0,1]$, the class $\mathcal F^\gamma_{\op{mono}}$ consists of the proper lsc functions $f:\R^d\to\overline{\R}$ such that $\overline{\partial} f$ is $\gamma$--tilted monotone. For a proper lsc function $f$, $\gamma^*_{\op{mono}}(f):=\sup\{\gamma\in[0,1]:\overline{\partial} f\text{ is }\gamma\text{--tilted monotone}\}$, with $\sup\emptyset:=0$.
\end{definition}

\subsection{Secant slopes: \texorpdfstring{$\gamma$}{gamma}--tilted secant convexity}
\label{subsec:def-jensen}

A function is convex exactly when the slopes of its secants increase: if $\x$ lies strictly between $\z$ and $\y$ on a line, the slope of the secant over $[\x,\y]$ is at least the slope of the secant over $[\z,\x]$, and we apply $\psi_\gamma$ to this latter slope.

\begin{definition}[$\gamma$--tilted secant convexity]
\label{def:gamma-MW-Jensen-convex}
Let $\gamma\in[0,1]$ and $f:\mathcal K\to\mathbb R \cup \{+\infty\}$. We say that $f$ is \emph{$\gamma$--tilted secant convex} if, for all $\y\ne\z$ in $\op{dom}(f)$ and $\lambda\in(0,1)$, the point $\x:=\lambda\y+(1-\lambda)\z$ lies in $\op{dom}(f)$ and the secant slopes
\begin{equation}\label{eq:secant-slopes}
s_-:=\frac{f(\x)-f(\z)}{\norm{\x-\z}},
\qquad
s_+:=\frac{f(\y)-f(\x)}{\norm{\y-\x}}
\end{equation}
satisfy
\begin{equation}\label{eq:secant-form}
s_+\ge\psi_\gamma(s_-)
\end{equation}
whenever $\psi_\gamma(s_-)$ is defined. For any such function we define
\[
\gamma^*_{\op{sec}}(f):=\sup\{\gamma\in[0,1]: f\text{ is }\gamma\text{--tilted secant convex}\},
\qquad\sup\emptyset:=0.
\]
\end{definition}

For $\gamma=1$, \eqref{eq:secant-form} is the increase of secant slopes. In general, \eqref{eq:secant-form} applies the distortion $\psi_\gamma$ to secant slopes, as~\eqref{eq:gamma-mwc} applies it to supports and~\eqref{eq:gamma-mw-mono} to subdifferentials. Requiring $\psi_\gamma(s_-)$ to be defined excludes only the case $\gamma=0$ and $s_-=0$. An equivalent form in terms of function values, which weakens Jensen's inequality, is given in Section~\ref{subsec:jensen-class} (Proposition~\ref{prop:secant-jensen}).

The three suprema $\gamma^*$, $\gamma^*_{\op{mono}}$, and $\gamma^*_{\op{sec}}$ are attained whenever they are positive. Each of the three conditions \eqref{eq:gamma-mwc}, \eqref{eq:gamma-mw-mono}, and \eqref{eq:secant-form} with parameter $\gamma$ implies the same condition with every smaller parameter, by Lemma~\ref{lem:psi_basics}(7). For fixed $z$, the value $\psi_\gamma(z)=\gamma^{\op{sign}(z)}z$ depends continuously on $\gamma\in(0,1]$, so a condition that holds for every $\gamma<\gamma_0$ with $\gamma_0>0$ also holds at $\gamma_0$. Hence, if $\gamma^*(f)>0$, then $f$ is $\gamma$--tilted convex exactly for $\gamma\in[0,\gamma^*(f)]$, and the same holds for $\gamma^*_{\op{mono}}$ and $\gamma^*_{\op{sec}}$.

For $\gamma\in[0,1]$ we write $\mathcal F^\gamma_{\op{sec}}$ for the class of proper lsc $\gamma$--tilted secant convex functions $f:\R^d\to\overline{\R}$.

\subsection{Envelope description of the secant class}
\label{subsec:abstract}

Proper lower semicontinuous convex functions have a fourth description: they are the pointwise suprema of affine functions. Applying the same distortion to the generators gives the tilted affine functions $\y\mapsto c+\psi_\gamma(\langle\y-\w,\z\rangle)$ and their suprema, the $\mathcal H_\gamma$--convex functions defined below. This produces no new class: Theorem~\ref{thm:H-convex-equivalent-def} shows that the proper $\mathcal H_\gamma$--convex functions are exactly the members of $\mathcal F^{\gamma^2}_{\op{sec}}$. The envelope form makes closure under suprema immediate, and it produces the nonsmooth examples of Section~\ref{subsec:examples-jensen}. We use the following standard notion of abstract convexity.

\begin{definition}
Let $\mathcal X$ be a set and let
$\mathcal H$ be a family of functions
$h:\mathcal X\to \overline{\mathbb R}$.
A function
$f:\mathcal X\to \overline{\mathbb R}$
is called \emph{$\mathcal H$--convex} if there exists a non-empty set 
$\mathcal U\subseteq\mathcal H$ such that
\[
f(\x)
=
\sup_{h\in\mathcal U} h(\x),
\qquad
\forall\,
\x\in\mathcal X.
\]
\end{definition}

Among proper functions on $\mathbb R^d$, $\mathcal H$--convexity for the family of affine functions is exactly lower semicontinuous convexity.

\begin{definition}
Fix $\gamma\in(0,1]$.
Define
\[
\mathcal H_\gamma
:=
\{
\y \mapsto c + \psi_\gamma(\langle \y-\w,\z\rangle) : \mathbb R^d \to \mathbb R
\mid
c\in\mathbb R, \z,\w\in\mathbb R^d
\}.
\]
A function $f : \mathbb R^d \to \overline{\mathbb R}$ is thus \emph{$\mathcal H_\gamma$--convex} if $f(\x) = \sup_{h \in \C{U}} h(\x)$ for all $\x \in \mathbb R^d$ and some non-empty $\C{U} \subseteq \C{H}_\gamma$.
\end{definition}

Each element of $\mathcal H_\gamma$ is piecewise affine, with gradient $\gamma\z$ on one side of the hyperplane $\{\y:\langle \y-\w,\z\rangle=0\}$ and gradient $\z/\gamma$ on the other.
The family $\mathcal H_1$ consists of the affine functions, so its proper suprema are exactly the proper lower semicontinuous convex functions.
The parameter is squared because the two slopes of a member of $\mathcal H_\gamma$ along $\z$ differ by the factor $\gamma^2$, and Example~\ref{ex:two-slope} shows that $\gamma^2$ is then the exact secant parameter. Equivalently, $\mathcal F^\gamma_{\op{sec}}$ consists of the proper suprema of members of $\mathcal H_{\sqrt\gamma}$.

\subsection{Endpoints}
\label{subsec:end-point}

At $\gamma=1$, the support, monotonicity, and secant conditions of Sections~\ref{subsec:def-mwc}--\ref{subsec:def-jensen} all characterize convexity, and $\mathcal H_1$ is the family of affine functions; at $\gamma=0$, the secant condition characterizes quasiconvexity. For the support and monotonicity conditions, retaining or omitting the inequality at a zero inner product gives the distinct endpoints $0^+$ and $0$, which under radial continuity characterize pseudoconvexity and quasiconvexity, respectively. The union of the positive-parameter classes is a strict subset of the pseudoconvex class (Theorem~\ref{thm:hierarchy}).
Theorems~\ref{thm:mwc_one}, \ref{thm:mwc_zero}, and~\ref{thm:mwc_zero_pseudo} restate the corresponding classical characterizations in terms of $\psi_\gamma$.

For the secant condition, both endpoints follow from Definition~\ref{def:gamma-MW-Jensen-convex}. At $\gamma=1$, $\psi_1$ is the identity, so~\eqref{eq:secant-form} is the increase of secant slopes, which characterizes convexity. At $\gamma=0$, $\psi_0(s_-)$ equals $0$ for $s_->0$ and $-\infty$ for $s_-<0$, so~\eqref{eq:secant-form} says that $f(\x)>f(\z)$ implies $f(\y)\ge f(\x)$, that is, $f(\x)\le\max\{f(\y),f(\z)\}$. Together with the requirement $\x\in\op{dom}(f)$, this is quasiconvexity.

\begin{theorem}\label{thm:mwc_one}
Let $f : \mathbb R^d \to \overline{\mathbb{R}}$ be a proper lsc function.
Then the following are equivalent.
\begin{itemize}
\item[(i)] For all $\x,\y \in \op{dom}(f)$ and $\v \in \overline{\partial} f(\x)$, we have
$f(\y)-f(\x)\ge \langle \y-\x,\v\rangle$.
\item[(ii)] The function $f$ is $1$--tilted convex.
\item[(iii)] The function $f$ is convex.
\item[(iv)] The operator $\overline{\partial} f$ is $1$--tilted monotone.
\item[(v)] The operator $\overline{\partial} f$ is monotone.
\end{itemize}
\end{theorem}

\begin{proof}
For every $z \in \mathbb{R}$ we have $\psi_1(z)=1^{\op{sign}(z)}z=z$, and this value is always defined.
Hence \eqref{eq:gamma-mwc} with $\gamma=1$ and $F=\overline{\partial} f$ reads
\[
f(\y)-f(\x)\ge\langle \y-\x,\v\rangle,
\qquad
\x,\y\in\op{dom}(f),\ \v\in\overline{\partial} f(\x),
\]
which is condition~(i), so (i) and (ii) are equivalent.
Likewise, \eqref{eq:gamma-mw-mono} with $\gamma=1$ and $F=\overline{\partial} f$ reads $\langle \y^*,\y-\x\rangle\ge\langle \x^*,\y-\x\rangle$, that is,
\[
\langle \y^*-\x^*,\y-\x\rangle\ge0,
\qquad
\x,\y\in\op{dom}(\overline{\partial} f),\ \x^*\in\overline{\partial} f(\x),\ \y^*\in\overline{\partial} f(\y),
\]
which is monotonicity of $\overline{\partial} f$, so (iv) and (v) are equivalent.
For a proper lsc convex function, the Clarke subdifferential is the usual convex subdifferential, so (iii) implies (i). Adding the inequalities in (i) with the two points interchanged shows that (i) implies (v). Finally, (v) implies (iii) by the classical monotonicity characterization of convexity; see \cite[Theorem~2.1(i)]{daniilidis1999subdifferentials}.
\end{proof}

\begin{definition}[Quasimonotonicity]
A set-valued operator $F:\mathbb R^d \rightrightarrows \mathbb R^d$ is \emph{quasimonotone} if, for all
$\x,\y\in\op{dom}(F)$, $\x^\ast\in F(\x)$, and $\y^\ast\in F(\y)$,
\[
\langle \x^\ast,\y-\x\rangle>0
\quad\Longrightarrow\quad
\langle \y^\ast,\y-\x\rangle\ge 0.
\]
\end{definition}

\begin{theorem}\label{thm:mwc_zero}
Let $f : \mathbb R^d \to \overline{\mathbb{R}}$ be a proper, lsc, and radially continuous function.
Then the following are equivalent.
\begin{enumerate}
\item[(i)] For all $\x, \y \in \op{dom}(f)$ and $\v \in \overline{\partial} f(\x)$ such that
$\langle \y - \x, \v \rangle > 0$,
we have
$f(\y) \ge f(\x)$.
\item[(ii)] The function $f$ is $0$--tilted convex.
\item[(iii)] The function $f$ is quasiconvex.
\item[(iv)] The operator $\overline{\partial} f$ is $0$--tilted monotone.
\item[(v)] The operator $\overline{\partial} f$ is quasimonotone.
\end{enumerate}
\end{theorem}

\begin{proof}
By definition, $\psi_0(z)=0$ for $z > 0$, $\psi_0(z)=-\infty$ for $z < 0$,  while $\psi_0(0)$ is undefined.
Fix $\x,\y\in\op{dom}(f)$ and $\v\in\overline{\partial} f(\x)$, and put $z:=\langle \y-\x,\v\rangle$.
Condition~(ii) imposes $f(\y)-f(\x)\ge\psi_0(z)$ only when $z\neq0$.
If $z<0$, this reads $f(\y)-f(\x)\ge-\infty$, which always holds because $f(\x),f(\y)\in\mathbb R$.
If $z>0$, it reads $f(\y)-f(\x)\ge0$.
Consequently, condition~(ii) is equivalent to requiring that
$f(\y)\ge f(\x)$
whenever
$\langle \y-\x,\v\rangle>0$,
which is condition~(i), so (i) and (ii) are equivalent.

Similarly, fix $\x,\y\in\op{dom}(\overline{\partial} f)$, $\x^\ast\in\overline{\partial} f(\x)$, and $\y^\ast\in\overline{\partial} f(\y)$.
With $\gamma=0$, inequality \eqref{eq:gamma-mw-mono} is imposed only when $\langle \x^\ast,\y-\x\rangle\neq0$.
It always holds when $\langle \x^\ast,\y-\x\rangle<0$, since its right-hand side is then $-\infty$; when $\langle \x^\ast,\y-\x\rangle>0$, its right-hand side is $0$.
So Definition~\ref{def:gamma-MW-mono} with $\gamma=0$ reduces to the implication
\[
\langle \x^\ast,\y-\x\rangle>0
\quad\Longrightarrow\quad
\langle \y^\ast,\y-\x\rangle\ge 0,
\]
which is exactly quasimonotonicity, so (iv) and (v) are equivalent.

Under radial continuity, (i) and (iii) are equivalent by \cite[Theorem~2.1 and Proposition~2.1]{aussel1998subdifferential}. The equivalence of (iii) and (v) is the classical quasimonotonicity characterization of quasiconvexity; see \cite[Theorem~2.1(ii)]{daniilidis1999subdifferentials}.
\end{proof}

\begin{remark}
Under stronger regularity, similar results hold for $\partial f$: if $f$ is locally Lipschitz, then~\cite{soleimanidamaneh07_chara_nonsm} shows that
\[
(i) \Leftrightarrow (ii) \Leftrightarrow (iii) \Rightarrow (iv) \Leftrightarrow (v).
\]
\end{remark}

\begin{definition}[Pseudoconvexity]
A differentiable function $f:\mathcal K\to\mathbb R$ is \emph{pseudoconvex} if, for all
$\x,\y\in\mathcal K$,
\[
\langle \y-\x,\nabla f(\x)\rangle \ge 0
\quad\Longrightarrow\quad
f(\y)\ge f(\x).
\]
More generally~\cite{hadjisavvas2006handbook}, a proper, lsc, and radially continuous function $f : \mathbb R^d \to \overline{\mathbb{R}}$ is \emph{pseudoconvex} if, for all $\x, \y \in \op{dom}(f)$ and $\vv \in \overline{\partial} f(\x)$, we have
\[
\langle \y-\x,\vv\rangle \ge 0
\quad\Longrightarrow\quad
f(\y)\ge f(\x).
\]
\end{definition}

If radial continuity is omitted, pseudoconvexity has several inequivalent definitions~\cite{hadjisavvas2006handbook}.

\begin{definition}[Pseudomonotonicity]
A set-valued operator $F:\mathbb R^d \rightrightarrows \mathbb R^d$ is \emph{pseudomonotone} if, for all
$\x,\y\in \op{dom}(F)$, $\x^\ast\in F(\x)$, and $\y^\ast\in F(\y)$,
\[
\langle \x^\ast,\y-\x\rangle \ge 0
\quad\Longrightarrow\quad
\langle \y^\ast,\y-\x\rangle \ge 0.
\]
\end{definition}

Define
\begin{align*}
\psi_{0^+}(z) 
:= \begin{cases}
0                 &\t{ if } z \ge 0; \\
-\infty           &\t{ if } z < 0.
\end{cases}
\end{align*}
The functions $\psi_0$ and $\psi_{0^+}$ differ only at $z = 0$, where $\psi_0$ is undefined.

\begin{definition}
\label{def:mwc-pseudo}
Let $f : \mathbb R^d \to \overline{\mathbb{R}}$ be a proper lsc function.
We say $f$ is $0^+$--tilted convex if, for all $\x, \y \in \op{dom}(f)$ and $\v \in \overline{\partial} f(\x)$, we have
\begin{align*}
f(\y) - f(\x)
\;\ge\;
\psi_{0^+}(\langle \y - \x, \v \rangle).
\end{align*}
A set-valued operator $F:\mathbb R^d \rightrightarrows \mathbb R^d$ is $0^+$--tilted monotone if, for all
$\x,\y\in \op{dom}(F)$, $\x^\ast\in F(\x)$, and $\y^\ast\in F(\y)$,
\[
\langle \y^\ast,\y-\x\rangle \ge \psi_{0^+}(\langle \x^\ast,\y-\x\rangle).
\]
\end{definition}

We write $\mathcal F^{0^+}$ for the class of proper lsc $0^+$--tilted convex functions.

\begin{theorem}\label{thm:mwc_zero_pseudo}
Let $f : \mathbb R^d \to \overline{\mathbb{R}}$ be a proper, lsc, and radially continuous function.
Then the following are equivalent.
\begin{enumerate}
\item[(i)] For all $\x,\y\in \op{dom}(f)$ and $\v \in \overline{\partial} f(\x)$ such that
$\langle \y-\x,\v\rangle \ge 0$,
we have $f(\y)\ge f(\x)$.
\item[(ii)] The function $f$ is $0^+$--tilted convex.
\item[(iii)] The function $f$ is pseudoconvex.
\item[(iv)] The operator $\overline{\partial} f$ is $0^+$--tilted monotone.
\item[(v)] The operator $\overline{\partial} f$ is pseudomonotone.
\end{enumerate}
\end{theorem}

\begin{proof}
Fix $\x,\y\in\op{dom}(f)$ and $\v\in\overline{\partial} f(\x)$, and put $z:=\langle \y-\x,\v\rangle$.
If $z<0$, then $\psi_{0^+}(z)=-\infty$, and the inequality $f(\y)-f(\x)\ge\psi_{0^+}(z)$ always holds.
If $z\ge0$, then $\psi_{0^+}(z)=0$, and the inequality reads $f(\y)-f(\x)\ge0$.
Hence condition~(ii) is equivalent to requiring that
$f(\y)\ge f(\x)$
whenever
$\langle \y-\x,\v\rangle \ge 0$,
which is condition~(i), and condition~(i) is the definition of pseudoconvexity of a proper, lsc, and radially continuous function, that is, condition~(iii).
Therefore, (i), (ii), and (iii) are equivalent.

Similarly, fix $\x,\y\in\op{dom}(\overline{\partial} f)$, $\x^\ast\in\overline{\partial} f(\x)$, and $\y^\ast\in\overline{\partial} f(\y)$.
The inequality $\langle \y^\ast,\y-\x\rangle\ge\psi_{0^+}(\langle \x^\ast,\y-\x\rangle)$ always holds when $\langle \x^\ast,\y-\x\rangle<0$, and it reads $\langle \y^\ast,\y-\x\rangle\ge0$ when $\langle \x^\ast,\y-\x\rangle\ge0$.
So $0^+$--tilted monotonicity in Definition~\ref{def:mwc-pseudo} is equivalent to the implication
\[
\langle \x^\ast,\y-\x\rangle \ge 0
\quad\Longrightarrow\quad
\langle \y^\ast,\y-\x\rangle \ge 0,
\]
which is exactly pseudomonotonicity, so (iv) and (v) are equivalent.

The equivalence of (iii) and (v) under radial continuity is the classical pseudomonotonicity characterization of pseudoconvexity; see \cite[Theorem~2.1(iii)]{daniilidis1999subdifferentials}.
\end{proof}

Smaller parameters give weaker conditions: for all $\gamma', \gamma \in (0, 1]$ with $\gamma' < \gamma$,
\begin{equation}\label{eq:psi-endpoint-chain}
\psi_0(z) \le \psi_{0^+}(z) \le \psi_{\gamma'}(z) \le \psi_{\gamma}(z)
\end{equation}
for all $z \in \mathbb R$ at which both sides of each inequality are defined.
The last inequality is Lemma~\ref{lem:psi_basics}(7).
The middle one holds because $\psi_{0^+}(z)=0\le\gamma'z=\psi_{\gamma'}(z)$ for $z\ge0$ and $\psi_{0^+}(z)=-\infty$ for $z<0$, and the first holds with equality because $\psi_0$ and $\psi_{0^+}$ agree wherever $\psi_0$ is defined.

Hence an inequality $f(\y)-f(\x)\ge\psi_\theta(\langle\y-\x,\v\rangle)$ implies the same inequality for every smaller parameter in \eqref{eq:psi-endpoint-chain}, and
\begin{equation}\label{eq:gradient-class-chain}
\mathcal F^1\subseteq\mathcal F^\gamma\subseteq\mathcal F^{\gamma'}\subseteq\mathcal F^{0^+}\subseteq\mathcal F^0,
\qquad 0<\gamma'<\gamma\le1.
\end{equation}
By Theorem~\ref{thm:mwc_one}, $\mathcal F^1$ is the class of proper lsc convex functions. By Theorems~\ref{thm:mwc_zero_pseudo} and~\ref{thm:mwc_zero}, a proper, lsc, and radially continuous function lies in $\mathcal F^{0^+}$ exactly when it is pseudoconvex, and in $\mathcal F^0$ exactly when it is quasiconvex. Theorem~\ref{thm:hierarchy} in Section~\ref{subsec:chain} shows that every inclusion in \eqref{eq:gradient-class-chain} is strict when $\gamma<1$.

\section{Structural Properties}
\label{sec:structure}

Section~\ref{subsec:mwc-props} shows that, like convex functions, members of $\mathcal F^\gamma$ have convex domains, are locally Lipschitz on the relative interior of their domains, are radially continuous, and their local minimizers are global. Sections~\ref{subsec:mono-class} and~\ref{subsec:jensen-class} compare $\mathcal F^\gamma_{\op{mono}}$ and $\mathcal F^\gamma_{\op{sec}}$ with $\mathcal F^\gamma$ and transfer these properties to them, and Section~\ref{subsec:H-equal} identifies the $\mathcal H_\gamma$--convex functions with $\mathcal F^{\gamma^2}_{\op{sec}}$. Section~\ref{subsec:separations} separates the classes and collects properties of convex functions that they lack, and Section~\ref{subsec:chain} combines the comparisons into one chain of inclusions and orders each class strictly in its parameter.

\subsection{Properties of \texorpdfstring{$\gamma$}{gamma}--tilted convex functions}
\label{subsec:mwc-props}

\subsubsection{Convex domains and global minimizers}
\label{subsec:domain-minimizers}

Convexity of the domain is a consequence of $\gamma$--tilted convexity rather than an assumption.

\begin{theorem}
\label{thm:mwc-domain-is-convex}
Let $f : \mathbb{R}^{d} \to \overline{\mathbb{R}}$ be proper, lsc, and $\gamma$--tilted convex for some $\gamma \in (0, 1]$.
Then $\op{dom}(f)$ is convex and $f$ is quasiconvex.
\end{theorem}

The proof is in Appendix~\ref{apx:mwc-domain-is-convex}.
For convex functions every local minimizer is a global minimizer; Theorem~\ref{thm:local-min-is-global-min}, proved in Appendix~\ref{apx:local-min-is-global-min}, extends this to $\gamma$--tilted convex functions and characterizes their minimizers by $\BF{0}\in\hat{\partial} f(\x)$.

\begin{theorem}
\label{thm:local-min-is-global-min}
Let $f:\mathbb R^d\to\overline{\mathbb R}$ be proper, lsc, and
$\gamma$--tilted convex for some $\gamma\in(0,1]$.
Then, for every $\x\in\op{dom}(f)$, the following are equivalent.
\begin{itemize}
\item[(i)] $\x$ is a global minimizer of $f$. 
\item[(ii)] $\x$ is a local minimizer of $f$. 
\item[(iii)] $\BF{0} \in \hat{\partial} f(\x)$.
\item[(iv)] $\BF{0} \in \overline{\partial} f(\x)$.
\end{itemize}
\end{theorem}

\subsubsection{Regularity}
\label{subsec:regularity}

The regularity and limit results of this section rest on one tool, proved in Appendix~\ref{apx:approx}. A proper lsc function $f$ is \emph{approximately $\gamma$--tilted convex} with respect to an operator $F$ (Definition~\ref{def:approx-mwc}) if, for every $\varepsilon>0$, each $\x\in\op{dom}(f)\cap\op{dom}(F)$ has a vector $\v\in F(\x)$ with $f(\y)-f(\x)\ge\psi_\gamma(\langle\v,\y-\x\rangle)-\varepsilon$ for all $\y\in\op{dom}(f)$; the vector may depend on $\varepsilon$. Every member of $\mathcal F^\gamma$ has this property with $\varepsilon=0$ and $F$ the Fr\'echet subdifferential, whose domain is dense in $\op{dom}(f)$. When $\mathcal K:=\op{dom}(f)$ is convex and $\op{dom}(F)$ is dense in $\mathcal K$, approximate supports already force local Lipschitz continuity on $\op{ri}(\mathcal K)$, with a constant that depends only on a local bound for $f$, and continuity along every line through $\op{ri}(\mathcal K)$. This gives Theorems~\ref{thm:gamma-MW-convex-is-locally-Lipschitz} and~\ref{thm:gamma-MW-continuous-on-lines} below and the characterizations of Theorem~\ref{thm:mwc-alternative-characterizations}; since the constant is uniform along a locally bounded sequence, it also gives the line and limit results of Section~\ref{subsec:lines-limits}.

Convex functions are locally Lipschitz on the relative interior of their domains, and so are $\gamma$--tilted convex functions.

\begin{theorem}\label{thm:gamma-MW-convex-is-locally-Lipschitz}
Let $f:\mathbb R^d\to\overline{\mathbb R}$ be proper, lsc, and
$\gamma$--tilted convex for some $\gamma\in(0,1]$, and set
$\mathcal K:=\op{dom}(f)$.  Then
$f|_{\op{ri}(\mathcal K)}$ is locally Lipschitz.
\end{theorem}

The proof is in Appendix~\ref{apx:gamma-MW-convex-is-locally-Lipschitz}.
Such a function need not be locally Lipschitz at the boundary, even when it is continuous: the convex function $f(x) = - \sqrt{x}$ on $[0,1]$ has $f'(x) = -1/(2\sqrt{x})\to-\infty$ as $x\to0^+$, so it is not locally Lipschitz at $0$.

A proper lsc convex function need not even be continuous on the boundary of its domain.
For example, take $f : \mathbb R^2 \to \overline{\mathbb R}$ defined by
\[
f(x,y) := \begin{cases}
x^2/y         &\text{ if } y > 0; \\
0             &\text{ if } x = y = 0; \\
\infty        &\text{ otherwise.} \\
\end{cases}
\]
This function is convex and lsc but not continuous at $(0,0)$. It is, however, radially continuous (Definition~\ref{def:radially-continuous}), as is every proper lsc convex function.
Theorem~\ref{thm:gamma-MW-continuous-on-lines} proves this property for $\gamma$--tilted convex functions on $\mathbb R$, and Section~\ref{subsec:lines-limits} extends it to $\mathbb R^d$.

\begin{theorem}\label{thm:gamma-MW-continuous-on-lines}
Let $f:\mathbb R\to\overline{\mathbb R}$ be proper, lsc, and
$\gamma$--tilted convex for some $\gamma\in(0,1]$, and set
$\mathcal K:=\op{dom}(f)$.  Then $f|_{\mathcal K}$ is 
continuous.
\end{theorem}

The proof is in Appendix~\ref{apx:gamma-MW-continuous-on-lines}.
\subsubsection{Equivalent characterizations}
\label{subsec:characterizations}

For $\gamma\in(0,1]$, the Clarke subdifferential in the definition of $\gamma$--tilted convexity can be replaced by the limiting or the Fr\'echet subdifferential, or by gradients at the points of the relative interior where they exist.

\begin{theorem}
\label{thm:mwc-alternative-characterizations}
Let $\gamma\in(0,1]$ and let $f:\mathbb R^d\to\overline{\mathbb R}$ be a proper lsc function with effective domain $\C{K} := \op{dom}(f)$.
Then the following are equivalent.
\begin{itemize}
\item[(i)] $f$ is $\gamma$--tilted convex.
\item[(ii)] $f$ is $\gamma$--tilted convex with respect to $\partial f$.
\item[(iii)] $f$ is $\gamma$--tilted convex with respect to $\hat{\partial} f$.
\item[(iv)] $\mathcal K$ is convex and $f$ is $\gamma$--tilted convex with respect to $\hat{\partial} f|_{\op{ri}(\C{K})}$, i.e., we have
\[
f(\y) - f(\x) \geq \psi_\gamma(\langle \vv, \y - \x \rangle),
\]
for all $\y \in \C{K}$, $\x \in \op{ri}(\C{K})$, and $\vv \in \hat{\partial} f(\x)$.
\item[(v)] $\mathcal K$ is convex, the set of points $\x \in \op{ri}(\C{K})$ where $\nabla f(\x)$ exists is dense in $\C{K}$ and $f$ is $\gamma$--tilted convex with respect to $\x \mapsto \{\nabla f(\x)\}: \op{ri}(\C{K}) \rightrightarrows \mathbb R^d$, i.e., we have
\[
f(\y) - f(\x) \geq \psi_\gamma(\langle \nabla f(\x), \y - \x \rangle),
\]
for all $\y \in \C{K}$, $\x \in \op{ri}(\C{K})$ where $\nabla f(\x)$ exists.
Here $\nabla f(\x)$ is considered with respect to the affine hull of $\C{K}$.
\end{itemize}
\end{theorem}

The proof is in Appendix~\ref{apx:mwc-alternative-characterizations}.
For functions that are continuously differentiable on the relative interior, the equivalence of (i) and (v) gives the following test.

\begin{corollary}\label{cor:mwc-smooth-test}
Let $\gamma\in(0,1]$ and let $f:\mathbb R^d\to\overline{\mathbb R}$ be a proper, lsc, and radially continuous function with a convex effective domain $\C{K} := \op{dom}(f)$ such that $f|_{\op{ri}(\mathcal K)}$ is continuously differentiable.
Then $f$ is $\gamma$--tilted convex if and only if
\[
f(\y) - f(\x) \geq \psi_\gamma(\langle \nabla f(\x), \y - \x \rangle),
\]
for all $\x, \y \in \op{ri}(\C{K})$.
Here $\nabla f(\x)$ is considered with respect to the affine hull of $\C{K}$.
\end{corollary}

\begin{proof}
Since $\C{K}$ is non-empty and convex, $\op{ri}(\C{K})$ is non-empty and $\overline{\op{ri}(\C{K})}=\overline{\C{K}}$ (Theorems~6.2 and~6.3 in~\cite{rockafellar15_convex_analy}), so $\op{ri}(\C{K})$ is dense in $\C{K}$.
By assumption, $\nabla f(\x)$ exists at every $\x\in\op{ri}(\C{K})$.

Suppose that $f$ is $\gamma$--tilted convex.
By Theorem~\ref{thm:mwc-alternative-characterizations}, (i) implies (v), so the inequality of the corollary holds for all $\x\in\op{ri}(\C{K})$ and $\y\in\C{K}$, and in particular for $\y\in\op{ri}(\C{K})$.

Conversely, suppose that the inequality holds for all $\x,\y\in\op{ri}(\C{K})$.
Fix $\x\in\op{ri}(\C{K})$ and $\y\in\C{K}$, and for $t\in(0,1]$ put $\y_t:=(1-t)\y+t\x$.
Then $\y_t\in\op{ri}(\C{K})$ by Theorem~6.1 in~\cite{rockafellar15_convex_analy}, and $\y_t-\x=(1-t)(\y-\x)$.
The assumed inequality at the pair $\x,\y_t$ therefore reads
\begin{equation}\label{eq:m1-c1-mwc-yt}
f(\y_t)-f(\x)
\ge
\psi_\gamma\bigl((1-t)\langle \nabla f(\x),\y-\x\rangle\bigr),
\qquad t\in(0,1].
\end{equation}
As $t\to0^+$, the points $\y_t$ on the segment $[\x,\y]$ converge to $\y$; since $\x,\y\in\C{K}$, radial continuity gives $f(\y_t)\to f(\y)$.
The function $\psi_\gamma$ is continuous by Lemma~\ref{lem:psi_basics}(1), so the right-hand side of \eqref{eq:m1-c1-mwc-yt} converges to $\psi_\gamma(\langle \nabla f(\x),\y-\x\rangle)$.
Passing to the limit in \eqref{eq:m1-c1-mwc-yt} gives
\[
f(\y)-f(\x)\ge\psi_\gamma(\langle \nabla f(\x),\y-\x\rangle),
\qquad \x\in\op{ri}(\C{K}),\ \y\in\C{K}.
\]
Together with the convexity of $\C{K}$ and the density of $\op{ri}(\C{K})$ in $\C{K}$, this is condition~(v) of Theorem~\ref{thm:mwc-alternative-characterizations}, so $f$ is $\gamma$--tilted convex.
\end{proof}

\subsubsection{Lines and limits}
\label{subsec:lines-limits}

A function is convex if and only if its restriction to every line is convex, and the same holds for $\gamma$--tilted convexity.

\begin{theorem}\label{thm:gamma-MW-convex-on-lines}
Let $\gamma \in (0,1]$ and let $f : \mathbb R^d \to \overline{\mathbb{R}}$ be a proper lsc function.
Then $f$ is $\gamma$--tilted convex if and only if, for every line $\ell$ that intersects $\op{dom}(f)$, the restriction
\[
f|_\ell : \ell \to \overline{\mathbb{R}}
\]
is $\gamma$--tilted convex, where $\ell$ is regarded as a one-dimensional Euclidean space.
\end{theorem}

The proof is in Appendix~\ref{apx:gamma-MW-convex-on-lines}.
Together with Theorem~\ref{thm:gamma-MW-continuous-on-lines}, it gives radial continuity in $\mathbb R^d$.

\begin{corollary}\label{cor:mwc-radially-continuous}
Let $f : \mathbb R^d \to \overline{\mathbb{R}}$ be a proper lsc function that is $\gamma$--tilted convex for some $\gamma \in (0,1]$.
Then it is radially continuous.
\end{corollary}

\begin{proof}
Fix $\x,\y\in\op{dom}(f)$.
If $\x=\y$, the segment $[\x,\y]$ is a single point and there is nothing to prove.
Otherwise put $\dd:=(\y-\x)/\|\y-\x\|$ and $g(t):=f(\x+t\dd)$ for $t\in\mathbb R$.
The map $t\mapsto\x+t\dd$ is an isometry from $\mathbb R$ onto the line $\ell$ through $\x$ and $\y$, so $g$ is the restriction $f|_\ell$ written in this coordinate.
The function $g$ is proper, since $g(0)=f(\x)\in\mathbb R$ and $g>-\infty$, and it is lsc as the composition of the lsc function $f$ with a continuous map.
By Theorem~\ref{thm:gamma-MW-convex-on-lines}, $g$ is $\gamma$--tilted convex, and hence, by Theorem~\ref{thm:gamma-MW-continuous-on-lines}, $g$ is continuous on $\op{dom}(g)$.
The set $\op{dom}(f)$ is convex by Theorem~\ref{thm:mwc-domain-is-convex}, so $[\x,\y]\subseteq\op{dom}(f)$ and $[0,\|\y-\x\|]\subseteq\op{dom}(g)$.
Hence $t\mapsto f(\x+t\dd)$ is continuous on $[0,\|\y-\x\|]$, i.e., $f$ is continuous on $[\x,\y]$.
\end{proof}

The values of a proper lsc convex function on the relative interior of its domain determine its values on the closure.
Theorem~\ref{thm:mwc-lsc-envelope} extends this to $\gamma$--tilted convex functions, and its proof in Appendix~\ref{apx:mwc-lsc-envelope} uses continuity on lines for the finite boundary values and lower semicontinuity for the infinite ones.

\begin{theorem}
\label{thm:mwc-lsc-envelope}
Let $\gamma\in(0,1]$, and let
$f,g:\mathbb R^d\to\overline{\mathbb R}$ be proper, lsc, and
$\gamma$--tilted convex.  Set $\mathcal K:=\op{dom}(f)$.  If
\[
f=g\qquad\text{on }\op{ri}(\mathcal K),
\]
then
\[
f=g\qquad\text{on }\overline{\mathcal K}.
\]
Moreover, we have
\[
f=\op{lsc}\bigl(f|_{\op{ri}(\mathcal K)}\bigr).
\]
\end{theorem}

If convex functions converge pointwise on a dense set, then they converge uniformly on compact subsets of the relative interior, and the limit is convex.
Theorem~\ref{thm:gamma-MW-convex-pointwise-limit}, proved in Appendix~\ref{apx:gamma-MW-convex-pointwise-limit}, gives the same for $\gamma$--tilted convex functions.

\begin{theorem}
\label{thm:gamma-MW-convex-pointwise-limit}
Let $\mathcal K\subseteq \mathbb R^d$ be a non-empty convex set and $\gamma \in (0, 1]$.
Let $f_n:\mathbb R^d \to \overline{\mathbb R}$ be proper, lsc, and $\gamma$--tilted convex functions that converge pointwise to finite values on a dense subset $\mathcal D \subseteq \mathcal K$.
Then there is a continuous function $f : \op{ri}(\C{K}) \to \mathbb R$ such that
\begin{enumerate}[label=(\roman*)]
\item $f_n \to f$ uniformly on compact subsets of
$\op{ri}(\mathcal K)$;
\item the function $\op{lsc}(f) : \mathbb R^d \to \overline{\mathbb R}$ is $\gamma$--tilted convex.
\end{enumerate}
Moreover, if there is an lsc function $g : \C{K} \to \overline{\mathbb R}$ such that $f_n \to g$ pointwise, then $\op{lsc}(g) = \op{lsc}(f)$ and so it is $\gamma$--tilted convex.
\end{theorem}

Applied to a constant sequence, Theorem~\ref{thm:gamma-MW-convex-pointwise-limit} shows that restrictions to convex sets stay in the class.

\begin{corollary}[Stability under convex restrictions]
\label{cor:gamma-MW-convex-restrictions}
Let $\gamma\in(0,1]$, let
$f:\mathbb R^d\to\overline{\mathbb R}$ be proper, lsc, and
$\gamma$--tilted convex, and let $\mathcal C\subseteq\mathbb R^d$ be a
convex set such that
$\mathcal C\cap\operatorname{dom}(f)\neq\emptyset$.  Then
\[
\op{lsc}(f|_{\mathcal C}):\mathbb R^d\to\overline{\mathbb R}
\]
is proper, lsc, and $\gamma$--tilted convex.
In particular, if $\mathcal C$ is closed, then the function
\[
f_{\mathcal C}(\x):=
\begin{cases}
f(\x),&\x\in\mathcal C,\\
+\infty,&\x\notin\mathcal C,
\end{cases}
\]
is proper, lsc, and $\gamma$--tilted convex.
\end{corollary}

\begin{proof}
The set $\mathcal K:=\mathcal C\cap\operatorname{dom}(f)$ is non-empty, and it is convex because $\operatorname{dom}(f)$ is convex by Theorem~\ref{thm:mwc-domain-is-convex}. The constant sequence $f_n:=f$ converges pointwise on $\mathcal K$ to the finite-valued function $g:=f|_{\mathcal K}$, which is lsc on $\mathcal K$ because for $\x\in\mathcal K$ the lower semicontinuity of $f$ gives $\liminf_{\y\in\mathcal K,\y\to\x}f(\y)\ge\liminf_{\y\to\x}f(\y)\ge f(\x)$. The final assertion of Theorem~\ref{thm:gamma-MW-convex-pointwise-limit}, applied with the dense set $\mathcal D:=\mathcal K$, therefore shows that $\op{lsc}(g)$ is $\gamma$--tilted convex.

We next show that $\op{lsc}(g)=\op{lsc}(f|_{\mathcal C})$.
The effective domains of $g$ and $f|_{\mathcal C}$ are both $\mathcal K$, so both envelopes equal $+\infty$ outside $\overline{\mathcal K}$.
For $\x\in\overline{\mathcal K}$, the points $\y\in\mathcal C\setminus\mathcal K$ satisfy $f(\y)=+\infty$ and do not lower the $\liminf$, so
\begin{equation}\label{eq:m1-restr-envelope}
\op{lsc}(f|_{\mathcal C})(\x)
=\liminf_{\y\in\mathcal C,\,\y\to\x}f(\y)
=\liminf_{\y\in\mathcal K,\,\y\to\x}f(\y)
=\op{lsc}(g)(\x).
\end{equation}
Hence $\op{lsc}(f|_{\mathcal C})$ is $\gamma$--tilted convex, and it is lsc as an lsc envelope. It never takes the value $-\infty$, because for $\x\in\overline{\mathcal K}$, \eqref{eq:m1-restr-envelope} and the lower semicontinuity of $f$ give $\op{lsc}(f|_{\mathcal C})(\x)\ge\liminf_{\y\to\x}f(\y)\ge f(\x)>-\infty$, while outside $\overline{\mathcal K}$ it equals $+\infty$. Since the envelope lies below $f|_{\mathcal C}$, which is finite on the non-empty set $\mathcal K$, it is proper, and the first assertion follows.

Let $\mathcal C$ be closed, so that $f_{\mathcal C}$ is the proper extension of $f|_{\mathcal C}$. It is lsc at every $\x\in\mathcal C$, since $f_{\mathcal C}\ge f$ gives $\liminf_{\y\to\x}f_{\mathcal C}(\y)\ge\liminf_{\y\to\x}f(\y)\ge f(\x)=f_{\mathcal C}(\x)$, and at every $\x\notin\mathcal C$, since $\mathbb R^d\setminus\mathcal C$ is open and $f_{\mathcal C}=+\infty$ on it. As noted after the definition of the lsc envelope, an lsc proper extension coincides with the lsc envelope, so $f_{\mathcal C}=\op{lsc}(f|_{\mathcal C})$, and the second assertion follows from the first.
\end{proof}

\subsubsection{Closure under supremum}
\label{subsec:suprema}

A pointwise supremum of convex functions is convex, and the same holds in $\mathcal F^\gamma$.

\begin{theorem}[Closure under supremum]
\label{thm:sup-closure}
Let $\gamma\in(0,1]$, let $\{f_i\}_{i\in I}\subseteq\mathcal F^\gamma$ be a non-empty family, and suppose that $s:=\sup_{i\in I}f_i$ is proper. Then $s\in\mathcal F^\gamma$.
\end{theorem}

The proof in Appendix~\ref{apx:sup-closure} treats the maximum of two functions on each line with Lebourg's mean value theorem, passes from finite maxima to countable suprema with Theorem~\ref{thm:gamma-MW-convex-pointwise-limit}, and replaces an arbitrary family by a countable subfamily with the same supremum.

\subsection{Properties of \texorpdfstring{$\gamma$}{gamma}--tilted monotone subdifferentials}
\label{subsec:mono-class}

$\gamma$--tilted monotonicity of the subdifferential implies $\gamma$--tilted convexity, and $\gamma$--tilted convexity implies $\gamma^2$--tilted monotonicity. The proof is in Appendix~\ref{apx:mwc-vs-mw-mono}.

\begin{theorem}\label{thm:mwc-vs-mw-mono}
Assume $f : \mathbb R^d \to \overline{\mathbb{R}}$ is a proper lsc function and $F : \mathbb R^d \rightrightarrows \mathbb R^d$ is a multi-valued operator where $\operatorname{dom}(F) \subseteq \operatorname{dom}(f)$.
Let $\gamma \in (0,1]$.
\begin{enumerate}[label=(\roman*)]
\item If $\hat{\partial} f$ is $\gamma$--tilted monotone, then $f$ is $\gamma$--tilted convex.
\item If $f$ is $\gamma$--tilted convex with respect to $F$, then $F$ is $\gamma^2$--tilted monotone. 
\end{enumerate}
In particular,
\[
\gamma^*_{\op{mono}}(f) \le \gamma^*(f) \le (\gamma^*_{\op{mono}}(f))^{1/2}.
\]
\end{theorem}

The class $\mathcal F^\gamma_{\op{mono}}$ has characterizations analogous to Theorem~\ref{thm:mwc-alternative-characterizations}, proved in Appendix~\ref{apx:mw-mono-alternative-characterizations}.

\begin{theorem}
\label{thm:mw-mono-alternative-characterizations}
Assume $f : \mathbb R^d \to \overline{\mathbb{R}}$ is a proper lsc function with effective domain $\C{K} := \op{dom}(f)$ and $\gamma \in (0,1]$.
Then the following are equivalent.
\begin{itemize}
\item[(i)] $\overline{\partial} f$ is $\gamma$--tilted monotone.
\item[(ii)] $\partial f$ is $\gamma$--tilted monotone.
\item[(iii)] $\hat{\partial} f$ is $\gamma$--tilted monotone.
\item[(iv)] $\mathcal K$ is convex, $f$ is radially continuous, $f|_{\op{ri}(\C{K})}$ is locally Lipschitz, and $\hat{\partial} f|_{\op{ri}(\C{K})}$ is $\gamma$--tilted monotone.
\item[(v)] $\mathcal K$ is convex, $f$ is radially continuous, $f|_{\op{ri}(\C{K})}$ is locally Lipschitz, and $\x \mapsto \{\nabla f(\x)\}: \op{ri}(\C{K}) \rightrightarrows \mathbb R^d$ is $\gamma$--tilted monotone, i.e., we have
\[
\langle \nabla f(\y), \y - \x \rangle \geq \psi_\gamma(\langle \nabla f(\x), \y - \x \rangle),
\]
for all $\x,\y \in \op{ri}(\C{K})$ where $\nabla f(\x)$ and $\nabla f(\y)$ exist.
Here $\nabla f$ is considered with respect to the affine hull of $\mathcal{K}$.
\end{itemize}
\end{theorem}

For functions that are continuously differentiable on the relative interior of their domain, the equivalence of (i) and (v) gives a first-order test.
\begin{corollary}\label{cor:mono-smooth-test}
Let $\gamma\in(0,1]$ and let $f:\mathbb R^d\to\overline{\mathbb R}$ be a proper, lsc, and radially continuous function with a convex effective domain $\C{K} := \op{dom}(f)$ such that $f|_{\op{ri}(\mathcal K)}$ is continuously differentiable.
Then $\overline{\partial}f$ is $\gamma$--tilted monotone if and only if
\[
\langle \nabla f(\y), \y - \x \rangle \geq \psi_\gamma(\langle \nabla f(\x), \y - \x \rangle),
\]
for all $\x,\y \in \op{ri}(\C{K})$.
Here $\nabla f$ is considered with respect to the affine hull of $\mathcal{K}$.
\end{corollary}

\begin{proof}
By assumption, $\C{K}$ is convex and $f$ is radially continuous.
The function $f|_{\op{ri}(\C{K})}$ is locally Lipschitz, because every $\x\in\op{ri}(\C{K})$ has a closed ball $B$ of the affine hull of $\C{K}$ centered at $\x$ with $B\subseteq\op{ri}(\C{K})$.
The continuous map $\nabla f$ is bounded on the compact set $B$, say by $L$, and the mean value theorem on the segment between any two points of the convex set $B$ gives $|f(\uu)-f(\w)|\le L\|\uu-\w\|$ for all $\uu,\w\in B$.
Finally, $\nabla f(\x)$ exists at every $\x\in\op{ri}(\C{K})$, so the inequality in condition~(v) of Theorem~\ref{thm:mw-mono-alternative-characterizations} is required for all $\x,\y\in\op{ri}(\C{K})$ and coincides with the inequality of the corollary.
Hence the inequality of the corollary is equivalent to condition~(v), and by that theorem it is equivalent to condition~(i), i.e., to $\gamma$--tilted monotonicity of $\overline{\partial}f$.
\end{proof}

\begin{remark}
By Theorems~\ref{thm:mwc-vs-mw-mono}, \ref{thm:mwc-alternative-characterizations}, and~\ref{thm:mw-mono-alternative-characterizations}, conditions (i)--(v) of Theorem~\ref{thm:mwc_one} remain equivalent when $\overline{\partial} f$ is replaced by $\partial f$ or $\hat{\partial} f$, which is also classical \cite[Theorem~12.17]{rockafellar1998variational}.
\end{remark}

Membership in $\mathcal F^\gamma_{\op{mono}}$ can also be checked on lines, where a restriction to a line is regarded as a function on a one-dimensional Euclidean space, extended by $+\infty$ outside its effective domain.
The proof in Appendix~\ref{apx:gamma-MW-monotone-on-lines}, like that of Theorem~\ref{thm:gamma-MW-monotone-pointwise-limit} below, uses the approximate supports of Section~\ref{subsec:regularity}, which members of $\mathcal F^\gamma_{\op{mono}}$ have because they lie in $\mathcal F^\gamma$ (Theorem~\ref{thm:mwc-vs-mw-mono}).

\begin{theorem}[Line restrictions of $\gamma$--tilted monotone subdifferentials]
\label{thm:gamma-MW-monotone-on-lines}
Let $\gamma\in(0,1]$ and let
$f:\mathbb R^d\to\overline{\mathbb R}$ be proper and lsc.  Then
$\overline{\partial} f$ is $\gamma$--tilted monotone if and only if, for every line
$\ell$ meeting $\operatorname{dom}(f)$, the operator
\[
  \overline{\partial}(f|_\ell)
\]
is $\gamma$--tilted monotone on $\ell$.
Moreover, when these equivalent conditions hold, the same conclusion is
true for the restriction of $f$ to any nonempty closed interval contained
in $\ell$ and meeting $\operatorname{dom}(f)$.
\end{theorem}

For closed convex sets, the line criterion also gives stability under restriction.

\begin{corollary}[Stability of $\gamma$--tilted monotone subdifferentials under closed convex restrictions]
\label{cor:gamma-MW-monotone-restrictions}
Let $\gamma\in(0,1]$, let $f:\mathbb R^d\to\overline{\mathbb R}$ be proper and lsc with $\overline{\partial}f$ $\gamma$--tilted monotone, and let $\mathcal C\subseteq\mathbb R^d$ be a closed convex set such that $\mathcal C\cap\operatorname{dom}(f)\neq\emptyset$. Then the function $f_{\mathcal C}$ that equals $f$ on $\mathcal C$ and $+\infty$ outside $\mathcal C$ is proper and lsc, and $\overline{\partial}f_{\mathcal C}$ is $\gamma$--tilted monotone.
\end{corollary}

\begin{proof}
The function $f_{\mathcal C}$ is finite on the non-empty set $\mathcal C\cap\operatorname{dom}(f)$ and never equals $-\infty$, so it is proper, and it is lsc by the argument in the last paragraph of the proof of Corollary~\ref{cor:gamma-MW-convex-restrictions}. Let $\ell$ be a line meeting $\operatorname{dom}(f_{\mathcal C})=\mathcal C\cap\operatorname{dom}(f)$. Then $\ell\cap\mathcal C$ is a nonempty closed interval contained in $\ell$ and meeting $\operatorname{dom}(f)$, and $f_{\mathcal C}|_\ell$ is the restriction of $f$ to this interval, so the final assertion of Theorem~\ref{thm:gamma-MW-monotone-on-lines} shows that $\overline{\partial}(f_{\mathcal C}|_\ell)$ is $\gamma$--tilted monotone. Since $\ell$ was arbitrary, Theorem~\ref{thm:gamma-MW-monotone-on-lines}, applied to $f_{\mathcal C}$, shows that $\overline{\partial}f_{\mathcal C}$ is $\gamma$--tilted monotone.
\end{proof}

The class $\mathcal F^\gamma_{\op{mono}}$ has the same stability under limits as $\mathcal F^\gamma$, with proof in Appendix~\ref{apx:gamma-MW-monotone-pointwise-limit}.

\begin{theorem}[Stability of $\gamma$--tilted monotone subdifferentials]
\label{thm:gamma-MW-monotone-pointwise-limit}
Let $\mathcal K\subseteq\mathbb R^d$ be a nonempty convex set and let
$\gamma\in(0,1]$.  Let
$f_n:\mathbb R^d\to\overline{\mathbb R}$ be proper lsc functions such
that $\overline{\partial} f_n$ is $\gamma$--tilted monotone for every $n$.  Suppose that
$f_n$ converges pointwise to finite values on a dense subset
$\mathcal D\subseteq\mathcal K$.  Then there is a continuous function
$f:\operatorname{ri}(\mathcal K)\to\mathbb R$ such that
\begin{enumerate}[label=(\roman*)]
\item $f_n\to f$ uniformly on compact subsets of
$\operatorname{ri}(\mathcal K)$;
\item if $\bar f:=\op{lsc}(f)$, then
$\overline{\partial}\bar f$ is $\gamma$--tilted monotone.
\end{enumerate}
Moreover, if there is an lsc function
$g:\mathcal K\to\overline{\mathbb R}$ such that $f_n\to g$ pointwise
on $\mathcal K$, then $\op{lsc}(g)=\bar f$, and hence 
$\overline{\partial}(\op{lsc}(g))$ is $\gamma$--tilted monotone.
\end{theorem}

The class $\mathcal F^\gamma_{\op{mono}}$ inherits the properties of Sections~\ref{subsec:domain-minimizers}--\ref{subsec:lines-limits} through Theorem~\ref{thm:mwc-vs-mw-mono}(i).

\begin{corollary}
\label{cor:mono-structural-properties}
Let $\gamma\in(0,1]$ and let
$f:\mathbb R^d\to\overline{\mathbb R}$ be proper and lsc.  Suppose that
$\overline{\partial}f$ is $\gamma$--tilted monotone, and set
$\mathcal K:=\operatorname{dom}(f)$.  Then:
\begin{enumerate}[label=(\roman*)]
\item $f$ is $\gamma$--tilted convex, $\mathcal K$ is convex, and $f$ is
quasiconvex;
\item $f|_{\operatorname{ri}(\mathcal K)}$ is locally Lipschitz, and $f$ is radially continuous;
\item for every $\x\in\mathcal K$, the following are equivalent:
$\x$ is a global minimizer of $f$, $\x$ is a local minimizer of $f$,
$\BF{0}\in\hat{\partial} f(\x)$, and
$\BF{0}\in\overline{\partial} f(\x)$;
\item $f=\op{lsc}\bigl(f|_{\operatorname{ri}(\mathcal K)}\bigr)$.
\end{enumerate}
\end{corollary}

\begin{proof}
Since $f$ is lsc on $\mathbb R^d$, it is locally lsc at every point of $\mathcal K$, and Proposition~\ref{prop:subdiff-basics}(i) gives $\hat{\partial} f(\x)\subseteq\overline{\partial}f(\x)$ for every $\x\in\mathcal K$.
Inequality~\eqref{eq:gamma-mw-mono} for $\overline{\partial}f$ therefore holds in particular for all $\x^*\in\hat{\partial} f(\x)$ and $\y^*\in\hat{\partial} f(\y)$, so $\hat{\partial} f$ is $\gamma$--tilted monotone.
Theorem~\ref{thm:mwc-vs-mw-mono}(i) now shows that $f$ is $\gamma$--tilted convex, so Theorem~\ref{thm:mwc-domain-is-convex} gives that $\mathcal K$ is convex and $f$ is quasiconvex, which proves~(i).
Assertion~(ii) follows from
Theorem~\ref{thm:gamma-MW-convex-is-locally-Lipschitz} and Corollary~\ref{cor:mwc-radially-continuous}, assertion~(iii) from
Theorem~\ref{thm:local-min-is-global-min}, and assertion~(iv) from
Theorem~\ref{thm:mwc-lsc-envelope}.
\end{proof}

\subsection{Properties of \texorpdfstring{$\gamma$}{gamma}--tilted secant convex functions}
\label{subsec:jensen-class}

Convexity can be stated through increasing secant slopes or through Jensen's inequality, which bounds the value at a convex combination by the weighted arithmetic mean of the two values. Tilted secant convexity also has a Jensen form, in which the arithmetic mean is replaced by a mean that multiplies by $\gamma$ the weight of the smaller value (Proposition~\ref{prop:secant-jensen}). The right side of this form is nondecreasing in both values, which makes it the convenient form for suprema and lower semicontinuous envelopes (Theorem~\ref{thm:gamma-MW-Jensen-basics}), and it extends to convex combinations of any finite number of points (Theorem~\ref{thm:gamma-MW-Jensen-equivalent-def}).

For $\gamma\in(0,1]$, $\lambda\in[0,1]$, and $a,b\in\mathbb R$, define
the mean
\begin{equation}
\label{eq:gamma-MW-Jensen-mean}
M_{\gamma,\lambda}(a,b)
:=
\begin{cases}
\displaystyle
\frac{\gamma\lambda a+(1-\lambda)b}
     {\gamma\lambda+(1-\lambda)},
& a\leq b,\\[2ex]
\displaystyle
\frac{\lambda a+\gamma(1-\lambda)b}
     {\lambda+\gamma(1-\lambda)},
& a>b.
\end{cases}
\end{equation}
We extend this definition to $\gamma=0$ by setting
\begin{equation}
\label{eq:zero-MW-Jensen-mean}
M_{0,\lambda}(a,b):=\max\{a,b\},
\qquad
\lambda\in[0,1].
\end{equation}
Thus, $M_{1,\lambda}(a,b)=\lambda a+(1-\lambda)b$, whereas
$M_{0,\lambda}(a,b)=\max\{a,b\}$.

For $\gamma>0$, this mean is the upper expectile of a random variable taking $a$ and $b$ with probabilities $\lambda$ and $1-\lambda$, at level $1/(1+\gamma)$~\citep{newey1987asymmetric,bellini2014generalized}. Equivalently, it uniquely minimizes
\[
\lambda\bigl[(a-t)_+^2+\gamma(a-t)_-^2\bigr]
+(1-\lambda)\bigl[(b-t)_+^2+\gamma(b-t)_-^2\bigr]
\]
over $t\in\mathbb R$, where $u_+:=\max\{u,0\}$ and $u_-:=\max\{-u,0\}$.

\begin{proposition}[Jensen form]
\label{prop:secant-jensen}
Let $\gamma\in[0,1]$ and $f:\mathcal K\to\mathbb R \cup \{+\infty\}$. Then $f$ is $\gamma$--tilted secant convex if and only if
\begin{equation}
\label{eq:gamma-MW-Jensen-convex}
f(\lambda\y+(1-\lambda)\z)
\leq
M_{\gamma,\lambda}\bigl(f(\y),f(\z)\bigr)
\qquad\text{for all }\y,\z\in\op{dom}(f)\text{ and }\lambda\in[0,1].
\end{equation}
\end{proposition}

\begin{proof}
We write $a:=f(\y)$ and $b:=f(\z)$ whenever $\y,\z\in\op{dom}(f)$, so that $a,b\in\mathbb R$.

For $\gamma\in(0,1]$, the denominators in~\eqref{eq:gamma-MW-Jensen-mean} are positive, since $\gamma\lambda+(1-\lambda)\ge\gamma\lambda+\gamma(1-\lambda)=\gamma$ and likewise $\lambda+\gamma(1-\lambda)\ge\gamma$. So each branch of~\eqref{eq:gamma-MW-Jensen-mean} is a weighted average of $a$ and $b$ with nonnegative weights that sum to one. Hence
\begin{equation}\label{eq:secant-mean-facts}
M_{\gamma,\lambda}(a,a)=a,\qquad M_{\gamma,1}(a,b)=a,\qquad M_{\gamma,0}(a,b)=b.
\end{equation}
For $\lambda=1$, the first branch is $\gamma a/\gamma=a$ and the second is $a/1=a$; for $\lambda=0$ the first branch is $b/1=b$ and the second is $\gamma b/\gamma=b$. For $\gamma=0$, \eqref{eq:zero-MW-Jensen-mean} gives $M_{0,\lambda}(a,b)=\max\{a,b\}$, which equals $a$ when $a=b$ and is at least $a$ and at least $b$. In all cases, $M_{\gamma,\lambda}(a,a)=a$, $M_{\gamma,1}(a,b)\ge a$, and $M_{\gamma,0}(a,b)\ge b$.

Let $\y,\z\in\op{dom}(f)$ and $\lambda\in[0,1]$. If $\y=\z$, then $\lambda\y+(1-\lambda)\z=\y$, and~\eqref{eq:gamma-MW-Jensen-convex} reads $a\le M_{\gamma,\lambda}(a,a)=a$. If $\lambda=1$, it reads $a\le M_{\gamma,1}(a,b)$, and if $\lambda=0$, it reads $b\le M_{\gamma,0}(a,b)$. By the facts above, all three hold, and Definition~\ref{def:gamma-MW-Jensen-convex} imposes nothing on these degenerate triples. So it suffices to fix $\y\ne\z$ in $\op{dom}(f)$ and $\lambda\in(0,1)$, put $\x:=\lambda\y+(1-\lambda)\z$, and show that
\begin{equation}\label{eq:secant-jensen-goal}
\Bigl[\x\in\op{dom}(f)\ \text{and \eqref{eq:secant-form} holds whenever $\psi_\gamma(s_-)$ is defined}\Bigr]
\iff
f(\x)\le M_{\gamma,\lambda}(a,b).
\end{equation}

The right side $M_{\gamma,\lambda}(a,b)$ of~\eqref{eq:secant-jensen-goal} is a real number, since $a$ and $b$ are real. If $\x\notin\op{dom}(f)$, then $f(\x)=+\infty$, so the right side of~\eqref{eq:secant-jensen-goal} fails; the left side fails as well, so both sides of~\eqref{eq:secant-jensen-goal} are false and the equivalence holds. From now on let $\x\in\op{dom}(f)$, and put $m:=f(\x)\in\mathbb R$. It remains to show that~\eqref{eq:secant-form}, required whenever $\psi_\gamma(s_-)$ is defined, is equivalent to $m\le M_{\gamma,\lambda}(a,b)$.

Since $\y-\x=(1-\lambda)(\y-\z)$ and $\x-\z=\lambda(\y-\z)$, we have $\norm{\y-\x}=(1-\lambda)\norm{\y-\z}$ and $\norm{\x-\z}=\lambda\norm{\y-\z}$. Put $c:=\lambda(1-\lambda)\norm{\y-\z}>0$. Multiplying the definitions~\eqref{eq:secant-slopes} of $s_+$ and $s_-$ by $c$ gives
\begin{equation}\label{eq:secant-values}
c\,s_+=\lambda(a-m),
\qquad
c\,s_-=(1-\lambda)(m-b).
\end{equation}
In particular, $s_+$ has the sign of $a-m$, and $s_-$ has the sign of $m-b$.

Suppose first that $\gamma>0$ and $a>b$, so that the second branch of~\eqref{eq:gamma-MW-Jensen-mean} applies. Multiplying $m\le M_{\gamma,\lambda}(a,b)$ by the positive denominator $\lambda+\gamma(1-\lambda)$ and rearranging, we get
\[
m\le M_{\gamma,\lambda}(a,b)
\iff
\lambda(a-m)\ge\gamma(1-\lambda)(m-b)
\iff
s_+\ge\gamma s_-,
\]
where the last step uses~\eqref{eq:secant-values} and divides by $c>0$. If $s_-\ge0$, then $\psi_\gamma(s_-)=\gamma s_-$, so $s_+\ge\gamma s_-$ is exactly~\eqref{eq:secant-form}. If $s_-<0$, then $m<b<a$, so $s_+>0$ by~\eqref{eq:secant-values}. Then $s_+>0>\gamma s_-$, so $m\le M_{\gamma,\lambda}(a,b)$ holds, and $s_+>0>s_-/\gamma=\psi_\gamma(s_-)$, so~\eqref{eq:secant-form} holds. In both cases the two conditions agree.

Suppose next that $\gamma>0$ and $a\le b$, so that the first branch of~\eqref{eq:gamma-MW-Jensen-mean} applies. Multiplying by the positive denominator $\gamma\lambda+(1-\lambda)$ and rearranging, we get
\[
m\le M_{\gamma,\lambda}(a,b)
\iff
\gamma\lambda(a-m)\ge(1-\lambda)(m-b)
\iff
\gamma s_+\ge s_-
\iff
s_+\ge s_-/\gamma.
\]
If $s_-\le0$, then $\psi_\gamma(s_-)=s_-/\gamma$, so $s_+\ge s_-/\gamma$ is exactly~\eqref{eq:secant-form}. If $s_->0$, then $m>b\ge a$, so $s_+<0$ by~\eqref{eq:secant-values}. Then $s_+<0<s_-/\gamma$, so $m\le M_{\gamma,\lambda}(a,b)$ fails, and $s_+<0<\gamma s_-=\psi_\gamma(s_-)$, so~\eqref{eq:secant-form} fails. In both cases the two conditions agree.

Finally, let $\gamma=0$. Then $M_{0,\lambda}(a,b)=\max\{a,b\}$, and $\psi_0(s_-)$ equals $0$ if $s_->0$, equals $-\infty$ if $s_-<0$, and is undefined if $s_-=0$. We use the signs of $s_\pm$ from~\eqref{eq:secant-values}.
If $s_->0$, then $m>b$, so $m\le\max\{a,b\}$ holds exactly when $m\le a$, that is, when $s_+\ge0=\psi_0(s_-)$; the two conditions agree.
If $s_-<0$, then $m<b\le\max\{a,b\}$, and $s_+\ge-\infty=\psi_0(s_-)$; both conditions hold.
If $s_-=0$, then $m=b\le\max\{a,b\}$, and~\eqref{eq:secant-form} is not required because $\psi_0(0)$ is undefined; both conditions hold.

These three cases cover all possibilities, which proves~\eqref{eq:secant-jensen-goal} and hence the proposition.
\end{proof}

The next theorem collects basic properties of $\gamma$--tilted secant convex functions, including closure under suprema in item~(iii) and a reduction to lines in item~(vi).

\begin{theorem}
\label{thm:gamma-MW-Jensen-basics}
Let $\gamma\in[0,1]$, $\mathcal{X} \subseteq \mathbb R^d$ be a convex set, and 
$f:\mathcal X\to\mathbb R \cup \{+\infty\}$ with effective domain $\mathcal K := \op{dom}(f)$. 
Then the following are true.
\begin{itemize}
\item[(i)] If $f$ is $\gamma$--tilted secant convex, then it is quasiconvex, and $\mathcal K$ is convex.
\item[(ii)] If $f$ is convex, then it is $\gamma$--tilted secant convex.
\item[(iii)] The set of $\gamma$--tilted secant convex functions over $\mathcal X$ is closed under arbitrary non-empty suprema. 
\item[(iv)] If $f$ is $\gamma$--tilted secant convex and $\mathcal C \subseteq \mathcal X$ is convex, the function
\[
f|_{\mathcal C} : \mathcal C \to \mathbb R \cup \{+\infty\}
\]
is $\gamma$--tilted secant convex.
\item[(v)] If $\gamma\in(0,1]$ and $f$ is $\gamma$--tilted secant convex, then 
\[
\op{lsc}(f) : \mathbb R^d \to \mathbb R \cup \{+\infty\}
\]
is $\gamma$--tilted secant convex and $f = \op{lsc}(f)$ on $\op{ri}(\mathcal K)$.
\item[(vi)] $f$ is $\gamma$--tilted secant convex if and only if, for every line
$\ell$ that intersects $\mathcal X$, the restriction
\[
f|_\ell : \ell \cap \mathcal{X} \to \mathbb R \cup \{+\infty\}
\]
is $\gamma$--tilted secant convex.
\end{itemize}
\end{theorem}

The proof is in Appendix~\ref{apx:gamma-MW-Jensen-basics}.
A supremum of lsc functions is lsc, so by item~(iii) a proper supremum of members of $\mathcal F^\gamma_{\op{sec}}$ lies in $\mathcal F^\gamma_{\op{sec}}$.

The binary mean in~\eqref{eq:gamma-MW-Jensen-mean} extends to $m$ values.
For an integer $m\geq2$, let
\[
\Delta_m
:=
\left\{
\boldsymbol\lambda=(\lambda_1,\ldots,\lambda_m)\in[0,1]^m
\,\middle|\,
\sum_{i=1}^m\lambda_i=1
\right\}
\]
denote the probability simplex.

\begin{definition}[Mean of $m$ values; upper expectile]
\label{def:finite-dimensional-gamma-MW-Jensen-mean}
Fix $m\geq2$, $\boldsymbol\lambda\in\Delta_m$, and
$\gamma\in(0,1]$.  For
$\a=(a_1,\ldots,a_m)\in\mathbb R^m$, define
\begin{equation}
\label{eq:gamma-MW-Jensen-subset-max}
M^{(m)}_{\gamma,\boldsymbol\lambda}(\a)
:=
\max_{S\subseteq\{1,\ldots,m\}}
\frac{
\sum_{i\in S}\lambda_i a_i
+\gamma\sum_{i\notin S}\lambda_i a_i
}{
\sum_{i\in S}\lambda_i
+\gamma\sum_{i\notin S}\lambda_i
}.
\end{equation}
For $\gamma=0$, define
\begin{equation}
\label{eq:finite-dimensional-zero-MW-Jensen-mean}
M^{(m)}_{0,\boldsymbol\lambda}(\a)
:=
\max_{1\leq i\leq m}a_i.
\end{equation}
\end{definition}

For $\gamma>0$, this is the expectile at level
$1/(1+\gamma)$ of the finite distribution assigning mass
$\lambda_i$ to $a_i$~\citep{newey1987asymmetric,bellini2014generalized}.
Equivalently, it is the unique minimizer over $t\in\mathbb R$ of
\[
\sum_{i=1}^m\lambda_i
\bigl[(a_i-t)_+^2+\gamma(t-a_i)_+^2\bigr],
\qquad r_+:=\max\{r,0\};
\]
see Lemma~\ref{lem:finite-dimensional-gamma-MW-Jensen-mean-characterization}.
Although~\eqref{eq:gamma-MW-Jensen-subset-max} maximizes over subsets,
the threshold characterization in that lemma computes the mean in
$O(m\log m)$ arithmetic and comparison operations by sorting the
values and scanning the resulting intervals.  At $\gamma=0$,
\eqref{eq:finite-dimensional-zero-MW-Jensen-mean} is a separate
convention: it takes the maximum over all coordinates, whereas the
positive-parameter limit ignores coordinates of weight zero.

For $m=2$ and
$\boldsymbol\lambda=(\lambda,1-\lambda)$, this definition recovers the
binary mean (by Lemma~\ref{lem:finite-dimensional-gamma-MW-Jensen-mean-characterization} for $\gamma>0$, and by the definitions for $\gamma=0$):
\begin{equation}
\label{eq:binary-as-finite-dimensional-gamma-MW-Jensen-mean}
M^{(2)}_{\gamma,(\lambda,1-\lambda)}(a,b)
=M_{\gamma,\lambda}(a,b),
\qquad
\gamma\in[0,1].
\end{equation}
Moreover, $M^{(m)}_{1,\boldsymbol\lambda}$ is the weighted
arithmetic mean, and $M^{(m)}_{0,\boldsymbol\lambda}$ is the
maximum.

\begin{theorem}[Characterizations of $\gamma$--tilted secant convexity]
\label{thm:gamma-MW-Jensen-equivalent-def}
Let $f:\mathbb R^d\to\overline{\mathbb R}$ be a proper lsc function
with effective domain $\mathcal K:=\operatorname{dom}(f)$, and
let $\gamma\in(0,1]$.  Then the following are equivalent.
\begin{itemize}
\item[(i)] The function $f$ is $\gamma$--tilted secant convex.

\item[(ii)] For every $\y,\z\in\mathcal K$ and
$\lambda\in(0,1)$, one has
\begin{equation}
\label{eq:gamma-MW-Jensen-equivalent-def}
\lambda\,
\overline{\psi}_{\sqrt{\gamma}}\bigl(f(\y)-f(\x)\bigr)
+(1-\lambda)\,
\overline{\psi}_{\sqrt{\gamma}}\bigl(f(\z)-f(\x)\bigr)
\geq0,
\qquad
\x=\lambda\y+(1-\lambda)\z.
\end{equation}

\item[(iii)] For every $m\geq2$, every
$\y_1,\ldots,\y_m\in\mathcal K$, and every
$\boldsymbol\lambda\in\Delta_m$, one has
\begin{equation}
\label{eq:gamma-MW-Jensen-finite-point}
f\left(\sum_{i=1}^m\lambda_i\y_i\right)
\leq
M^{(m)}_{\gamma,\boldsymbol\lambda}
\bigl(f(\y_1),\ldots,f(\y_m)\bigr).
\end{equation}

\item[(iv)] For every $m\geq2$, every
$\y_1,\ldots,\y_m\in\mathcal K$, and every
$\lambda_1,\ldots,\lambda_m > 0$ where
$\sum_{i=1}^m\lambda_i=1$, one has
\begin{equation}
\label{eq:gamma-MW-Jensen-finite-point-psi}
\sum_{i=1}^m
\lambda_i\,
\overline{\psi}_{\sqrt{\gamma}}
\left(f(\y_i)-f(\x)\right)
\geq0,
\qquad
\x=\sum_{i=1}^m\lambda_i\y_i.
\end{equation}

\item[(v)] $\mathcal K$ is convex and there is a multi-valued operator $F : \mathbb R^d \rightrightarrows \mathbb R^d$ such that $\op{ri}(\C{K}) \subseteq \op{dom}(F)$ and $f$ is $\sqrt{\gamma}$--tilted convex with respect to $F$.
\end{itemize}
Furthermore, the equivalence of (i), (ii), (iii), and (iv) holds even if we do not assume $f$ is lsc.
\end{theorem}

The proof is in Appendix~\ref{apx:gamma-MW-Jensen-equivalent-def}.

The next theorem, proved in Appendix~\ref{apx:gamma-MW-Jensen-vs-gamma-MW-convex}, places $\gamma$--tilted secant convexity between the other two notions: it implies $\gamma$--tilted convexity and follows from $\gamma$--tilted monotonicity of the subdifferential.

\begin{theorem}
\label{thm:gamma-MW-Jensen-vs-gamma-MW-convex}
Let $f:\mathbb R^d\to\overline{\mathbb R}$ be a proper lsc function
and let $\gamma\in(0,1]$.
\begin{enumerate}[label=(\roman*)]
\item If $f$ is $\gamma$--tilted secant convex, then $f$ is
$\gamma$--tilted convex.
\item If $\overline{\partial}f$ is $\gamma$--tilted monotone, then $f$
is $\gamma$--tilted secant convex.
\end{enumerate}
In particular,
\[
\gamma^*_{\op{mono}}(f)
\leq
\gamma^*_{\op{sec}}(f)
\leq
\gamma^*(f).
\]
\end{theorem}

The class $\mathcal F^\gamma_{\op{sec}}$ inherits the properties of Sections~\ref{subsec:domain-minimizers}--\ref{subsec:lines-limits} through Theorem~\ref{thm:gamma-MW-Jensen-vs-gamma-MW-convex}(i).

\begin{corollary}
\label{cor:gamma-MW-Jensen-structural-properties}
Let $\gamma\in(0,1]$ and let
$f:\mathbb R^d\to\overline{\mathbb R}$ be a proper lsc
$\gamma$--tilted secant convex function with effective domain
$\mathcal K:=\operatorname{dom}(f)$.  Then the following assertions
hold:
\begin{enumerate}[label=(\roman*)]
\item $f|_{\operatorname{ri}(\mathcal K)}$ is locally Lipschitz, and
$f$ is radially continuous;
\item for every $\x\in\mathcal K$, the following are equivalent:
$\x$ is a global minimizer of $f$, $\x$ is a local minimizer of $f$,
$\BF{0}\in\hat{\partial}f(\x)$, and
$\BF{0}\in\overline{\partial}f(\x)$;
\item
$f=\op{lsc}\bigl(f|_{\operatorname{ri}(\mathcal K)}\bigr)$.
\end{enumerate}
Moreover, $f|_{\operatorname{ri}(\mathcal K)}$ is locally Lipschitz even if we do not assume that $f$ is lsc.
\end{corollary}

\begin{proof}
By Theorem~\ref{thm:gamma-MW-Jensen-vs-gamma-MW-convex}\textup{(i)},
$f$ is $\gamma$--tilted convex.  Assertion~\textup{(i)} follows from
Theorem~\ref{thm:gamma-MW-convex-is-locally-Lipschitz} and
Corollary~\ref{cor:mwc-radially-continuous}; assertion~\textup{(ii)}
follows from Theorem~\ref{thm:local-min-is-global-min}; and
assertion~\textup{(iii)} follows from
Theorem~\ref{thm:mwc-lsc-envelope}.
For the final claim, assume only that $f$ is proper and $\gamma$--tilted secant convex, so that $\mathcal K$ is convex by Theorem~\ref{thm:gamma-MW-Jensen-basics}(i) and $\operatorname{ri}(\mathcal K)\neq\emptyset$. By Theorem~\ref{thm:gamma-MW-Jensen-basics}(v), the envelope $g:=\op{lsc}(f)$, a function from $\mathbb R^d$ to $\mathbb R\cup\{+\infty\}$, is $\gamma$--tilted secant convex and
\begin{equation}\label{eq:m1-jsp-agree}
g=f\qquad\text{on }\operatorname{ri}(\mathcal K).
\end{equation}
It is lsc, and it is proper because it never takes the value $-\infty$ and is finite on $\operatorname{ri}(\mathcal K)$ by \eqref{eq:m1-jsp-agree}. Its effective domain $\mathcal K':=\operatorname{dom}(g)$ satisfies $\mathcal K\subseteq\mathcal K'\subseteq\overline{\mathcal K}$, because $g\le f$ on $\mathcal K$ and $g=+\infty$ outside $\overline{\mathcal K}$, so $\overline{\mathcal K'}=\overline{\mathcal K}$. The set $\mathcal K'$ is convex by Theorem~\ref{thm:gamma-MW-Jensen-basics}(i) applied to $g$, and two convex sets with the same closure have the same relative interior (Corollary~6.3.1 in~\cite{rockafellar15_convex_analy}), so $\operatorname{ri}(\mathcal K')=\operatorname{ri}(\mathcal K)$. Assertion~(i), applied to the proper lsc $\gamma$--tilted secant convex function $g$, shows that $g|_{\operatorname{ri}(\mathcal K)}$ is locally Lipschitz, and by \eqref{eq:m1-jsp-agree} so is $f|_{\operatorname{ri}(\mathcal K)}$.
\end{proof}

The class $\mathcal F^\gamma_{\op{sec}}$ is stable under limits as well, with proof in Appendix~\ref{apx:gamma-MW-Jensen-pointwise-limit}.

\begin{theorem}[Stability of $\gamma$--tilted secant convexity under limits]
\label{thm:gamma-MW-Jensen-pointwise-limit}
Let $\mathcal K\subseteq\mathbb R^d$ be a non-empty convex set and let
$\gamma\in(0,1]$.  Let
$f_n:\mathbb R^d\to\overline{\mathbb R}$ be proper
$\gamma$--tilted secant convex functions that converge pointwise to finite
values on a dense subset $\mathcal D\subseteq\mathcal K$.
Then there is a continuous function
$f:\operatorname{ri}(\mathcal K)\to\mathbb R$ such that
\begin{enumerate}[label=(\roman*)]
\item $f_n\to f$ uniformly on compact subsets of
$\operatorname{ri}(\mathcal K)$;
\item $f$ is $\gamma$--tilted secant convex.
\end{enumerate}
Moreover, if there is a function
$g:\mathcal K\to\overline{\mathbb R}$ such that $f_n\to g$ pointwise
on $\mathcal K$, then $g$ never takes the value $-\infty$ and is $\gamma$--tilted secant convex.
\end{theorem}

\subsection{Envelopes of tilted affine functions}
\label{subsec:H-equal}

The tilted affine functions of $\mathcal H_\gamma$ play the role for $\gamma^2$--tilted secant convexity that affine functions play for convexity.

\begin{theorem}
\label{thm:H-convex-equivalent-def}
Let $f:\mathbb R^d \to \overline{\mathbb R}$ be a proper lsc function and let $\gamma\in(0,1]$.
Then $f$ is $\mathcal H_\gamma$--convex if and only if it is
$\gamma^2$--tilted secant convex.
\end{theorem}

The proof is in Appendix~\ref{apx:H-convex-equivalent-def}. Each member of $\mathcal H_\gamma$ is $\gamma$--tilted convex with respect to a constant operator, hence $\gamma^2$--tilted secant convex by Theorem~\ref{thm:gamma-MW-Jensen-equivalent-def}(v), and Theorem~\ref{thm:gamma-MW-Jensen-basics}(iii) passes to suprema. Conversely, at each $\x$ in the relative interior of the domain, the operator of Theorem~\ref{thm:gamma-MW-Jensen-equivalent-def}(v) gives a member of $\mathcal H_\gamma$ that minorizes $f$ and equals $f(\x)$ at $\x$, and $f$ is determined by its values on the relative interior (Corollary~\ref{cor:gamma-MW-Jensen-structural-properties}(iii)).

\subsection{Separations and negative results}
\label{subsec:separations}

Three functions separate the classes from one another, and three properties of convex functions fail in them: closure under addition, classical second-order regularity, and, for $\mathcal F^\gamma_{\op{mono}}$, closure under maxima.

\subsubsection{Separating the three classes}
\label{subsec:separate-classes}

For $\gamma\in(0,1)$, three functions separate the classes $\mathcal F^{\sqrt\gamma}$, $\mathcal F^\gamma_{\op{mono}}$, $\mathcal F^\gamma_{\op{sec}}$, and $\mathcal F^\gamma$. Example~\ref{ex:sep-1} attains $\gamma^*_{\op{mono}}=\gamma^*$, Example~\ref{ex:sep-2} attains $\gamma^*=(\gamma^*_{\op{mono}})^{1/2}$, and Example~\ref{ex:sep-3} gives $\gamma^*_{\op{sec}}<\gamma^*$.
Their proofs are in Appendices~\ref{apx:example-1}, \ref{apx:example-2}, and~\ref{apx:example-3}.

\begin{example}
\label{ex:sep-1}
Fix $\gamma\in(0,1)$ and define $f:\mathbb R\to\overline{\mathbb R}$ by
\[
f(x):=
\begin{cases}
x, & -1\le x\le0,\\
\gamma x, & 0\le x\le1,\\
+\infty, & x\notin[-1,1].
\end{cases}
\]
Then $f$ is proper and lsc, and
\[
\gamma^*_{\op{mono}}(f)
=\gamma^*_{\op{sec}}(f)
=\gamma^*(f)
=\gamma.
\]
Therefore
\[
f \notin \mathcal F^{\sqrt{\gamma}},
\qquad
f \in \mathcal F_{\op{mono}}^\gamma
\subseteq \mathcal F_{\op{sec}}^\gamma
\subseteq \mathcal F^\gamma.
\]
\end{example}

\begin{example}
\label{ex:sep-2}
Fix $\gamma \in (0,1)$ and define $c := \frac{1-\gamma}{\gamma}$.
Define $f :\mathbb R\to\overline{\mathbb R}$ by
\[
f(x) := \dfrac{x}{1+cx}
\]
for $x\in[0,1]$ and $f(x) = +\infty$ elsewhere.
Then $f$ is proper and lsc, its restriction to $[0,1]$ is continuously
differentiable, and
\[
\gamma^*_{\op{sec}}(f)
= \gamma^*(f)
= \bigl(\gamma^*_{\op{mono}}(f)\bigr)^{1/2}
= \gamma.
\]
Therefore
\[
f \notin \mathcal F_{\op{mono}}^\gamma,
\qquad
f \in \mathcal F_{\op{sec}}^{\gamma}
\subseteq \mathcal F^\gamma
\subseteq \mathcal F_{\op{mono}}^{\gamma^2}.
\]
\end{example}

\begin{example}
\label{ex:sep-3}
Fix $\gamma\in(0,1)$ and set $s:=\gamma$ and $a:=\frac{s}{2-s}$.
Define the continuous function $q:[0,1]\to(0,\infty)$ by
\[
q(x):=
\begin{cases}
1-\dfrac{2(1-s)}{s(2-s)}x, & 0\le x\le s,\\[2mm]
a, & s\le x\le1,
\end{cases}
\]
and define $f:\mathbb R\to\overline{\mathbb R}$ by
\[
f(x):= \int_0^x q(t)\,dt
\]
for $x\in[0,1]$ and $f(x) = +\infty$ elsewhere.
Then $f$ is proper and lsc, its restriction to $[0,1]$ is continuously
differentiable, and
\begin{align*}
\gamma^*(f)
&=\gamma,
\quad
\gamma^*_{\op{mono}}(f)
=\frac{\gamma}{2-\gamma},
\quad
\gamma^*_{\op{sec}}(f)
=\gamma\left[
1-
\left(
\frac{1-\gamma}
{1+\sqrt{(1-\gamma)(2-\gamma)}}
\right)^2
\right]
<\gamma.
\end{align*}
In particular,
\[
f\notin \mathcal F_{\op{sec}}^{\gamma},
\qquad
f\in\mathcal F^{\gamma} \subseteq \mathcal F_{\op{sec}}^{\gamma^2}.
\]
\end{example}

\subsubsection{Closure under sums}
\label{subsec:sums}

For every $0<\gamma<1$, none of the three classes is closed under addition. Both results below use the concave quadratic $h_c(t)=-t^2-ct$, whose parameters in the three classes are explicit.

\begin{example}[A concave quadratic]\label{ex:quadratic}
For $c>0$, the function $h_c(t)=-t^2-ct$ on $[0,1]$ satisfies
\[
\gamma^*(h_c)=\gamma^*_{\op{sec}}(h_c)=\frac c{c+1},\qquad\gamma^*_{\op{mono}}(h_c)=\frac c{c+2}.
\]
Equivalently, for $\gamma\in(0,1)$, $h_c$ is $\gamma$--tilted convex, and also $\gamma$--tilted secant convex, if and only if $c\ge\gamma/(1-\gamma)$, and its Clarke subdifferential is $\gamma$--tilted monotone if and only if $c\ge2\gamma/(1-\gamma)$.
\end{example}

The proof is in Appendix~\ref{apx:quadratic}. The parameter of $h_c$ in $\mathcal F^\gamma_{\op{mono}}$ is smaller than in the other two classes, so $h_c$ also separates $\mathcal F^\gamma_{\op{mono}}$ from $\mathcal F^\gamma_{\op{sec}}$.

\begin{example}[Non-closure under addition]
\label{ex:sum-not-closed}
For any $\gamma\in(0,1)$, there exist two functions $f_1,f_2:[0,1]\to\R$ that lie in $\mathcal F^\gamma_{\op{mono}}$, $\mathcal F^\gamma_{\op{sec}}$, and $\mathcal F^\gamma$, whose sum is not quasiconvex and therefore lies in none of the three classes for any positive parameter.
\end{example}
\begin{proof}
Let $c=2\gamma/(1-\gamma)$, and let $f_1=h_c$ and $f_2(t)=h_c(1-t)$ on $[0,1]$, so that $f_1\in\mathcal F^\gamma_{\op{mono}}$ by Example~\ref{ex:quadratic}. Both defining formulas are polynomials on $\R$, so their restrictions to $[0,1]$ have proper, lsc, and radially continuous extensions. For $s,t\in(0,1)$, Corollary~\ref{cor:mono-smooth-test} applied to $f_1$ gives
\begin{equation}\label{eq:m6-sum-hc}
h_c'(t)(t-s)\ge\psi_\gamma\bigl(h_c'(s)(t-s)\bigr).
\end{equation}
Let $x,y\in(0,1)$, and put $s=1-x$ and $t=1-y$. Since $f_2'(r)=-h_c'(1-r)$ and $y-x=-(t-s)$,
\[
f_2'(y)(y-x)=h_c'(t)(t-s),\qquad f_2'(x)(y-x)=h_c'(s)(t-s).
\]
By \eqref{eq:m6-sum-hc}, $f_2'(y)(y-x)\ge\psi_\gamma\bigl(f_2'(x)(y-x)\bigr)$, and Corollary~\ref{cor:mono-smooth-test} gives $f_2\in\mathcal F^\gamma_{\op{mono}}$. Theorem~\ref{thm:gamma-MW-Jensen-vs-gamma-MW-convex} then gives $f_1,f_2\in\mathcal F^\gamma_{\op{sec}}\subseteq\mathcal F^\gamma$.

The linear terms $-ct$ and $-c(1-t)$ add up to $-c$, so
\begin{equation}\label{eq:m6-sum-value}
f_1(t)+f_2(t)=-t^2-(1-t)^2-c\qquad(t\in[0,1]).
\end{equation}
By \eqref{eq:m6-sum-value}, $f_1+f_2$ equals $-1-c$ at $t=0$ and at $t=1$, and it equals $-\frac12-c>-1-c$ at $t=\frac12$. The sublevel set $\{t:f_1(t)+f_2(t)\le-1-c\}$ thus contains $0$ and $1$ but not their midpoint, so $f_1+f_2$ is not quasiconvex. For $\eta\in(0,1]$, every member of $\mathcal F^\eta$ is quasiconvex by Theorem~\ref{thm:mwc-domain-is-convex}, and $\mathcal F^\eta_{\op{mono}}\subseteq\mathcal F^\eta_{\op{sec}}\subseteq\mathcal F^\eta$ by Theorem~\ref{thm:gamma-MW-Jensen-vs-gamma-MW-convex}. Hence $f_1+f_2$ lies in none of the three classes with parameter $\eta$.
\end{proof}

Every smooth function on the cube whose curvature is bounded below is a convex function plus $n$ concave quadratics that lie in all three classes. In Theorem~\ref{thm:decomposition}, $\one$ denotes the all-ones vector.

\begin{theorem}[Decomposition with controlled negative curvature]\label{thm:decomposition}
Fix $0<\gamma<1$ and $c=2\gamma/(1-\gamma)$. Suppose that $f$ is twice continuously differentiable on a neighborhood of $[0,1]^n$ and that $\nabla^2f(\x)\succeq-\tau I$ on the cube for some $\tau\ge0$. Then
\begin{equation}\label{eq:decomposition}
f=q+\sum_{i=1}^np_i,\qquad
p_i(\x)=-\frac\tau2\bigl(x_i^2+cx_i\bigr),\qquad
q(\x)=f(\x)+\frac\tau2\norm{\x}^2+\frac{\tau c}2\,\one^\top\x,
\end{equation}
where $q$ is convex and every $p_i$ on $[0,1]^n$ lies in $\mathcal F^\gamma_{\op{mono}}$, $\mathcal F^\gamma_{\op{sec}}$, and $\mathcal F^\gamma$. If $\norm{\nabla f}\le L$ and $|f|\le B$ on the cube, then
\[
\norm{\nabla p_i}\le\tfrac\tau2(c+2),\quad
|p_i|\le\tfrac\tau2(c+1),\quad
\norm{\nabla q}\le L+\tfrac\tau2(c+2)\sqrt n,\quad
|q|\le B+\tfrac{\tau n}2(c+1).
\]
\end{theorem}

The proof is in Appendix~\ref{apx:decomposition}.
Given an evaluation oracle for $f$ and a bound $\nabla^2f\succeq-\tau I$, the theorem reduces minimizing $f$ on the cube to minimizing a sum of $n+1$ members of any of the three classes, whose terms are evaluated with one evaluation of $f$ and $O(n)$ arithmetic operations. Minimizing such sums is therefore as hard as minimizing a general smooth function on the cube with curvature bounded below.

\subsubsection{Second-order conditions and their limitations}
\label{subsec:second-order}

For a $C^2$ function on an open convex set, the classical second-order
characterization of convexity is
\[
f \text{ is convex}
\quad\Longleftrightarrow\quad
\nabla^2 f(\x)\succeq0
\quad\text{for every }\x.
\]
For a $C^1$ convex function on an open interval, the first derivative
is monotone and therefore differentiable almost everywhere.  More
generally, Alexandrov's theorem gives a second-order expansion almost
everywhere for a finite convex function on an open convex set
\citep[Theorem~13.51]{rockafellar1998variational}.  An almost-everywhere
condition on the classical Hessian alone does not characterize
convexity: for example, $-|x|$ has classical second derivative zero
almost everywhere but is not convex.

The next example shows that the one-dimensional $C^1$ assertion fails
for every $\gamma<1$, even in the smallest of the three classes;
``second derivative'' there means the classical derivative of the
first derivative.

\begin{example}[Failure of classical second-order regularity]
\label{ex:second-order}
For every $\gamma\in[0,1)$ there is a proper lsc function
$f_\gamma:\mathbb R\to\overline{\mathbb R}$ with effective domain
$[0,1]$ such that
\[
f_\gamma\in\mathcal F_{\op{mono}}^\gamma
\cap\mathcal F_{\op{sec}}^\gamma\cap\mathcal F^\gamma,
\]
and $f_\gamma|_{(0,1)}$ is continuously differentiable but its
classical second derivative exists at no point of $(0,1)$.
In fact, all three memberships hold with
$\kappa_\gamma=(1+3\gamma)/(3+\gamma)>\gamma$.
\end{example}

\begin{proof}
Let $w:[0,1]\to[-1,1]$ be continuous and nowhere differentiable.
Set
\[
\delta_\gamma:=\frac{1-\gamma}{2(1+\gamma)},\qquad
h_\gamma:=1+\delta_\gamma w,
\]
and define
\[
f_\gamma(x):=
\begin{cases}
\displaystyle\int_0^x h_\gamma(t)\,dt,&x\in[0,1],\\[2mm]
+\infty,&x\notin[0,1].
\end{cases}
\]
Extend $h_\gamma$ continuously to $\mathbb R$ by keeping it constant
on each of $(-\infty,0]$ and $[1,\infty)$.  Its antiderivative is
$C^1$ on $\mathbb R$, and on $[0,1]$ its derivative satisfies
\[
0<m:=1-\delta_\gamma\le h_\gamma\le M:=1+\delta_\gamma.
\]
Let $\kappa=m/M$. For $x,y\in(0,1)$ put $d=y-x$, $u=f_\gamma'(x)d$, and $v=f_\gamma'(y)d$. If $d\ge0$, then $v\ge md=\kappa Md\ge\kappa u=\psi_\kappa(u)$. If $d<0$, then $u\le md<0$ and $v\ge Md=md/\kappa\ge u/\kappa=\psi_\kappa(u)$. The proper extension of $f_\gamma$ is proper, lsc, and radially continuous, because $f_\gamma$ is continuous on $[0,1]$, so Corollary~\ref{cor:mono-smooth-test} gives $f_\gamma\in\mathcal F_{\op{mono}}^{m/M}$.
Since
\[
\frac mM=\frac{1+3\gamma}{3+\gamma},\qquad
\frac mM-\gamma=\frac{1-\gamma^2}{3+\gamma}>0,
\]
Theorem~\ref{thm:gamma-MW-Jensen-vs-gamma-MW-convex} at the positive
parameter $m/M$, followed by nesting in the parameter, gives the
asserted memberships, including those at $\gamma=0$.
The function is proper and lsc because it is continuous on its closed
effective domain.  On $(0,1)$ its first derivative is
$1+\delta_\gamma w$, which is nowhere differentiable because $\delta_\gamma>0$;
this proves the final assertion.
\end{proof}

Example~\ref{ex:second-order} concerns classical second derivatives of $C^1$ functions. At $\gamma=0$, second-order characterizations of quasiconvexity are classical~\citep{cambini2009generalized}.

\subsubsection{Maxima in \texorpdfstring{$\mathcal F^\gamma_{\mathrm{mono}}$}{F-mono}}
\label{subsec:mono-max}

For every $0<\gamma<1$, the class $\mathcal F^\gamma_{\op{mono}}$ is not closed under maxima, whereas $\mathcal F^\gamma$ and $\mathcal F^\gamma_{\op{sec}}$ are closed under proper suprema.

\begin{example}[A maximum outside $\mathcal F^\gamma_{\op{mono}}$]
\label{ex:mono-max}
Let $\gamma\in(0,1)$ and $\sigma\in(\gamma,1)$, put $\delta:=\sigma-\gamma$, and let $\mathcal K:=[L,U]$ with
\[
L:=-\frac{\delta}{2(1-\sigma)},\qquad U:=2+\frac{\delta}{2\gamma(1-\sigma)}.
\]
The functions
\[
f_1(t):=\min\bigl\{\sigma t-\delta,\ \gamma \sigma(t-2)+2\sigma-\delta\bigr\},\qquad
f_2(t):=\min\{t,\ \gamma t\}
\]
on $\mathcal K$ belong to $\mathcal F^\gamma_{\op{mono}}$, while
$h:=\max\{f_1,f_2\}$ on $\mathcal K$ satisfies
\[
\gamma^*_{\op{mono}}(h)=\gamma\sigma<\gamma,\qquad
\gamma^*_{\op{sec}}(h)=\gamma^*(h)=\gamma.
\]
In particular, $h\notin\mathcal F^\gamma_{\op{mono}}$.
\end{example}

Each of $f_1$ and $f_2$ is concave with one kink, at which its slope drops by the factor $\gamma$. Their maximum has slope $1$ near the left endpoint and slope $\gamma \sigma$ near the right endpoint, and this drop by the factor $\gamma \sigma<\gamma$ violates \eqref{eq:gamma-mw-mono}. The proof is in Appendix~\ref{apx:mono-max}.

Theorem~\ref{thm:mwc-vs-mw-mono} bounds how far the monotonicity parameter can drop under suprema in $\mathcal F^\gamma_{\op{mono}}$. Let $\{f_i\}_{i\in I}\subseteq\mathcal F^\gamma_{\op{mono}}$ have a proper supremum $s$. Each $\hat\partial f_i$ is $\gamma$--tilted monotone, since $\hat\partial f_i\subseteq\overline\partial f_i$ by Proposition~\ref{prop:subdiff-basics}(i), so Theorem~\ref{thm:mwc-vs-mw-mono}(i) gives $f_i\in\mathcal F^\gamma$, Theorem~\ref{thm:sup-closure} gives $s\in\mathcal F^\gamma$, and Theorem~\ref{thm:mwc-vs-mw-mono}(ii) with $F=\overline\partial s$ gives $s\in\mathcal F^{\gamma^2}_{\op{mono}}$. The exponent $2$ cannot be lowered: as $\sigma\downarrow\gamma$, the maximum $h$ in Example~\ref{ex:mono-max} has $\gamma^*_{\op{mono}}(h)=\gamma\sigma$, which tends to $\gamma^2$.

\subsection{Relations between the classes}
\label{subsec:chain}

Theorems~\ref{thm:mwc-vs-mw-mono} and~\ref{thm:gamma-MW-Jensen-vs-gamma-MW-convex} combine into one chain, in which the parameter drops by at most a square.

\begin{corollary}
\label{cor:relations}
For $\gamma\in(0,1]$,
\[
\mathcal F^{\sqrt{\gamma}}
\subseteq
\mathcal F_{\op{mono}}^\gamma
\subseteq
\mathcal F_{\op{sec}}^\gamma
\subseteq
\mathcal F^\gamma.
\]
Equivalently, if $f: \mathbb{R}^d \to \overline{\mathbb{R}}$ is a proper lsc function, then
\[
(\gamma^*(f))^2
\le
\gamma^*_{\op{mono}}(f)
\le
\gamma^*_{\op{sec}}(f)
\le
\gamma^*(f).
\]
\end{corollary}

\begin{proof}
If $f\in\mathcal F^{\sqrt\gamma}$, then applying
Theorem~\ref{thm:mwc-vs-mw-mono}(ii) to
$F=\overline\partial f$ gives
$f\in\mathcal F_{\op{mono}}^\gamma$.
Theorem~\ref{thm:gamma-MW-Jensen-vs-gamma-MW-convex}(ii) gives the
next inclusion, and part~(i) of that theorem gives the last one.
Thus
\begin{equation}
\label{eq:m1-rel-classes}
\mathcal F^{\sqrt\gamma}
\subseteq\mathcal F_{\op{mono}}^\gamma
\subseteq\mathcal F_{\op{sec}}^\gamma
\subseteq\mathcal F^\gamma,
\qquad \gamma\in(0,1].
\end{equation}
For a fixed proper lsc function $f$, the same-parameter inclusions
imply $\gamma^*_{\op{mono}}(f)\le\gamma^*_{\op{sec}}(f)\le\gamma^*(f)$
by taking suprema over admissible positive parameters.
The first inclusion, applied with $\gamma=\theta^2$, gives
$f\in\mathcal F_{\op{mono}}^{\theta^2}$ whenever
$f\in\mathcal F^\theta$ and $\theta>0$.
Taking suprema and using continuity of the increasing map
$\theta\mapsto\theta^2$ yields
$(\gamma^*(f))^2\le\gamma^*_{\op{mono}}(f)$.
If an admissible positive-parameter set is empty, its parameter is
zero by definition and the corresponding inequality is immediate.
\end{proof}

By Theorem~\ref{thm:H-convex-equivalent-def}, the class $\mathcal F^\gamma_{\op{sec}}$ in the chain is also the class of proper suprema of members of $\mathcal H_{\sqrt\gamma}$.

When $\gamma=1$, all classes in the chain equal the class of proper lsc convex functions by Theorem~\ref{thm:mwc_one}. For $\gamma\in(0,1)$, Examples~\ref{ex:sep-1}, \ref{ex:sep-2}, and~\ref{ex:sep-3} show that each inclusion in the chain is strict.

Each of the three families is also strictly ordered in its parameter, and all of them lie strictly between the convex and the pseudoconvex functions.

\begin{theorem}[Strict hierarchy]\label{thm:hierarchy}
Let $0<\gamma'<\gamma<1$, and let $\mathcal C^\theta$ stand for $\mathcal F^\theta_{\op{mono}}$ for every $\theta$, for $\mathcal F^\theta_{\op{sec}}$ for every $\theta$, or for $\mathcal F^\theta$ for every $\theta$. Then $\mathcal C^1$ is the class of proper lsc convex functions, and
\begin{equation}\label{eq:hierarchy}
\mathcal C^1\subsetneq\mathcal C^\gamma\subsetneq\mathcal C^{\gamma'}
\subsetneq\bigcup_{0<\theta\le1}\mathcal C^\theta=\bigcup_{0<\theta\le1}\mathcal F^\theta
\subsetneq\mathcal F^{0^+}\subsetneq\mathcal F^0 .
\end{equation}
\end{theorem}

\begin{proof}
By Theorem~\ref{thm:mwc_one}, $\mathcal F^1$ and $\mathcal F^1_{\op{mono}}$ consist of the proper lsc convex functions, and so does $\mathcal F^1_{\op{sec}}$, because $1$--tilted secant convexity is convexity. Each of the three defining conditions with parameter $\theta$ implies the same condition with every smaller parameter (Section~\ref{subsec:def-jensen}), so $\mathcal C^\theta\subseteq\mathcal C^{\theta'}$ for $\theta'<\theta$. Corollary~\ref{cor:relations} gives $\mathcal F^{\sqrt\theta}\subseteq\mathcal C^\theta\subseteq\mathcal F^\theta$ for $\theta\in(0,1]$, so the two unions in \eqref{eq:hierarchy} agree, and \eqref{eq:gradient-class-chain} gives $\bigcup_\theta\mathcal F^\theta\subseteq\mathcal F^{0^+}\subseteq\mathcal F^0$.

For $\theta\in(0,1)$, let $f_\theta$ be the function of Example~\ref{ex:sep-1} with parameter $\theta$. Its three parameters equal $\theta$ and are attained (Section~\ref{subsec:def-jensen}), so $f_\theta\in\mathcal C^\theta$ and $f_\theta\notin\mathcal C^\eta$ for $\eta>\theta$. Hence $f_\gamma\in\mathcal C^\gamma\setminus\mathcal C^1$, $f_{\gamma'}\in\mathcal C^{\gamma'}\setminus\mathcal C^\gamma$, and $f_{\gamma'/2}\in\mathcal C^{\gamma'/2}\setminus\mathcal C^{\gamma'}$, which proves the first three strict inclusions.

Let $s(t)=\sqrt t$ on $[0,1]$. Its proper extension is proper and lsc, and $s$ is continuous on $[0,1]$, hence radially continuous. For $x\in(0,1)$, $s$ is continuously differentiable near $x$, so $\overline\partial s(x)=\{1/(2\sqrt x)\}$ by Proposition~\ref{prop:subdiff-basics}(iv). Since $\sqrt t$ is continuously differentiable near $1$, Proposition~\ref{prop:subdiff-basics}(v) with $\mathcal K=[0,1]$ gives $\overline\partial s(1)=1/2+N_{[0,1]}(1)=[1/2,\infty)$. Every $v\in\overline\partial s(x)$ with $x\in(0,1]$ is therefore positive, so $v(y-x)\ge0$ forces $y\ge x$ and $s(y)\ge s(x)$, while $s(y)\ge0=s(0)$ for every $y\in[0,1]$. This is condition~(i) of Theorem~\ref{thm:mwc_zero_pseudo}, so $s\in\mathcal F^{0^+}$. For $\theta\in(0,1]$ and $x\in(0,1)$, inequality \eqref{eq:gamma-mwc} with $v=1/(2\sqrt x)$ and $y=1$ reads
\begin{equation}\label{eq:sqrt-witness}
1-\sqrt x\ \ge\ \frac{\theta(1-x)}{2\sqrt x},
\end{equation}
which fails for small $x$, because its right side tends to $+\infty$ as $x\downarrow0$. Hence $s\notin\mathcal F^\theta$ for every $\theta\in(0,1]$.

Let $g(t)=(t-1/2)^3$ on $[0,1]$. Its proper extension is proper and lsc, and $g$ is continuous and nondecreasing, so it is radially continuous and quasiconvex, and Theorem~\ref{thm:mwc_zero} gives $g\in\mathcal F^0$. Proposition~\ref{prop:subdiff-basics}(iv) gives $\overline\partial g(1/2)=\{0\}$, and with $x=1/2$, $v=0$, and $y=1/4$ we have $v(y-x)=0$ but $g(y)=-1/64<0=g(x)$. So condition~(i) of Theorem~\ref{thm:mwc_zero_pseudo} fails, and $g\notin\mathcal F^{0^+}$.
\end{proof}

\section{Examples}
\label{sec:examples}

Throughout this section, every feasible set is nonempty. We group the examples by the result that establishes their membership: Lemma~\ref{lem:slope-ratio} and Example~\ref{ex:radial} in Section~\ref{subsec:examples-mono}, function values and envelopes in Section~\ref{subsec:examples-jensen}, and increasing transformations of convex costs in Section~\ref{subsec:examples-mw}. Several examples of the last two subsections also have $\gamma$--tilted monotone generalized gradients and therefore lie in all three classes by Corollary~\ref{cor:relations}, whereas membership through function values can hold at a larger parameter than monotonicity, as the function $t/(1+ct)$ of Example~\ref{ex:sep-2} shows. Table~\ref{tab:examples} lists the parameters of the examples by subsection, together with those of the separating functions of Section~\ref{subsec:separations}, and the proofs are in Appendix~\ref{apx:examples}. Algorithmic uses of these examples rely on the access model of Section~\ref{sec:offline-opt}.

\subsection{Examples from monotone generalized gradients}
\label{subsec:examples-mono}

A function of one linear coordinate lies in $\mathcal F^\theta_{\op{mono}}$ when its slope changes by a bounded factor, with $\theta$ the ratio of the smallest to the largest slope. The proof is in Appendix~\ref{apx:slope-ratio}.

\begin{lemma}[Slope ratios]\label{lem:slope-ratio}
Let $\mathcal K\subseteq\R^d$ be nonempty, closed, and convex, let $\a\in\R^d$, and let $J=\{\ip\a\x:\x\in\mathcal K\}$. Let $H$ be continuously differentiable on an open interval containing $J$, and suppose that $H'$ has constant sign on $J$ with $0<A\le|H'(s)|\le B$ for $s\in J$. Then $f(\x)=H(\ip\a\x)$ on $\mathcal K$ belongs to $\mathcal F^{A/B}_{\op{mono}}$.
\end{lemma}

A concave polynomial in one coordinate with a large enough linear term is an example.

\begin{example}\label{ex:poly-coordinate}
Let $k\ge2$ be an integer, $a>0$, and $1\le i\le n$. The function $f(\x)=-x_i^k-ax_i$ on $[0,1]^n$ has exact monotonicity parameter $\gamma^*_{\op{mono}}(f)=a/(a+k)$. In particular, for $\gamma\in(0,1)$ and $a\ge k\gamma/(1-\gamma)$, it lies in all three classes with parameter $\gamma$.
\end{example}

Radial functions give nonconvex examples with a single minimizer surrounded by regions of negative curvature. Composing with an affine map covers functions of a residual.

\begin{example}[Radial functions]\label{ex:radial}
Let $\varphi:[0,\infty)\to\R$ be continuously differentiable with $\varphi'(0)=0$, let $R\in(0,\infty]$, suppose that $\varphi'(r)>0$ for every real $r\in(0,R]$, and let
\[
\delta=\inf\Bigl\{\frac{\varphi'(s)}{\varphi'(r)}:0<r\le s\le R\Bigr\}.
\]
Let $\mathcal K\subseteq\R^d$ be closed and convex, let $A\in\R^{m\times d}$ and $\BF b\in\R^m$, and suppose that $\norm{A\x-\BF b}\le R$ on $\mathcal K$. If $\delta>0$, then $f(\x)=\varphi(\norm{A\x-\BF b})$ on $\mathcal K$ belongs to $\mathcal F^\delta_{\op{mono}}$. In particular, if $R=\infty$, then $\varphi(\norm\x)$ lies in all three classes on $\R^d$ with parameter $\delta$.
\end{example}

The number $\delta$ measures how far the radial slope can drop when moving outward, and it equals $1$ when $\varphi'$ is nondecreasing, that is, when $\varphi(\norm\x)$ is convex.

\begin{example}[A quadratic bowl with a radial oscillation]\label{ex:bowl-cosine}
Let $b,\omega>0$, let $\kappa=b\omega^2$, and let $f(\x)=\frac12\norm\x^2-b\cos(\omega\norm\x)$ on $\R^d$. If $\kappa\le1$, then $f$ is convex. Let $\kappa_0>1$ be the unique solution of $2\pi-\arccos(-1/\kappa)=\sqrt{\kappa^2-1}$. If $1<\kappa<\kappa_0$, then $f$ is not convex, because along a line through the origin $f''(t)=1+\kappa\cos(\omega t)$ is negative at $t=\pi/\omega$, and $f$ belongs to $\mathcal F^{\gamma_\kappa}_{\op{mono}}$, and hence to all three classes, with
\[
\gamma_\kappa=\frac{2\pi-s_\kappa-\sqrt{\kappa^2-1}}{s_\kappa+\sqrt{\kappa^2-1}},\qquad s_\kappa=\arccos(-1/\kappa).
\]
The monotonicity parameter is exact: $\gamma^*_{\op{mono}}(f)=\gamma_\kappa$. For $\kappa=2$, $s_\kappa=2\pi/3$ and $\gamma_2=(4\pi-3\sqrt3)/(2\pi+3\sqrt3)$. This covers $f(t)=\frac12t^2-2\cos t$ of Figure~\ref{fig:example}, for which $d=1$, $b=2$, and $\omega=1$.
\end{example}

\subsection{Examples from function values and envelopes}
\label{subsec:examples-jensen}

By Theorem~\ref{thm:H-convex-equivalent-def}, a proper lsc function is $\gamma$--tilted secant convex exactly when it is a pointwise supremum of members of $\mathcal H_{\sqrt\gamma}$, the tilted affine functions $\x\mapsto c+\psi_{\sqrt\gamma}(\ip{\x-\w}{\z})$. Membership in $\mathcal F^\gamma_{\op{sec}}$ can therefore be shown either through an inequality on function values, as in the first examples below, or by writing the function as such a supremum.

Ratios of a cost to a positive output, such as cost per unit of throughput, are classical nonconvex objectives \citep{charnes1962programming,dinkelbach1967nonlinear,schaible1976fractional}. Their parameter is the ratio of the smallest to the largest value of the denominator.

\begin{example}[Fractional objectives]\label{ex:fractional}
Let $\mathcal K$ be compact and convex, and let $W$ and $Z$ be continuously differentiable on an open set containing $\mathcal K$, with $Z>0$ on $\mathcal K$ and $\gamma=\min_{\mathcal K}Z/\max_{\mathcal K}Z$. Suppose that $W$ is convex and $Z$ is affine, or that $W\ge0$ is convex and $Z$ is concave. Then $f=W/Z$ on $\mathcal K$ is $\gamma$--tilted secant convex, hence $\gamma$--tilted convex, with a $\gamma^2$--tilted monotone Clarke subdifferential, and for all $\x,\y\in\mathcal K$,
\begin{equation}\label{eq:fractional-identity}
f(\y)-f(\x)\ \ge\ \frac{Z(\x)}{Z(\y)}\,\ip{\nabla f(\x)}{\y-\x},
\end{equation}
with equality when $W$ and $Z$ are both affine.
\end{example}

The bound can be exact: $t\mapsto t/(1+ct)$ on $[0,1]$ with $c=(1-\gamma)/\gamma$ has $\gamma^*=\gamma^*_{\op{sec}}=\gamma$ and $\gamma^*_{\op{mono}}=\gamma^2$, and here $\min_{[0,1]}Z/\max_{[0,1]}Z=\gamma$ for $Z(t)=1+ct$ (Example~\ref{ex:sep-2} in Section~\ref{subsec:separate-classes}).

\begin{example}[Risk per unit of return and energy per bit]\label{ex:ratios}
(a) Let $\Sigma\succ0$ be a covariance matrix, and let $\mu_i>0$ denote expected excess returns relative to a deterministic risk-free benchmark. On the probability simplex
\[
\Delta_n=\{\x\in\mathbb R^n_+:\textstyle\sum_i x_i=1\},
\]
the inverse Sharpe ratio $f(\x)=\sqrt{\x^\top\Sigma\x}/\ip{\mu}{\x}$ is $\gamma$--tilted secant convex with $\gamma=\min_i\mu_i/\max_i\mu_i$.

(b) Let $\mathcal K\subseteq\mathbb R^n_+$ be compact and convex, let $\BF c\ge\BF0$ and $P_0>0$, and let $R$ be concave and continuously differentiable on an open convex neighborhood of $\mathcal K$, with $R>0$ on $\mathcal K$. The energy per bit $f(\x)=(\ip{\BF c}{\x}+P_0)/R(\x)$ is $\gamma$--tilted secant convex with $\gamma=\min_{\mathcal K}R/\max_{\mathcal K}R$ \citep{zappone2015energy}. Examples include sums of $\log(1+\operatorname{SNR}_j(\x))$ with affine $\operatorname{SNR}_j$ on a neighborhood where the logarithms are defined, provided the feasible set ensures a positive total rate.

For sequences of either type, the hindsight objective is a sum of ratios and need not retain the structure of a single fractional objective. General sum-of-ratios optimization is NP-hard \citep{schaible2003fractional}.
\end{example}

The remaining examples are suprema of members of $\mathcal H_\gamma$, much as convex functions arise as suprema of affine functions.

\begin{example}\label{ex:sup-H}
Let $\varnothing\ne\{h_i\}_{i\in I}\subseteq\mathcal H_\gamma$, let $f=\sup_{i\in I}h_i$, and suppose that $f$ is finite at some point. Then $f$ is $\mathcal H_\gamma$--convex, hence $\gamma^2$--tilted secant convex and $\gamma^2$--tilted convex.
\end{example}

\begin{example}[Distorted convex functions and worst-margin costs]\label{ex:distorted-convex}
Let $\phi:\R^d\to\R$ be convex. For each $\x_0$ and $\g_0\in\partial\phi(\x_0)$, superadditivity and monotonicity of $\psi_\gamma$ (Lemma~\ref{lem:psi_basics}) give $\psi_\gamma(\phi(\x))\ge\psi_\gamma(\phi(\x_0))+\psi_\gamma(\ip{\g_0}{\x-\x_0})$ for all $\x$, with equality at $\x=\x_0$. Hence $\psi_\gamma\circ\phi$ is a supremum of members of $\mathcal H_\gamma$ and is $\gamma^2$--tilted secant convex. Its Clarke subdifferential is also $\gamma^2$--tilted monotone, so it lies in all three classes with parameter $\gamma^2$. For a nonempty closed convex set $C$ and $r_0>0$, the choice $\phi=\operatorname{dist}(\cdot,C)-r_0$ penalizes distance at rate $1/\gamma$ within distance $r_0$ of $C$ and at rate $\gamma$ beyond it.

For labeled data $(\mathbf a_i,b_i)$ with $b_i\in\{-1,1\}$, let $\ell(\x)=\max_i\{-b_i\ip{\mathbf a_i}{\x}\}$. This function is convex, and monotonicity of $\psi_\gamma$ gives
\[
\psi_\gamma(\ell(\x))=\max_i\psi_\gamma(-b_i\ip{\mathbf a_i}{\x}).
\]
Thus the displayed distortion has the class memberships proved above. For minimization, fix a bounded feasible set such as $\mathcal K=\{\x:\norm\x\le1\}$, on which minimizing $\ell$ is equivalent to maximizing the minimum signed margin. The distortion has slope $1/\gamma$ at negative arguments and slope $\gamma$ at positive arguments, and since it is strictly increasing, it preserves the set of minimizers of $\ell$ on every fixed $\mathcal K$. Restricting to $\mathcal K$ keeps the three memberships: for $\mathcal F_{\op{sec}}$ by the definition, for $\mathcal F$ by Corollary~\ref{cor:gamma-MW-convex-restrictions}, and for $\mathcal F_{\op{mono}}$ by Corollary~\ref{cor:gamma-MW-monotone-restrictions}.
\end{example}

\begin{example}[Two-slope costs]\label{ex:two-slope}
Let $0<b\le a$ and $f(t)=at$ for $t\le0$ and $f(t)=bt$ for $t\ge0$. With $\theta=\sqrt{b/a}$ and $\nu=\sqrt{ab}$, $f(t)=\psi_\theta(\nu t)$, so $f\in\mathcal H_\theta$, and Example~\ref{ex:distorted-convex} with $\phi(t)=\nu t$ shows that $\overline\partial f$ is $(b/a)$--tilted monotone. On $\R$, $\gamma^*(f)=b/a$ (Appendix~\ref{apx:examples}), so $\gamma^*_{\op{mono}}(f)=\gamma^*_{\op{sec}}(f)=\gamma^*(f)=b/a$. Such a function models a marginal cost that drops from $a$ to $b$ beyond a threshold, as in a quantity discount. It is convex only when $a=b$, and because its derivative drops at $0$, adding a quadratic does not make it convex either, in contrast with the additive relaxation of weak convexity \citep{davis2019stochastic}.
\end{example}

\subsection{Increasing transformations of convex costs}
\label{subsec:examples-mw}

Robust estimation applies an increasing transformation to a convex cost, which keeps its minimizers and bounds the influence of large residuals. Example~\ref{ex:transform} bounds $\gamma^*$ of such a composition through secants of the transformation relative to its derivative.

\begin{example}[Increasing transformations of convex costs]\label{ex:transform}
Let $\mathcal K$ be closed and convex, let $g$ be convex and continuously differentiable on an open set containing $\mathcal K$ with $g(\mathcal K)\subseteq J$, and let $\phi$ be continuously differentiable with $\phi'>0$ on an open interval containing $J$. Suppose that $0<a\le1\le b$ and, for all $u,v\in J$,
\begin{equation}\label{eq:relative-secants}
\phi(v)-\phi(u)\ \ge\
\begin{cases}
a\,\phi'(u)(v-u),&v\ge u,\\
b\,\phi'(u)(v-u),&v<u.
\end{cases}
\end{equation}
Then $\phi\circ g$ is $\theta$--tilted convex on $\mathcal K$ with $\theta=\min\{a,1/b\}$.
\end{example}

Examples~\ref{ex:radial} and~\ref{ex:transform} apply to two standard robust regression losses, which are radial functions of a residual.

\begin{example}[Cauchy and Geman--McClure losses]\label{ex:cauchy}
Let $r(\x)=\ip{\a}{\x}-\beta$ be an affine residual with $|r|\le R$ on a closed convex set $\mathcal K$, let $\sigma>0$, and let $q=R^2/\sigma^2$.
\begin{enumerate}[label=(\alph*)]
\item The Cauchy loss $\log(1+r(\x)^2/\sigma^2)$ is convex if $q\le1$ and belongs to $\mathcal F^\theta_{\op{mono}}$ with $\theta=2\sqrt q/(1+q)$ if $q\ge1$.
\item The Geman--McClure loss $r(\x)^2/(\sigma^2+r(\x)^2)$ is convex if $q\le1/3$ and belongs to $\mathcal F^\theta_{\op{mono}}$ with $\theta=16\sqrt q/\bigl(3\sqrt3\,(1+q)^2\bigr)$ if $q\ge1/3$. It is also $1/(1+q)$--tilted convex (Example~\ref{ex:transform}). For $q\ge1/3$, this second bound is at least $\theta$ exactly when $q\ge(101+16\sqrt{37})/27$.
\end{enumerate}
As functions of the scalar residual $r$, the Cauchy and Geman--McClure profiles have negative second derivative for $|r|>\sigma$ and $|r|>\sigma/\sqrt3$, respectively. Their affine compositions therefore have negative curvature along any feasible line segment on which the residual varies and remains beyond the corresponding threshold. Both are standard bounded-influence losses in robust regression and vision \citep{black1996unification,barron2019general}, and summing them over data points gives the robust M-estimation objective.
\end{example}

\begin{table}[t]
\centering
\renewcommand{\arraystretch}{1.15}
\caption{Parameters of the examples and of the separating functions of Section~\ref{subsec:separations}. Each entry is a lower bound on the parameter of its column, and the entries marked exact are equalities. By Corollary~\ref{cor:relations}, a lower bound $\theta$ on $\gamma^*_{\op{mono}}$ is also a lower bound on $\gamma^*_{\op{sec}}$ and $\gamma^*$, and a lower bound $\theta$ on $\gamma^*$ gives the lower bound $\theta^2$ on the other two. Here $\theta_{\mathrm C}=2\sqrt q/(1+q)$ and $\theta_{\mathrm{GM}}=16\sqrt q/(3\sqrt3(1+q)^2)$.}
\label{tab:examples}
\begin{tabularx}{\textwidth}{@{}>{\RaggedRight\arraybackslash}p{0.36\textwidth}>{\RaggedRight\arraybackslash}X>{\RaggedRight\arraybackslash}X>{\RaggedRight\arraybackslash}X>{\RaggedRight\arraybackslash}p{0.13\textwidth}@{}}
\toprule
Function & $\gamma^*_{\op{mono}}$ & $\gamma^*_{\op{sec}}$ & $\gamma^*$ & Reference\\
\midrule
\multicolumn{5}{@{}l}{\emph{Section~\ref{subsec:examples-mono}}}\\
$-x_i^k-ax_i$ on $[0,1]^n$ & $a/(a+k)$, exact & $a/(a+k)$ & $a/(a+k)$ & Ex.~\ref{ex:poly-coordinate}\\
$\varphi(\norm{A\x-\BF b})$ & $\delta$ & $\delta$ & $\delta$ & Ex.~\ref{ex:radial}\\
$\frac12\norm\x^2-b\cos(\omega\norm\x)$, $1<\kappa<\kappa_0$ & $\gamma_\kappa$, exact & $\gamma_\kappa$ & $\gamma_\kappa$ & Ex.~\ref{ex:bowl-cosine}\\
\midrule
\multicolumn{5}{@{}l}{\emph{Section~\ref{subsec:examples-jensen}}}\\
$W/Z$, convex over affine, $\gamma=\min Z/\max Z$ & $\gamma^2$ & $\gamma$ & $\gamma$ & Ex.~\ref{ex:fractional}\\
Inverse Sharpe ratio; energy per bit & $\gamma^2$ & $\gamma$ & $\gamma$ & Ex.~\ref{ex:ratios}\\
Supremum of members of $\mathcal H_\gamma$ & $\gamma^4$ & $\gamma^2$ & $\gamma^2$ & Ex.~\ref{ex:sup-H}\\
$\psi_\gamma\circ\phi$, $\phi$ convex; worst-margin costs & $\gamma^2$ & $\gamma^2$ & $\gamma^2$ & Ex.~\ref{ex:distorted-convex}\\
Two-slope cost, slopes $a\ge b$ & $b/a$, exact & $b/a$, exact & $b/a$, exact & Ex.~\ref{ex:two-slope}\\
\midrule
\multicolumn{5}{@{}l}{\emph{Section~\ref{subsec:examples-mw}}}\\
$\phi\circ g$ as in Ex.~\ref{ex:transform} & $\theta^2$ & $\theta^2$ & $\theta=\min\{a,1/b\}$ & Ex.~\ref{ex:transform}\\
Cauchy loss, $q\ge1$ & $\theta_{\mathrm C}$ & $\theta_{\mathrm C}$ & $\theta_{\mathrm C}$ & Ex.~\ref{ex:cauchy}\\
Geman--McClure loss, $q\ge1/3$ & $\theta_{\mathrm{GM}}$ & $\theta_{\mathrm{GM}}$ & larger of $\theta_{\mathrm{GM}}$ and $1/(1+q)$ & Ex.~\ref{ex:cauchy}\\
\midrule
\multicolumn{5}{@{}l}{\emph{Separating functions, Section~\ref{subsec:separations}}}\\
$x$ on $[-1,0]$, $\gamma x$ on $[0,1]$ & $\gamma$, exact & $\gamma$, exact & $\gamma$, exact & Ex.~\ref{ex:sep-1}\\
$-t^2-ct$ on $[0,1]$ & $c/(c+2)$, exact & $c/(c+1)$, exact & $c/(c+1)$, exact & Ex.~\ref{ex:quadratic}\\
$t/(1+ct)$ on $[0,1]$, $c=(1-\gamma)/\gamma$ & $\gamma^2$, exact & $\gamma$, exact & $\gamma$, exact & Ex.~\ref{ex:sep-2}\\
Function of Example~\ref{ex:sep-3} & $\gamma/(2-\gamma)$, exact & $\gamma(1-r^2)<\gamma$, exact, where $r=\tfrac{1-\gamma}{1+\sqrt{(1-\gamma)(2-\gamma)}}$ & $\gamma$, exact & Ex.~\ref{ex:sep-3}\\
Maximum $h$ of Example~\ref{ex:mono-max}, $\sigma\in(\gamma,1)$ & $\gamma\sigma$, exact & $\gamma$, exact & $\gamma$, exact & Ex.~\ref{ex:mono-max}\\
\bottomrule
\end{tabularx}
\end{table}

\section{Polynomial-Time Offline Optimization}
\label{sec:offline-opt}

An $L$--smooth function from any of the three classes can be minimized to additive accuracy $\varepsilon$ with a number of gradient and projection calls polynomial in $L$, $D$, $1/\gamma$, and $1/\varepsilon$. We work with the largest class $\mathcal F^\gamma$, and Corollary~\ref{cor:relations} transfers the results to the two smaller classes. Smooth members are tilted convex in the sense of Mart\'inez-Rubio~\cite{martinezrubio2022global}, hence quasar-convex~\cite{martinezrubio2025smooth}, and the bounds of this section are those of known first-order methods for quasar-convex functions. The access model is shared with the online problem in Sections~\ref{sec:linearization} and~\ref{sec:hardness}.

Let $\mathcal K\subseteq\R^d$ be a nonempty compact convex set of diameter at most a known bound $D$, with a known feasible rational starting point, and we assume efficient Euclidean projection onto $\mathcal K$. A \emph{value-and-support oracle} returns, at a query point $\z\in\mathcal K$,
\begin{equation}\label{eq:oracle-model}
\mathcal Q_f(\z)=\bigl(f(\z),q_f(\z)\bigr).
\end{equation}
The vector $q_f(\z)$ is a \emph{support vector}: inequality~\eqref{eq:gamma-mwc} holds at $\x=\z$ with $\v=q_f(\z)$ for every $\y\in\mathcal K$, as it does for every element of $\overline\partial f(\z)$. We assume that $f$ is $G$--Lipschitz on $\mathcal K$ and that $\norm{q_f(\z)}\le G$ at every feasible point. If $f$ is differentiable on an open set containing $\mathcal K$, then $\nabla f(\z)$ is a Fr\'echet subgradient of the proper extension of $f|_{\mathcal K}$, hence a Clarke subgradient by Proposition~\ref{prop:subdiff-basics}(i), and we use $q_f=\nabla f$, so that \eqref{eq:oracle-model} is a first-order oracle. We call $f$ \emph{$L$--smooth on $\mathcal K$} when $\nabla f$ is $L$--Lipschitz on $\mathcal K$. A \emph{linear minimization oracle} returns, for a query vector $\BF c$, a point of $\argmin_{\y\in\mathcal K}\ip{\BF c}{\y}$; it is used only by the Frank--Wolfe method below. Query complexity counts oracle and projection calls.

Write $f^\star=\min_{\mathcal K}f$; an \emph{$\varepsilon$--minimizer} is a feasible point $\widehat\x$ with $f(\widehat\x)\le f^\star+\varepsilon$. For smooth objectives, the results of Mart\'inez-Rubio~\cite{martinezrubio2025smooth} apply directly.

\begin{corollary}[Query Complexity]
\label{cor:tilted-pgd}
Suppose that $f$ is differentiable on an open set containing $\mathcal K$, that $\nabla f$ is $L$--Lipschitz on $\mathcal K$, and that $f|_{\mathcal K}\in\mathcal F^\gamma$.
\begin{enumerate}[label=(\roman*)]
\item Let $\x_0\in\mathcal K$, and let $D_0\le D$ be the diameter of $\{\x\in\mathcal K:f(\x)\le f(\x_0)\}$. The projected gradient iterates $\x_{t+1}=\Pi_{\mathcal K}\bigl(\x_t-L^{-1}\nabla f(\x_t)\bigr)$ satisfy
\begin{equation}\label{eq:tilted-pgd-rate}
f(\x_t)-f^\star\le\frac{20LD_0^2}{(t+1)\gamma^2}\qquad(t\ge1),
\end{equation}
so $O(1+LD^2/(\gamma^2\varepsilon))$ gradient evaluations and projections give an $\varepsilon$--minimizer. The step size does not depend on $\gamma$.
\item With a linear minimization oracle, the Frank--Wolfe iterates of~\cite[Theorem~9]{martinezrubio2025smooth} satisfy $f(\x_t)-f^\star\le6LD^2/\bigl((t+1)\gamma^2\bigr)$ for $t\ge1$.
\item The accelerated method of~\cite[Theorem~5]{martinezrubio2025smooth} finds an $\varepsilon$--minimizer with
\begin{equation}\label{eq:tilted-accelerated-rate}
\widetilde O\Bigl(\frac1\gamma\sqrt{\frac{LD^2}{\varepsilon}}\Bigr)
\end{equation}
first-order queries, where $\widetilde O$ hides logarithmic factors.
\end{enumerate}
\end{corollary}

\begin{proof}
Since $f$ is continuous and $\mathcal K$ is compact, $f$ attains its minimum on $\mathcal K$ at some $\x^\star$. As noted above, $\nabla f(\x)$ is a support vector at every $\x\in\mathcal K$, so $f(\y)-f(\x)\ge\psi_\gamma(\ip{\nabla f(\x)}{\y-\x})$ for all $\x,\y\in\mathcal K$. This is tilted convexity on $\mathcal K$ with $\gamma_n=\gamma_p=\gamma$~\cite[Eq.~(3)]{martinezrubio2022global}, and tilted-convex functions are $\gamma$--quasar-convex in $\mathcal K$ with center $\x^\star$~\cite[Section~2]{martinezrubio2025smooth}. The other hypotheses of Theorems~5, 8, and~9 of~\cite{martinezrubio2025smooth}, namely $L$--smoothness on $\mathcal K$ and compactness and convexity of $\mathcal K$, hold by assumption. Theorem~8 gives \eqref{eq:tilted-pgd-rate}, Theorem~9 gives (ii), and Theorem~5 gives (iii). The right side of \eqref{eq:tilted-pgd-rate} is at most $\varepsilon$ once $t+1\ge20LD^2/(\gamma^2\varepsilon)$, and each iteration in (i) uses one gradient evaluation and one projection.
\end{proof}

\section{\texorpdfstring{$\gamma$}{gamma}-Approximate Online Optimization}
\label{sec:linearization}

We give a polynomial-time online algorithm for nonpositive $\gamma$--tilted convex losses whose $\gamma$--regret is $O(DG\sqrt T)$; as in Section~\ref{sec:offline-opt}, results stated for $\mathcal F^\gamma$ apply to the two smaller classes. The algorithm passes one vector per round to an online learner for linear losses, following the reduction of Pedramfar and Aggarwal~\cite{pedramfar2024linear} (Section~\ref{subsec:lin-prelim}). We prove that a suitable vector with coefficient $\gamma$ exists for every continuous nonpositive loss in $\mathcal F^\gamma$ at every point with a support vector (Theorem~\ref{thm:linearization}), compute it with polynomially many oracle calls (Theorem~\ref{thm:query}), and derive the regret bound (Corollary~\ref{cor:ombq}).

\subsection{Online model and approximate regret}
\label{subsec:online}

A learner and an adversary interact for $T$ rounds on $\mathcal K$. In round $t$ the learner chooses $\x_t\in\mathcal K$ from its past observations and incurs the loss $f_t(\x_t)$; it then receives the oracle \eqref{eq:oracle-model} for $f_t$, which it may query at any points of $\mathcal K$ before choosing $\x_{t+1}$. Our upper bounds hold for every realized loss sequence, including sequences that the adversary adapts to the learner's past actions, whereas the lower bound of Section~\ref{sec:hardness} uses independent losses drawn from a fixed distribution.

The losses satisfy $-B\le f_t\le0$ on $\mathcal K$ for some $B>0$. Under this normalization, multiplying the benchmark $\min_{\uu}\sum_tf_t(\uu)\le0$ by a coefficient $\alpha<1$ raises it, so a smaller $\alpha$ gives a weaker benchmark. For $\alpha\in(0,1]$, an interval $I=[s,e]$ of rounds, and comparators $\uu_s,\dots,\uu_e\in\mathcal K$, let
\begin{equation}\label{eq:regret-path}
\Reg_\alpha(I;\uu_{s:e})=\sum_{t=s}^e\bigl[f_t(\x_t)-\alpha f_t(\uu_t)\bigr].
\end{equation}
The static $\alpha$--regret on $I$ is
\begin{equation}\label{eq:static}
\SReg_\alpha(I)=\sum_{t\in I}f_t(\x_t)-\alpha\min_{\uu\in\mathcal K}\sum_{t\in I}f_t(\uu),
\end{equation}
and we write $\SReg_\alpha(T)$ for $\SReg_\alpha([1,T])$, which is the $\alpha$--regret $\operatorname{Reg}_\alpha(T)$ of Section~\ref{sec:intro}.

Since the lower bound of Section~\ref{sec:hardness} concerns computation, the online results are stated in the bit model. Query points are rational, and values, support vectors, and feasible approximate projections are available to any requested accuracy in time polynomial in the input length and the requested number of bits. Numerical parameters are rational or are given by procedures that output certified rational intervals of width at most $2^{-b}$ in time polynomial in the input length and $b$, and a nonrational $\gamma$ also comes with a rational lower bound $0<\gamma_{\rm lb}\le\gamma$ of polynomial encoding length; we then call $\gamma$ \emph{effectively specified}. Running times are polynomial in the numerical quantities $D$, $G$, and $1/\gamma$, not only in their encoding lengths. If $D=0$ or $G=0$, every feasible point has the same value; otherwise $D$ and $G$ are positive rational upper bounds.

A coefficient $\alpha\in(0,1]$ is \emph{achievable} for a class of losses if an online algorithm, possibly randomized, that is given the horizon $T$ and runs in time polynomial in $N$ and $T$ satisfies
\begin{equation}\label{eq:achievable}
\mathbb E\bigl[\SReg_\alpha(T)\bigr]\le P(N)\,T^{1-\rho}
\end{equation}
for every horizon $T$ and every sequence of losses from the class. Here $P$ is a fixed polynomial and $\rho>0$ is a fixed constant. The number $N$ bounds the encoding length of $\mathcal K$ and of a single loss, the descriptions of the parameter-approximation procedures, the encoding lengths of all numerical bounds, and the numerical values $B,D,G$ of the normalization and of Section~\ref{sec:offline-opt}. It does not depend on $T$ or on the number of losses, and the oracle \eqref{eq:oracle-model} answers to any requested accuracy, as in the bit model above. The \emph{online coefficient} is $\sup(\{0\}\cup\{\alpha\in(0,1]:\alpha\text{ is achievable}\})$.

\subsection{Lower-linearizability and the reduction to linear losses}
\label{subsec:lin-prelim}

Pedramfar and Aggarwal~\cite{pedramfar2024linear} reduce online optimization of nonlinear losses to online linear optimization whenever each loss admits a comparison vector that is valid against all comparators. In the minimization form, a class $\mathcal F$ of functions on $\mathcal K$ is \emph{lower-linearizable} if there are maps $g:\mathcal F\times\mathcal K\to\R^d$ and $\chi:\mathcal K\to\mathcal K$ and constants $\alpha\in(0,1]$ and $\beta>0$ such that
\begin{equation}\label{eq:lower-linearizable-general}
f(\chi(\x))-\alpha f(\y)\le\beta\,\ip{g(f,\x)}{\x-\y}\qquad\text{for all }f\in\mathcal F\text{ and }\x,\y\in\mathcal K.
\end{equation}
The map $\chi$ transforms the point proposed by the linear learner into the point that is played. We use the special case $\chi=\mathrm{id}$ and $\beta=1$, and call $\mathcal F$ \emph{lower-linearizable with coefficient $\alpha$} if
\begin{equation}\label{eq:lower-linearizable}
f(\x)-\alpha f(\y)\le\ip{g(f,\x)}{\x-\y}\qquad\text{for all }f\in\mathcal F\text{ and }\x,\y\in\mathcal K.
\end{equation}
Convex functions satisfy \eqref{eq:lower-linearizable} with $\alpha=1$ and $g(f,\x)\in\partial f(\x)$.

Let $\mathcal A$ be a deterministic online algorithm for linear losses $\z\mapsto\ip{\a_t}{\z}$ on $\mathcal K$ with $\norm{\a_t}\le G$. A \emph{pathwise regret bound} for $\mathcal A$ is a function $\mathcal B_{\mathcal A}$ such that
\begin{equation}\label{eq:linear-bound}
\sum_{t\in I}\ip{\a_t}{\x_t-\uu_t}\le\mathcal B_{\mathcal A}(I;\uu_{s:e})
\end{equation}
for every realized sequence $\a_1,\a_2,\dots$, every interval $I=[s,e]$, and every comparator path. The vector $\a_t$ may depend on the action $\x_t$ that $\mathcal A$ plays in round $t$; Algorithm~\ref{alg:OMBQ} implements this reduction.

\begin{algorithm}[t]
\caption{Reduction to online linear optimization (minimization form of OMBQ~\cite{pedramfar2024linear})}
\label{alg:OMBQ}
\begin{algorithmic}[1]
\Require Online algorithm $\mathcal A$ for linear losses on $\mathcal K$; procedure $\mathcal G$ that maps a loss oracle and a point to a vector
\For{$t=1,2,\dots,T$}
\State Play the action $\x_t$ proposed by $\mathcal A$ and incur $f_t(\x_t)$.
\State Compute $\a_t=\mathcal G(\mathcal Q_{f_t},\x_t)$.
\State Feed the linear loss $\z\mapsto\ip{\a_t}{\z}$ to $\mathcal A$.
\EndFor
\end{algorithmic}
\end{algorithm}

If $\a_t=g(f_t,\x_t)$ for a map $g$ satisfying \eqref{eq:lower-linearizable}, then applying \eqref{eq:lower-linearizable} with $\x=\x_t$ and $\y=\uu_t$ and summing over $t\in I$ gives $\Reg_\alpha(I;\uu_{s:e})\le\mathcal B_{\mathcal A}(I;\uu_{s:e})$. Every regret bound of $\mathcal A$ for linear losses thus becomes an $\alpha$--regret bound for $\mathcal F$ with the same rate. For nonpositive losses in $\mathcal F^\gamma$ we show that $\alpha=\gamma$ is possible with $\norm{g}\le G$, and that an additive approximation of $g$ can be computed from the oracle.

\subsection{A comparison vector with coefficient \texorpdfstring{$\gamma$}{gamma}}
\label{subsec:comparison-field}

In this subsection $\mathcal K$ is any convex set, $f:\mathcal K\to\R$ is continuous and nonpositive with $f\in\mathcal F^\gamma$, $q_f(\x)$ is a support vector at the base point $\x$, and $r=-f\ge0$. Lemma~\ref{lem:interpolation}, which integrates \eqref{eq:gamma-mwc} along a segment, is the source of the coefficient $\gamma$.

\begin{lemma}[Interpolation for nonnegative rewards]\label{lem:interpolation}
For $\y_0,\y_1\in\mathcal K$ and $t\in[0,1]$,
\begin{equation}\label{eq:reward-interpolation}
r\bigl((1-t)\y_0+t\y_1\bigr)\ge\gamma\bigl((1-t)r(\y_0)+t\,r(\y_1)\bigr).
\end{equation}
More precisely, if $A=r(\y_0)\le r(\y_1)=B_0$, then
\begin{equation}\label{eq:interpolation-strong}
r\bigl((1-t)\y_0+t\y_1\bigr)\ge B_0-(B_0-A)(1-t)^\gamma.
\end{equation}
\end{lemma}
\begin{proof}
If $\y_0=\y_1$, inequality \eqref{eq:reward-interpolation} reads $r(\y_0)\ge\gamma r(\y_0)$, which holds because $r\ge0$ and $\gamma\le1$, and \eqref{eq:interpolation-strong} reads $r(\y_0)\ge r(\y_0)$. Assume $\y_0\ne\y_1$. Let $\widetilde f$ be the proper extension of $f$, let $\widehat g(t)=\widetilde f(\y_0+t(\y_1-\y_0))$ for $t\in\R$, and let $\phi(t)=r((1-t)\y_0+t\y_1)$ for $t\in[0,1]$, so that $\widehat g=-\phi$ on $[0,1]$.

By Theorem~\ref{thm:gamma-MW-convex-on-lines}, the restriction of $\widetilde f$ to the line through $\y_0$ and $\y_1$ is $\gamma$--tilted convex. Let $c=\norm{\y_1-\y_0}>0$. We identify this line with $\R$ through the isometry $\tau\mapsto\y_0+\tau(\y_1-\y_0)/c$, so that $k(\tau)=\widetilde f(\y_0+\tau(\y_1-\y_0)/c)$ is $\gamma$--tilted convex on $\R$ and $\widehat g(t)=k(ct)$. The map $t\mapsto ct$ is a linear bijection of $\R$, so the definitions of the subdifferentials give $\overline\partial\widehat g(t)=c\,\overline\partial k(ct)$. Let $t,t'$ lie in the domain of $\widehat g$ and let $w\in\overline\partial\widehat g(t)$, so that $w=cu$ with $u\in\overline\partial k(ct)$. Inequality \eqref{eq:gamma-mwc} for $k$ gives
\begin{equation}\label{eq:po-line-mwc}
\widehat g(t')-\widehat g(t)=k(ct')-k(ct)\ge\psi_\gamma\bigl((ct'-ct)u\bigr)=\psi_\gamma\bigl((t'-t)w\bigr).
\end{equation}
By \eqref{eq:po-line-mwc}, $\widehat g$ is $\gamma$--tilted convex on $\R$. Its domain $\{t:\y_0+t(\y_1-\y_0)\in\mathcal K\}$ is an interval containing $[0,1]$, because $\mathcal K$ is convex. By Theorem~\ref{thm:gamma-MW-convex-is-locally-Lipschitz}, $\widehat g$ is locally Lipschitz on the interior of its domain, which contains $(0,1)$. A locally Lipschitz function is Lipschitz on each compact subset, so $\phi=-\widehat g$ is Lipschitz on every interval $[t_0,t_1]\subset(0,1)$ and, by Rademacher's theorem, differentiable at almost every $t\in(0,1)$. Since $f$ is continuous on $\mathcal K$, $\phi$ is continuous on $[0,1]$.

Let $t\in(0,1)$ be a point where $\phi$ is differentiable. Then $\widehat g=-\phi$ on a neighborhood of $t$, so $\widehat g$ is differentiable at $t$ with derivative $-\phi'(t)$. Proposition~\ref{prop:subdiff-basics}(iii) gives $-\phi'(t)\in\partial\widehat g(t)$, and Proposition~\ref{prop:subdiff-basics}(i) gives $-\phi'(t)\in\overline\partial\widehat g(t)$. We apply \eqref{eq:po-line-mwc} at $t$ with $t'=1$ and $w=-\phi'(t)$. Since $\widehat g(1)=-\phi(1)=-B_0$ and $\widehat g(t)=-\phi(t)$, it reads
\begin{equation}\label{eq:interp-support}
\psi_\gamma\bigl(-(1-t)\phi'(t)\bigr)\le\phi(t)-B_0.
\end{equation}
If $\phi(t)<B_0$, the right side of \eqref{eq:interp-support} is negative. Since $\psi_\gamma\ge0$ on $[0,\infty)$, this forces $-(1-t)\phi'(t)<0$. On negative arguments $\psi_\gamma(z)=z/\gamma$, so the left side of \eqref{eq:interp-support} equals $-(1-t)\phi'(t)/\gamma$. Multiplying \eqref{eq:interp-support} by $-\gamma/(1-t)<0$ gives
\begin{equation}\label{eq:interp-ode}
\phi'(t)\ge\frac{\gamma\,(B_0-\phi(t))}{1-t}\qquad\text{whenever }\phi(t)<B_0.
\end{equation}

Let $w(t)=\max\{B_0-\phi(t),0\}$. It is continuous on $[0,1]$, Lipschitz on every $[t_0,t_1]\subset(0,1)$, and $w(0)=B_0-A$. At every $t\in(0,1)$ where $\phi$ and $w$ are both differentiable,
\begin{equation}\label{eq:interp-w}
w'(t)\le-\frac{\gamma\,w(t)}{1-t}.
\end{equation}
Indeed, if $w(t)>0$, then $\phi(t)<B_0$ and, by continuity, $w=B_0-\phi$ near $t$, so \eqref{eq:interp-ode} gives $w'(t)=-\phi'(t)\le-\gamma w(t)/(1-t)$. If $w(t)=0$, then $t$ is an interior minimizer of the nonnegative function $w$, so $w'(t)=0$, which equals the right side of \eqref{eq:interp-w}. Since $\phi$ and $w$ are Lipschitz on every $[t_0,t_1]\subset(0,1)$, both are differentiable almost everywhere, so \eqref{eq:interp-w} holds at almost every $t\in(0,1)$. The function $u(t)=w(t)(1-t)^{-\gamma}$ is a product of two Lipschitz functions on every $[t_0,t_1]\subset(0,1)$, hence absolutely continuous there. By the product rule and \eqref{eq:interp-w},
\begin{equation}\label{eq:po-u-decreasing}
u'(t)=(1-t)^{-\gamma}w'(t)+\gamma(1-t)^{-\gamma-1}w(t)=(1-t)^{-\gamma}\Bigl(w'(t)+\frac{\gamma\,w(t)}{1-t}\Bigr)\le0
\end{equation}
almost everywhere. For $0<t_0<t<1$, absolute continuity and \eqref{eq:po-u-decreasing} give $u(t)-u(t_0)=\int_{t_0}^tu'(\sigma)\,d\sigma\le0$, that is,
\begin{equation}\label{eq:po-u-compare}
w(t)(1-t)^{-\gamma}\le w(t_0)(1-t_0)^{-\gamma}.
\end{equation}
The right side of \eqref{eq:po-u-compare} tends to $w(0)=B_0-A$ as $t_0\downarrow0$, because $w$ is continuous at $0$. Hence $w(t)\le(B_0-A)(1-t)^\gamma$. Since $w\ge B_0-\phi$, we obtain $\phi(t)\ge B_0-(B_0-A)(1-t)^\gamma$ for $t\in(0,1)$. At $t=0$ and $t=1$ the right side equals $\phi(0)=A$ and $\phi(1)=B_0$, which proves \eqref{eq:interpolation-strong}.

The function $\sigma\mapsto\sigma^\gamma$ is concave on $[0,\infty)$ because $\gamma\in(0,1]$, so it lies below its tangent $1+\gamma(\sigma-1)$ at $\sigma=1$. With $\sigma=1-t$ this gives
\begin{equation}\label{eq:po-tangent}
(1-t)^\gamma\le1-\gamma t.
\end{equation}
Since $B_0-A\ge0$, substituting \eqref{eq:po-tangent} into \eqref{eq:interpolation-strong} gives
\[
\phi(t)\ge B_0-(B_0-A)(1-\gamma t)=A+\gamma t(B_0-A)=\gamma\bigl((1-t)A+tB_0\bigr)+(1-\gamma)A\ge\gamma\bigl((1-t)A+tB_0\bigr),
\]
where the last step uses $A\ge0$ and $\gamma\le1$. This is \eqref{eq:reward-interpolation} when $r(\y_0)\le r(\y_1)$. If $r(\y_0)>r(\y_1)$, applying this case to the pair $(\y_1,\y_0)$ with parameter $1-t$ gives the same point $(1-t)\y_0+t\y_1$ and the bound $\gamma(t\,r(\y_1)+(1-t)r(\y_0))$.
\end{proof}

For vectors $p$ and $p'$, let $[p,p']$ denote the closed segment between them. The comparison vector is a rescaling of the support vector $q_f(\x)$ by a factor in $[\gamma,1]$.

\begin{theorem}[Comparison vector with coefficient $\gamma$]\label{thm:linearization}
Let $f:\mathcal K\to\R$ be continuous and nonpositive with $f\in\mathcal F^\gamma$, and let $\x\in\mathcal K$. The set
\begin{equation}\label{eq:g-feasible}
\bigl\{\v\in[\gamma q_f(\x),q_f(\x)]:\ f(\x)-\gamma f(\y)\le\ip{\v}{\x-\y}\ \text{for every }\y\in\mathcal K\bigr\}
\end{equation}
is nonempty, convex, and compact. Let $g(f,\x)$ be its element of minimum norm. Then
\begin{equation}\label{eq:gamma-linearization}
f(\x)-\gamma f(\y)\le\ip{g(f,\x)}{\x-\y}\quad(\y\in\mathcal K),
\qquad \norm{g(f,\x)}\le\norm{q_f(\x)}.
\end{equation}
In particular, the continuous nonpositive functions on $\mathcal K$ in $\mathcal F^\gamma$ form a lower-linearizable class with coefficient $\gamma$, and $\norm{g(f,\x)}\le G$ under the oracle bound of Section~\ref{sec:offline-opt}.
\end{theorem}
\begin{proof}
Write $q=q_f(\x)$, $p=-q$, and $r=-f\ge0$, and for $\y\in\mathcal K$ let $s_\y=\ip{p}{\y-\x}$ and $c_\y=\gamma r(\y)-r(\x)$. Since $\ip{\y-\x}{q}=-s_\y$, inequality \eqref{eq:gamma-mwc} at $\x$ with the vector $q$ reads $f(\y)-f(\x)\ge\psi_\gamma(-s_\y)$, that is, $r(\y)-r(\x)\le-\psi_\gamma(-s_\y)$. If $s_\y\ge0$, then $-s_\y\le0$ and $-\psi_\gamma(-s_\y)=s_\y/\gamma$. If $s_\y<0$, then $-s_\y>0$ and $-\psi_\gamma(-s_\y)=\gamma s_\y$. Hence
\begin{equation}\label{eq:reward-support}
r(\y)-r(\x)\le
\begin{cases}
s_\y/\gamma,& s_\y\ge0,\\
\gamma s_\y,& s_\y<0.
\end{cases}
\end{equation}
For a real number $t$, the identities $f=-r$ and $\ip{q}{\x-\y}=s_\y$ give
\begin{equation}\label{eq:scalar-constraint}
f(\x)-\gamma f(\y)-\ip{t\,q}{\x-\y}=-r(\x)+\gamma r(\y)-t\,s_\y=c_\y-t\,s_\y,
\end{equation}
so $t\,q$ satisfies the constraint of \eqref{eq:g-feasible} for $\y$ exactly when $c_\y\le t\,s_\y$.

If $q=\BF0$, then $s_\y=0$ for all $\y$, \eqref{eq:reward-support} gives $r(\y)\le r(\x)$, and $c_\y\le\gamma r(\x)-r(\x)=-(1-\gamma)r(\x)\le0$. So \eqref{eq:g-feasible} equals $\{\BF0\}=[\gamma q,q]$, and all claims hold with $g(f,\x)=\BF0$. Assume from now on that $q\ne\BF0$, so that $[\gamma q,q]=\{t\,q:t\in[\gamma,1]\}$, and sort the comparators by the sign of $s_\y$, using \eqref{eq:reward-support} in each case.
\begin{itemize}
\item If $s_\y=0$, then $r(\y)\le r(\x)$, so $c_\y\le-(1-\gamma)r(\x)\le0=t\,s_\y$, and the constraint holds for every $t$.
\item If $s_\y>0$, dividing $c_\y\le t\,s_\y$ by $s_\y$ shows that the constraint reads $t\ge c_\y/s_\y$. Since $r(\y)\le r(\x)+s_\y/\gamma$ and $r(\x)\ge0$,
\begin{equation}\label{eq:po-lower-bound}
c_\y\le\gamma r(\x)+s_\y-r(\x)=s_\y-(1-\gamma)r(\x)\le s_\y,\qquad\text{so}\qquad \frac{c_\y}{s_\y}\le1 .
\end{equation}
\item If $s_\y<0$, dividing $c_\y\le t\,s_\y$ by $s_\y<0$ shows that the constraint reads $t\le c_\y/s_\y$. Since $r(\y)-r(\x)\le\gamma s_\y$ and $r(\y)\ge0$,
\begin{equation}\label{eq:po-upper-bound}
c_\y=\bigl(r(\y)-r(\x)\bigr)-(1-\gamma)r(\y)\le\gamma s_\y,\qquad\text{so}\qquad \frac{c_\y}{s_\y}\ge\gamma .
\end{equation}
\end{itemize}

To compare lower and upper bounds, let $\y_+,\y_-\in\mathcal K$ with $s_+=s_{\y_+}>0>s_-=s_{\y_-}$, let $\lambda=-s_-/(s_+-s_-)\in(0,1)$, and let $\z=\lambda\y_++(1-\lambda)\y_-\in\mathcal K$. Since $\y\mapsto s_\y$ is affine and $1-\lambda=s_+/(s_+-s_-)$,
\[
s_\z=\lambda s_++(1-\lambda)s_-=\frac{-s_-s_++s_+s_-}{s_+-s_-}=0,
\]
so $r(\z)\le r(\x)$ by the first case. Lemma~\ref{lem:interpolation} with $\y_0=\y_-$, $\y_1=\y_+$, and $t=\lambda$ gives $r(\z)\ge\gamma(\lambda r(\y_+)+(1-\lambda)r(\y_-))$. Combining the two bounds on $r(\z)$ with the definition of $c_\y$,
\begin{equation}\label{eq:pairing}
\lambda c_{\y_+}+(1-\lambda)c_{\y_-}=\gamma\bigl(\lambda r(\y_+)+(1-\lambda)r(\y_-)\bigr)-r(\x)\le r(\z)-r(\x)\le0.
\end{equation}
Multiplying \eqref{eq:pairing} by $s_+-s_->0$ and using $\lambda(s_+-s_-)=-s_-$ and $(1-\lambda)(s_+-s_-)=s_+$ gives $-s_-c_{\y_+}+s_+c_{\y_-}\le0$, that is, $s_+c_{\y_-}\le s_-c_{\y_+}$. Dividing by $s_+s_-<0$ reverses the inequality and gives
\begin{equation}\label{eq:po-ratio-compare}
\frac{c_{\y_+}}{s_+}\le\frac{c_{\y_-}}{s_-}.
\end{equation}
So every lower bound is at most every upper bound. Let
\begin{equation}\label{eq:po-tstar}
t^*=\max\Bigl\{\gamma,\ \sup_{s_\y>0}\frac{c_\y}{s_\y}\Bigr\},
\end{equation}
with the supremum over the empty set equal to $-\infty$. By \eqref{eq:po-lower-bound}, $t^*\in[\gamma,1]$, and by \eqref{eq:po-tstar}, $t^*$ is at least every lower bound. By \eqref{eq:po-upper-bound} and \eqref{eq:po-ratio-compare}, both $\gamma$ and every lower bound are at most every upper bound, so $t^*$ is at most every upper bound. Since the first case imposes no restriction, \eqref{eq:scalar-constraint} shows that $t^*q$ lies in \eqref{eq:g-feasible}.

The set \eqref{eq:g-feasible} is the intersection of the compact segment $[\gamma q,q]$ with the closed half-spaces $\{\v:c_\y\le\ip{\v}{\x-\y}\}$, which are all of $\R^d$ when $\y=\x$ because $c_\x\le0$. So it is convex and compact. The continuous function $\norm\cdot^2$ attains its minimum on it, and the minimizer is unique: if $\v_1\ne\v_2$ both attain the minimum $m$, then $\norm{(\v_1+\v_2)/2}^2=\frac12\norm{\v_1}^2+\frac12\norm{\v_2}^2-\frac14\norm{\v_1-\v_2}^2<m$. This defines $g(f,\x)$, and its membership in \eqref{eq:g-feasible} is the first part of \eqref{eq:gamma-linearization}. Every $t\,q$ with $t\in[\gamma,1]$ has norm $t\norm q\le\norm q$, which gives the second part. The final sentence of the theorem follows from \eqref{eq:lower-linearizable} with $\alpha=\gamma$ and $\norm{q_f(\x)}\le G$.
\end{proof}

The factor in $[\gamma,1]$ that defines $g(f,\x)$ depends on the values of $f$ on all of $\mathcal K$, so in general it cannot be read off from the oracle at $\x$. Because one vector serves every comparator, it can be passed to the linear learner before the comparator is known.

\subsection{Computing the comparison vector}
\label{subsec:query}

The set \eqref{eq:g-feasible} is cut out by one constraint for each comparator $\y\in\mathcal K$, so we search the segment $[\gamma q_f(\x),q_f(\x)]$ by bisection on the scale factor and test each candidate by approximately maximizing its constraint violation over $\y$. In this subsection $\mathcal K$ is compact with diameter at most $D$, and $f:\mathcal K\to\R$ is nonpositive and $G$--Lipschitz with $f\in\mathcal F^\gamma$, with the access of Section~\ref{sec:offline-opt}. For a base point $\x$ and a vector $\v$, the \emph{violation} of $\v$ is
\begin{equation}\label{eq:violation}
\Phi_\x(\v)=\max_{\y\in\mathcal K}\bigl[f(\x)-\gamma f(\y)-\ip{\v}{\x-\y}\bigr].
\end{equation}
It is convex and $D$--Lipschitz in $\v$, and a vector $\v\in[\gamma q_f(\x),q_f(\x)]$ lies in \eqref{eq:g-feasible} exactly when $\Phi_\x(\v)\le0$. The function maximized in \eqref{eq:violation} is not concave in $\y$, so convex optimization does not evaluate $\Phi_\x$ directly. Lemma~\ref{lem:violation-oracle} instead splits the range of $f$ into levels and, on each level $a$, maximizes the linear function $\ip{\v}{\y-\x}$ over the sublevel set $\{\y\in\mathcal K:f(\y)\le a\}$.

\begin{lemma}[Evaluating the violation]\label{lem:violation-oracle}
Let $\x\in\mathcal K$, $\norm{\v}\le G$, and $h>0$. A deterministic procedure returns a point $\y_\v\in\mathcal K$ whose value $V_\v=f(\x)-\gamma f(\y_\v)-\ip{\v}{\x-\y_\v}$ satisfies
\begin{equation}\label{eq:violation-accuracy}
\Phi_\x(\v)-4h\le V_\v\le\Phi_\x(\v),
\end{equation}
using
\begin{equation}\label{eq:candidate-complexity}
O\bigl((1+\gamma DG/h)\,\bigl[1+(DG/h)^2\bigr]\bigr)
\end{equation}
value, support, and projection calls.
\end{lemma}

In the proof, given in Appendix~\ref{apx:violation}, a projected method at level $a$ moves against $q_f(\z_j)$ when $f(\z_j)$ exceeds $a+h/\gamma$ and along $\v$ otherwise. Inequality \eqref{eq:gamma-mwc} shows that either move makes progress at least $h$ toward a maximizer of $\ip{\v}{\y}$ on $\{f\le a\}$, and the projected-subgradient potential \citep{ben2001lectures} bounds the number of steps. The levels are spaced $h/\gamma$ apart because $f(\y)$ enters \eqref{eq:violation} with weight $\gamma$, and this spacing keeps inverse powers of $\gamma$ out of \eqref{eq:candidate-complexity}.

Algorithm~\ref{alg:query} combines Lemma~\ref{lem:violation-oracle} with bisection on the scale factor. It maintains an interval of scale factors that contains a feasible one, which exists initially by Theorem~\ref{thm:linearization}. When a candidate $t\,q_f(\x)$ has a large violation, the witness $\y_\v$ gives a constraint that every feasible scale factor satisfies and $t$ violates, and the sign of $\ip{q_f(\x)}{\x-\y_\v}$ determines on which side of $t$ the feasible scale factors lie.

\begin{algorithm}[t]
\caption{Computing a comparison vector with coefficient $\gamma$}
\label{alg:query}
\begin{algorithmic}[1]
\Require Oracle $\mathcal Q_f$, point $\x\in\mathcal K$, parameter $\gamma\in(0,1]$, tolerance $\varepsilon>0$, bounds $D,G>0$
\State Query $f(\x)$ and $q_f(\x)$. Return $\BF0$ if $q_f(\x)=\BF0$, and return $q_f(\x)$ if $\gamma=1$.
\State Set $\underline t=\gamma$, $\overline t=1$, and $h=\varepsilon/16$.
\While{$DG(\overline t-\underline t)>\varepsilon/4$}
\State Set $t=(\underline t+\overline t)/2$ and $\v=t\,q_f(\x)$.
\State Obtain $\y_\v$ and $V_\v$ from Lemma~\ref{lem:violation-oracle} with accuracy $h$.
\If{$V_\v\le\varepsilon/2$} \State \Return $\v$. \EndIf
\If{$\ip{q_f(\x)}{\x-\y_\v}>0$} \State Set $\underline t=t$. \Else \State Set $\overline t=t$. \EndIf
\EndWhile
\State \Return $\frac{\underline t+\overline t}2\,q_f(\x)$.
\end{algorithmic}
\end{algorithm}

\begin{theorem}[Computing the comparison vector]\label{thm:query}
Let $\x\in\mathcal K$ and $\varepsilon>0$, and let $f:\mathcal K\to\R$ be nonpositive and $G$--Lipschitz with $f\in\mathcal F^\gamma$.
\begin{enumerate}[label=(\roman*),leftmargin=*]
\item \emph{Exact oracle.} Algorithm~\ref{alg:query} returns $\widehat g\in[\gamma q_f(\x),q_f(\x)]$ with
\begin{equation}\label{eq:approx-linearization}
f(\x)-\gamma f(\y)\le\ip{\widehat g}{\x-\y}+\varepsilon\qquad(\y\in\mathcal K),
\end{equation}
using
\begin{equation}\label{eq:query-complexity}
O\bigl((1+\gamma DG/\varepsilon)\log(2+DG/\varepsilon)\bigl[1+(DG/\varepsilon)^2\bigr]\bigr)
\end{equation}
oracle calls.
\item \emph{Inexact oracle.} Suppose that the oracle returns values $\widehat f(\z)$ with $|\widehat f(\z)-f(\z)|\le\zeta$ and vectors $\a_\z$ with $\norm{\a_\z}\le G$ and
\begin{equation}\label{eq:noisy-oracle}
f(\y)-f(\z)\ge\psi_\gamma\bigl(\ip{\a_\z}{\y-\z}\bigr)-\xi\qquad(\y,\z\in\mathcal K),
\end{equation}
where $\xi,\zeta\ge0$ are known. A variant of Algorithm~\ref{alg:query} with the same number of calls returns $\widehat g\in[\gamma\a_\x,\a_\x]$ with
\begin{equation}\label{eq:noisy-comparison}
f(\x)-\gamma f(\y)\le\ip{\widehat g}{\x-\y}+\varepsilon+(1+\gamma)\xi+(2+4\gamma)\zeta\qquad(\y\in\mathcal K).
\end{equation}
\item \emph{Bit model.} Suppose that $B\ge\sup_{\mathcal K}|f|$ is known. For rational $\gamma$, 
part (ii) has a rational implementation returning $\widehat g\in[\gamma\a,\a]$, where $\a$ is rational, 
$\norm{\a-\a_\x}\le\gamma\varepsilon/(16D)$,  $\norm\a\le G$, 
and the output satisfies~\eqref{eq:noisy-comparison}. 
For an effectively specified $\gamma$ (Section~\ref{subsec:online}), the implementation returns a rational vector satisfying~\eqref{eq:noisy-comparison} at the original $\gamma$, 
with $\norm{\widehat g}\le G$. 
If $B>0$, it uses a rational $\gamma'\in[\gamma/2,\gamma]$ and guarantees $\widehat g\in[\gamma'\a,\a]$; if $B=0$, it may return zero. 
Running time is polynomial when $1/\varepsilon,D,G$ are polynomially bounded and the required numerical bounds and precisions have polynomial encoding length.
\end{enumerate}
The number of oracle calls has no inverse-power dependence on $\gamma$. For $\gamma=1$, the support vector $q_f(\x)$ itself satisfies \eqref{eq:approx-linearization} with $\varepsilon=0$.
\end{theorem}

The proof is in Appendix~\ref{apx:query-proof}. In part (ii), the defect $\xi$ measures how far $\a_\z$ is from satisfying the support inequality: if $\a_\z$ lies within distance $\nu$ of a support vector, then \eqref{eq:noisy-oracle} holds with $\xi=D\nu/\gamma$ because $\psi_\gamma$ is $(1/\gamma)$--Lipschitz, and the rational implementation in part (iii) relies on this conversion.

\FloatBarrier
\subsection{Regret bounds}
\label{subsec:regret}

We run Algorithm~\ref{alg:OMBQ} with the procedure $\mathcal G$ of Theorem~\ref{thm:query} at tolerance $\varepsilon_t$ in round $t$. If the oracle for $f_t$ has support defect $\xi_t$ and value error $\zeta_t$ as in Theorem~\ref{thm:query}(ii), let
\begin{equation}\label{eq:round-error}
\mathrm{err}_t=\varepsilon_t+(1+\gamma)\xi_t+(2+4\gamma)\zeta_t,\qquad \mathrm{Err}(I)=\sum_{t\in I}\mathrm{err}_t.
\end{equation}
Errors from rational arithmetic are included in $\varepsilon_t$ and in the bound $\mathcal B_{\mathcal A}$ of the linear learner.

\begin{corollary}[Static $\gamma$--regret]\label{cor:ombq}
Let $f_1,\dots,f_T:\mathcal K\to\R$ be nonpositive and $G$--Lipschitz with $f_t\in\mathcal F^\gamma$. For every interval $I=[s,e]$ and every comparator path, Algorithm~\ref{alg:OMBQ} with the vectors of Theorem~\ref{thm:query} satisfies
\begin{equation}\label{eq:ombq-transfer}
\Reg_\gamma(I;\uu_{s:e})\le\mathcal B_{\mathcal A}(I;\uu_{s:e})+\mathrm{Err}(I).
\end{equation}
With projected online gradient descent with step size $D/(G\sqrt T)$ as $\mathcal A$, this gives $\SReg_\gamma(T)\le DG\sqrt T+\mathrm{Err}([1,T])$. If $\mathrm{err}_t\le c/\sqrt t$ for every $t$, then every interval $I=[s,e]$ satisfies
\begin{equation}\label{eq:interval-error-condition}
\mathrm{Err}(I)\le2c\bigl(\sqrt e-\sqrt{s-1}\bigr)\le2c\sqrt{|I|},
\end{equation}
and with an exact oracle the tolerances $\varepsilon_t=c/\sqrt t$ satisfy this condition. Round $t$ uses the number of oracle calls in \eqref{eq:query-complexity} with $\varepsilon=\varepsilon_t$. With inverse-polynomial tolerances, the algorithm runs in polynomial time in the bit model.
\end{corollary}

Inequality \eqref{eq:ombq-transfer}, proved in Appendix~\ref{apx:linear-regret}, follows by summing \eqref{eq:noisy-comparison} with $\x=\x_t$ and $\y=\uu_t$ over $t\in I$ and applying \eqref{eq:linear-bound}. For a known horizon and $DG>0$, choose rational per-round comparison errors with $\mathrm{err}_t\le DG/(2\sqrt T)$ and charge at most another $DG\sqrt T/2$ to rational updates of the linear learner, so that $\SReg_\gamma(T)\le2DG\sqrt T$. The required precisions have polynomial encoding length under Section~\ref{subsec:online}, so $\gamma$ is achievable on $\mathcal F^\gamma$. If $D=0$ or $G=0$, every feasible point has the same loss and the approximate regret is nonpositive.

\begin{remark}[Moving comparators and intervals]\label{rem:dynamic-adaptive}
Since inequality \eqref{eq:ombq-transfer} holds for every interval and every comparator path, every deterministic linear learner whose regret bound holds for all realized sequences of vectors gives the same bound on $\gamma$--regret, up to $\mathrm{Err}(I)$. Projected online gradient descent with a step size tuned to a known bound $P$ on the path length $\sum_t\norm{\uu_{t+1}-\uu_t}$ gives $O\bigl(G\sqrt{T(D^2+DP)}\bigr)$ against all such paths \citep{zinkevich2003online}. The strongly adaptive learner of Daniely, Gonen, and Shalev-Shwartz~\cite{daniely2015strongly}, played at the weighted average of its experts' points, gives $O\bigl(DG\sqrt{|I|}\log T\bigr)$ on every interval $I$ against a fixed comparator, because the loss of a linear function at the weighted average equals the weighted average of the losses.
\end{remark}

\section{Hardness Beyond \texorpdfstring{$\gamma$}{gamma}: A Common Framework}
\label{sec:hardness}
\label{sec:online-hardness}

We give a class-independent criterion for hardness of approximate online
optimization and verify it for the smallest of our three classes at a
fixed parameter.  By Corollary~\ref{cor:relations},
\begin{equation}
\label{eq:hardness-three-class-inclusion}
\mathcal F_{\op{mono}}^\gamma
\subseteq \mathcal F_{\op{sec}}^\gamma
\subseteq \mathcal F^\gamma.
\end{equation}
For every fixed $\gamma\in(0,1)$, we prove that no coefficient
$\alpha>\gamma$ is achievable on any of these classes unless $\NP=\RP$.
The proof has two parts: a gap property that does not mention
generalized convexity, and a scalar-profile criterion that places the
same hard family in all three classes.  Each loss in this family is
smooth and can be minimized by one linear optimization.

\subsection{The gap property}

The losses are continuous functions $f_t:\mathcal K\to[-1,0]$ on a
compact rational convex set $\mathcal K$, and regret is the static
$\alpha$--regret \eqref{eq:static}.  Computations use the bit model of
Section~\ref{subsec:online}, with rational actions of polynomial bit
length, and $P$, $\rho$, and $N$ are as in \eqref{eq:achievable}.
Lipschitz continuity and smoothness of a loss refer to its restriction
to $\mathcal K$.

\begin{definition}[Online gap property]
\label{def:online-gap-property}
Let $\beta\in[0,1)$, and let $\mathcal L(\mathcal K)$ be a class of
continuous losses $f:\mathcal K\to[-1,0]$ for each compact rational
convex set $\mathcal K$.  The class has the
\emph{$\beta$--online gap property} if, for every fixed
$\alpha\in(\beta,1]$, a deterministic polynomial-time reduction from
SAT produces the following data:
\begin{enumerate}[label=(\roman*)]
\item a set $\mathcal K$, a nonempty polynomial-size index set $I$,
and explicit losses $f_i\in\mathcal L(\mathcal K)$ for $i\in I$;
\item rational numbers $\Lambda>0$ and $0\le\tau<\alpha$, and a
score $Q:\mathcal K\to[0,1]$, such that
\begin{equation}
\label{eq:abstract-hardness-average-loss}
\frac1{|I|}\sum_{i\in I}f_i(\x)=-\frac{Q(\x)}{\Lambda};
\end{equation}
\item on a satisfiable formula, a point $\x^*\in\mathcal K$ exists
with $Q(\x^*)=1$;
\item on an unsatisfiable formula, $Q(\x)\le\tau$ for every rational
$\x\in\mathcal K$.
\end{enumerate}
All descriptions have polynomial encoding length; the numerical
quantities $\Lambda$ and $(\alpha-\tau)^{-1}$ are polynomially
bounded in the SAT input size.  The score is rational and exactly
computable in polynomial time at rational points, so comparison with
$\tau$ is exact.
Loss evaluation and every stipulated feasibility or loss oracle are
polynomial-time rational computations.  Uniform indices from $I$
can be sampled in expected polynomial time, with a polynomially
bounded implementation obtained by capping rejection sampling.
\end{definition}

\begin{theorem}[Class-independent hardness criterion]
\label{thm:online-gap-hardness-criterion}
Suppose $\mathcal L$ has the $\beta$--online gap property.  Fix
$\alpha\in(\beta,1]$.  If a randomized polynomial-time online
algorithm with rational actions satisfies either
\begin{equation}
\label{eq:hardness-expected-regret-assumption}
\mathbb E[\SReg_\alpha(T)]\le P(N)T^{1-\rho},
\end{equation}
which says that $\alpha$ is achievable, or
\begin{equation}
\label{eq:hardness-high-prob-regret-assumption}
\mathbb P\{\SReg_\alpha(T)\le P(N)T^{1-\rho}\}\ge 8/9,
\end{equation}
for every horizon and every admissible loss sequence, then $\NP=\RP$.
It suffices to assume the guarantee for independent uniform samples
from each finite family produced by the reduction.
\end{theorem}

In the proof (Appendix~\ref{apx:hardness-proof}), independent samples
on a satisfiable instance satisfy
\[
\mathbb E\!\left[
f_{J_t}(\x_t)-\alpha f_{J_t}(\x^*)
\,\middle|\,\text{past and learner's random tape}\right]
=\frac{\alpha-Q(\x_t)}{\Lambda}.
\]
Summed over the rounds, this identity shows that a low-regret guarantee
forces some iterate to have score above $\tau$ with inverse-polynomial
probability after polynomially many rounds.  An unsatisfiable instance
never produces such an iterate, so testing the scores of the iterates
gives a one-sided randomized algorithm for SAT.  The argument uses no
property of generalized convexity.

\subsection{A scalar-profile criterion}

Theorem~\ref{thm:direct-hard-profile-criterion} reduces the gap property to a
one-dimensional inequality and to membership of lifted losses in the class.
For $p,q\ge1$, let
\begin{equation}
\label{eq:hard-profile-local-domain}
\mathcal K_{p,q}
=\prod_{u=1}^p\left\{\x_u\in\mathbb R_+^q:
\sum_{a=1}^q x_{u,a}\le1\right\}.
\end{equation}
For $\ell\ge1$, block indices $u_j\in[p]$, and subsets $R_j\subseteq[q]$,
define
\begin{equation}
\label{eq:hard-profile-local-map}
z_{\boldsymbol u,\boldsymbol R}(\x)
=\frac1\ell\sum_{j=1}^\ell\sum_{a\in R_j}x_{u_j,a}.
\end{equation}
Repeated indices are allowed, and $0\le z\le1$ on
$\mathcal K_{p,q}$.  Writing $z(\x)=\langle\a,\x\rangle$, we have
$\a\ge0$ and $\|\a\|_2^2\le q$.

\begin{theorem}[Direct hard-profile criterion]
\label{thm:direct-hard-profile-criterion}
Let $\beta\in[0,1)$.  Suppose that for each rational
$\theta\in(\beta,1)$ there are a nondecreasing Lipschitz function
$H_\theta:[0,1]\to[0,1]$ and rational constants
$0\le a_\theta<\theta$ and $b_\theta>0$ such that
\begin{equation}
\label{eq:hard-profile-scalar-bounds}
H_\theta(0)=0,\quad H_\theta(1)=1,\quad
H_\theta(s)\le s,\quad
H_\theta(s)\le a_\theta s+b_\theta s^2
\quad(0\le s\le1).
\end{equation}
Suppose also that for every $p,q,\ell$ there is a rational scale
$S_{\theta,p,q,\ell}\ge1$, polynomially bounded in $p+q+\ell$ for fixed
$\theta$, such that all losses
\begin{equation}
\label{eq:hard-profile-lifted-loss}
f_{\boldsymbol u,\boldsymbol R}(\x)
=-\frac{H_\theta(z_{\boldsymbol u,\boldsymbol R}(\x))}
{S_{\theta,p,q,\ell}}
\end{equation}
belong to $\mathcal L(\mathcal K_{p,q})$.  The profiles, scales, losses,
and stipulated oracles must have polynomial-time rational
implementations, including exact profile evaluation on rational inputs.
Then $\mathcal L$ has the $\beta$--online gap property.
\end{theorem}

The proof in Appendix~\ref{apx:hard-profile-criterion} reduces from
Label Cover.  An instance $\Gamma$ of Label Cover consists of a bipartite
graph with left vertices $V_L$, right vertices $V_R$, and edges $E$,
finite label sets $\Sigma_L$ and $\Sigma_R$, and a map
$\pi_e:\Sigma_L\to\Sigma_R$ for each edge $e$.  A labeling assigns a
label in $\Sigma_L$ to each left vertex and a label in $\Sigma_R$ to
each right vertex, and it satisfies an edge $e=(u,v)$ when $\pi_e$
sends the label of $u$ to the label of $v$.  The value
$\operatorname{val}(\Gamma)$ is the largest fraction of edges that one
labeling satisfies.  For every fixed $\delta\in(0,1)$, a
polynomial-time reduction maps satisfiable SAT formulas to instances
of value $1$ and unsatisfiable ones to instances of value at most
$\delta$ \citep{moshkovitz2010two,moshkovitz2015projection}.

The feasible set $\mathcal K_\Gamma$ is a product of simplices, with one
block $\{\x_u\in\mathbb R_+^{\Sigma_L}:\sum_a x_{u,a}\le1\}$ for each
left vertex $u$, and a labeling of the left vertices corresponds to
the one-hot point that puts weight one on the label of each $u$.
For a right vertex $v$ and a label $b\in\Sigma_R$, the projected mass
$z_{v,b}(\x)$ averages, over the edges $e=(u,v)$ at $v$, the weight
that $\x_u$ places on labels sent to $b$ by $\pi_e$, so that
$\sum_b z_{v,b}\le1$.  With $w_v$ the fraction of edges incident to
$v$, the score is
\[
Q_\Gamma(\x)=\sum_v w_v\sum_b H_\theta(z_{v,b}(\x)).
\]
It equals one at the one-hot point of a labeling that satisfies every
edge, and rounding $\x$ deterministically to a labeling
(Appendix~\ref{apx:hard-profile-criterion}) gives
\[
Q_\Gamma(\x)\le a_\theta+b_\theta\operatorname{val}(\Gamma).
\]
On an unsatisfiable instance, the linear term $a_\theta s$ of
\eqref{eq:hard-profile-scalar-bounds} still contributes up to
$a_\theta$, since it does not depend on whether labels agree across
edges; the quadratic term contributes only through agreement, and
rounding bounds that contribution by $b_\theta\operatorname{val}(\Gamma)$.
Choosing $\delta$ small enough that $a_\theta+b_\theta\delta<\alpha$
establishes the gap for any $\alpha>\theta$, and $\theta$ can be taken
arbitrarily close to $\beta$.  The only class-specific step is
membership of~\eqref{eq:hard-profile-lifted-loss}.

\subsection{Profiles with a sharp monotonicity parameter}

The quadratic profile $h_\theta(s)=\theta s+(1-\theta)s^2$ satisfies
$-h_\theta=(1-\theta)h_c$ with $c=\theta/(1-\theta)$.  Multiplying a
function by a positive constant does not change its parameters, so
Example~\ref{ex:quadratic} gives
\begin{equation}
\label{eq:quadratic-mono-obstruction}
\gamma^*_{\op{mono}}(-h_\theta+\delta_{[0,1]})
=\frac{c}{c+2}=\frac{\theta}{2-\theta}.
\end{equation}
Consequently, requiring this quadratic to belong to
$\mathcal F_{\op{mono}}^\gamma$ forces
$\theta\ge2\gamma/(1+\gamma)$, which is too large for a gap approaching
$\gamma$. The score gap is controlled by the initial slope of the
profile, whereas membership in the monotonicity class is controlled
by the ratio of its smallest and largest slopes; for the quadratic
these quantities differ by the factor $2-\theta$. We choose a polynomial
profile whose largest slope approaches one while its
smallest-to-largest slope ratio remains fixed at $\kappa$, so its initial
slope approaches $\kappa$ and the score threshold can
approach $\gamma$ from above. When the extreme slopes occur at the
endpoints, the slope-ratio criterion is exact:

\begin{lemma}[Derivative ratios determine $\gamma^*_{\op{mono}}$]
\label{lem:hardness-derivative-ratio}
Let $H$ be $C^1$ on a neighborhood of $[0,1]$, and suppose
$0<A\le H'(s)\le B$ on $[0,1]$ with $H'(0)=A$ and $H'(1)=B$.
Then $\Psi=-H+\delta_{[0,1]}$ has
$\gamma^*_{\op{mono}}(\Psi)=A/B$.
\end{lemma}

Within the normalized convex-profile construction considered here,
the threshold cannot fall below $\gamma$. Let $H$ be a convex profile that is $C^1$ near $[0,1]$, with $H(0)=0$, $H(1)=1$, and
$H'(0)>0$.  Convexity makes $H'$ nondecreasing, so
Lemma~\ref{lem:hardness-derivative-ratio} gives
$\gamma^*_{\op{mono}}(-H+\delta_{[0,1]})=H'(0)/H'(1)$.  Convexity also
gives $1=H(1)-H(0)\le H'(1)$ and $H(s)\ge H(0)+H'(0)s=H'(0)s$.
Membership in $\mathcal F_{\op{mono}}^\gamma$ means
$H'(0)\ge\gamma H'(1)\ge\gamma$, and therefore
\begin{equation}
\label{eq:ph-profile-floor}
H(s)\ge\gamma s\qquad(0\le s\le1).
\end{equation}
At a point whose blocks each carry total weight one,
$\sum_b z_{v,b}(\x)=1$ for every $v$, so \eqref{eq:ph-profile-floor}
gives $Q_\Gamma(\x)\ge\gamma\sum_v w_v\sum_b z_{v,b}(\x)=\gamma$, on
satisfiable and unsatisfiable instances alike.  The reduction can lower the
unsatisfiable score to just above $\gamma$ but no further, which
matches the coefficient $\gamma$ achieved by Corollary~\ref{cor:ombq}.

For rational $\kappa\in(0,1)$ and a positive odd integer $m$, define
\begin{equation}
\label{eq:hardness-polynomial-profile}
H_{\kappa,m}(s)
=\frac{(m+1)s-(1-\kappa)\bigl[1-(1-s)^{m+1}\bigr]}
{m+\kappa},\qquad s\in[0,1],
\end{equation}
and set
\begin{equation}
\label{eq:hardness-polynomial-constants}
B=\frac{m+1}{m+\kappa},\qquad
A=\kappa B,\qquad C=\frac{m(1-\kappa)B}{2}.
\end{equation}

\begin{lemma}[A polynomial hard profile]
\label{lem:hardness-tight-monotone-profile}
The profile~\eqref{eq:hardness-polynomial-profile} is increasing and
convex on $[0,1]$, satisfies
\begin{equation}
\label{eq:hardness-polynomial-bounds}
H(0)=0,\quad H(1)=1,\quad
0\le H(s)\le s,\quad H(s)\le As+Cs^2,
\end{equation}
and has $A\le H'(s)\le B$ and $0\le H''(s)\le2C$ there.
For $\Psi=-H+\delta_{[0,1]}$,
\begin{equation}
\label{eq:hardness-exact-mono-parameter}
\gamma^*_{\op{mono}}(\Psi)=\kappa.
\end{equation}
Moreover, $A\downarrow\kappa$ as $m\to\infty$ through odd integers.
The polynomial $H$ is convex on all of $\mathbb R$.
\end{lemma}

The proof is in Appendix~\ref{apx:hardness-profile}.  To apply
Theorem~\ref{thm:direct-hard-profile-criterion} with $\beta=\gamma$,
fix a rational $\theta\in(\gamma,1)$, choose rational
$\kappa\in(\gamma,\theta)$, and choose an odd integer
\begin{equation}
\label{eq:hardness-degree-choice}
m>\frac{\kappa(1-\theta)}{\theta-\kappa}.
\end{equation}
Since $A=\kappa(m+1)/(m+\kappa)$, the inequality $A<\theta$ is
equivalent to $m(\theta-\kappa)>\kappa(1-\theta)$, so
\eqref{eq:hardness-degree-choice} gives $A<\theta$.  Use
$H_\theta=H_{\kappa,m}$,
$a_\theta=A$, $b_\theta=C$, and
\begin{equation}
\label{eq:hardness-common-normalization}
S_{\theta,p,q,\ell}=1+q(B+2C).
\end{equation}
The lifted losses are $1$--Lipschitz and $1$--smooth on
$\mathcal K_{p,q}$, and their proper extensions have $\kappa$--tilted monotone Clarke subdifferentials, including at the boundary, and
therefore belong to $\mathcal F_{\op{mono}}^\gamma$.

\begin{theorem}[The common hardness threshold]
\label{thm:hardness}
\label{thm:online-hardness-monotone-above-gamma}
Fix $\gamma\in(0,1)$ and $\alpha=\gamma+\varepsilon\le1$, where
$\varepsilon>0$ is fixed.  Consider the classes
\[
\mathcal F_{\op{mono}}^\gamma,\qquad
\mathcal F_{\op{sec}}^\gamma,\qquad
\mathcal F^\gamma.
\]
For any one of them, suppose a randomized polynomial-time algorithm
satisfies either \eqref{eq:hardness-expected-regret-assumption} or
\eqref{eq:hardness-high-prob-regret-assumption} for all sequences of
nonpositive losses with Lipschitz constant at most one.
Then $\NP=\RP$.

The conclusion already holds on products of simplices for
restrictions of rational, $C^\infty$, concave polynomials whose
Hessians have rank at most one.  On the feasible set, each loss is
$[-1,0]$-valued, $1$--Lipschitz, $1$--smooth, and coordinatewise
nonincreasing.  Its degree and number of relevant coordinates are
bounded by constants depending only on $(\gamma,\alpha)$, and one
linear optimization gives an exact minimizer.  The losses may be
independent uniform samples from a polynomial-size family.

For $\mathcal F^\gamma$ alone, the same hardness conclusion holds
even when every loss is a single concave quadratic.
\end{theorem}

The proof is in Appendix~\ref{apx:hardness-application}.  The profile
and its lifted losses verify the common criterion for
$\mathcal F_{\op{mono}}^\gamma$; inclusion
\eqref{eq:hardness-three-class-inclusion} transfers it to the other
two classes.  The final assertion follows from the same criterion
using $h_\theta(s)=\theta s+(1-\theta)s^2$.
All parameters are fixed before the SAT reduction: the polynomial
degree and the Label Cover alphabet and degree bounds may depend on
$(\gamma,\alpha)$, but not on the SAT input size.

\begin{corollary}[Exact coefficients and the endpoint classes]
\label{cor:endpoints}
\label{cor:online-hardness-pseudo-quasi}
Under the oracle assumptions of Corollary~\ref{cor:ombq}, the online
coefficient of each of the three classes is $\gamma$, unless $\NP=\RP$.
For pseudoconvex and for quasiconvex nonpositive smooth losses, no
positive coefficient is achievable unless $\NP=\RP$.
\end{corollary}

For the first claim, Corollary~\ref{cor:ombq} achieves the coefficient $\gamma$ on
$\mathcal F^\gamma$, and by \eqref{eq:hardness-three-class-inclusion}
the same algorithm applies to the two subclasses, while
Theorem~\ref{thm:hardness} excludes every larger coefficient on each of
the three classes.  The products of simplices used by the reduction meet the
domain assumptions of Section~\ref{sec:linearization}: projection
decomposes over the blocks, and the diameter is at most $\sqrt{pq}$
(Appendix~\ref{apx:hardness-application}).  For the second claim, given any proposed positive coefficient on
pseudoconvex or quasiconvex losses, choose $0<\gamma<\alpha$; the
constructed losses also belong to these endpoint classes, as verified
directly in Appendix~\ref{apx:hardness-application}.  All lower bounds
are for fixed $(\gamma,\alpha)$; no uniform polynomial dependence on
$1/(\alpha-\gamma)$ is asserted.  At $\gamma=1$ there is no coefficient
in $(\gamma,1]$ to exclude.

\section{Conclusion}
For every fixed, effectively specified $0<\gamma<1$, the three formulations define distinct classes with the same optimal polynomial-time online coefficient: an online algorithm on the largest class achieves coefficient $\gamma$, and the hardness result on the smallest class rules out every larger coefficient unless $\NP=\RP$, already for smooth losses, whereas known methods for constrained quasar-convex functions minimize smooth members to any accuracy. The smooth support inequality is the equal-parameter tilted-convex condition, and its proper lower semicontinuous extension has convex effective domains, local regularity on relative interiors, and equivalent formulations through different generalized gradients. The secant class is exactly the class of proper $\mathcal H_{\sqrt\gamma}$--convex functions, and the examples show that the parameter losses in the inclusions are sharp and that none of the classes is closed under addition.

The online algorithm uses polynomially many queries to each revealed loss, and whether coefficient $\gamma$ can be achieved with one gradient query per round, or with only a value at the played point, remains open. Offline, it is open whether the accelerated rate of~\cite{martinezrubio2025smooth} is optimal on the three classes, since the lower bound for quasar-convex functions~\citep{hinder2020near} need not apply to them.

\section*{Use of AI Tools}
ChatGPT (OpenAI) assisted with checking the proofs, reviewing the manuscript, and proposing  revisions. Claude Opus (Anthropic) assisted with exposition. The authors take full responsibility for the statements, proofs, references, and other content of the submitted paper.

\clearpage
\appendix

\section{Useful lemmas}
\label{apx:useful}

We use the following chain rule for subdifferentials; see, e.g., Theorem~10.6 in~\cite{rockafellar1998variational}.
\begin{theorem}
\label{thm:chain-rule}
Let $\mathcal X \subseteq \mathbb R^m$.
Suppose $g = f(F(\x))$ for a proper lsc function $f : \mathcal X \to \overline{\mathbb R}$ and a continuously differentiable function $F : \mathbb R^n \to \mathbb R^m$.
Then, at any point $\x \in \op{dom}(g) = F^{-1}(\op{dom}(f))$, we have
\[
\hat{\partial} g(\x) \supseteq \nabla F(\x)^\top \hat{\partial} f(F(\x)).
\]
Moreover, if $f$ is locally Lipschitz
\footnote{
The statement of Theorem~10.6 in~\cite{rockafellar1998variational} uses the assumption that the only vector $\v \in \partial^\infty f(F(\x))$ with $\nabla F(\x)^\top \v = 0$ is $\v = 0$.
However, if $f$ is locally Lipschitz at $F(\x)$, then $\partial^\infty f(F(\x)) = \{ 0 \}$ (e.g., see Theorem~9.13 in the same book) and this condition is trivially satisfied.
}
at $F(\x)$, then
\[
\partial g(\x) \subseteq \nabla F(\x)^{\top} \partial f(F(\x)).
\]
\end{theorem}

A common Lipschitz constant turns pointwise convergence on a dense set into uniform convergence on bounded sets; see, e.g., Proposition~9.59 in~\cite{rockafellar1998variational}.
\begin{theorem}\label{thm:lipschitz-limit}
Let $\mathcal X \subseteq \mathbb R^m$.
Suppose that the functions $f_k : \mathcal X \to \mathbb R$, $k\ge1$, are Lipschitz continuous with the same constant $L \geq 0$ and converge pointwise on a dense subset $\mathcal D \subseteq \mathcal X$.
Then they converge uniformly on bounded subsets of $\mathcal X$ to a function $f : \mathcal X \to \mathbb R$ which is Lipschitz continuous on $\mathcal X$ with the same constant $L$.
Moreover, for any point $\mathbf x \in \op{int}(\mathcal X)$ and $\mathbf v \in \partial f(\mathbf x)$, there are sequences $\mathbf x_k \to \mathbf \x$ and $\mathbf v_k \in \partial f_k(\mathbf x_k)$ such that $\mathbf v_k \to \mathbf v$.
\end{theorem}

We also use Lebourg's mean value theorem for Lipschitz functions; see, e.g.,~Theorem~2.3.7 in~\cite{clarke90_optim_nonsm_analy}.

\begin{theorem}[Lebourg's mean value theorem for Lipschitz functions]
\label{thm:Lipschitz-mean-value}
Let $\mathcal K \subseteq \mathbb R^d$ be convex, let $\x,\y\in\mathcal K$ be distinct, and let $f : \mathcal K \to \mathbb R$ be Lipschitz on an open neighborhood of the line segment $[\x,\y]$.
Then there is a point $\z \in (\x, \y)$ and $\vv \in \overline{\partial} f(\z) = \op{conv}(\partial f(\z))$ such that
\footnote{
See Proposition~\ref{prop:subdiff-basics} for the equality $\overline{\partial} f(\z) = \op{conv}(\partial f(\z))$.
}
\[
f(\y) - f(\x) = \langle \vv, \y - \x \rangle.
\]
\end{theorem}

Lebourg's theorem extends to the case where $f$ is only continuous at the end-points.

\begin{corollary}\label{cor:Lipschitz-interior-mean-value}
Let $\mathcal K \subseteq \mathbb R^d$ be convex, let $\x,\y\in\mathcal K$ be distinct, and let $f : \mathcal K \to \mathbb R$ be continuous on $[\x,\y]$ and locally Lipschitz on an open set containing $(\x,\y)$.
Then there is a point $\z \in (\x, \y)$ and $\vv \in \overline{\partial} f(\z) = \op{conv}(\partial f(\z))$ such that
\[
f(\y) - f(\x) = \langle \vv, \y - \x \rangle.
\]
\end{corollary}
\begin{proof}
For $\lambda \in [0,1]$, let $\z_\lambda := (1-\lambda) \x + \lambda \y$.
Define the function $\phi : [0,1] \to \mathbb R$ by
\[
\phi(\lambda) := f(\z_\lambda) - (1-\lambda) f(\x) -\lambda f(\y).
\]
Then $\phi$ is continuous because $f$ is continuous on $[\x,\y]$, and $\phi(0) = \phi(1) = 0$.
We claim that there are real numbers $a,b$ with
\begin{equation}
\label{eq:pb-phi-equal}
0 < a < b < 1,
\qquad
\phi(a) = \phi(b).
\end{equation}
If $\phi \equiv 0$, any $0<a<b<1$ satisfies \eqref{eq:pb-phi-equal}.
Otherwise, choose $c \in (0,1)$ with $\phi(c) \neq 0$.
The value $\phi(c)/2$ lies strictly between $\phi(0)=0$ and $\phi(c)$, and also strictly between $\phi(c)$ and $\phi(1)=0$.
By the intermediate value theorem there are $a \in (0,c)$ and $b \in (c,1)$ with $\phi(a) = \phi(b) = \phi(c)/2$, which gives \eqref{eq:pb-phi-equal}.

Writing out $\phi(a) = \phi(b)$ gives
\[
f(\z_a) - (1-a) f(\x) - a f(\y)
=
f(\z_b) - (1-b) f(\x) - b f(\y).
\]
Rearranging, $f(\z_b) - f(\z_a) = (b-a)\bigl(f(\y) - f(\x)\bigr)$, that is,
\begin{equation}
\label{eq:pb-secant-ab}
f(\y) - f(\x)
=
\frac{f(\z_b) - f(\z_a)}{b-a}.
\end{equation}
Because $[\z_a,\z_b]$ is compact and lies in $(\x,\y)$, there is $r>0$ such that its closed $r$--neighborhood $C_r$ lies in the open set on which $f$ is locally Lipschitz. The set $C_r$ is compact and convex, so covering it by finitely many balls on which $f$ is Lipschitz and splitting each segment in $C_r$ at the boundaries of these balls shows that $f$ is Lipschitz on $C_r$, and hence on the open $r$--neighborhood of $[\z_a,\z_b]$.
Lebourg's theorem (Theorem~\ref{thm:Lipschitz-mean-value}) applied to $f$ on $[\z_a,\z_b]$ therefore gives $\z \in (\z_a, \z_b) \subseteq (\x, \y)$ and $\vv \in \op{conv}(\partial f(\z))$ such that
\begin{equation}
\label{eq:pb-lebourg-ab}
f(\z_b) - f(\z_a) = \langle \vv, \z_b - \z_a \rangle.
\end{equation}
By definition of $\z_\lambda$, we have $\z_b - \z_a = (b-a)(\y - \x)$.
Substituting \eqref{eq:pb-lebourg-ab} into \eqref{eq:pb-secant-ab} gives
\[
f(\y) - f(\x)
=
\frac{f(\z_b) - f(\z_a)}{b-a}
=
\left\langle \vv, \frac{\z_b - \z_a}{b-a} \right\rangle
=
\langle \vv, \y - \x \rangle.
\qedhere
\]
\end{proof}

For general lsc functions, an approximate mean-value theorem holds~\citep{mordukhovich2006variational1}; we include a finite-dimensional proof.

\begin{theorem}[Approximate mean values for lsc functions]
\label{thm:approx-mean-value}
Let $f : \mathbb R^d \to \overline{\mathbb R}$ be a proper lsc function.
Let $\mathbf x \neq \mathbf y$ in $\C{X} := \op{dom}(f)$ and, for $\lambda \in [0,1]$, let $\mathbf z_\lambda = \lambda \mathbf x + (1-\lambda)\mathbf y$.
Then there is $\lambda\in[0,1)$ with $\mathbf z_\lambda\in\op{dom}(f)$ and sequences $\mathbf x_k \xrightarrow{f} \mathbf z_\lambda$ and $\mathbf v_k \in \hat{\partial} f(\mathbf x_k)$ such that
\[
\frac{1}{1-\lambda} \limsup_{k \to \infty} \langle \mathbf v_k, \mathbf x_k-\mathbf x \rangle 
\le 
f(\mathbf y) - f(\mathbf x),
\]
and
\[
\limsup_{k \to \infty} \langle \mathbf v_k, \mathbf y - \mathbf x \rangle 
\le 
f(\mathbf y) - f(\mathbf x).
\]
Similarly, there is $\lambda'\in(0,1]$ with $\mathbf z_{\lambda'}\in\op{dom}(f)$ and sequences $\mathbf x'_k \xrightarrow{f} \mathbf z_{\lambda'}$ and $\mathbf v'_k \in \hat{\partial} f(\mathbf x'_k)$ such that
\[
f(\mathbf y) - f(\mathbf x)
\leq
\frac{1}{\lambda'} \liminf_{k \to \infty} \langle \mathbf v'_k, \mathbf y - \mathbf x'_k \rangle,
\]
and
\[
f(\mathbf y) - f(\mathbf x)
\le
\liminf_{k \to \infty} \langle \mathbf v'_k, \mathbf y - \mathbf x \rangle.
\]
\end{theorem}
\begin{proof}
Put $\dd:=\y-\x$, $S:=[\x,\y]$, and
$q:=(f(\y)-f(\x))/\|\dd\|^2$. Define
\[
\varphi(\uu):=f(\uu)-f(\x)-q\langle\dd,\uu-\x\rangle,
\]
so that $\varphi(\x)=\varphi(\y)=0$.
The function $\varphi$ is proper and lsc. It therefore has a finite
minimum $m\le0$ on the compact segment $S$.
For $\epsilon>0$, write
\[
\varphi_\epsilon(\uu):=\varphi(\uu)
                    -\epsilon\langle\dd,\uu-\x\rangle.
\]
For all sufficiently small $\epsilon$, the minimizers of
$\varphi_\epsilon$ on $S$ stay a fixed positive distance from $\x$.
Indeed, if $m<0$, lower semicontinuity at $\x$ gives a neighborhood
where $\varphi>m/2$, while
$\min_S\varphi_\epsilon\le m$; the affine perturbation is uniformly
smaller than $-m/4$ on $S$ for small enough $\epsilon$.
If $m=0$, the unique minimizer of $\varphi_\epsilon$ on $S$ is $\y$,
since $\varphi\ge0$ there and
$\langle\dd,\uu-\x\rangle\le\|\dd\|^2$, with equality only at $\y$.

Let $C:=\{\uu:\operatorname{dist}(\uu,S)\le1\}$, a compact neighborhood
of $S$. For fixed $\epsilon>0$ and each integer $n\ge1$, choose a minimizer
$\uu_{\epsilon,n}$ over $C$ of
\[
\varphi_\epsilon(\uu)+n\operatorname{dist}(\uu,S)^2.
\]
These minimizers exist because the objective is lsc, bounded below on
$C$, and finite at $\y$. Comparison with a minimizer on $S$ shows
$\operatorname{dist}(\uu_{\epsilon,n},S)\to0$ as $n\to\infty$.
Every cluster point minimizes $\varphi_\epsilon$ on $S$: lower
semicontinuity and the nonnegative penalty give this after passing to
the limit in the same comparison.

Choose $\epsilon_k\downarrow0$. By the preceding two paragraphs,
we can choose $n_k\to\infty$ so that
$\uu_k:=\uu_{\epsilon_k,n_k}$ lies in the interior of $C$,
$\operatorname{dist}(\uu_k,S)\to0$, and $\uu_k$ stays a fixed positive
distance from $\x$. Passing to a subsequence, assume
$\uu_k\to\z\in S\setminus\{\x\}$.
The minimizing inequality and $\min_S\varphi_{\epsilon_k}\le m$ imply
\[
\varphi(\uu_k)
\le m+\epsilon_k\langle\dd,\uu_k-\x\rangle.
\]
Since $\varphi(\z)\ge m$ and $\varphi$ is lsc, it follows that
$\varphi(\uu_k)\to\varphi(\z)=m$ and hence $\uu_k\xrightarrow{f}\z$.

Let $\mathbf p_k:=\Pi_S(\uu_k)$. For large $k$, $\mathbf p_k\ne\x$.
The squared distance to a nonempty closed convex set is continuously
differentiable, with gradient $2(\uu-\Pi_S(\uu))$.
The necessary condition for the local minimum at $\uu_k$, followed by
the smooth-sum rule in Proposition~\ref{prop:subdiff-basics}(v), gives
\[
\mathbf r_k:=\epsilon_k\dd-2n_k(\uu_k-\mathbf p_k)
\in\hat\partial\varphi(\uu_k).
\]
Projection onto the segment gives
\[
\langle\uu_k-\mathbf p_k,\dd\rangle\ge0,
\qquad
\langle\uu_k-\mathbf p_k,\mathbf p_k-\x\rangle\ge0.
\]
Both inner products vanish if $\mathbf p_k$ lies in the open segment; if
$\mathbf p_k=\y$, they are nonnegative by the projection inequality.
Consequently,
\[
\langle\mathbf r_k,\dd\rangle\le\epsilon_k\|\dd\|^2,
\qquad
\langle\mathbf r_k,\uu_k-\x\rangle
\le\epsilon_k\langle\dd,\uu_k-\x\rangle.
\]
Writing $\z=\lambda\x+(1-\lambda)\y$ with $\lambda\in[0,1)$ and setting
$\mathbf v_k:=\mathbf r_k+q\dd\in\hat\partial f(\uu_k)$ therefore gives
\[
\limsup_k\langle\mathbf v_k,\dd\rangle
\le f(\y)-f(\x),
\qquad
\limsup_k\langle\mathbf v_k,\uu_k-\x\rangle
\le(1-\lambda)(f(\y)-f(\x)).
\]
Taking $\mathbf x_k:=\uu_k$ proves the first pair of inequalities.
Interchanging $\x$ and $\y$ proves the second pair.
\end{proof}
The $\limsup$ and $\liminf$ versions are equivalent: each follows from the other by swapping $\x$ and $\y$.

\section{Approximate \texorpdfstring{$\gamma$}{gamma}--tilted convexity}
\label{apx:approx}

This appendix studies a weaker notion than $\gamma$--tilted convexity. Its results use only Section~\ref{sec:background}, and the proofs in Appendix~\ref{apx:structure} rely on them: those of Theorems~\ref{thm:gamma-MW-convex-is-locally-Lipschitz}, \ref{thm:gamma-MW-continuous-on-lines}, \ref{thm:mwc-alternative-characterizations}, \ref{thm:gamma-MW-convex-on-lines}, \ref{thm:gamma-MW-convex-pointwise-limit}, \ref{thm:gamma-MW-monotone-on-lines}, \ref{thm:gamma-MW-monotone-pointwise-limit}, and~\ref{thm:gamma-MW-Jensen-equivalent-def}, and through them the rest of Section~\ref{sec:structure}. Section~\ref{subsec:regularity} summarizes the notion and its role in the main text.

\begin{definition}[Approximate $\gamma$--tilted convexity]
\label{def:approx-mwc}
Let $\gamma\in(0,1]$, let $f : \mathbb R^d \to \overline{\mathbb R}$ be a proper lsc function, and let
$F:\mathbb R^d \rightrightarrows \mathbb R^d$ be a multivalued operator.
We say that $f$ is \emph{approximately $\gamma$--tilted convex with respect to
$F$} if, for every $\x \in \op{dom}(f) \cap \op{dom}(F)$ and every $\epsilon>0$,
there exists $\v\in F(\x)$ such that
\[
  f(\y)-f(\x)
  \geq\psi_\gamma(\langle\y-\x,\v\rangle)-\epsilon
  \qquad\forall\y \in \op{dom}(f).
\]
\end{definition}

Here the operator $F$ need not be a
subdifferential, and the supporting vector may depend on $\epsilon$; in
particular, exact $\gamma$--tilted convexity with respect to $F$ implies
approximate $\gamma$--tilted convexity with respect to $F$. If the domain is
convex and approximate supports exist on a dense subset of it, then the function is locally
Lipschitz on the relative interior of its domain and continuous along every line meeting that relative interior.

\subsection{Local boundedness}

Approximately $\gamma$--tilted convex functions are bounded
above on finitely generated subsets of their domain.

\begin{lemma}
\label{lem:approximate-gamma-supports-locally-bounded}
Let $\gamma\in(0,1]$, let
$f:\mathbb R^d\to\overline{\mathbb R}$ be proper and lsc, and set
$\mathcal K:=\op{dom}(f)$.  Assume that $\mathcal K$ is convex.  Let
$F:\mathbb R^d\rightrightarrows\mathbb R^d$ be such that
$\mathcal K\cap\op{dom}(F)$ is dense in $\mathcal K$, and suppose that $f$
is approximately $\gamma$--tilted convex with respect to $F$ in the sense of
Definition~\ref{def:approx-mwc}.
If $\y_1,\ldots,\y_m\in\mathcal K$ satisfy
\[
  \operatorname{aff}\{\y_1,\ldots,\y_m\}
  =\operatorname{aff}(\mathcal K),
\]
then
\[
  f(\x)\leq\max_{1\leq i\leq m}f(\y_i)
  \qquad
  \forall\x\in\operatorname{conv}\{\y_1,\ldots,\y_m\}.
\]
\end{lemma}

\begin{proof}
Let
\[
  C:=\operatorname{conv}\{\y_1,\ldots,\y_m\},
  \qquad
  M^+:=\max_{1\leq i\leq m}f(\y_i).
\]
Each $\y_i$ lies in $\mathcal K$, so $M^+\in\mathbb R$.  Convexity of
$\mathcal K$ gives $C\subseteq\mathcal K$.  We first prove that
\begin{equation}
  f(\x)\le M^+
  \qquad
  \forall\x\in\mathcal K\cap\operatorname{dom}(F)\cap\operatorname{ri}(C).
  \label{eq:pc1-bdd-dense-points}
\end{equation}
Fix such an $\x$ and write
$\x=\sum_{i=1}^m\lambda_i\y_i$, where $\lambda_i\geq0$ and
$\sum_i\lambda_i=1$.  Suppose that $f(\x)>M^+$.  Choose
$0<\epsilon<f(\x)-M^+$.  Definition~\ref{def:approx-mwc}, applied at
$\x\in\op{dom}(f)\cap\op{dom}(F)$, gives a vector $\v\in F(\x)$ with
$f(\y)-f(\x)\ge\psi_\gamma(\langle\y-\x,\v\rangle)-\epsilon$ for all
$\y\in\mathcal K$.  Taking $\y=\y_i$ and using
$f(\y_i)\le M^+<f(\x)-\epsilon$ gives
\begin{equation}
  \psi_\gamma(\langle\y_i-\x,\v\rangle)
  \leq f(\y_i)-f(\x)+\epsilon<0
  \qquad (1\leq i\leq m).
  \label{eq:pc1-bdd-psi}
\end{equation}
Applying the increasing inverse $\overline{\psi}_\gamma$
to~\eqref{eq:pc1-bdd-psi} and using
$\overline{\psi}_\gamma(t)=\gamma t$ for $t<0$, we obtain
\begin{equation}
  \langle\y_i-\x,\v\rangle
  \leq\gamma\bigl(f(\y_i)-f(\x)+\epsilon\bigr)
  \qquad (1\leq i\leq m).
  \label{eq:pc1-bdd-linear}
\end{equation}
Multiply~\eqref{eq:pc1-bdd-linear} by $\lambda_i$ and sum over $i$.
The left-hand sides add up to
$\langle\sum_i\lambda_i\y_i-\x,\v\rangle=0$, and using
$\sum_i\lambda_i=1$ and $f(\y_i)\le M^+$ on the right gives
\[
  0
  \leq\gamma\left(\sum_{i=1}^m\lambda_i f(\y_i)-f(\x)+\epsilon\right)
  \leq\gamma\bigl(M^+-f(\x)+\epsilon\bigr)<0,
\]
where the last inequality is the choice of $\epsilon$.  This
contradiction proves~\eqref{eq:pc1-bdd-dense-points}.

Next we show that
$D:=\mathcal K\cap\operatorname{dom}(F)\cap\operatorname{ri}(C)$ is dense
in $C$.  The set $C$ is nonempty and convex, so $\operatorname{ri}(C)$
is dense in $C$ (Theorem~6.3 in~\cite{rockafellar15_convex_analy}).  It
therefore suffices to show that every $\mathbf r\in\operatorname{ri}(C)$
lies in the closure of $D$.  By the definition of the relative interior,
there is $\delta>0$ with
$\mathbb B(\mathbf r,\delta)\cap\operatorname{aff}(C)\subseteq C$.  If
$\w\in\mathbb B(\mathbf r,\delta/2)\cap\operatorname{aff}(C)$, then
$\mathbb B(\w,\delta/2)\cap\operatorname{aff}(C)\subseteq C$, so
$\w\in\operatorname{ri}(C)$.  By hypothesis,
$\operatorname{aff}(C)=\operatorname{aff}(\mathcal K)\supseteq\mathcal K$,
and hence
\begin{equation}
  \mathbb B(\mathbf r,\delta/2)\cap\mathcal K
  \subseteq\operatorname{ri}(C).
  \label{eq:pc1-bdd-ri}
\end{equation}
Let $0<\eta<\delta/2$.  Since $\mathbf r\in C\subseteq\mathcal K$ and
$\mathcal K\cap\op{dom}(F)$ is dense in $\mathcal K$, the ball
$\mathbb B(\mathbf r,\eta)$ contains a point of
$\mathcal K\cap\op{dom}(F)$.  By~\eqref{eq:pc1-bdd-ri}, this point lies in
$D$.  As $\eta$ was arbitrary, $\mathbf r$ lies in the closure of $D$.

Finally, the sublevel set $\{\y\in\mathbb R^d:f(\y)\leq M^+\}$ is
closed by lower semicontinuity.  By~\eqref{eq:pc1-bdd-dense-points} it
contains $D$, so it contains the closure of $D$, which contains $C$.
\end{proof}

\subsection{Local Lipschitz continuity}

The local Lipschitz constant depends only on a local bound for $f$ and on the local geometry of its effective domain.

\begin{theorem}[Local Lipschitz regularity from approximate $\gamma$--tilted convexity]
\label{thm:approximate-gamma-supports-local-Lipschitz}
Let $\gamma\in(0,1]$, let
$f:\mathbb R^d\to\overline{\mathbb R}$ be proper and lsc, and set
$\mathcal K:=\op{dom}(f)$.  Assume that $\mathcal K$ is convex.  Let
$F:\mathbb R^d\rightrightarrows\mathbb R^d$ be such that
$\mathcal K\cap\op{dom}(F)$ is dense in $\mathcal K$, and suppose that $f$
is approximately $\gamma$--tilted convex with respect to $F$ in the sense of
Definition~\ref{def:approx-mwc}.
Then $f$ is locally Lipschitz on $\operatorname{ri}(\mathcal K)$.
More precisely, put $A:=\operatorname{aff}(\mathcal K)$. If
$\x_0\in\operatorname{ri}(\mathcal K)$ and $R>0$ satisfy
$\overline{\mathbb B(\x_0,3R)}\cap A\subseteq\mathcal K$,
then
\[
        M
        :=\sup\left\{|f(\x)|:
        \x\in\overline{\mathbb B(\x_0,2R)}\cap A\right\}
        <\infty,
    \]
    and
    \begin{equation}
        |f(\x)-f(\y)|
        \leq \frac{2M}{\gamma^2R}\|\x-\y\|
  \qquad
  \forall \x,\y\in\mathbb B(\x_0,R)\cap A.
  \label{eq:approximate-gamma-support-Lipschitz-bound}
\end{equation}
The translation-invariant form is
\begin{equation}
|f(\x)-f(\y)|\le\frac{\Delta}{\gamma^2R}\|\x-\y\|,
\qquad \x,\y\in\mathbb B(\x_0,R)\cap A,
\label{eq:approximate-gamma-support-oscillation-bound}
\end{equation}
where
\[
\Delta:=\sup_{\overline{\mathbb B(\x_0,2R)}\cap A}f
       -\inf_{\overline{\mathbb B(\x_0,2R)}\cap A}f.
\]
\end{theorem}

\begin{proof}
    If $\dim A=0$, then $A=\{\x_0\}$, so $M=|f(\x_0)|<\infty$
    and~\eqref{eq:approximate-gamma-support-Lipschitz-bound} only
    concerns $\x=\y=\x_0$.  Suppose therefore that $k:=\dim A\geq1$.
    Let $V:=A-A$, a linear subspace of dimension $k$, and fix an
    orthonormal basis $\mathbf e_1,\ldots,\mathbf e_k$ of $V$.

    First we verify that $f$ is bounded on
    $\overline{\mathbb B(\x_0,2R)}\cap A$. Choose finitely many points
    $\y_1,\ldots,\y_m\in\overline{\mathbb B(\x_0,3R)}\cap A$ such that the
    polytope
    \[
        C:=\operatorname{conv}\{\y_1,\ldots,\y_m\}
    \]
    has affine hull $A$ and contains
    $\overline{\mathbb B(\x_0,2R)}\cap A$; we build them from cubes.  For
    $\mathbf p\in A$, consider the cube
    \begin{equation}
      Q_{\mathbf p}
      :=\Bigl\{\mathbf p+\sum_{i=1}^k s_i\mathbf e_i:
      |s_i|\le R/\sqrt k\ (1\le i\le k)\Bigr\}\subseteq A.
      \label{eq:pc1-lip-cube}
    \end{equation}
    It is the convex hull of its $2^k$ vertices
    $\mathbf p+(R/\sqrt k)\sum_i\sigma_i\mathbf e_i$ with
    $\sigma_i\in\{-1,1\}$.  By orthonormality, each vertex has distance
    exactly $R$ from $\mathbf p$.  The vertices have affine hull $A$,
    because the difference of two vertices that differ only in
    $\sigma_i$ is $\pm(2R/\sqrt k)\mathbf e_i$.  The cube
    $Q_{\mathbf p}$ contains the set
    $O_{\mathbf p}:=\{\x\in A:|\langle\x-\mathbf p,\mathbf e_i\rangle|<R/\sqrt k
    \ (1\le i\le k)\}$, which is open relative to $A$ and contains
    $\mathbf p$.  The sets $O_{\mathbf p}$ with
    $\mathbf p\in\overline{\mathbb B(\x_0,2R)}\cap A$ cover the compact
    set $\overline{\mathbb B(\x_0,2R)}\cap A$, so finitely many of them,
    say for $\mathbf p_1,\ldots,\mathbf p_N$, already cover it.  Let
    $\y_1,\ldots,\y_m$ be the vertices of
    $Q_{\mathbf p_1},\ldots,Q_{\mathbf p_N}$.  Every such vertex $\y$ of
    $Q_{\mathbf p_j}$ lies in $A$ and satisfies
    $\|\y-\x_0\|\le\|\y-\mathbf p_j\|+\|\mathbf p_j-\x_0\|\le R+2R=3R$.
    The hull $C$ contains every $Q_{\mathbf p_j}$, and hence
    $\overline{\mathbb B(\x_0,2R)}\cap A$.  Its affine hull lies in $A$
    and contains the affine hull $A$ of the vertices of
    $Q_{\mathbf p_1}$.
    Since $\y_i\in\overline{\mathbb B(\x_0,3R)}\cap A\subseteq\mathcal K$
    and $\operatorname{aff}\{\y_1,\ldots,\y_m\}=A=\operatorname{aff}(\mathcal K)$,
    Lemma~\ref{lem:approximate-gamma-supports-locally-bounded} gives
    \[
      f(\x)\leq\max_{1\leq i\leq m}f(\y_i)
      \qquad\forall\x\in C.
    \]
    Thus $f$ is bounded above on
    $\overline{\mathbb B(\x_0,2R)}\cap A$. On the other hand, this set is
    compact and contained in $\mathcal K$, and $f$ is finite and lsc on
    $\mathcal K$, so $f$ attains a finite minimum on it and is
    bounded below there. Consequently, $M<\infty$.

    Next, fix
    $\x\in\operatorname{dom}(F)\cap\mathbb B(\x_0,R)\cap A$, and let
    $\epsilon>0$.  The point $\x$ lies in $\mathcal K$.  Choose
    $\v\in F(\x)$ as in Definition~\ref{def:approx-mwc}, so that
    \begin{equation}
      f(\y)-f(\x)\ge\psi_\gamma(\langle\y-\x,\v\rangle)-\epsilon
      \qquad\forall\y\in\mathcal K.
      \label{eq:pc1-lip-approx}
    \end{equation}
    For $\y\in\mathcal K\subseteq A$ we have $\y-\x\in V$.  Hence
    replacing $\v$ by its orthogonal projection onto $V$ does not change
    any inner product in~\eqref{eq:pc1-lip-approx}, and we may assume
    $\v\in V$.  If $\v\neq0$, then let
    \[
        \w:=\x+R\frac{\v}{\|\v\|}
        \in\overline{\mathbb B(\x_0,2R)}\cap A;
    \]
    indeed $\w\in A$ because $\v\in V$, and
    $\|\w-\x_0\|\le\|\x-\x_0\|+R<2R$.  The point $\x$ also lies in
    $\overline{\mathbb B(\x_0,2R)}\cap A$, so $|f(\w)|,|f(\x)|\le M$.
    Since $\langle\w-\x,\v\rangle=R\|\v\|>0$ and
    $\psi_\gamma(s)=\gamma s$ for $s>0$, applying
    \eqref{eq:pc1-lip-approx} with $\y=\w$ yields
    \[
        2M
        \geq f(\w)-f(\x)
        \geq\psi_\gamma(R\|\v\|)-\epsilon
        =\gamma R\|\v\|-\epsilon.
    \]
    The same bound on $\|\v\|$ is automatic when $\v=0$, and therefore
    \begin{equation}
        \|\v\|\leq\frac{2M+\epsilon}{\gamma R}.
        \label{eq:approximate-support-vector-bound}
    \end{equation}
    We also use the elementary estimate
    \begin{equation}
      \psi_\gamma(s)\geq-\gamma^{-1}|s|
      \qquad\forall s\in\mathbb R,
      \label{eq:pc1-lip-psi-lower}
    \end{equation}
    which holds because $\psi_\gamma(s)=\gamma s\ge0$ for $s\ge0$ and
    $\psi_\gamma(s)=\gamma^{-1}s=-\gamma^{-1}|s|$ for $s<0$.
    Let $\y\in\mathbb B(\x_0,R)\cap A$.  Then $\y\in\mathcal K$.  By
    \eqref{eq:pc1-lip-approx}, \eqref{eq:pc1-lip-psi-lower}, the
    Cauchy--Schwarz inequality, and
    \eqref{eq:approximate-support-vector-bound},
    \[
        f(\y)-f(\x)
        \geq-\frac{1}{\gamma}\|\v\|\,\|\y-\x\|-\epsilon
        \geq-\frac{2M+\epsilon}{\gamma^2R}\|\y-\x\|-\epsilon.
    \]
    The right-hand side no longer depends on $\v$.  Letting
    $\epsilon\downarrow0$, we conclude that
    \begin{equation}
        f(\y)-f(\x)
        \geq-\frac{2M}{\gamma^2R}\|\y-\x\|
        \label{eq:dense-basepoint-one-sided-bound}
    \end{equation}
    whenever
    $\x\in\operatorname{dom}(F)\cap\mathbb B(\x_0,R)\cap A$ and
    $\y\in\mathbb B(\x_0,R)\cap A$.

    Finally, take arbitrary $\x,\y\in\mathbb B(\x_0,R)\cap A$.  Because
    $\mathbb B(\x_0,R)\cap A\subseteq\mathcal K\subseteq A$, we have
    $\mathbb B(\x_0,R)\cap\mathcal K=\mathbb B(\x_0,R)\cap A$.  For each
    $n$, the open set $\mathbb B(\x,1/n)\cap\mathbb B(\x_0,R)$ contains
    $\x\in\mathcal K$, so by density it contains a point $\x_n$ of
    $\mathcal K\cap\op{dom}(F)$.  Then
    $\x_n\in\operatorname{dom}(F)\cap\mathbb B(\x_0,R)\cap A$ and
    $\x_n\to\x$.  By lower semicontinuity,
    $f(\x)\le\liminf_n f(\x_n)$.  Applying
    \eqref{eq:dense-basepoint-one-sided-bound} to $\x_n$ and $\y$ gives
    \[
        f(\y)-f(\x)
        \geq f(\y)-\liminf_{n\to\infty}f(\x_n)
        =\limsup_{n\to\infty}\bigl(f(\y)-f(\x_n)\bigr)
        \geq-\frac{2M}{\gamma^2R}\lim_{n\to\infty}\|\y-\x_n\|
        =-\frac{2M}{\gamma^2R}\|\y-\x\|.
    \]
    Interchanging $\x$ and $\y$ proves
    \eqref{eq:approximate-gamma-support-Lipschitz-bound}.

    For every $\x_0\in\operatorname{ri}(\mathcal K)$ there is
    $\varepsilon>0$ with
    $\mathbb B(\x_0,\varepsilon)\cap A\subseteq\mathcal K$, and
    $R:=\varepsilon/4$ satisfies
    $\overline{\mathbb B(\x_0,3R)}\cap A\subseteq\mathcal K$.  So
    \eqref{eq:approximate-gamma-support-Lipschitz-bound} holds on a
    neighborhood of $\x_0$ relative to $A$.  Since $\x_0$ was
    arbitrary, $f$ is locally Lipschitz on $\operatorname{ri}(\mathcal K)$.

    Approximate tilted convexity is unchanged by adding a constant,
    so we may apply~\eqref{eq:approximate-gamma-support-Lipschitz-bound} to $f-c$, where
    $c$ is the midpoint of the supremum and infimum of $f$ on
    $\overline{\mathbb B(\x_0,2R)}\cap A$. The corresponding absolute-value
    bound is $\Delta/2$, which proves~\eqref{eq:approximate-gamma-support-oscillation-bound}.
\end{proof}

\subsection{Dense approximate \texorpdfstring{$\gamma$}{gamma}--tilted convexity implies exact \texorpdfstring{$\gamma$}{gamma}--tilted convexity over the relative interior}

\begin{proposition}
\label{prop:dense-approx-mwc-relative-interior}
Let $\gamma\in(0,1]$, let
$f:\mathbb R^d\to\overline{\mathbb R}$ be proper and lsc, and set
$\mathcal K:=\op{dom}(f)$.  Assume that $\mathcal K$ is convex.  Let
$F:\mathbb R^d\rightrightarrows\mathbb R^d$ be such that
$\mathcal K\cap\op{dom}(F)$ is dense in $\mathcal K$, and suppose that $f$
is approximately $\gamma$--tilted convex with respect to $F$ in the sense of
Definition~\ref{def:approx-mwc}.  Then there is an operator
$\widetilde F:\mathbb R^d\rightrightarrows\mathbb R^d$ with
\[
  \op{dom}(\widetilde F)=\operatorname{ri}(\mathcal K)
\]
such that $f$ is $\gamma$--tilted convex with respect to $\widetilde F$.
\end{proposition}

\begin{proof}
Put $A:=\operatorname{aff}(\mathcal K)$ and $V:=A-A$. By
Theorem~\ref{thm:approximate-gamma-supports-local-Lipschitz}, $f$ is locally
Lipschitz, and hence continuous, on $\operatorname{ri}(\mathcal K)$.

Fix $\z\in\operatorname{ri}(\mathcal K)$, and choose $R>0$ such that
$\overline{\mathbb B(\z,3R)}\cap A\subseteq\mathcal K$; this is possible
by the definition of the relative interior.  For each $j$, the open ball
$\mathbb B(\z,\min\{R,1/j\})$ contains $\z\in\mathcal K$.  Since
$\mathcal K\cap\op{dom}(F)$ is dense in $\mathcal K$, this ball contains
a point $\z_j\in\mathcal K\cap\op{dom}(F)$.  Then $\z_j\to\z$ and
$\z_j\in\mathbb B(\z,R)\cap A$.  Moreover
$\mathbb B(\z_j,2R)\cap A\subseteq\mathbb B(\z,3R)\cap A\subseteq\mathcal K$,
so $\z_j\in\operatorname{ri}(\mathcal K)$.  Choose a sequence
$\epsilon_j\downarrow0$.  Definition~\ref{def:approx-mwc}, applied at
$\z_j$ with $\epsilon_j$, gives vectors $\uu_j\in F(\z_j)$ such that
\[
  f(\y)-f(\z_j)
  \ge\psi_\gamma(\langle\y-\z_j,\uu_j\rangle)-\epsilon_j
  \qquad\forall\y\in\mathcal K.
\]
Let $\v_j$ be the orthogonal projection of $\uu_j$ onto $V$. Since
$\y-\z_j\in V$ for every $\y\in\mathcal K$, we have
$\langle\y-\z_j,\uu_j\rangle=\langle\y-\z_j,\v_j\rangle$, and therefore
\begin{equation}
  f(\y)-f(\z_j)
  \ge\psi_\gamma(\langle\y-\z_j,\v_j\rangle)-\epsilon_j
  \qquad\forall\y\in\mathcal K.
  \label{eq:pc1-dense-approx}
\end{equation}

By Theorem~\ref{thm:approximate-gamma-supports-local-Lipschitz}, applied
with $\x_0:=\z$,
\[
  M:=\sup\{|f(\w)|:\w\in
  \overline{\mathbb B(\z,2R)}\cap A\}<\infty.
\]
If $\v_j\neq0$, then
\[
  \w_j:=\z_j+R\frac{\v_j}{\|\v_j\|}
  \in\overline{\mathbb B(\z,2R)}\cap A,
\]
because $\v_j\in V$ and $\|\w_j-\z\|\le\|\z_j-\z\|+R<2R$.  The point
$\z_j$ lies in the same set, so $f(\w_j)-f(\z_j)\le2M$.  Since
$\langle\w_j-\z_j,\v_j\rangle=R\|\v_j\|>0$ and $\psi_\gamma(s)=\gamma s$
for $s>0$, inequality~\eqref{eq:pc1-dense-approx} with $\y=\w_j$ gives
\[
  2M\geq f(\w_j)-f(\z_j)
  \geq\gamma R\|\v_j\|-\epsilon_j.
\]
Hence $\|\v_j\|\le(2M+\epsilon_j)/(\gamma R)\le(2M+\epsilon_1)/(\gamma R)$.
This bound is automatic when $\v_j=0$, so $(\v_j)$ is bounded.
After passing to a subsequence, $\v_j\to\v_{\z}$ for some
$\v_{\z}\in\mathbb R^d$.

Fix $\y\in\mathcal K$ and let $j\to\infty$ in~\eqref{eq:pc1-dense-approx}
along this subsequence.  Since $f$ is continuous on
$\operatorname{ri}(\mathcal K)$ and $\z_j,\z\in\operatorname{ri}(\mathcal K)$,
we have $f(\z_j)\to f(\z)$.  Also
$\langle\y-\z_j,\v_j\rangle\to\langle\y-\z,\v_{\z}\rangle$, $\psi_\gamma$
is continuous, and $\epsilon_j\to0$.  Therefore
\begin{equation}
  f(\y)-f(\z)
  \geq\psi_\gamma(\langle\y-\z,\v_{\z}\rangle)
  \qquad\forall\y\in\mathcal K.
  \label{eq:pc1-dense-exact}
\end{equation}

Define $\widetilde F(\z):=\{\v_{\z}\}$ on
$\operatorname{ri}(\mathcal K)$ and $\widetilde F(\z):=\varnothing$
otherwise. Then $\op{dom}(\widetilde F)=\operatorname{ri}(\mathcal K)$.
If $\x,\y\in\op{dom}(f)$ and $\v\in\widetilde F(\x)$, then
$\x\in\operatorname{ri}(\mathcal K)$ and $\v=\v_{\x}$,
so~\eqref{eq:pc1-dense-exact} with $\z=\x$ is
inequality~\eqref{eq:gamma-mwc}.  Thus $f$ is exactly $\gamma$--tilted convex
with respect to $\widetilde F$.
\end{proof}

\subsection{Continuity along lines}

\begin{theorem}\label{thm:gamma-MW-continuous-on-lines--strong-version}
Let $\gamma\in(0,1]$, let
$f:\mathbb R^d\to\overline{\mathbb R}$ be proper and lsc, and set
$\mathcal K:=\op{dom}(f)$.  Assume that $\mathcal K$ is convex.  Let
$F:\mathbb R^d\rightrightarrows\mathbb R^d$ be such that
$\mathcal K\cap\op{dom}(F)$ is dense in $\mathcal K$, and suppose that $f$
is approximately $\gamma$--tilted convex with respect to $F$ in the sense of
Definition~\ref{def:approx-mwc}.
Then, for every line $\ell$ that intersects $\operatorname{ri}(\mathcal K)$,
the restriction
\[
f|_{\ell \cap \mathcal K}
\]
is continuous.
\end{theorem}

\begin{proof}
By Theorem~\ref{thm:approximate-gamma-supports-local-Lipschitz}, $f$ is
locally Lipschitz, and hence continuous, on $\operatorname{ri}(\mathcal K)$.
By Proposition~\ref{prop:dense-approx-mwc-relative-interior}, there is an
operator $\widetilde F:\mathbb R^d\rightrightarrows\mathbb R^d$ with
$\op{dom}(\widetilde F)=\operatorname{ri}(\mathcal K)$ such that
$f$ is $\gamma$--tilted convex with respect to $\widetilde F$, that is,
\begin{equation}
  f(\uu)-f(\w)\ge\psi_\gamma(\langle\uu-\w,\v\rangle)
  \qquad
  \forall\w\in\operatorname{ri}(\mathcal K),\
  \v\in\widetilde F(\w),\
  \uu\in\mathcal K.
  \label{eq:pc1-lines-exact}
\end{equation}
We also use the line segment principle
(Theorem~6.1 in~\cite{rockafellar15_convex_analy}):
\begin{equation}
  \w\in\operatorname{ri}(\mathcal K),\ \uu\in\overline{\mathcal K}
  \quad\Longrightarrow\quad
  (1-t)\uu+t\w\in\operatorname{ri}(\mathcal K)
  \quad\forall t\in(0,1].
  \label{eq:pc1-segment-principle}
\end{equation}

Now fix a line $\ell$ meeting $\operatorname{ri}(\mathcal K)$ at a point
$\mathbf p$.  The set $\ell\cap\mathcal K$ is a convex subset of $\ell$,
that is, an interval.  If it is a singleton, there is nothing to prove.
Otherwise, let $\mathbf q\in\ell\cap\mathcal K$ be a point that is not
an endpoint of this interval.  If $\mathbf q\ne\mathbf p$, the interval
contains a point $\mathbf r$ such that $\mathbf q$ lies strictly between
$\mathbf p$ and $\mathbf r$, so $\mathbf q=(1-t)\mathbf r+t\mathbf p$
for some $t\in(0,1)$.  By~\eqref{eq:pc1-segment-principle},
$\mathbf q\in\operatorname{ri}(\mathcal K)$.  Thus all non-endpoint
points of $\ell\cap\mathcal K$ belong to $\operatorname{ri}(\mathcal K)$.
Near such a point $\mathbf q$, all points of $\ell\cap\mathcal K$ are
non-endpoint points, so the restriction of $f$ is continuous at
$\mathbf q$ by continuity of $f$ on $\operatorname{ri}(\mathcal K)$.
Now let $\x$ be an endpoint of this interval
that belongs to $\mathcal K$.  If $\x\in\operatorname{ri}(\mathcal K)$,
continuity at $\x$ is again known.  Otherwise $\x\ne\mathbf p$; put
$\y:=\mathbf p\in\ell\cap\operatorname{ri}(\mathcal K)$ and
\[
  \z_t:=(1-t)\x+t\y,
  \qquad
  g(t):=f(\z_t).
\]
Since $\x$ is an endpoint, $\ell\cap\mathcal K\subseteq\{\z_t:t\ge0\}$,
so continuity of $f|_{\ell\cap\mathcal K}$ at $\x$ means
$g(t)\to g(0)$ as $t\downarrow0$.
By~\eqref{eq:pc1-segment-principle},
$\z_t\in\operatorname{ri}(\mathcal K)$ for $t\in(0,1]$, so $g$ is finite
and continuous on $(0,1]$. Lower semicontinuity of $f$ at $\x$ gives, in
the extended real sense,
\[
  a:=\limsup_{t\downarrow0}g(t)
  \geq b:=\liminf_{t\downarrow0}g(t)
  \geq g(0).
\]
Here $g(0)=f(\x)\in\mathbb R$.  We show that both inequalities are
equalities.

First suppose that $a>b$.  Then $b$ is finite.  Choose a strictly
decreasing sequence $t_n\downarrow0$ in $(0,1)$ such that
$g(t_n)\to b$. We claim that, for every $n$,
\begin{equation}
  g(s)\leq\max\{g(t_{n+1}),g(t_n)\}
  \qquad\forall s\in[t_{n+1},t_n].
  \label{eq:line-segment-maximum}
\end{equation}
This is immediate at the endpoints.  Suppose that it fails at some
$s\in(t_{n+1},t_n)$, and choose $\v_{\z_s}\in\widetilde F(\z_s)$, which
is possible because $\z_s\in\operatorname{ri}(\mathcal K)$.
By~\eqref{eq:pc1-lines-exact} with $\w=\z_s$ and $\uu=\z_{t_i}$,
\begin{equation}
  \psi_\gamma(\langle\z_{t_i}-\z_s,\v_{\z_s}\rangle)
  \leq g(t_i)-g(s)<0
  \qquad(i\in\{n,n+1\}).
  \label{eq:pc1-lines-neg}
\end{equation}
Since $\psi_\gamma(r)$ has the sign of $r$, both inner products
in~\eqref{eq:pc1-lines-neg} are negative.  Put
$\lambda:=(s-t_{n+1})/(t_n-t_{n+1})\in(0,1)$, so that
$s=(1-\lambda)t_{n+1}+\lambda t_n$.  The map $t\mapsto\z_t$ is affine,
so $\z_s=(1-\lambda)\z_{t_{n+1}}+\lambda\z_{t_n}$, and hence
\[
  0
  =\left\langle
  (1-\lambda)(\z_{t_{n+1}}-\z_s)
  +\lambda(\z_{t_n}-\z_s),\v_{\z_s}\right\rangle
  <0,
\]
a contradiction.  This proves~\eqref{eq:line-segment-maximum}.

On the other hand, choose $s_n\downarrow0$ in $(0,t_1]$ such that
$g(s_n)\to a$.  The intervals $[t_{k+1},t_k]$, $k\ge1$, cover $(0,t_1]$
because $t_k\downarrow0$.  So for each $n$ there is an index $k(n)\ge1$
such that $s_n\in[t_{k(n)+1},t_{k(n)}]$.  Since
$t_{k(n)+1}\le s_n\to0$ and every $t_k$ is positive, $k(n)\to\infty$.
Equation~\eqref{eq:line-segment-maximum}, with $n$ replaced by $k(n)$,
then yields
\[
  g(s_n)\leq\max\{g(t_{k(n)+1}),g(t_{k(n)})\}\longrightarrow b.
\]
Hence $a=\lim_n g(s_n)\le b$, contradicting $a>b$, so $a=b$.

Finally, suppose that $a=b>g(0)$, so that
$g(t)\to a\in(g(0),+\infty]$ as $t\downarrow0$.  Put
$\rho:=\gamma^2/(1+\gamma^2)\in(0,1)$, and choose $t_0\in(0,1)$ such
that $g(t)>g(0)$ for all $t\in(0,t_0]$.  Let $t_n:=\rho^n t_0$.  Then
$t_n\in(0,t_0]$, so
\begin{equation}
  g(t_n)>g(0)
  \qquad\forall n\ge0.
  \label{eq:pc1-lines-above}
\end{equation}
The definition of $\rho$ gives $(1+\gamma^2)t_{n+1}=\gamma^2t_n$, that
is,
\begin{equation}
  \frac{t_{n+1}}{\gamma}=\gamma(t_n-t_{n+1}).
  \label{eq:pc1-lines-rho}
\end{equation}
For each $n$, choose
$\v_{n+1}\in\widetilde F(\z_{t_{n+1}})$ and put
$q_{n+1}:=\langle\y-\x,\v_{n+1}\rangle$.  We have
$\x-\z_{t_{n+1}}=-t_{n+1}(\y-\x)$ and
$\z_{t_n}-\z_{t_{n+1}}=(t_n-t_{n+1})(\y-\x)$.
So~\eqref{eq:pc1-lines-exact} with $\w=\z_{t_{n+1}}$, and with $\uu=\x$
or $\uu=\z_{t_n}$, gives
\begin{align}
  g(0)-g(t_{n+1})
  &\geq\psi_\gamma(-t_{n+1}q_{n+1}),
  \label{eq:pc1-lines-to-x}\\
  g(t_n)-g(t_{n+1})
  &\geq\psi_\gamma\bigl((t_n-t_{n+1})q_{n+1}\bigr).
  \label{eq:pc1-lines-to-tn}
\end{align}
By~\eqref{eq:pc1-lines-above}, the left-hand side
of~\eqref{eq:pc1-lines-to-x} is negative.  Hence
$\psi_\gamma(-t_{n+1}q_{n+1})<0$, which implies $q_{n+1}>0$.  So the
argument of $\psi_\gamma$ is negative in~\eqref{eq:pc1-lines-to-x} and
positive in~\eqref{eq:pc1-lines-to-tn}.  Using $\psi_\gamma(r)=r/\gamma$
for $r<0$ and $\psi_\gamma(r)=\gamma r$ for $r>0$, we get
\begin{align*}
  g(0)-g(t_{n+1})
  &\geq-\frac{t_{n+1}}{\gamma}q_{n+1},\\
  g(t_n)-g(t_{n+1})
  &\geq\gamma(t_n-t_{n+1})q_{n+1}.
\end{align*}
When the last two inequalities are added, the terms with $q_{n+1}$ cancel
by~\eqref{eq:pc1-lines-rho}, which yields
$g(0)+g(t_n)-2g(t_{n+1})\ge0$, that is,
\begin{equation}
  g(t_{n+1})-g(0)
  \leq\frac12\bigl(g(t_n)-g(0)\bigr).
  \label{eq:pc1-lines-halving}
\end{equation}
All values here are finite because $\z_{t_n}\in\mathcal K$.  Induction
on $n$ with~\eqref{eq:pc1-lines-halving} and~\eqref{eq:pc1-lines-above}
gives
\[
  0<g(t_n)-g(0)\le2^{-n}\bigl(g(t_0)-g(0)\bigr)
  \qquad\forall n\ge0.
\]
So $g(t_n)\to g(0)$, contrary to $g(t_n)\to a>g(0)$.
Thus $a=b=g(0)$, proving continuity at $\x$. Applying the same argument to
each endpoint of $\ell\cap\mathcal K$ completes the proof.
\end{proof}

\section{Proofs for Section~\ref{sec:structure}}
\label{apx:structure}

\subsection{Proof of Theorem~\ref{thm:mwc-domain-is-convex}}
\label{apx:mwc-domain-is-convex}

Each point of the domain is approximated by points with controlled regular subgradients.

\begin{lemma}
\label{lem:mwc-attentive-regular-subgradients}
Let $f:\mathbb R^d\to\overline{\mathbb R}$ be proper and lsc.
For every $\z\in\op{dom}(f)$, there are sequences
\begin{equation}
\z_k\xrightarrow{f}\z,
\qquad
\v_k\in\widehat\partial f(\z_k),
\qquad
\langle\v_k,\z_k-\z\rangle\longrightarrow0.
\label{eq:domain-convex-attentive-subgradients}
\end{equation}
\end{lemma}

\begin{proof}
For each integer $k\ge1$, minimize
$f(\w)+k\|\w-\z\|^2$ over $\overline{\mathbb B(\z,1)}$, and let
$\z_k$ be a minimizer. A minimizer exists because the objective is lsc
on a compact set and finite at $\z$. The function $f$ is bounded below
on that set, so comparison with $\z$ gives $\z_k\to\z$.
The same comparison gives
\[
f(\z_k)+k\|\z_k-\z\|^2\le f(\z).
\]
Lower semicontinuity at $\z$ now implies $f(\z_k)\to f(\z)$ and
$k\|\z_k-\z\|^2\to0$.
For all sufficiently large $k$, the point $\z_k$ is in the interior
of the ball and hence is an unconstrained local minimizer of the
regularized objective. The Fr\'echet necessary condition and the
smooth-sum rule give
\[
\v_k:=-2k(\z_k-\z)\in\widehat\partial f(\z_k).
\]
Discarding the finitely many initial indices, we obtain
$\langle\v_k,\z_k-\z\rangle=-2k\|\z_k-\z\|^2\to0$,
which proves the claim.
\end{proof}

By Proposition~\ref{prop:subdiff-basics}-(i),
$\widehat{\partial}f\subseteq\overline{\partial}f$.  Hence every
$\gamma$--tilted convex function in the sense of
Definition~\ref{def:gamma-MWC} is $\gamma$--tilted convex with respect to
$\widehat{\partial}f$.  It therefore suffices to prove the following
slightly stronger version of Theorem~\ref{thm:mwc-domain-is-convex}.

\begin{theorem}
\label{thm:mwc-domain-is-convex--stronger-version}
Let $f : \mathbb{R}^{d} \to \overline{\mathbb{R}}$ be proper, lsc, and $\gamma$--tilted convex with respect to $\hat{\partial} f$ for some $\gamma \in (0, 1]$.
Then $\op{dom}(f)$ is convex and $f$ is quasiconvex.
\end{theorem}

\begin{proof}
Fix distinct $\x,\y\in\op{dom}(f)$, and put
$\z_t:=t\x+(1-t)\y$ and $M:=\max\{f(\x),f(\y)\}$.
First we show that
\begin{equation}
f(\z_t)\le M\qquad\text{whenever }\z_t\in\op{dom}(f).
\label{eq:domain-convex-segment-bound}
\end{equation}
Suppose $f(\z_t)>M$, which forces $0<t<1$.
Lemma~\ref{lem:mwc-attentive-regular-subgradients} gives
$\w_k\xrightarrow{f}\z_t$ and $\v_k\in\hat\partial f(\w_k)$ with
$\langle\v_k,\w_k-\z_t\rangle\to0$.
Apply the support inequality at $\w_k$ with targets $\x$ and $\y$,
and then apply the increasing inverse $\overline\psi_\gamma$ of
$\psi_\gamma$. Taking the corresponding convex combination gives
\[
\langle\v_k,\z_t-\w_k\rangle
\le t\overline\psi_\gamma(f(\x)-f(\w_k))
 +(1-t)\overline\psi_\gamma(f(\y)-f(\w_k)).
\]
The left-hand side tends to zero, whereas the right-hand side tends
to a strictly negative number, because both value differences at
$\z_t$ are negative. This proves~\eqref{eq:domain-convex-segment-bound}.

Set $I:=\{t\in[0,1]:\z_t\in\op{dom}(f)\}$.
By~\eqref{eq:domain-convex-segment-bound},
$I=\{t\in[0,1]:f(\z_t)\le M\}$, so lower semicontinuity makes $I$
closed. It contains $0$ and $1$. If $I\ne[0,1]$, choose a component
$(a,b)$ of its complement. Then $\z_a,\z_b\in\op{dom}(f)$, while
$(\z_a,\z_b)$ contains no point of $\op{dom}(f)$.
Put $\dd:=\z_b-\z_a\ne\BF0$ and
$\Delta:=f(\z_b)-f(\z_a)$. Choose $A\in\mathbb R$ so large that
\begin{equation}
\psi_\gamma(A-\Delta)>-\Delta.
\label{eq:domain-convex-choice-of-A}
\end{equation}
Define
\[
\tau(\w):=\frac{\langle\w-\z_a,\dd\rangle}{\|\dd\|^2},
\qquad q(\w):=A\tau(\w)(\tau(\w)-1),
\qquad g:=f+q.
\]
The function $g$ is proper and lsc, has the same domain as $f$, and
satisfies $g(\z_a)=f(\z_a)$ and $g(\z_b)=f(\z_b)$.
Apply Theorem~\ref{thm:approx-mean-value} to $g$ with endpoints
$\z_a,\z_b$. Its intermediate point belongs to
$\op{dom}(g)\cap(\z_a,\z_b]$, so it must be $\z_b$.
Thus there are $\w_k\xrightarrow{g}\z_b$ and
$\uu_k\in\hat\partial g(\w_k)$ such that
\[
\limsup_k\langle\uu_k,\w_k-\z_a\rangle\le\Delta.
\]
The smooth-sum rule gives
$\v_k:=\uu_k-\nabla q(\w_k)\in\hat\partial f(\w_k)$.
Since $q$ is continuously differentiable,
$\w_k\xrightarrow{f}\z_b$ and
\[
\langle\nabla q(\w_k),\w_k-\z_a\rangle
\longrightarrow\langle\nabla q(\z_b),\dd\rangle=A.
\]
Consequently,
\[
\liminf_k\langle\v_k,\z_a-\w_k\rangle\ge A-\Delta.
\]
The support inequality for $f$ at $\w_k$ with target $\z_a$ now yields
\[
-\Delta
=\lim_k(f(\z_a)-f(\w_k))
\ge\liminf_k\psi_\gamma(\langle\v_k,\z_a-\w_k\rangle)
\ge\psi_\gamma(A-\Delta),
\]
contrary to~\eqref{eq:domain-convex-choice-of-A}.
Therefore $I=[0,1]$. Together with~\eqref{eq:domain-convex-segment-bound},
this proves that the effective domain and every sublevel set are convex.
The equal-endpoint case is immediate.
\end{proof}

\subsection{Proof of Theorem~\ref{thm:local-min-is-global-min}}
\label{apx:local-min-is-global-min}

\begin{proof}
A global minimizer is a local minimizer, so (i) implies (ii).  Since $f$ is lsc,
Proposition~\ref{prop:subdiff-basics}-(i) gives
$\hat{\partial}f(\x)\subseteq\overline{\partial}f(\x)$, so (iii)
implies (iv).

\noindent\textbf{(ii) $\Rightarrow$ (iii).}
If $\x$ is a local minimizer of the extended-real-valued function $f$,
then there is $\delta>0$ such that $f(\y)\ge f(\x)$ whenever
$\|\y-\x\|<\delta$.  Hence, for every $\y\in\op{dom}(f)$ with
$0<\|\y-\x\|<\delta$,
\[
  \frac{f(\y)-f(\x)-\langle\BF{0},\y-\x\rangle}
       {\|\y-\x\|}
  =\frac{f(\y)-f(\x)}{\|\y-\x\|}\ge0.
\]
Consequently,
\[
  \liminf_{\substack{\y\to\x\\
                      \y\in\op{dom}(f),\ \y\neq\x}}
  \frac{f(\y)-f(\x)-\langle\BF{0},\y-\x\rangle}
       {\|\y-\x\|}
  \geq0.
\]
By the definition of the Fr\'echet subdifferential,
$\BF{0}\in\hat{\partial}f(\x)$.

\noindent\textbf{(iv) $\Rightarrow$ (i).}
Assume $\BF{0}\in\overline{\partial} f(\x)$.  Then the $\gamma$--tilted convexity of $f$, applied at $\x$ with $\v=\BF{0}$, yields
\[
  f(\y)-f(\x)
  \geq\psi_\gamma(\langle\y-\x,\BF{0}\rangle)
  =\psi_\gamma(0)=0
  \qquad\forall\y\in\op{dom}(f).
\]
If $\y\notin\op{dom}(f)$, then $f(\y)=+\infty\ge f(\x)$ because $f$ is
proper.  Thus $\x$ is a global minimizer of $f$.
\end{proof}

\subsection{Proofs of Theorems~\ref{thm:gamma-MW-convex-is-locally-Lipschitz} and~\ref{thm:gamma-MW-continuous-on-lines}}
\label{apx:gamma-MW-convex-is-locally-Lipschitz}
\label{apx:gamma-MW-continuous-on-lines}

Both theorems are special cases of the results of Appendix~\ref{apx:approx}.

\begin{proof}[Proof of Theorem~\ref{thm:gamma-MW-convex-is-locally-Lipschitz}]
Let $f$ be as in the theorem and take $F:=\overline{\partial}f$.
The set $\mathcal K$ is convex by Theorem~\ref{thm:mwc-domain-is-convex}.
By Proposition~\ref{prop:subdiff-basics}-(i), every point of $\mathcal K$
is a limit of points where $\hat{\partial}f$, and hence $F$, is nonempty,
so $\mathcal K\cap\op{dom}(F)$ is dense in $\mathcal K$.  Exact
$\gamma$--tilted convexity with respect to $F$ implies approximate
$\gamma$--tilted convexity with respect to $F$, with the same $\v$ for every
$\epsilon$. Theorem~\ref{thm:approximate-gamma-supports-local-Lipschitz} therefore shows that $f$ is locally Lipschitz on $\operatorname{ri}(\mathcal K)$.
\end{proof}

\begin{proof}[Proof of Theorem~\ref{thm:gamma-MW-continuous-on-lines}]
Let $f$ be as in the theorem.  Its
domain $\mathcal K$ is convex by Theorem~\ref{thm:mwc-domain-is-convex}.
The $\gamma$--tilted convexity of $f$ implies exact, and hence approximate,
$\gamma$--tilted convexity with respect to $\widehat{\partial}f$, whose domain
is dense in $\mathcal K$ by Proposition~\ref{prop:subdiff-basics}-(i).
For $d=1$ the only line is $\ell=\mathbb R$, which meets
$\operatorname{ri}(\mathcal K)$ because $\mathcal K$ is nonempty and
convex, and $f|_{\ell\cap\mathcal K}=f|_{\mathcal K}$. Theorem~\ref{thm:gamma-MW-continuous-on-lines--strong-version} therefore shows that $f|_{\mathcal K}$ is continuous.
\end{proof}

\subsection{Proof of Theorem~\ref{thm:mwc-alternative-characterizations}: Alternative First-order Characterization}
\label{apx:mwc-alternative-characterizations}

Since $f$ is lsc, Proposition~\ref{prop:subdiff-basics}-(i) gives
\begin{equation}
\label{eq:pc2-subdiff-inclusions}
\widehat{\partial}f(\x)\subseteq\partial f(\x)\subseteq\overline{\partial}f(\x),
\qquad\x\in\mathcal K.
\end{equation}
If $f$ is $\gamma$--tilted convex with respect to an operator, then it is
$\gamma$--tilted convex with respect to every operator whose values are
subsets of the values of the first one.  Hence
\eqref{eq:pc2-subdiff-inclusions} gives the implications
(i) $\implies$ (ii) $\implies$ (iii).  If (iii) holds, then
Theorem~\ref{thm:mwc-domain-is-convex--stronger-version} shows that
$\mathcal K$ is convex, and the inequality in (iv) is the inequality
of (iii) restricted to $\x\in\op{ri}(\mathcal K)$.  This proves
(iii) $\implies$ (iv).

To see that (i) $\implies$ (v), assume (i) and put
$A:=\op{aff}(\mathcal K)$.  Theorem~\ref{thm:mwc-domain-is-convex}
shows that $\mathcal K$ is convex, and
Theorem~\ref{thm:gamma-MW-convex-is-locally-Lipschitz} shows that
$f|_{\op{ri}(\mathcal K)}$ is locally Lipschitz.  By Rademacher's
theorem in the Euclidean space $A$, the points of
$\op{ri}(\mathcal K)$ where the relative gradient $\nabla f$ exists
have full measure in the relatively open set $\op{ri}(\mathcal K)$, so
they are dense in $\op{ri}(\mathcal K)$.  Since $\mathcal K$ is convex,
$\op{ri}(\mathcal K)$ is dense in $\mathcal K$ (Theorem~6.3
in~\cite{rockafellar15_convex_analy}), and hence these points are
dense in $\mathcal K$.  Let $\x\in\op{ri}(\mathcal K)$ be such a point.
Every $\y\in\op{dom}(f)=\mathcal K$ lies in $A$, so the
differentiability of $f|_A$ at $\x$ gives
\[
\lim_{\substack{\y\to\x\\ \y\in\mathcal K,\ \y\neq\x}}
\frac{f(\y)-f(\x)-\langle\nabla f(\x),\y-\x\rangle}{\|\y-\x\|}
=0.
\]
By the definition of $\widehat{\partial}f$ and by
\eqref{eq:pc2-subdiff-inclusions}, this yields
\begin{equation}
\label{eq:pc2-gradient-is-regular}
\nabla f(\x)\in\widehat{\partial}f(\x)\subseteq\overline{\partial}f(\x).
\end{equation}
Condition (i) with $\vv=\nabla f(\x)$ now gives the inequality in (v).
It remains to prove (ii) $\implies$ (i), (iv) $\implies$ (ii), and
(v) $\implies$ (iv).

\subsubsection{Horizon subdifferentials and normal cones}

We use normal cones to control the horizon subgradients of $\gamma$--tilted convex functions.

\begin{definition}
Let $\C{X} \subseteq \mathbb R^d$ and $\x \in \C{X}$.
A vector $\vv \in \mathbb R^d$ is \emph{regular normal to $\C{X}$}, written $\vv \in \hat{N}_{\C{X}}(\x)$, if 
\[
\limsup_{\substack{\y\to\x\\
                    \y\in\C{X},\ \y\neq\x}}
\frac{\langle\vv,\y-\x\rangle}{\|\y-\x\|}
\leq 0.
\]
Furthermore, a vector $\vv \in \mathbb R^d$ is \emph{normal to $\C{X}$}, written
$\vv \in N_{\C{X}}(\x)$, if there are sequences $\x_k\in\C{X}$ and
$\vv_k\in\mathbb R^d$ such that
\[
  \x_k\to\x,
  \qquad
  \vv_k\to\vv,
  \qquad
  \vv_k\in\hat N_{\C{X}}(\x_k).
\]
\end{definition}

For functions that are $\gamma$--tilted convex with respect to
their regular subdifferentials, horizon subgradients are normal to the effective domain.

\begin{lemma}
\label{lem:horizon-subdiff}
Let $f:\mathbb R^d\to\overline{\mathbb R}$ be proper, lsc, and
$\gamma$--tilted convex with respect to $\widehat{\partial}f$ for some
$\gamma\in(0,1]$, and set
$\mathcal K:=\op{dom}(f)$.
Then, for any $\x \in \C{K}$, we have
\[
\partial^{\infty} f(\x)
\subseteq 
\{ \vv \mid \forall \y \in \C{K}, \langle \vv, \y-\x \rangle \le 0 \}
= 
\hat{N}_{\C{K}}(\x)
= N_{\C{K}}(\x).
\]
In particular, if $V = \op{aff}(\C{K}) - \op{aff}(\C{K})$, then
\[
V \cap \partial^{\infty} f(\x) \cap -N_{\C{K}}(\x) = \{ \BF{0} \}.
\]
\end{lemma}

\begin{proof}
Theorem~\ref{thm:mwc-domain-is-convex--stronger-version} shows that
$\mathcal K$ is convex.
Let $\vv\in\partial^\infty f(\x)$.  By the definition of the horizon
subdifferential, there are sequences
\begin{equation}
\label{eq:pc2-horizon-sequences}
  \x_k\xrightarrow{f}\x,
  \qquad
  \lambda_k>0,
  \qquad
  \lambda_k\to0,
  \qquad
  \vv_k\in\widehat{\partial}f(\x_k),
  \qquad
  \lambda_k\vv_k\to\vv.
\end{equation}
Fix $\y\in\mathcal K$.  Each $\x_k$ lies in $\op{dom}(f)=\mathcal K$,
so the $\gamma$--tilted convexity of $f$ with respect to
$\widehat{\partial}f$ gives
\[
  f(\y)-f(\x_k)
  \geq\psi_\gamma(\langle\vv_k,\y-\x_k\rangle).
\]
By Lemma~\ref{lem:psi_basics}, $\overline{\psi}_\gamma$ is increasing
and $\overline{\psi}_\gamma(\psi_\gamma(a))=a$.  Applying
$\overline{\psi}_\gamma$ to both sides and multiplying by
$\lambda_k>0$ yields
\begin{equation}
\label{eq:pc2-horizon-scaled}
  \langle\lambda_k\vv_k,\y-\x_k\rangle
  \leq
  \lambda_k\overline{\psi}_\gamma\bigl(f(\y)-f(\x_k)\bigr).
\end{equation}
Since $f(\x_k)\to f(\x)$ and $\overline{\psi}_\gamma$ is continuous,
$\overline{\psi}_\gamma(f(\y)-f(\x_k))$ converges to the real number
$\overline{\psi}_\gamma(f(\y)-f(\x))$.  As $\lambda_k\to0$, the
right-hand side of \eqref{eq:pc2-horizon-scaled} tends to zero, while
by \eqref{eq:pc2-horizon-sequences} the left-hand side tends to
$\langle\vv,\y-\x\rangle$.  Hence
\begin{equation}
\label{eq:pc2-horizon-polar}
  \langle\vv,\y-\x\rangle\leq0
  \qquad\forall\y\in\mathcal K.
\end{equation}

For $\x'\in\mathcal K$, put
\[
  P(\x'):=\{\vv:\langle\vv,\y-\x'\rangle\leq0
    \ \forall\y\in\mathcal K\}.
\]
By \eqref{eq:pc2-horizon-polar}, $\partial^\infty f(\x)\subseteq P(\x)$.
According to Theorem~6.9 in~\cite{rockafellar1998variational}, we have
\[
  P(\x)
  =\widehat N_{\mathcal K}(\x)
  =N_{\mathcal K}(\x).
\]
These equalities also
follow directly from the definitions and the convexity of
$\mathcal K$.  If $\vv\in P(\x')$, then every quotient
in the definition of $\widehat N_{\mathcal K}(\x')$ is nonpositive, so
$\vv\in\widehat N_{\mathcal K}(\x')$.  Conversely, let
$\vv\in\widehat N_{\mathcal K}(\x')$ and
$\y\in\mathcal K\setminus\{\x'\}$.  By convexity, the points
$\y_t:=\x'+t(\y-\x')$ with $t\in(0,1]$ lie in $\mathcal K$, they tend
to $\x'$ as $t\downarrow0$, and
\[
  \frac{\langle\vv,\y_t-\x'\rangle}{\|\y_t-\x'\|}
  =\frac{\langle\vv,\y-\x'\rangle}{\|\y-\x'\|}.
\]
The limit superior in the definition of
$\widehat N_{\mathcal K}(\x')$ is at least this constant, so
$\langle\vv,\y-\x'\rangle\leq0$.  Thus
$\widehat N_{\mathcal K}(\x')=P(\x')$ for every $\x'\in\mathcal K$.
Constant sequences show that
$\widehat N_{\mathcal K}(\x)\subseteq N_{\mathcal K}(\x)$.  Finally, let
$\vv\in N_{\mathcal K}(\x)$, with $\x_k\to\x$ and $\vv_k\to\vv$ as in
the definition.  Since $\vv_k\in\widehat N_{\mathcal K}(\x_k)=P(\x_k)$,
we have $\langle\vv_k,\y-\x_k\rangle\leq0$ for every
$\y\in\mathcal K$, and letting $k\to\infty$ gives $\vv\in P(\x)$.
This completes the proof of the first claim.

For the second claim, let
$\vv\in V\cap\partial^\infty f(\x)\cap(-N_{\mathcal K}(\x))$.
The first claim gives $\langle\vv,\y-\x\rangle\leq0$ for all
$\y\in\mathcal K$.  Since $-\vv\in N_{\mathcal K}(\x)=P(\x)$, we also
have $\langle-\vv,\y-\x\rangle\leq0$ for all $\y\in\mathcal K$.  Hence
\[
  \langle\vv,\y-\x\rangle=0
  \qquad\forall\y\in\mathcal K.
\]
Because $\x\in\mathcal K$, the vectors $\y-\x$ with $\y\in\mathcal K$
span $V$.  Thus $\vv\in V^\perp$.  Since also $\vv\in V$, we obtain
$\langle\vv,\vv\rangle=0$, that is, $\vv=\BF{0}$.
Conversely, $\BF{0}$ lies in $V$, in the cone $N_{\mathcal K}(\x)$, and in the closed cone $\partial^\infty f(\x)$ (Theorem~8.6 in~\cite{rockafellar1998variational}), which proves the equality.
\end{proof}

\subsubsection{Proof of (ii) \texorpdfstring{$\implies$}{implies} (i)}

\begin{proof}
Assume (ii).  By \eqref{eq:pc2-subdiff-inclusions}, $f$ is also
$\gamma$--tilted convex with respect to $\widehat{\partial}f$, so
Lemma~\ref{lem:horizon-subdiff} applies to $f$.
Fix $\x \in \C{K}$.
Let $V \subseteq \mathbb R^d$ be the set of vectors $\vv$ such that
\[
f(\y) - f(\x)
\ge
\psi_\gamma(\langle \y - \x, \vv \rangle)
,\qquad \forall \y \in \C{K}.
\]
By Lemma~\ref{lem:psi_basics}, $\psi_\gamma$ and
$\overline{\psi}_\gamma$ are increasing bijections that are inverse to
each other.  Hence $\psi_\gamma(a)\leq b$ holds if and only if
$a\leq\overline{\psi}_\gamma(b)$, and
\begin{align}
V
&=
\left\{ \vv \in \mathbb{R}^d
\mid
\overline{\psi}_\gamma(f(\y) - f(\x))
\ge
\langle \y - \x, \vv \rangle
,\ \forall \y \in \C{K}
\right\} \notag\\
&= \bigcap_{\y \in \C{K}}
\left\{ \vv \in \mathbb{R}^d
\mid
\overline{\psi}_\gamma(f(\y) - f(\x))
\ge
\langle \y - \x, \vv \rangle
\right\}.
\label{eq:pc2-V-intersection}
\end{align}
For each $\y\neq\x$, the set on the right of
\eqref{eq:pc2-V-intersection} is a closed half-space, and for
$\y=\x$ it is $\mathbb R^d$.  So $V$ is convex and closed.
Condition (ii) states that $\partial f(\x) \subseteq V$, so the
convexity of $V$ gives
\begin{equation}
\label{eq:pc2-conv-in-V}
\op{conv}(\partial f(\x)) \subseteq V.
\end{equation}
Now let $\vv\in V$ and $\mathbf n\in\hat{N}_{\C{K}}(\x)$.
Lemma~\ref{lem:horizon-subdiff} gives
$\langle\y-\x,\mathbf n\rangle\leq0$ for every $\y\in\mathcal K$, and
therefore
\[
\langle \y - \x, \vv+\mathbf n \rangle
\leq
\langle \y - \x, \vv \rangle
\leq
\overline{\psi}_\gamma(f(\y) - f(\x))
,\qquad \forall \y \in \C{K}.
\]
Thus $V + \hat{N}_{\C{K}}(\x) \subseteq V$.  Together with
\eqref{eq:pc2-conv-in-V}, this gives
$\op{conv}(\partial f(\x)) + \hat{N}_{\C{K}}(\x) \subseteq V$.
Since $V$ is closed, we obtain
\begin{equation}
\label{eq:pc2-closure-in-V}
\overline{\op{conv}(\partial f(\x)) + \hat{N}_{\C{K}}(\x)} \subseteq V.
\end{equation}
Lemma~\ref{lem:horizon-subdiff} also gives
$\partial^\infty f(\x)\subseteq\hat{N}_{\C{K}}(\x)$, and
$\hat{N}_{\C{K}}(\x)$ is convex, being an intersection of closed
half-spaces.  Hence
$\op{conv}(\partial^\infty f(\x))\subseteq\hat{N}_{\C{K}}(\x)$.  By the
definition of $\overline{\partial} f$ and by
\eqref{eq:pc2-closure-in-V},
\[
\overline{\partial} f(\x)
=
\overline{\op{conv}(\partial f(\x)) + \op{conv}(\partial^\infty f(\x))}
\subseteq
\overline{\op{conv}(\partial f(\x)) + \hat{N}_{\C{K}}(\x)}
\subseteq
V.
\]
Hence, by the definition of $V$, for all
$\vv \in \overline{\partial} f(\x)$ we have
\[
f(\y) - f(\x)
\ge
\psi_\gamma(\langle \y - \x, \vv \rangle)
,\qquad \forall \y \in \C{K}.
\]
Since $\x\in\mathcal K$ was arbitrary, $f$ is $\gamma$--tilted convex.
\end{proof}

\subsubsection{Proof of (iv) \texorpdfstring{$\implies$}{implies} (ii)}

We first approximate regular subgradients at points of the domain by limiting subgradients at nearby points of the relative interior.

\begin{lemma}
\label{lem:approx-from-relint}
Under the assumptions of Theorem~\ref{thm:mwc-alternative-characterizations}-(iv), for any $\x \in \mathcal K$, $\vv \in \hat{\partial} f(\x)$, and $\y \in \op{ri}(\C{K})$, there exist sequences $\z_n \in \op{ri}(\C{K})$ and $\vv_n \in \partial f(\z_n)$ such that $\z_n \xrightarrow{f} \x$ and
\[
\langle \vv, \y - \x \rangle
\;\le\;
\liminf_{n\to\infty}
\langle \vv_n, \y - \x \rangle.
\]
\end{lemma}

\begin{proof}
If $\x = \y$, then $\x\in\op{ri}(\mathcal K)$, and
$\vv\in\hat{\partial} f(\x)\subseteq\partial f(\x)$ by
\eqref{eq:pc2-subdiff-inclusions}.  Choosing $\z_n = \x$
and $\vv_n = \vv$ for all $n \geq 1$ makes both sides of the required
inequality zero.
Assume $\x \neq \y$, and put $A:=\op{aff}(\mathcal K)$.
For $t \in (0,1]$, define
\[
\y_t := \x + t(\y - \x).
\]
Since $\x\in\mathcal K$ and $\y\in\op{ri}(\mathcal K)$, Theorem~6.1
in~\cite{rockafellar15_convex_analy} shows that
\begin{equation}
\label{eq:pc2-half-open-segment}
(1-s)\x+s\y\in\op{ri}(\mathcal K),
\qquad s\in(0,1].
\end{equation}

Let $\x'\in\op{ri}(\mathcal K)$.  Proposition~\ref{prop:subdiff-basics}-(i)
gives points $\x_i\to\x'$ in $\op{dom}(f)=\mathcal K$ with
$\hat{\partial} f(\x_i)\neq\emptyset$.  Since $\op{ri}(\C{K})$ is open
in $A$ and $\mathcal K\subseteq A$, these points lie in
$\op{ri}(\mathcal K)$ for all large $i$.  Therefore the set of points
of $\op{ri}(\C{K})$ where $\hat{\partial} f$ is non-empty is dense in
$\op{ri}(\C{K})$.
Since $\mathcal K$ is convex, $\op{ri}(\mathcal K)$ is dense in
$\mathcal K$, so this set is dense in $\mathcal K$.  Put
$F:=\widehat{\partial}f|_{\op{ri}(\mathcal K)}$.  Then
$\mathcal K\cap\op{dom}(F)$ is dense in $\mathcal K$, and
condition~(iv) states that $f$ is $\gamma$--tilted convex, hence
approximately $\gamma$--tilted convex, with respect to $F$.  Thus the
hypotheses of Theorems~\ref{thm:approximate-gamma-supports-local-Lipschitz}
and~\ref{thm:gamma-MW-continuous-on-lines--strong-version} hold, and
they give
\begin{equation}
\label{eq:pc2-iv-regularity}
\begin{gathered}
f \text{ is locally Lipschitz on } \op{ri}(\mathcal K),\\
f|_{\ell\cap\mathcal K} \text{ is continuous for every line } \ell
\text{ that meets } \op{ri}(\mathcal K).
\end{gathered}
\end{equation}
The line through $\x$ and $\y$ meets $\op{ri}(\mathcal K)$ at $\y$, so
$f$ is continuous on $[\x,\y]$.

Fix $t\in(0,1]$.  By \eqref{eq:pc2-half-open-segment}, the open
segment $(\x,\y_t)$ lies in $\op{ri}(\mathcal K)$, where $f$ is locally
Lipschitz, and $f$ is continuous on $[\x,\y_t]\subseteq[\x,\y]$.  We
apply Corollary~\ref{cor:Lipschitz-interior-mean-value} to $f|_A$ in
the Euclidean space $A$, at the points $\x$ and $\y_t$.  It gives a
point $\z_t \in (\x, \y_t) \subseteq \op{ri}(\C{K})$ and a vector
$\overline{\vv}_t$ in the convex hull of the limiting subdifferential
of $f|_A$ at $\z_t$ such that
\begin{equation}
\label{eq:pc2-mean-value-t}
f(\y_t) - f(\x)
= \langle \overline{\vv}_t, \y_t - \x \rangle.
\end{equation}
Since $\op{dom}(f)=\mathcal K\subseteq A$, the quotients defining the
regular subdifferentials of $f$ and of $f|_A$ at a point of
$\mathcal K$ are the same.  Hence every regular subgradient of $f|_A$
is a regular subgradient of $f$, and, passing to limits, every
limiting subgradient of $f|_A$ is a limiting subgradient of $f$.  In
particular, $\overline{\vv}_t \in \op{conv}(\partial f(\z_t))$.

Since $\vv \in \hat{\partial} f(\x)$, we have
\begin{equation}
\label{eq:pc2-regular-at-x}
\liminf_{\substack{\z \to \x\\ \z \in \mathcal K,\ \z \neq \x}}
\frac{
f(\z) - f(\x) - \langle \vv, \z - \x \rangle
}{
\|\z - \x\|
}
\ge 0.
\end{equation}
The points $\y_t$ lie in $\mathcal K\setminus\{\x\}$ and tend to $\x$
as $t\downarrow0$, so the lower limit along $\y_t$ is at least the
left-hand side of \eqref{eq:pc2-regular-at-x}.  Using
\eqref{eq:pc2-mean-value-t} and $\y_t - \x = t (\y - \x)$, we see that
\[
0
\le
\liminf_{t \downarrow 0}
\frac{
f(\y_t) - f(\x) - \langle \vv, \y_t - \x \rangle
}{
\|\y_t - \x\|
}
=
\liminf_{t \downarrow 0}
\frac{
\langle \overline{\vv}_t - \vv, \y_t - \x \rangle
}{
\|\y_t - \x\|
}
=
\liminf_{t \downarrow 0}
\frac{
\langle \overline{\vv}_t - \vv, \y - \x \rangle
}{
\|\y - \x\|
}.
\]
Multiplying by $\|\y-\x\|>0$ and adding $\langle\vv,\y-\x\rangle$ gives
\begin{equation}
\label{eq:pc2-liminf-conv}
\langle \vv, \y - \x \rangle
\le
\liminf_{t \downarrow 0}
\langle \overline{\vv}_t, \y - \x \rangle.
\end{equation}

Fix $t\in(0,1]$ again.  Since
$\overline{\vv}_t\in\op{conv}(\partial f(\z_t))$, write
$\overline{\vv}_t=\sum_{i=1}^m\theta_i\vv_{t,i}$, where
$\vv_{t,i}\in\partial f(\z_t)$, $\theta_i>0$, and
$\sum_{i=1}^m\theta_i=1$.  Then
\[
\langle\overline{\vv}_t,\y-\x\rangle
=\sum_{i=1}^m\theta_i\langle\vv_{t,i},\y-\x\rangle.
\]
If every $\vv_{t,i}$ satisfied
$\langle\vv_{t,i},\y-\x\rangle<\langle\overline{\vv}_t,\y-\x\rangle$,
then multiplying these inequalities by $\theta_i$ and summing would give
$\langle\overline{\vv}_t,\y-\x\rangle<\langle\overline{\vv}_t,\y-\x\rangle$.
Hence we can choose one of the $\vv_{t,i}$, denoted by
$\vv_t\in\partial f(\z_t)$, such that
\begin{equation}
\label{eq:pc2-select-component}
\langle \vv_t, \y - \x \rangle
\ge
\langle \overline{\vv}_t, \y - \x \rangle.
\end{equation}
Combining \eqref{eq:pc2-liminf-conv} and
\eqref{eq:pc2-select-component} gives
\[
\langle \vv, \y - \x \rangle
\le
\liminf_{t \downarrow 0}
\langle \overline{\vv}_t, \y - \x \rangle
\le
\liminf_{t \downarrow 0}
\langle \vv_t, \y - \x \rangle.
\]
Choose any sequence $t_n\downarrow0$ and set
$\z_n:=\z_{t_n}\in\op{ri}(\mathcal K)$ and
$\vv_n:=\vv_{t_n}\in\partial f(\z_n)$.  The lower limit along the
sequence $t_n$ is at least the lower limit as $t\downarrow0$, so the
last display gives the required inequality for $(\vv_n)$.  Moreover,
$\|\z_n-\x\|\leq\|\y_{t_n}-\x\|=t_n\|\y-\x\|\to0$, and the continuity
of $f$ on $[\x,\y]$ gives $f(\z_n)\to f(\x)$.  Thus
$\z_n\xrightarrow{f}\x$.
\end{proof}

\begin{proof}[Proof of (iv) $\implies$ (ii)]
Assume (iv).  We show that
\[
f(\mathbf y)-f(\mathbf x)
\;\ge\;
\psi_\gamma(\langle \mathbf v,\mathbf y-\mathbf x\rangle),
\]
for all $\mathbf x,\mathbf y \in \mathcal K$ and $\mathbf v \in \partial f(\mathbf x)$.
Put $A:=\op{aff}(\mathcal K)$.  The proof of
Lemma~\ref{lem:approx-from-relint} shows that condition~(iv) implies
\eqref{eq:pc2-iv-regularity}.

\noindent\textit{Case 1: $\mathbf x,\mathbf y \in \op{ri}(\mathcal K)$ and $\mathbf v \in \partial f(\mathbf x)$. }

By definition of the limiting subdifferential, there exist
\[
\mathbf x_k\to\mathbf x,\qquad
f(\mathbf x_k)\to f(\mathbf x),\qquad
\vv_k\in\hat\partial f(\mathbf x_k),\qquad
\vv_k\to\mathbf v.
\]
Each $\x_k$ lies in $\op{dom}(f)=\mathcal K\subseteq A$.  Since
$\mathbf x\in\op{ri}(\mathcal K)$ and $\op{ri}(\mathcal K)$ is open in
$A$, after discarding finitely many terms we have
$\mathbf x_k\in\op{ri}(\mathcal K)$.
Condition (iv), applied at $\x_k$ with $\vv_k\in\hat\partial f(\x_k)$,
gives
\[
f(\y) - f(\x_k) \geq \psi_\gamma(\langle \vv_k, \y - \x_k \rangle).
\]
As $k \to \infty$, we have $f(\x_k)\to f(\x)$ and
$\langle \vv_k, \y - \x_k \rangle\to\langle \vv, \y - \x \rangle$.
Since $\psi_\gamma$ is continuous, we obtain
\[
f(\y) - f(\x) \geq \psi_\gamma(\langle \vv, \y - \x \rangle).
\]

\noindent\textit{Case 2: $\x \in \C{K}$, $\y \in \op{ri}(\C{K})$ and $\vv \in \hat{\partial} f(\x)$.}

Set $\dd := \y - \x$.
By Lemma~\ref{lem:approx-from-relint}, there exist $\z_n \in \op{ri}(\C{K})$ and $\vv_n \in \partial f(\z_n)$ such that $\z_n \xrightarrow{f} \x$ and
\begin{equation}
\label{eq:pc2-case2-liminf}
\langle \vv, \dd \rangle
\;\le\;
\liminf_{n \to \infty} \langle \vv_n, \dd \rangle.
\end{equation}
Define $\y_n := \z_n + \dd$.  Since $\z_n\in A$ and $\dd\in A-A$, we
have $\y_n\in A$, and $\y_n \to \x+\dd=\y$.  Since
$\y \in \op{ri}(\C{K})$ and $\op{ri}(\C{K})$ is open in $A$, after
discarding finitely many terms we may assume that
$\y_n \in \op{ri}(\C{K})$.
Case~1, applied to $\z_n,\y_n\in\op{ri}(\C{K})$ and
$\vv_n\in\partial f(\z_n)$, together with $\y_n - \z_n = \dd$, gives
\begin{equation}
\label{eq:pc2-case2-n}
f(\y_n) - f(\z_n)
\;\ge\;
\psi_\gamma(\langle \vv_n, \dd \rangle).
\end{equation}
By \eqref{eq:pc2-iv-regularity}, $f$ is continuous on
$\op{ri}(\mathcal K)$, so $f(\y_n)\to f(\y)$.  Also
$f(\z_n)\to f(\x)$, because $\z_n\xrightarrow{f}\x$.  Hence the
left-hand side of \eqref{eq:pc2-case2-n} converges to the real number
$f(\y)-f(\x)$.  Let $L:=\liminf_{n\to\infty}\langle\vv_n,\dd\rangle$.
By \eqref{eq:pc2-case2-liminf}, $L\geq\langle\vv,\dd\rangle$.  If
$L=+\infty$, then $\psi_\gamma(\langle\vv_n,\dd\rangle)\to+\infty$,
which contradicts \eqref{eq:pc2-case2-n}.  So $L$ is finite.  Choose a
subsequence along which $\langle\vv_n,\dd\rangle\to L$.  Passing to the
limit in \eqref{eq:pc2-case2-n} along it, and using that $\psi_\gamma$
is continuous and increasing, we get
\[
f(\y) - f(\x)
\;\ge\;
\psi_\gamma(L)
\;\ge\;
\psi_\gamma(\langle \vv, \dd \rangle)
= \psi_\gamma(\langle \vv, \y - \x \rangle).
\]
Thus the desired inequality holds for all $\vv \in \hat\partial f(\x)$ whenever $\y \in \op{ri}(\C{K})$.

\noindent\textit{Case 3: $\x \in \C{K}$, $\y \in \op{ri}(\C{K})$ and $\vv \in \partial f(\x)$.}

By definition of the limiting subdifferential, there exist
\[
\mathbf x_k\to\mathbf x,\qquad
f(\mathbf x_k)\to f(\mathbf x),\qquad
\vv_k\in\hat\partial f(\mathbf x_k),\qquad
\vv_k\to\mathbf v.
\]
Each $\x_k$ lies in $\op{dom}(f)=\mathcal K$.
Applying Case~2 to each $(\mathbf x_k,\vv_k)$ and the fixed $\y \in \op{ri}(\C{K})$ yields
\[
f(\mathbf y)-f(\mathbf x_k)
\;\ge\;
\psi_\gamma(\langle \vv_k,\mathbf y-\mathbf x_k\rangle).
\]
Letting $k\to\infty$ and using continuity of $\psi_\gamma$, we obtain
\[
f(\y)-f(\x)
\geq
\psi_\gamma(\langle \vv,\y-\x\rangle).
\]

\noindent\textit{Case 4: $\x,\y \in \C{K}$ and $\vv \in \partial f(\x)$.}

Fix $\y_0\in\op{ri}(\mathcal K)$; this set is non-empty because
$\mathcal K$ is non-empty and convex.  If $\y=\y_0$, Case~3 applies.
Otherwise, set
$\y_k:=(1-k^{-1})\y+k^{-1}\y_0$ for $k\geq1$.  By Theorem~6.1
in~\cite{rockafellar15_convex_analy}, $\y_k\in\op{ri}(\mathcal K)$.
All $\y_k$ lie on the line through $\y$ and $\y_0$, which meets
$\op{ri}(\mathcal K)$ at $\y_0$.  By \eqref{eq:pc2-iv-regularity}, $f$
is continuous on the intersection of this line with $\mathcal K$, so
$f(\y_k)\to f(\y)$.
Case~3, applied to $\x$, $\y_k$, and $\vv$, gives
\[
f(\y_k) - f(\x)
\;\ge\;
\psi_\gamma(\langle \vv, \y_k - \x\rangle).
\]
Letting $k \to \infty$ and using continuity of $\psi_\gamma$ gives
\[
f(\y)-f(\x)
\;\ge\;
\psi_\gamma(\langle \vv, \y - \x \rangle).
\]
Thus $f$ is $\gamma$--tilted convex with respect to $\partial f$, proving
(ii).
\end{proof}

\subsubsection{Proof of (v) \texorpdfstring{$\implies$}{implies} (iv)}

\begin{proof}
Assume (v).  Put $A:=\op{aff}(\mathcal K)$ and $W:=A-A$, and interpret
gradients relative to $A$.  Let $G$ be the operator
$\x\mapsto\{\nabla_A f(\x)\}$, defined at the points
$\x\in\op{ri}(\mathcal K)$ where $\nabla_A f(\x)$ exists.  By
condition~(v), $\mathcal K$ is convex, $\mathcal K\cap\op{dom}(G)$ is
dense in $\mathcal K$, and $f$ is $\gamma$--tilted convex, hence
approximately $\gamma$--tilted convex, with respect to $G$.
Theorem~\ref{thm:approximate-gamma-supports-local-Lipschitz} therefore
shows that $f$ is locally Lipschitz on $\op{ri}(\mathcal K)$.

Fix $\x\in\op{ri}(\mathcal K)$, $\y\in\mathcal K$, and
$\vv\in\widehat{\partial}f(\x)$, and let $\vv_A$ be the orthogonal
projection of $\vv$ onto $W$.  Every $\z\in\op{dom}(f)=\mathcal K$
satisfies $\z-\x\in W$, so
\begin{equation}
\label{eq:pc2-projection}
\langle\vv,\z-\x\rangle=\langle\vv_A,\z-\x\rangle,
\qquad \z\in\mathcal K.
\end{equation}
Hence the quotients in the definition of the regular subdifferential
do not change when $\vv$ is replaced by $\vv_A$, and
$\vv_A\in\widehat{\partial}_A f(\x)$, where the subdifferential is
taken relative to $A$.

Let $\overline{\nabla}_A f(\x)$ be the set of
Proposition~\ref{prop:subdiff-basics}-(ii) for $f|_A$ in the Euclidean
space $A$, and let $\mathbf w\in\overline{\nabla}_A f(\x)$.  Then there
is a sequence $\x_k\to\x$ in $A$ such that the relative gradients
$\nabla_A f(\x_k)$ exist and converge to $\mathbf w$.  Since
$\op{ri}(\mathcal K)$ is open in $A$, we have
$\x_k\in\op{ri}(\mathcal K)$ for all large $k$.  Condition~(v) gives
\[
f(\y) - f(\x_k)
\geq
\psi_\gamma(\langle \nabla_A f(\x_k), \y - \x_k \rangle).
\]
Since $f$ is continuous at $\x$ and $\psi_\gamma$ is continuous,
letting $k\to\infty$ gives
\[
f(\y)-f(\x)
\geq
\psi_\gamma(\langle\mathbf w,\y-\x\rangle).
\]
Applying the increasing inverse $\overline{\psi}_\gamma$
(Lemma~\ref{lem:psi_basics}) yields
\begin{equation}
\label{eq:pc2-v-iv-halfspace}
\overline{\psi}_\gamma(f(\y)-f(\x))
\geq
\langle\mathbf w,\y-\x\rangle,
\qquad
\mathbf w\in\overline{\nabla}_A f(\x).
\end{equation}
The set of $\mathbf w\in W$ satisfying \eqref{eq:pc2-v-iv-halfspace} is
a closed half-space of $W$ or all of $W$, hence convex.  It contains
$\overline{\nabla}_A f(\x)$, so it contains
$\op{conv}(\overline{\nabla}_A f(\x))$.  Since $f|_A$ is locally
Lipschitz at $\x$, Proposition~\ref{prop:subdiff-basics}-(i) and~(ii),
applied in the Euclidean space $A$, give
\[
\vv_A\in\widehat{\partial}_A f(\x)
\subseteq\overline{\partial}_A f(\x)
=\op{conv}(\overline{\nabla}_A f(\x)).
\]
Therefore \eqref{eq:pc2-v-iv-halfspace} holds with $\mathbf w=\vv_A$.
By \eqref{eq:pc2-projection} with $\z=\y$, we have
$\langle\vv_A,\y-\x\rangle=\langle\vv,\y-\x\rangle$, and hence
$\overline{\psi}_\gamma(f(\y)-f(\x))\geq\langle\vv,\y-\x\rangle$.
Applying the increasing function $\psi_\gamma$ to this inequality
gives
\[
f(\y)-f(\x)\geq\psi_\gamma(\langle\vv,\y-\x\rangle).
\]
Together with the convexity of $\mathcal K$, this proves (iv).
\end{proof}

\subsection{Proof of Theorem~\ref{thm:gamma-MW-convex-on-lines}}
\label{apx:gamma-MW-convex-on-lines}

\begin{proof}
\noindent\textbf{Forward direction.}
Suppose first that $f$ is $\gamma$--tilted convex, and put
$\mathcal K:=\operatorname{dom}(f)$.  By
Theorem~\ref{thm:mwc-domain-is-convex}, the set $\mathcal K$ is convex.
Fix a line $\ell$ meeting $\mathcal K$, write
\[
\ell=\{\mathbf b+t\mathbf d:t\in\mathbb R\},
\qquad \|\mathbf d\|=1,
\]
and set
\[
I:=\{t\in\mathbb R:\mathbf b+t\mathbf d\in\mathcal K\},
\qquad
g(t):=f(\mathbf b+t\mathbf d).
\]
Since $\|\mathbf d\|=1$, the map $t\mapsto\mathbf b+t\mathbf d$ is an
isometry of $\mathbb R$ onto $\ell$, and $g$ is $f|_\ell$ written in this
coordinate; so it suffices to show that $g$ is $\gamma$--tilted convex.
The function $g:\mathbb R\to\overline{\mathbb R}$ is proper and lsc,
because it is the composition of $f$ with a continuous affine map, and
$I=\operatorname{dom}(g)$ is an interval, because it is the preimage of
the convex set $\mathcal K$ under an affine map.  If $I=\{t_0\}$ is a
singleton, the only instance of~\eqref{eq:gamma-mwc} for $g$ compares
$t_0$ with itself, and both sides vanish; so $g$ is $\gamma$--tilted convex.
Suppose therefore that $I$ is nondegenerate.
Then $\ell$ contains two distinct points of $\mathcal K$, so
$\ell\subseteq\operatorname{aff}(\mathcal K)$.  In what follows we
identify $\operatorname{aff}(\mathcal K)$ with a Euclidean space and take
subdifferentials of $f$ relative to this space.  The representative of a
relative subgradient in the direction space of
$\operatorname{aff}(\mathcal K)$ is also an ambient subgradient, because
$f=+\infty$ outside $\operatorname{aff}(\mathcal K)$.  Conversely, any
normal component of an ambient subgradient is orthogonal to differences of
points in $\mathcal K$ and hence does not affect the support inequalities
below.

Choose $\mathbf a\in\operatorname{ri}(\mathcal K)$.  For
$\epsilon\in(0,1)$ define
\[
\tau_\epsilon(t):=(1-\epsilon)(\mathbf b+t\mathbf d)+\epsilon\mathbf a,
\qquad
g_\epsilon(t):=f(\tau_\epsilon(t))
\quad(t\in\overline I),
\]
and extend $g_\epsilon$ by $+\infty$ outside $\overline I$.  We use the
standard relative-interior property of the convex set $\mathcal K$:
\begin{equation}
\label{eq:pc3-ri-segment}
(1-\epsilon)\mathbf z+\epsilon\mathbf a\in\operatorname{ri}(\mathcal K)
\qquad
\text{for all }\mathbf z\in\overline{\mathcal K}\text{ and }\epsilon\in(0,1].
\end{equation}
For $t\in\overline I$ the point $\mathbf b+t\mathbf d$ is a limit of
points of $\mathcal K$, so it lies in $\overline{\mathcal K}$.  Hence
\eqref{eq:pc3-ri-segment} gives
\[
\tau_\epsilon(\overline I)\subseteq\operatorname{ri}(\mathcal K).
\]
Theorem~\ref{thm:gamma-MW-convex-is-locally-Lipschitz} shows that $f$ is
locally Lipschitz, and hence continuous, on $\operatorname{ri}(\mathcal K)$.
Thus $g_\epsilon$ is
continuous on the closed interval $\overline I$, and its extension by
$+\infty$ is proper and lsc, with
$\operatorname{dom}(g_\epsilon)=\overline I$.  If
$t\in\operatorname{int}(I)$, this extension agrees near $t$ with the
composition $f\circ\tau_\epsilon$, and $f$ is locally Lipschitz at
$\tau_\epsilon(t)$.  The derivative of $\tau_\epsilon$ is
$(1-\epsilon)\mathbf d$.  Hence the second part of
Theorem~\ref{thm:chain-rule} shows that every
$p\in\partial g_\epsilon(t)$ has the form
\begin{equation}
\label{eq:pc3-chain-geps}
p=(1-\epsilon)\langle\mathbf v,\mathbf d\rangle
\qquad\text{for some }\mathbf v\in\partial f(\tau_\epsilon(t)).
\end{equation}
For every $s\in\overline I$ we have
\begin{equation}
\label{eq:pc3-tau-diff}
\tau_\epsilon(s)-\tau_\epsilon(t)=(1-\epsilon)(s-t)\mathbf d,
\end{equation}
and both $\tau_\epsilon(s)$ and $\tau_\epsilon(t)$ lie in
$\operatorname{ri}(\mathcal K)\subseteq\operatorname{dom}(f)$.  The vector
$\mathbf v$ in~\eqref{eq:pc3-chain-geps} belongs to
$\partial f(\tau_\epsilon(t))\subseteq\overline{\partial}f(\tau_\epsilon(t))$
by Proposition~\ref{prop:subdiff-basics}-(i).  Consequently, the
$\gamma$--tilted convexity of $f$, followed by \eqref{eq:pc3-tau-diff} and
\eqref{eq:pc3-chain-geps}, gives
\begin{equation}
\label{eq:pc3-geps-support}
g_\epsilon(s)-g_\epsilon(t)
\geq
\psi_\gamma\!\left(
\left\langle\mathbf v,\tau_\epsilon(s)-\tau_\epsilon(t)\right\rangle
\right)
=\psi_\gamma((s-t)p)
\qquad(t\in\operatorname{int}(I),\ p\in\partial g_\epsilon(t),\ s\in\overline I).
\end{equation}
The effective domain $\overline I$ of $g_\epsilon$ is convex,
$\operatorname{ri}(\operatorname{dom}(g_\epsilon))=\operatorname{int}(I)$,
and $\hat\partial g_\epsilon(t)\subseteq\partial g_\epsilon(t)$ by
Proposition~\ref{prop:subdiff-basics}-(i).  Thus
\eqref{eq:pc3-geps-support} is condition~(iv) of
Theorem~\ref{thm:mwc-alternative-characterizations} for $g_\epsilon$, and
$g_\epsilon$ is $\gamma$--tilted convex.

\smallskip
\noindent\emph{Local uniform convergence.}
Fix $t\in I$ and put $\x:=\mathbf b+t\mathbf d$.  If $\x=\mathbf a$, then
$\tau_\epsilon(t)=\mathbf a$ and $g_\epsilon(t)=g(t)$ for every
$\epsilon$.  Otherwise, $\tau_\epsilon(t)=(1-\epsilon)\x+\epsilon\mathbf a$
lies on the segment $[\x,\mathbf a]\subseteq\mathcal K$ and tends to $\x$
as $\epsilon\downarrow0$.  The line through $\x$ and $\mathbf a$ meets
$\operatorname{ri}(\mathcal K)$ at $\mathbf a$, and $f$ is $\gamma$--tilted convex with respect to $\hat\partial f$, whose domain is dense in
$\mathcal K$ by Proposition~\ref{prop:subdiff-basics}-(i).  So
Theorem~\ref{thm:gamma-MW-continuous-on-lines--strong-version} shows that
$f$ is continuous on the intersection of this line with $\mathcal K$.
In both cases,
\begin{equation}
\label{eq:pc3-geps-pointwise}
g_\epsilon(t)\longrightarrow g(t)
\qquad(\epsilon\downarrow0,\ t\in I).
\end{equation}
We next show directly that this convergence is locally uniform on
$\operatorname{int}(I)$.  Fix $t_0\in\operatorname{int}(I)$ and $R>0$
such that
\[
[t_0-3R,t_0+3R]\subseteq I.
\]
Pointwise convergence~\eqref{eq:pc3-geps-pointwise} at $t_0$ and
$t_0\pm2R$ gives $U>0$ such that,
for all sufficiently small $\epsilon$,
\begin{equation}
\label{eq:pc3-U-bound}
|g_\epsilon(t_0)|\leq U,
\qquad
|g_\epsilon(t_0-2R)|,|g_\epsilon(t_0+2R)|\leq U.
\end{equation}
By Theorem~\ref{thm:mwc-domain-is-convex}, $g_\epsilon$ is quasiconvex.
Every point of $[t_0-2R,t_0+2R]$ is a convex combination of $t_0-2R$ and
$t_0+2R$, so \eqref{eq:pc3-U-bound} gives
\begin{equation}
\label{eq:pc3-geps-upper}
g_\epsilon(t)
\leq\max\{g_\epsilon(t_0-2R),g_\epsilon(t_0+2R)\}
\leq U
\qquad(t\in[t_0-2R,t_0+2R]).
\end{equation}
Choose
$p_\epsilon\in\partial g_\epsilon(t_0)$.  Such a subgradient exists by
Proposition~\ref{prop:subdiff-basics}-(ii), because $g_\epsilon$ is locally
Lipschitz at $t_0\in\operatorname{ri}(\operatorname{dom}(g_\epsilon))$ by
Theorem~\ref{thm:gamma-MW-convex-is-locally-Lipschitz}.  If $p_\epsilon\neq0$, put
\[
w_\epsilon:=t_0+R\frac{p_\epsilon}{|p_\epsilon|}.
\]
Then $(w_\epsilon-t_0)p_\epsilon=R|p_\epsilon|>0$, so
$\psi_\gamma((w_\epsilon-t_0)p_\epsilon)=\gamma R|p_\epsilon|$.  Since
$w_\epsilon\in[t_0-R,t_0+R]$, \eqref{eq:pc3-geps-upper} gives
$g_\epsilon(w_\epsilon)\leq U$, and \eqref{eq:pc3-U-bound} gives
$g_\epsilon(t_0)\geq-U$.  Combining these bounds with
\eqref{eq:pc3-geps-support} at $t=t_0$ and $s=w_\epsilon$ gives
\[
2U
\geq g_\epsilon(w_\epsilon)-g_\epsilon(t_0)
\geq\gamma R|p_\epsilon|,
\]
that is,
\begin{equation}
\label{eq:pc3-peps-bound}
|p_\epsilon|\leq\frac{2U}{\gamma R}.
\end{equation}
This bound is automatic if $p_\epsilon=0$.  Next, the definition of
$\psi_\gamma$ gives $\psi_\gamma(z)=z/\gamma$ for $z\leq0$ and
$\psi_\gamma(z)=\gamma z\geq0$ for $z>0$, hence
\begin{equation}
\label{eq:pc3-psi-lower}
\psi_\gamma(z)\geq-\frac{|z|}{\gamma}
\qquad(z\in\mathbb R).
\end{equation}
Let $t\in[t_0-2R,t_0+2R]$, so that $|t-t_0|\leq2R$.  By
\eqref{eq:pc3-geps-support} with $s=t$, then \eqref{eq:pc3-psi-lower},
then \eqref{eq:pc3-U-bound} and \eqref{eq:pc3-peps-bound},
\[
g_\epsilon(t)
\geq g_\epsilon(t_0)
-\frac1\gamma|p_\epsilon|\,|t-t_0|
\geq-U-\frac1\gamma\cdot\frac{2U}{\gamma R}\cdot2R
=-U-\frac{4U}{\gamma^2}.
\]
Together with \eqref{eq:pc3-geps-upper}, and with $M_0:=U+4U/\gamma^2$,
this gives
\begin{equation}
\label{eq:pc3-geps-M0}
|g_\epsilon(t)|\leq M_0
\qquad(t\in[t_0-2R,t_0+2R])
\end{equation}
for all sufficiently small $\epsilon$.  Theorem~\ref{thm:approximate-gamma-supports-local-Lipschitz}
applies to
$g_\epsilon$ at $t_0$ with radius $R$, because the domain
$\overline I$ is convex and contains $[t_0-3R,t_0+3R]$, and $g_\epsilon$ is
$\gamma$--tilted convex with respect to $\hat\partial g_\epsilon$, whose domain
is dense in $\overline I$ by Proposition~\ref{prop:subdiff-basics}-(i).
By \eqref{eq:pc3-geps-M0}, the bound $M$ in that theorem is at most $M_0$.
Therefore, with $L:=2M_0/(\gamma^2R)$,
\begin{equation}
\label{eq:pc3-geps-Lip}
|g_\epsilon(t)-g_\epsilon(t')|\leq L|t-t'|
\qquad(t,t'\in X:=(t_0-R,t_0+R))
\end{equation}
for all sufficiently small $\epsilon$.

Let $\epsilon_k\downarrow0$.  By \eqref{eq:pc3-geps-Lip} and
\eqref{eq:pc3-geps-pointwise}, the restrictions of $g_{\epsilon_k}$ to the
bounded interval $X$ are $L$--Lipschitz for large $k$ and converge
pointwise on $X$ to $g$.  Theorem~\ref{thm:lipschitz-limit} therefore
shows that $g$ is $L$--Lipschitz on $X$ and
\begin{equation}
\label{eq:pc3-geps-unif}
\sup_{t\in X}|g_{\epsilon_k}(t)-g(t)|\longrightarrow0
\qquad(k\to\infty).
\end{equation}
The whole family converges in the same way: if
$\sup_X|g_\epsilon-g|$ did not tend to $0$ as $\epsilon\downarrow0$,
there would be $\eta>0$ and $\epsilon_k\downarrow0$ with
$\sup_X|g_{\epsilon_k}-g|\geq\eta$ for all $k$, contradicting
\eqref{eq:pc3-geps-unif}.  Every compact subset of
$\operatorname{int}(I)$ is covered by finitely many intervals $X$ of this
form.  Hence $g_\epsilon\to g$ locally uniformly on
$\operatorname{int}(I)$.

\smallskip
\noindent\emph{Passing to the limit.}
Fix $t\in\operatorname{int}(I)$, $s\in I$, and
$p\in\partial g(t)$.  Take $t_0:=t$ in the previous paragraph, with the
corresponding $R$, $L$, and $X=(t-R,t+R)$, and fix a sequence
$\epsilon_k\downarrow0$.  Since $X$ is open, the subdifferentials of the
restrictions to $X$ agree with those of $g_{\epsilon_k}$ and $g$ at points
of $X$.  The last part of Theorem~\ref{thm:lipschitz-limit}, applied on
$X$, gives
\begin{equation}
\label{eq:pc3-tk}
t_k\to t,
\qquad
p_k\in\partial g_{\epsilon_k}(t_k),
\qquad
p_k\to p.
\end{equation}
After discarding finitely many terms, $t_k\in X\subseteq\operatorname{int}(I)$,
so \eqref{eq:pc3-geps-support} gives
\begin{equation}
\label{eq:pc3-geps-support-k}
g_{\epsilon_k}(s)-g_{\epsilon_k}(t_k)
\geq\psi_\gamma((s-t_k)p_k).
\end{equation}
We let $k\to\infty$ in \eqref{eq:pc3-geps-support-k}.  At the fixed point
$s$ we have $g_{\epsilon_k}(s)\to g(s)$ by \eqref{eq:pc3-geps-pointwise}.
Since $g$ is $L$--Lipschitz on $X$,
\[
|g_{\epsilon_k}(t_k)-g(t)|
\leq\sup_{X}|g_{\epsilon_k}-g|+L|t_k-t|,
\]
and the right-hand side tends to $0$ by \eqref{eq:pc3-geps-unif} and
\eqref{eq:pc3-tk}.  The right-hand side of
\eqref{eq:pc3-geps-support-k} tends to $\psi_\gamma((s-t)p)$ by
\eqref{eq:pc3-tk} and the continuity of $\psi_\gamma$
(Lemma~\ref{lem:psi_basics}).  We obtain
\begin{equation}
\label{eq:pc3-g-line-support}
g(s)-g(t)\geq\psi_\gamma((s-t)p).
\end{equation}
The function $g$ is proper and lsc, and its effective domain $I$ is
convex with $\operatorname{ri}(I)=\operatorname{int}(I)$.  By
Proposition~\ref{prop:subdiff-basics}-(i),
$\hat{\partial}g(t)\subseteq\partial g(t)$, so
\eqref{eq:pc3-g-line-support} holds for every
$p\in\hat{\partial}g(t)$ with $t\in\operatorname{ri}(I)$ and every
$s\in I$.  This is condition~(iv) of
Theorem~\ref{thm:mwc-alternative-characterizations}, which shows that $g$ is
$\gamma$--tilted convex.

\medskip
\noindent\textbf{Reverse direction.}
Suppose every line restriction of $f$ meeting its
effective domain is $\gamma$--tilted convex.  The effective domain of each
such restriction is convex by Theorem~\ref{thm:mwc-domain-is-convex}.
Thus, for any two points of $\operatorname{dom}(f)$, the line through
them contains the entire segment between them in its effective domain;
hence $\operatorname{dom}(f)$ is convex.  Fix
$\mathbf x\in\operatorname{ri}(\operatorname{dom}(f))$,
$\mathbf y\in\operatorname{dom}(f)$, and
$\mathbf v\in\hat{\partial}f(\mathbf x)$.  If $\mathbf x=\mathbf y$, the
required inequality reads $0\geq\psi_\gamma(0)=0$.  Otherwise, define
\[
\mathbf d:=\frac{\mathbf y-\mathbf x}{\|\mathbf y-\mathbf x\|},
\qquad
q(t):=f(\mathbf x+t\mathbf d).
\]
The function $q$ is the restriction of $f$ to the line through $\mathbf x$
and $\mathbf y$ in a unit-speed parametrization, so it is $\gamma$--tilted convex by assumption.  The first part of Theorem~\ref{thm:chain-rule},
applied with $F(t):=\mathbf x+t\mathbf d$ and $\nabla F(0)=\mathbf d$,
gives
\[
\langle\mathbf v,\mathbf d\rangle
\in\hat{\partial}q(0).
\]
Since
$\hat{\partial}q(0)\subseteq\overline{\partial}q(0)$ by
Proposition~\ref{prop:subdiff-basics}-(i), the $\gamma$--tilted convexity of
$q$ between the points $0$ and $\|\mathbf y-\mathbf x\|$ of its domain
gives
\[
f(\mathbf y)-f(\mathbf x)
=q(\|\mathbf y-\mathbf x\|)-q(0)
\geq
\psi_\gamma\!\left(
\|\mathbf y-\mathbf x\|\langle\mathbf v,\mathbf d\rangle
\right)
=\psi_\gamma(\langle\mathbf v,\mathbf y-\mathbf x\rangle).
\]
This is condition~(iv) of
Theorem~\ref{thm:mwc-alternative-characterizations}, which proves that $f$ is
$\gamma$--tilted convex.
\end{proof}

\subsection{Proof of Theorem~\ref{thm:mwc-lsc-envelope}}
\label{apx:mwc-lsc-envelope}

\begin{proof}
By Theorem~\ref{thm:mwc-domain-is-convex}, $\mathcal K$ is convex.  It is
nonempty because $f$ is proper, so $\op{ri}(\mathcal K)\ne\emptyset$.  Fix
$\x\in\overline{\mathcal K}$.  If $\x\in\op{ri}(\mathcal K)$, then
$f(\x)=g(\x)$ by assumption.  Otherwise, choose
$\y\in\op{ri}(\mathcal K)$ and define
\[
\z_t:=(1-t)\x+t\y,
\qquad t\in(0,1].
\]
By the line segment principle~\eqref{eq:pc1-segment-principle},
$\z_t\in\op{ri}(\mathcal K)$ for every $t\in(0,1]$, and hence
\begin{equation}
f(\z_t)=g(\z_t)
\qquad\forall t\in(0,1].
\label{eq:pc1-env-equal}
\end{equation}
We claim that
\begin{equation}
\lim_{t\downarrow0}f(\z_t)=f(\x)
\qquad\text{and}\qquad
\lim_{t\downarrow0}g(\z_t)=g(\x).
\label{eq:pc1-env-limits}
\end{equation}
We give the argument for $f$; the one for $g$ is the same, since $g$ is
also proper, lsc, and $\gamma$--tilted convex, and $g(\y)=f(\y)\in\mathbb R$.
Let $\ell$ be the line through $\x$ and $\y$; note that $\x\ne\y$ since
$\x\notin\op{ri}(\mathcal K)$.  The line meets $\op{dom}(f)$ at $\y$.
By Theorem~\ref{thm:gamma-MW-convex-on-lines}, the restriction
$f|_\ell$, regarded as a function on a one-dimensional Euclidean space,
is $\gamma$--tilted convex.  It is also proper and lsc.
Theorem~\ref{thm:gamma-MW-continuous-on-lines} therefore shows that
$f|_\ell$ is continuous on its effective domain
$\ell\cap\op{dom}(f)$.  If $\x\in\op{dom}(f)$, the points $\z_t$
belong to $\ell\cap\op{dom}(f)$ and tend to $\x$, so
$f(\z_t)\to f(\x)$.  If $\x\notin\op{dom}(f)$, then $f(\x)=+\infty$, and
lower semicontinuity of $f$ at $\x$ gives
$\liminf_{t\downarrow0}f(\z_t)\ge f(\x)=+\infty$, so again
$f(\z_t)\to f(\x)$.  This proves~\eqref{eq:pc1-env-limits}.
By~\eqref{eq:pc1-env-limits} and~\eqref{eq:pc1-env-equal},
\[
f(\x)
=\lim_{t\downarrow0}f(\z_t)
=\lim_{t\downarrow0}g(\z_t)
=g(\x).
\]
Since $\x\in\overline{\mathcal K}$ was arbitrary, this proves
$f=g$ on $\overline{\mathcal K}$.

For the final assertion, set
\[
h:=\op{lsc}\bigl(f|_{\op{ri}(\mathcal K)}\bigr).
\]
The domain of $f|_{\op{ri}(\mathcal K)}$ is $\op{ri}(\mathcal K)$, and
its closure is $\overline{\mathcal K}$
(Theorem~6.3 in~\cite{rockafellar15_convex_analy}).  By the definition of
the lsc envelope,
\begin{equation}
h(\x)=\liminf_{\w\in\op{ri}(\mathcal K),\ \w\to\x}f(\w)
\quad(\x\in\overline{\mathcal K}),
\qquad
h(\x)=+\infty
\quad(\x\notin\overline{\mathcal K}).
\label{eq:pc1-env-h}
\end{equation}
Let $\x\in\overline{\mathcal K}$.  The lower limit
in~\eqref{eq:pc1-env-h} is taken over the subset
$\op{ri}(\mathcal K)$, so it is at least
$\liminf_{\w\to\x}f(\w)$, which is at least $f(\x)$ by lower
semicontinuity of $f$.  Hence
\begin{equation}
f(\x)\leq h(\x)
\qquad\forall\x\in\overline{\mathcal K}.
\label{eq:pc1-env-lower}
\end{equation}
Now let $\x\in\mathcal K$, choose $\y\in\op{ri}(\mathcal K)$, and set
$\z_t=(1-t)\x+t\y$.  The points $\z_t$ lie in $\op{ri}(\mathcal K)$ for
$t\in(0,1]$ by~\eqref{eq:pc1-segment-principle}, and $\z_t\to\x$ as
$t\downarrow0$.  If $\x\notin\op{ri}(\mathcal K)$,
then~\eqref{eq:pc1-env-limits} gives $f(\z_t)\to f(\x)$.  If
$\x\in\op{ri}(\mathcal K)$, the same limit holds because $f$ is
continuous on $\op{ri}(\mathcal K)$ by
Theorem~\ref{thm:gamma-MW-convex-is-locally-Lipschitz}.
By~\eqref{eq:pc1-env-h}, in both cases
\[
h(\x)
\leq\lim_{t\downarrow0}f(\z_t)
=f(\x).
\]
Together with~\eqref{eq:pc1-env-lower}, this gives $h=f$ on
$\mathcal K$.  If
$\x\in\overline{\mathcal K}\setminus\mathcal K$, then $f(\x)=+\infty$,
and~\eqref{eq:pc1-env-lower} gives $h(\x)=+\infty$.  Finally, both
$f$ and $h$ equal $+\infty$ outside $\overline{\mathcal K}$: the
function $f$ because these points lie outside $\mathcal K$, and $h$
by~\eqref{eq:pc1-env-h}.  Hence
\[
f=\op{lsc}\bigl(f|_{\op{ri}(\mathcal K)}\bigr).
\qedhere
\]
\end{proof}

\subsection{Proof of Theorem~\ref{thm:gamma-MW-convex-pointwise-limit}}
\label{apx:gamma-MW-convex-pointwise-limit}

\begin{proof}
\noindent\textbf{Local uniform convergence.}
Put $A:=\operatorname{aff}(\mathcal K)$ and
$V:=\operatorname{span}(\mathcal K-\mathcal K)$, and identify $A$ with
a Euclidean space.  For $\x_0\in A$ and $r>0$ we write
\[
\mathbb B_V(\x_0,r):=\{\x_0+\w:\w\in V,\ \|\w\|<r\}=\mathbb B(\x_0,r)\cap A
\]
for the open ball of $A$, and $\overline{\mathbb B_V(\x_0,r)}$ for its
closure.  Choose $\mathbf d_0\in\mathcal D$.  Since
$f_n(\mathbf d_0)$ converges to a finite value, $\mathbf d_0$ belongs to
$\operatorname{dom}(f_n)$ for all sufficiently large $n$.  Thus, after
discarding finitely many terms, every restriction $f_n|_A$ is proper and
lsc.  To see that it is $\gamma$--tilted convex, consider any line in $A$
meeting $\operatorname{dom}(f_n)$.  The forward implication of
Theorem~\ref{thm:gamma-MW-convex-on-lines} shows that the restriction of
$f_n$ to this line is $\gamma$--tilted convex.  These are precisely the line
restrictions of $f_n|_A$, so the reverse implication of the same theorem,
applied in the Euclidean space $A$, shows that $f_n|_A$ is
$\gamma$--tilted convex.  We continue to denote these restrictions by $f_n$.

Fix a compact set
$C\subseteq\operatorname{ri}(\mathcal K)$ and a point $\mathbf x_0\in C$.
Since $\operatorname{ri}(\mathcal K)$ is open relative to $A$, there is
$R>0$ with
\begin{equation}
\label{eq:pc3-4R}
\overline{\mathbb B_V(\mathbf x_0,4R)}
\subseteq\operatorname{ri}(\mathcal K).
\end{equation}
We use the density of $\mathcal D$ to choose points
$\y_1,\ldots,\y_m\in\mathcal D$ and
$\mathbf a\in\mathcal D\cap\mathbb B_V(\mathbf x_0,R/2)$ such that
\begin{equation}
\label{eq:pc3-3R-in-P}
\overline{\mathbb B_V(\mathbf x_0,3R)}
\subseteq P:=\operatorname{conv}\{\y_1,\ldots,\y_m\}.
\end{equation}
If $V=\{\mathbf 0\}$, then $\mathcal K=\mathcal D=\{\mathbf x_0\}$, and we
take $m=1$ and $\y_1=\mathbf a=\mathbf x_0$.  Otherwise, the relative
sphere $S:=\{\mathbf x_0+\w:\w\in V,\ \|\w\|=4R\}$ is compact and lies in
$\mathcal K$ by~\eqref{eq:pc3-4R}.  Cover $S$ by finitely many open balls
of radius $R/4$ centered at points $\mathbf q_1,\ldots,\mathbf q_m\in S$,
and choose $\y_i\in\mathcal D$ with $\|\y_i-\mathbf q_i\|<R/4$.  Choose
$\mathbf a\in\mathcal D\cap\mathbb B_V(\mathbf x_0,R/2)$, which is possible
because $\mathbf x_0\in\mathcal K$.  Let $\uu\in V$ with $\|\uu\|=1$.  The point
$\mathbf x_0+4R\uu\in S$ lies within $R/4$ of some $\mathbf q_i$, hence
within $R/2$ of $\y_i$, and therefore
\begin{equation}
\label{eq:pc3-support-P}
\max_{1\leq i\leq m}\langle\y_i-\mathbf x_0,\uu\rangle
>4R-\frac R2>3R.
\end{equation}
If some $\z\in\overline{\mathbb B_V(\mathbf x_0,3R)}$ were not in the
compact convex set $P\subseteq A$, separation in $A$ would give a unit
vector $\uu\in V$ with
$\langle\z-\mathbf x_0,\uu\rangle>\max_i\langle\y_i-\mathbf x_0,\uu\rangle$.
By \eqref{eq:pc3-support-P} this would force
$\|\z-\mathbf x_0\|>3R$, which is impossible.  This proves
\eqref{eq:pc3-3R-in-P}.

The relatively open balls $\mathbb B_V(\mathbf x_0,R)$, $\mathbf x_0\in C$,
cover $C$, and a
finite subcollection still covers $C$.  It therefore suffices to work on
one such ball
\[
B:=\mathbb B_V(\mathbf x_0,R),
\qquad
K_0:=\overline{\mathbb B_V(\mathbf x_0,2R)}.
\]
Since $f_n$ converges to finite values at the selected points, after
discarding finitely many terms there is $M>0$ such that
\begin{equation}
\label{eq:pc3-M}
|f_n(\mathbf a)|\leq M,
\qquad
|f_n(\y_i)|\leq M\quad(1\leq i\leq m).
\end{equation}
For these indices, the points $\y_i$ lie in $\operatorname{dom}(f_n)$,
which is convex by Theorem~\ref{thm:mwc-domain-is-convex}, so
$P\subseteq\operatorname{dom}(f_n)$.  The same theorem shows that $f_n$ is
quasiconvex.  Applying quasiconvexity repeatedly to a convex
combination of $\y_1,\ldots,\y_m$ and using \eqref{eq:pc3-M} gives
\begin{equation}
\label{eq:pc3-fn-upper}
f_n(\z)\leq\max_{1\leq i\leq m}f_n(\y_i)\leq M
\qquad\forall\z\in P.
\end{equation}
By \eqref{eq:pc3-3R-in-P}, the set $\operatorname{dom}(f_n)\subseteq A$
contains the ball $\mathbb B_V(\mathbf x_0,3R)$, which is open in $A$.
Hence $A$ is the affine hull of $\operatorname{dom}(f_n)$, and the point
$\mathbf a$ lies in $\operatorname{ri}(\operatorname{dom}(f_n))$.  The function
$f_n$ is locally Lipschitz at $\mathbf a$ by
Theorem~\ref{thm:gamma-MW-convex-is-locally-Lipschitz}, so
Proposition~\ref{prop:subdiff-basics}-(ii) provides
$\mathbf p_n\in\partial f_n(\mathbf a)\subseteq V$.  If
$\mathbf p_n\neq0$, put
\[
\w_n:=\mathbf a+R\frac{\mathbf p_n}{\|\mathbf p_n\|}.
\]
Then $\|\w_n-\mathbf x_0\|\leq\|\mathbf a-\mathbf x_0\|+R<3R/2$, so
$\w_n\in P$ by \eqref{eq:pc3-3R-in-P}.  Moreover,
$\langle\mathbf p_n,\w_n-\mathbf a\rangle=R\|\mathbf p_n\|>0$, so
$\psi_\gamma(\langle\mathbf p_n,\w_n-\mathbf a\rangle)=\gamma R\|\mathbf p_n\|$.
The support inequality at $\mathbf a$, together with
\eqref{eq:pc3-fn-upper} at $\z=\w_n$ and \eqref{eq:pc3-M}, gives
\[
2M
\geq f_n(\w_n)-f_n(\mathbf a)
\geq\gamma R\|\mathbf p_n\|,
\]
that is,
\begin{equation}
\label{eq:pc3-pn-bound}
\|\mathbf p_n\|\leq\frac{2M}{\gamma R}.
\end{equation}
This bound is automatic when
$\mathbf p_n=0$.  Let $\z\in K_0$.  Then $\z\in P$ by
\eqref{eq:pc3-3R-in-P}, and
$\|\z-\mathbf a\|\leq\|\z-\mathbf x_0\|+\|\mathbf x_0-\mathbf a\|\leq5R/2$.
The support inequality at $\mathbf a$, followed by
\eqref{eq:pc3-psi-lower}, the Cauchy--Schwarz inequality,
\eqref{eq:pc3-M}, and \eqref{eq:pc3-pn-bound}, gives
\[
f_n(\z)
\geq f_n(\mathbf a)+\psi_\gamma(\langle\mathbf p_n,\z-\mathbf a\rangle)
\geq f_n(\mathbf a)
-\frac{1}{\gamma}\|\mathbf p_n\|\,\|\z-\mathbf a\|
\geq-M-\frac1\gamma\cdot\frac{2M}{\gamma R}\cdot\frac{5R}{2}
=-M-\frac{5M}{\gamma^2}.
\]
Combining this with \eqref{eq:pc3-fn-upper}, and with
$M_0:=M+5M/\gamma^2$, we get
\begin{equation}
\label{eq:pc3-M0}
\sup_{\z\in K_0}|f_n(\z)|\leq M_0.
\end{equation}
Because $\overline{\mathbb B_V(\mathbf x_0,3R)}$ is contained in the
effective domain of $f_n$, and $f_n$ is $\gamma$--tilted convex with respect
to $\hat\partial f_n$, whose domain is dense in $\operatorname{dom}(f_n)$
by Proposition~\ref{prop:subdiff-basics}-(i),
Theorem~\ref{thm:approximate-gamma-supports-local-Lipschitz} applies with
center $\mathbf x_0$ and radius $R$.  By \eqref{eq:pc3-M0}, the bound $M$
in that theorem is at most $M_0$, so it gives the common Lipschitz bound
\begin{equation}
\label{eq:pc3-fn-Lip}
|f_n(\x)-f_n(\y)|
\leq\frac{2M_0}{\gamma^2R}\|\x-\y\|
\qquad\forall\x,\y\in B.
\end{equation}
The ball $B$ is open in $A$ and contained in $\mathcal K$, and
$\mathcal D$ is dense in $\mathcal K$; hence $\mathcal D\cap B$ is dense
in $B$.  Since $f_n$ converges on $\mathcal D\cap B$,
Theorem~\ref{thm:lipschitz-limit} gives a function $f_B$ on $B$,
Lipschitz with the constant in \eqref{eq:pc3-fn-Lip}, such that
\begin{equation}
\label{eq:pc3-unif-B}
f_n\to f_B\quad\text{uniformly on }B.
\end{equation}
Every $\x\in\operatorname{ri}(\mathcal K)$ lies in such a ball (take
$C=\{\x\}$), and pointwise limits are unique.  Hence
$f(\x):=\lim_nf_n(\x)$ defines a function
$f:\operatorname{ri}(\mathcal K)\to\mathbb R$ with $f=f_B$ on each
ball $B$ constructed above.  In particular, $f$ is Lipschitz on a
neighborhood in $A$ of each point, and so it is continuous.  Applying
\eqref{eq:pc3-unif-B}
to the finite collection of balls covering $C$ proves uniform convergence
of $f_n$ to $f$ on $C$.  This proves~(i).

\medskip
\noindent\textbf{$\gamma$--tilted convexity of $\op{lsc}(f)$.}
Extend $f$ by $+\infty$ outside $\operatorname{ri}(\mathcal K)$ and set
\[
\bar f:=\op{lsc}(f).
\]
At $\x\in\op{ri}(\mathcal K)$, only points of $\op{ri}(\mathcal K)$
contribute finite values to $\liminf_{\y\to\x}f(\y)$, and $f$ is
continuous there; hence $\bar f=f$ on $\op{ri}(\mathcal K)$.
We first verify the support inequality on $\operatorname{ri}(\mathcal K)$,
taking all subdifferentials relative to $A$ until stated otherwise.
Fix
\[
\mathbf x,\mathbf y\in\operatorname{ri}(\mathcal K),
\qquad
\mathbf v\in\partial \bar f(\mathbf x)=\partial f(\mathbf x),
\]
where the equality holds because $\bar f=f$ on the set
$\operatorname{ri}(\mathcal K)$, which is open in $A$.  Apply the
construction of the first part with $\mathbf x_0:=\mathbf x$.  It gives a
ball $B=\mathbb B_V(\mathbf x,R)$ on which, after discarding finitely many
terms, the functions $f_n$ share the Lipschitz constant in
\eqref{eq:pc3-fn-Lip} and converge uniformly to $f$ by
\eqref{eq:pc3-unif-B}.  By the last part of
Theorem~\ref{thm:lipschitz-limit}, applied on $B$, there exist
\begin{equation}
\label{eq:pc3-xn-vn}
\mathbf x_n\to\mathbf x,
\qquad
\mathbf v_n\in\partial f_n(\mathbf x_n),
\qquad
\mathbf v_n\to\mathbf v.
\end{equation}
For large $n$ we have $\mathbf x_n\in B$ and
$\mathbf y\in\operatorname{dom}(f_n)$, because $f_n(\mathbf y)\to
f(\mathbf y)\in\mathbb R$.  Moreover,
$\mathbf v_n\in\overline{\partial}f_n(\mathbf x_n)$ by
Proposition~\ref{prop:subdiff-basics}-(i).  Since each $f_n$ is
$\gamma$--tilted convex,
\begin{equation}
\label{eq:pc3-fn-support}
f_n(\mathbf y)-f_n(\mathbf x_n)
\ge
\psi_\gamma\!\left(
\langle \mathbf v_n,\mathbf y-\mathbf x_n\rangle
\right).
\end{equation}
We let $n\to\infty$ in \eqref{eq:pc3-fn-support}.  First,
$f_n(\mathbf y)\to f(\mathbf y)$.  Second,
\[
|f_n(\mathbf x_n)-f(\mathbf x)|
\leq\sup_{B}|f_n-f|+|f(\mathbf x_n)-f(\mathbf x)|\longrightarrow0
\]
by \eqref{eq:pc3-unif-B}, \eqref{eq:pc3-xn-vn}, and the continuity of
$f$.  Third, the right-hand side converges by \eqref{eq:pc3-xn-vn} and
the continuity of $\psi_\gamma$.  This yields
\begin{equation}
\label{eq:pc3-f-support}
f(\mathbf y)-f(\mathbf x)
\ge
\psi_\gamma\!\left(
\langle \mathbf v,\mathbf y-\mathbf x\rangle
\right)
\qquad(\mathbf x,\mathbf y\in\operatorname{ri}(\mathcal K),\
\mathbf v\in\partial f(\mathbf x)).
\end{equation}
This inequality also shows that $\bar f$ does not take the value $-\infty$.
Fix $\mathbf x\in\operatorname{ri}(\mathcal K)$ and
$\mathbf v\in\partial f(\mathbf x)$; such $\mathbf v$ exists by
Proposition~\ref{prop:subdiff-basics}-(ii), since $f$ is Lipschitz near
$\mathbf x$.  Let $\z\in\mathbb R^d$.  If
$\z\notin\overline{\mathcal K}$, some neighborhood of $\z$ misses
$\operatorname{ri}(\mathcal K)$, so $\bar f(\z)=+\infty$.  If
$\z\in\overline{\mathcal K}$, then $\bar f(\z)$ is the $\liminf$ of
$f(\y)$ as $\y\to\z$ with $\y\in\operatorname{ri}(\mathcal K)$, and
\eqref{eq:pc3-f-support} together with the continuity of $\psi_\gamma$
gives
\[
\bar f(\z)\geq f(\mathbf x)+\psi_\gamma(\langle\mathbf v,\z-\mathbf x\rangle)
>-\infty.
\]
Hence $\bar f$ is proper, since it is finite on the nonempty set
$\operatorname{ri}(\mathcal K)$, and it is lsc by construction.

Each $f_n$ is quasiconvex by
Theorem~\ref{thm:mwc-domain-is-convex}.  If
$\x',\y'\in\operatorname{ri}(\mathcal K)$ and $t\in[0,1]$, then
$(1-t)\x'+t\y'\in\operatorname{ri}(\mathcal K)$, and letting $n\to\infty$
in $f_n((1-t)\x'+t\y')\leq\max\{f_n(\x'),f_n(\y')\}$ gives
\begin{equation}
\label{eq:pc3-f-qc}
f((1-t)\x'+t\y')\leq\max\{f(\x'),f(\y')\}.
\end{equation}
Thus $f$ is quasiconvex on $\operatorname{ri}(\mathcal K)$.  Consequently,
$\operatorname{dom}(\bar f)$ is convex: if
$\mathbf a,\mathbf b\in\operatorname{dom}(\bar f)$, then, since $f=+\infty$
outside $\operatorname{ri}(\mathcal K)$, we can choose sequences
$\mathbf a_k,\mathbf b_k\in\operatorname{ri}(\mathcal K)$ such that
\[
\mathbf a_k\to\mathbf a,\quad f(\mathbf a_k)\to \bar f(\mathbf a),
\qquad
\mathbf b_k\to\mathbf b,\quad f(\mathbf b_k)\to \bar f(\mathbf b).
\]
Then, for
$t\in[0,1]$, the points $(1-t)\mathbf a_k+t\mathbf b_k$ converge to
$(1-t)\mathbf a+t\mathbf b$.  The lsc envelope is bounded by the
$\liminf$ along any convergent sequence, so \eqref{eq:pc3-f-qc} gives
\[
\bar f((1-t)\mathbf a+t\mathbf b)
\leq
\liminf_{k\to\infty}
f((1-t)\mathbf a_k+t\mathbf b_k)
\leq
\liminf_{k\to\infty}\max\{f(\mathbf a_k),f(\mathbf b_k)\}
=
\max\{\bar f(\mathbf a),\bar f(\mathbf b)\}<+\infty.
\]
Moreover, if $\mathbf y\in\operatorname{dom}(\bar f)$, choose
$\mathbf y_k\in\operatorname{ri}(\mathcal K)$ such that
$\mathbf y_k\to\mathbf y$ and $f(\mathbf y_k)\to \bar f(\mathbf y)$.
Applying \eqref{eq:pc3-f-support} at $\mathbf y_k$, letting
$k\to\infty$, and using $\bar f(\mathbf x)=f(\mathbf x)$ gives
\begin{equation}
\label{eq:pc3-fbar-support}
\bar f(\mathbf y)-\bar f(\mathbf x)
\geq
\psi_\gamma\!\left(
\langle\mathbf v,\mathbf y-\mathbf x\rangle
\right)
\end{equation}
for every $\mathbf x\in\operatorname{ri}(\mathcal K)$ and
$\mathbf v\in\partial \bar f(\mathbf x)$.  We have
$\operatorname{ri}(\mathcal K)\subseteq\operatorname{dom}(\bar f)
\subseteq\overline{\mathcal K}$, and $\operatorname{dom}(\bar f)$ is
convex.  A convex set lying between a convex set and its closure has the
same relative interior, so
$\operatorname{ri}(\operatorname{dom}(\bar f))=\operatorname{ri}(\mathcal K)$.
Therefore \eqref{eq:pc3-fbar-support}
also verifies the ambient support inequality in
Theorem~\ref{thm:mwc-alternative-characterizations}-(iv).  To check this, let
$\mathbf u\in\hat{\partial}\bar f(\mathbf x)$ be an ambient regular
subgradient and $\mathbf y\in\operatorname{dom}(\bar f)$.  Since
$\operatorname{dom}(\bar f)\subseteq A$, we have
$\mathbf y-\mathbf x\in V$, and hence
\begin{equation}
\label{eq:pc3-proj}
\langle\mathbf u,\mathbf y-\mathbf x\rangle
=\langle P_V\mathbf u,\mathbf y-\mathbf x\rangle.
\end{equation}
The definition of $\hat\partial\bar f(\mathbf x)$ involves only points of
$\operatorname{dom}(\bar f)$, so \eqref{eq:pc3-proj} shows that $P_V\mathbf u$
is a relative regular subgradient, and it lies in
$\partial\bar f(\mathbf x)$ by Proposition~\ref{prop:subdiff-basics}-(i).
Then \eqref{eq:pc3-fbar-support} with $\mathbf v=P_V\mathbf u$ and
\eqref{eq:pc3-proj} give
$\bar f(\mathbf y)-\bar f(\mathbf x)\geq
\psi_\gamma(\langle\mathbf u,\mathbf y-\mathbf x\rangle)$.
Theorem~\ref{thm:mwc-alternative-characterizations} shows that $\bar f$ is
$\gamma$--tilted convex.  This proves~(ii).

\medskip
\noindent\textbf{The additional pointwise limit.}
Finally, consider the case where $f_n(\x) \to g(\x)$ for all $\x \in \C{K}$ for some lsc function $g$.
Both $g$ and $f$ are pointwise limits of $f_n$ on $\operatorname{ri}(\mathcal K)$, so
$g=f$ on $\operatorname{ri}(\mathcal K)$.  Extend $g$ by $+\infty$
outside $\mathcal K$ and set
\[
\bar g:=\op{lsc}(g).
\]
Because $g$ is lsc on $\mathcal K$, we have $\bar g=g$ on $\mathcal K$;
in particular, $\bar g=f=\bar f$ on $\operatorname{ri}(\mathcal K)$.

\medskip
\noindent\textbf{$\gamma$--tilted convexity of $\op{lsc}(g)$.}
Fix
\[
\mathbf x \in \operatorname{ri}(\mathcal K),
\qquad
\mathbf v\in\partial \bar g(\mathbf x)=\partial f(\mathbf x),
\]
where the equality holds because $\bar g=f$ on the set
$\operatorname{ri}(\mathcal K)$, which is open in $A$.  The sequences
$\mathbf x_n$ and $\mathbf v_n$ in \eqref{eq:pc3-xn-vn} depend only on
$\mathbf x$, on $\mathbf v\in\partial f(\mathbf x)$, and on the
convergence $f_n\to f$ near $\mathbf x$, so they are available here.
For any $\mathbf y\in\mathcal K$ with $g(\mathbf y)<+\infty$, we have
$f_n(\mathbf y)\to g(\mathbf y)<+\infty$, so
$\mathbf y\in\operatorname{dom}(f_n)$ for all sufficiently large $n$.
For these $n$, the $\gamma$--tilted convexity of $f_n$ gives
\eqref{eq:pc3-fn-support}.  We let $n\to\infty$ in
\eqref{eq:pc3-fn-support}, using pointwise convergence at $\mathbf y$,
the limit $f_n(\mathbf x_n)\to f(\mathbf x)=g(\mathbf x)$ obtained above,
\eqref{eq:pc3-xn-vn}, and the continuity of $\psi_\gamma$.  This yields
\begin{equation}
\label{eq:pc3-g-support}
g(\mathbf y)-g(\mathbf x)
\ge
\psi_\gamma\!\left(
\langle \mathbf v,\mathbf y-\mathbf x\rangle
\right),
\end{equation}
which in particular rules out $g(\mathbf y)=-\infty$, since the
right-hand side is finite.  Here
$g(\mathbf x)=f(\mathbf x)=\bar g(\mathbf x)$.  If
$\mathbf y\in\operatorname{dom}(\bar g)$, then, since $g=+\infty$ outside
$\operatorname{dom}(g)$, we can choose
$\mathbf y_k\in\operatorname{dom}(g)$ such that
$\mathbf y_k\to\mathbf y$ and $g(\mathbf y_k)\to\bar g(\mathbf y)$.
Applying \eqref{eq:pc3-g-support} at $\mathbf y_k$ and letting
$k\to\infty$, we obtain
\begin{equation}
\label{eq:pc3-gbar-support}
\bar g(\mathbf y)-\bar g(\mathbf x)
\ge
\psi_\gamma\!\left(
\langle\mathbf v,\mathbf y-\mathbf x\rangle
\right).
\end{equation}

Next we verify the domain condition in
Theorem~\ref{thm:mwc-alternative-characterizations}.  Let
$\y',\z'\in\operatorname{dom}(g)$ and $t\in[0,1]$, and put
$\w:=(1-t)\y'+t\z'\in\mathcal K$.  Letting $n\to\infty$ in
$f_n(\w)\leq\max\{f_n(\y'),f_n(\z')\}$ gives
$g(\w)\leq\max\{g(\y'),g(\z')\}<+\infty$.  Thus $g$ is quasiconvex and
$\operatorname{dom}(g)$ is convex.  The sequence argument
used for $\bar f$, with $\operatorname{ri}(\mathcal K)$ replaced by
$\operatorname{dom}(g)$, then shows that $\operatorname{dom}(\bar g)$ is convex.
The function $\bar g$ cannot take the value $-\infty$:
for fixed $\mathbf x$ and $\mathbf v$ as above,
\eqref{eq:pc3-g-support} bounds $g(\y)$ below by
$g(\mathbf x)+\psi_\gamma(\langle\mathbf v,\y-\mathbf x\rangle)$, and this
bound stays bounded below as $\mathbf y$ approaches any fixed
point of $\overline{\mathcal K}$; outside $\overline{\mathcal K}$ we have
$\bar g=+\infty$.
Thus $\bar g$ is proper and lsc.  Moreover,
\[
\operatorname{ri}(\operatorname{dom}(\bar g))
=\operatorname{ri}(\mathcal K),
\]
because
$\operatorname{ri}(\mathcal K)\subseteq\operatorname{dom}(\bar g)
\subseteq\overline{\mathcal K}$ and $\operatorname{dom}(\bar g)$ is convex.
As for $\bar f$, projecting ambient
regular subgradients onto $V$ as in \eqref{eq:pc3-proj} shows that
\eqref{eq:pc3-gbar-support}
verifies condition~(iv) of
Theorem~\ref{thm:mwc-alternative-characterizations}.  Hence $\bar g$ is
$\gamma$--tilted convex.

\medskip
\noindent\textbf{Equality of the lsc envelopes.}
The functions $\bar f$ and $\bar g$ are therefore proper, lsc,
$\gamma$--tilted convex, and agree on $\operatorname{ri}(\mathcal K)$.  Since
$\operatorname{ri}(\mathcal K)\subseteq\operatorname{dom}(\bar f)
\subseteq\overline{\mathcal K}$, we have
\[
\operatorname{ri}(\operatorname{dom}(\bar f))=\operatorname{ri}(\mathcal K),
\qquad
\overline{\operatorname{dom}(\bar f)}=\overline{\mathcal K}.
\]
Theorem~\ref{thm:mwc-lsc-envelope}, applied to the pair $\bar f,\bar g$,
therefore gives $\bar g=\bar f$ on
$\overline{\mathcal K}$.  Moreover, both $\bar f=\op{lsc}(f)$ and
$\bar g=\op{lsc}(g)$ equal $+\infty$ outside
$\overline{\mathcal K}$, because $f$ and $g$ equal $+\infty$ outside
$\mathcal K$.  Hence
\[
\op{lsc}(g)=\bar g=\bar f=\op{lsc}(f)
\qquad\text{on }\mathbb R^d,
\]
as claimed.
\end{proof}

\subsection{Proof of Theorem~\ref{thm:sup-closure}}
\label{apx:sup-closure}

\begin{proof}
\noindent\textbf{Two functions on the real line.}
Let $f,g:\R\to\overline{\R}$ belong to $\mathcal F^\gamma$ and suppose that $h:=\max\{f,g\}$ is proper. The epigraph of $h$ is the intersection of the closed epigraphs of $f$ and $g$, so $h$ is lsc. Its domain $I:=\op{dom}(h)=\op{dom}(f)\cap\op{dom}(g)$ is non-empty, and it is an interval because $\op{dom}(f)$ and $\op{dom}(g)$ are intervals by Theorem~\ref{thm:mwc-domain-is-convex}. If $I=\{x\}$ is a single point, then the only pair in \eqref{eq:gamma-mwc} is $y=x$, for which both sides vanish, so $h\in\mathcal F^\gamma$.

Assume that $I$ has non-empty interior. Then $\op{ri}(I)=\op{int}(I)=\op{int}(\op{dom}(f))\cap\op{int}(\op{dom}(g))$. By Theorem~\ref{thm:gamma-MW-convex-is-locally-Lipschitz}, $f$ and $g$ are locally Lipschitz on the interiors of their domains, so $h$ is locally Lipschitz on $\op{int}(I)$. A locally Lipschitz function on an open interval is differentiable almost everywhere, so the points of $\op{int}(I)$ at which $h'$ exists are dense in $\op{int}(I)$ and hence in $I$.

Fix $x\in\op{int}(I)$ at which $h'(x)$ exists, and fix $y\in I$. Choose $\tau>0$ with $[x,x+\tau]\subseteq\op{int}(I)$ and put $t_n:=\tau/n$. For each $n$, the value $h(x+t_n)$ equals $f(x+t_n)$ or $g(x+t_n)$. After exchanging $f$ and $g$ and passing to a subsequence, we may assume that
\begin{equation}\label{eq:sup-active}
h(x+t_n)=f(x+t_n)\qquad\text{for all }n .
\end{equation}
The functions $f$ and $h$ are continuous at $x$, so letting $n\to\infty$ in \eqref{eq:sup-active} gives $h(x)=f(x)$. Together with \eqref{eq:sup-active}, this gives
\begin{equation}\label{eq:sup-quotient}
\frac{f(x+t_n)-f(x)}{t_n}=\frac{h(x+t_n)-h(x)}{t_n}\ \longrightarrow\ h'(x)\qquad(n\to\infty).
\end{equation}
The compact segment $[x,x+\tau]$ lies in the open interval $\op{int}(\op{dom}(f))$, on which $f$ is locally Lipschitz, so $f$ is Lipschitz on an open neighborhood of $[x,x+\tau]$. Theorem~\ref{thm:Lipschitz-mean-value} on $[x,x+t_n]$ gives $\xi_n\in(x,x+t_n)$ and $w_n\in\overline\partial f(\xi_n)$ with
\begin{equation}\label{eq:sup-lebourg}
f(x+t_n)-f(x)=w_n\,t_n ,
\end{equation}
where we use that the Clarke subdifferential at an interior point of the domain depends only on the values of $f$ near that point. Since $f$ is $\gamma$--tilted convex and $\xi_n,y\in\op{dom}(f)$,
\begin{equation}\label{eq:sup-at-xi}
f(y)-f(\xi_n)\ \ge\ \psi_\gamma\bigl(w_n(y-\xi_n)\bigr).
\end{equation}
By \eqref{eq:sup-quotient} and \eqref{eq:sup-lebourg}, $w_n\to h'(x)$. Moreover $\xi_n\to x$, $f(\xi_n)\to f(x)$ by continuity of $f$ at $x$, and $\psi_\gamma$ is continuous by Lemma~\ref{lem:psi_basics}. Letting $n\to\infty$ in \eqref{eq:sup-at-xi} and using $h\ge f$ and $h(x)=f(x)$, we obtain
\begin{equation}\label{eq:sup-h-ineq}
h(y)-h(x)\ \ge\ f(y)-f(x)\ \ge\ \psi_\gamma\bigl(h'(x)(y-x)\bigr).
\end{equation}
The domain $I$ is convex, the points of $\op{ri}(I)$ at which $h'$ exists are dense in $I$, and \eqref{eq:sup-h-ineq} holds for each such point $x$ and every $y\in I$. This is condition~(v) of Theorem~\ref{thm:mwc-alternative-characterizations}, so $h\in\mathcal F^\gamma$.

\medskip
\noindent\textbf{Finitely many functions on $\R^d$.}
Let $f,g\in\mathcal F^\gamma$ be functions on $\R^d$ such that $h:=\max\{f,g\}$ is proper. As before, $h$ is lsc. Let $\ell$ be a line that meets $\op{dom}(h)=\op{dom}(f)\cap\op{dom}(g)$. Then $\ell$ meets $\op{dom}(f)$ and $\op{dom}(g)$, and the forward implication of Theorem~\ref{thm:gamma-MW-convex-on-lines} shows that $f|_\ell$ and $g|_\ell$ are $\gamma$--tilted convex. Both restrictions are proper and lsc, so they belong to $\mathcal F^\gamma$ on the line $\ell$. Their maximum $h|_\ell$ is finite at a point of $\ell$, so it is proper, and the previous paragraphs show that $h|_\ell$ is $\gamma$--tilted convex. The reverse implication of Theorem~\ref{thm:gamma-MW-convex-on-lines} then gives $h\in\mathcal F^\gamma$. Let $f_1,\ldots,f_m\in\mathcal F^\gamma$ have a proper maximum. For $k\le m$, the maximum $\max\{f_1,\ldots,f_k\}$ lies between $f_1>-\infty$ and $\max\{f_1,\ldots,f_m\}$, so it is also proper, and induction on $k$ shows that
\begin{equation}\label{eq:sup-finite}
\max\{f_1,\ldots,f_m\}\in\mathcal F^\gamma .
\end{equation}

\medskip
\noindent\textbf{Countable families.}
Let $(f_n)_{n\ge1}\subseteq\mathcal F^\gamma$ have a proper supremum $s$. Set $s_n:=\max\{f_1,\ldots,f_n\}$. Each $s_n$ is proper, since $f_1\le s_n\le s$, so $s_n\in\mathcal F^\gamma$ by \eqref{eq:sup-finite}. Let $\mathcal K:=\op{dom}(s)$, which is non-empty. To see that $\mathcal K$ is convex, let $\y,\z\in\mathcal K$ and $\w\in[\y,\z]$. Each $s_n$ is quasiconvex by Theorem~\ref{thm:mwc-domain-is-convex}, so
\begin{equation}\label{eq:sup-quasiconvex}
s_n(\w)\le\max\{s_n(\y),s_n(\z)\}\le\max\{s(\y),s(\z)\}<+\infty ,
\end{equation}
and taking the supremum over $n$ in \eqref{eq:sup-quasiconvex} gives $s(\w)<+\infty$. For $\x\in\mathcal K$ we have $f_1(\x)\le s_n(\x)\le s(\x)<+\infty$, so the values $s_n(\x)$ are finite and increase to $s(\x)$. The function $g:=s|_{\mathcal K}$ is lsc on $\mathcal K$, because $s$ is lsc as a supremum of lsc functions. Theorem~\ref{thm:gamma-MW-convex-pointwise-limit}, applied to the sequence $(s_n)$ with the convex set $\mathcal K$ and the dense subset $\mathcal D:=\mathcal K$, shows in its final assertion that $\op{lsc}(g)$ is $\gamma$--tilted convex. The proper extension of $g$ is $s$, which is lsc, so $\op{lsc}(g)=s$. Hence $s\in\mathcal F^\gamma$.

\medskip
\noindent\textbf{Arbitrary families.}
Let $\{f_i\}_{i\in I}\subseteq\mathcal F^\gamma$ have a proper supremum $s$. For each $i$ the set $U_i:=\R^{d+1}\setminus\op{epi}(f_i)$ is open, because $f_i$ is lsc, and
\begin{equation}\label{eq:sup-epi}
\R^{d+1}\setminus\op{epi}(s)=\bigcup_{i\in I}U_i ,
\end{equation}
because $\op{epi}(s)=\bigcap_{i\in I}\op{epi}(f_i)$. Let $\mathcal B$ be the countable collection of open balls in $\R^{d+1}$ with rational centers and rational radii. For each ball $B\in\mathcal B$ that lies in some $U_i$, choose one such index $i_B$. Every point of an open set $U_i$ lies in a ball of $\mathcal B$ contained in $U_i$, so the countably many sets $U_{i_B}$ also cover the set in \eqref{eq:sup-epi}. This set is non-empty, since it contains $(\x,s(\x)-1)$ for any $\x\in\op{dom}(s)$, so at least one index $i_B$ is chosen. Enumerate the chosen indices as $i_1,i_2,\ldots$, repeating one index if there are only finitely many. Then $\op{epi}(s)=\bigcap_n\op{epi}(f_{i_n})=\op{epi}(\sup_nf_{i_n})$, so $s=\sup_nf_{i_n}$, and the countable case gives $s\in\mathcal F^\gamma$.
\end{proof}

\subsection{Proof of Theorem~\ref{thm:mwc-vs-mw-mono}}
\label{apx:mwc-vs-mw-mono}

\begin{proof}

\noindent\textbf{(i):}
Assume that $\hat{\partial} f$ is $\gamma$--tilted monotone. 
Fix $\mathbf{x},\mathbf{y}\in\operatorname{dom}(f)$ and
$\mathbf{x}^*\in\hat{\partial}f(\mathbf{x})$.  If $\mathbf x=\mathbf y$,
both sides of \eqref{eq:gamma-mwc} are $0$, so suppose $\x \neq \y$.  By
Theorem~\ref{thm:approx-mean-value}, there exist $\lambda \in [0,1)$ and
sequences $\x_k \to \lambda \x + (1-\lambda) \y$ and
$\v_k \in \hat{\partial} f(\x_k)$ such that
\begin{equation}
\label{eq:pb-i-meanvalue}
f(\mathbf y)-f(\mathbf x)
\;\ge\;
\frac{1}{1 - \lambda}\limsup_{k\to\infty}
\langle \mathbf v_k,\x_k - \x\rangle.
\end{equation}
Since $\x^* \in \hat\partial f(\x)$ and $\v_k \in \hat\partial f(\x_k)$, inequality
\eqref{eq:gamma-mw-mono} for $\hat{\partial} f$ with $(\y,\y^*)$ replaced by $(\x_k,\v_k)$ gives
\begin{equation}
\label{eq:pb-i-mono}
\langle \mathbf v_k,\x_k - \x\rangle
\;\ge\;
\psi_\gamma\!\left(
\langle \mathbf{x}^*,\x_k - \x\rangle
\right)
\qquad \forall k.
\end{equation}
The limit of $\x_k$ is $\lambda \x + (1-\lambda) \y = \x + (1-\lambda)(\y-\x)$, so
$\langle \x^*, \x_k - \x \rangle \to (1-\lambda)\langle \x^*, \y - \x \rangle$.
Since $\psi_\gamma$ is continuous (Lemma~\ref{lem:psi_basics}(1)), the right-hand side of \eqref{eq:pb-i-mono} converges to $\psi_\gamma\bigl((1-\lambda)\langle \x^*, \y - \x \rangle\bigr)$.
Combining \eqref{eq:pb-i-meanvalue} and \eqref{eq:pb-i-mono}, we obtain
\begin{align*}
f(\mathbf y)-f(\mathbf x)
&\ge
\frac{1}{1 - \lambda}\limsup_{k\to\infty}
\langle \mathbf v_k,\x_k - \x\rangle \\
&\ge
\frac{1}{1 - \lambda}\limsup_{k\to\infty}
\psi_\gamma(\langle \mathbf{x}^*,\x_k - \x\rangle) \\
&=
\frac{1}{1 - \lambda}
\psi_\gamma(\langle \mathbf{x}^*,(1-\lambda)(\y-\x)\rangle) \\
&=
\psi_\gamma(\langle \mathbf{x}^*,\y-\x\rangle).
\end{align*}
The last equality holds because $1-\lambda>0$ and $\psi_\gamma$ is positively homogeneous (Lemma~\ref{lem:psi_basics}(1)).
Since $\mathbf{x},\mathbf{y}\in\operatorname{dom}(f)$ and
$\mathbf{x}^*\in\hat{\partial}f(\mathbf{x})$ were arbitrary, $f$ is
$\gamma$--tilted convex with respect to $\hat{\partial}f$.
Thus, by the implication (iii)$\Rightarrow$(i) of Theorem~\ref{thm:mwc-alternative-characterizations}, it is $\gamma$--tilted convex.

\medskip

\noindent\textbf{(ii):}
Assume that $f$ is $\gamma$--tilted convex with respect to $F$. Fix
$\mathbf{x},\mathbf{y}\in\operatorname{dom}(F)$,
$\mathbf{x}^*\in F(\mathbf{x})$,
and $\mathbf{y}^*\in F(\mathbf{y})$.
Since $\op{dom}(F) \subseteq \op{dom}(f)$, we may apply \eqref{eq:gamma-mwc} at the pair $(\x, \y)$ with $\v = \x^*$ and at the pair $(\y, \x)$ with $\v = \y^*$.
This gives
\begin{equation}
\label{eq:pb-ii-xy}
f(\y) - f(\x) \geq \psi_\gamma(\langle \mathbf{x}^*,\y - \x\rangle)
\end{equation}
and
\begin{equation}
\label{eq:pb-ii-yx}
f(\x) - f(\y) \geq \psi_\gamma(\langle \mathbf{y}^*,\x - \y\rangle).
\end{equation}
Multiplying \eqref{eq:pb-ii-yx} by $-1$ and using Lemma~\ref{lem:psi_basics}(3) gives
\[
f(\y) - f(\x) \leq - \psi_\gamma(- \langle \mathbf{y}^*,\y - \x\rangle)
= \overline{\psi}_\gamma(\langle \mathbf{y}^*,\y - \x\rangle).
\]
Together with \eqref{eq:pb-ii-xy}, this yields
\begin{equation}
\label{eq:pb-ii-sandwich}
\overline{\psi}_\gamma(\langle \mathbf{y}^*,\y - \x\rangle)
\geq \psi_\gamma(\langle \mathbf{x}^*,\y - \x\rangle).
\end{equation}
Applying the increasing function $\psi_\gamma$ to both sides of \eqref{eq:pb-ii-sandwich}, and using $\psi_\gamma \circ \overline{\psi}_\gamma = \op{id}$ (Lemma~\ref{lem:psi_basics}(6)) on the left and $\psi_\gamma \circ \psi_\gamma = \psi_{\gamma^2}$ (Lemma~\ref{lem:psi_basics}(4)) on the right, gives
\[
\langle \mathbf{y}^*,\y - \x\rangle
= \psi_\gamma(\overline{\psi}_\gamma(\langle \mathbf{y}^*,\y - \x\rangle))
\geq \psi_\gamma(\psi_\gamma(\langle \mathbf{x}^*,\y - \x\rangle))
= \psi_{\gamma^2}(\langle \mathbf{x}^*,\y - \x\rangle),
\]
which is \eqref{eq:gamma-mw-mono} with parameter $\gamma^2$.
Hence $F$ is $\gamma^2$--tilted monotone.

\medskip

For the inequalities between the optimal parameters, let $\gamma_0 := \gamma^*_{\op{mono}}(f)$.
If $\gamma_0 > 0$, then $\overline\partial f$ is $\gamma_0$--tilted monotone.
By Proposition~\ref{prop:subdiff-basics}(i), $\hat\partial f(\x) \subseteq \overline\partial f(\x)$ at every $\x \in \op{dom}(f)$, so $\hat\partial f$ is also $\gamma_0$--tilted monotone.
Part~(i) then shows that $f$ is $\gamma_0$--tilted convex, so $\gamma^*_{\op{mono}}(f) \le \gamma^*(f)$.
This inequality is trivial if $\gamma_0 = 0$.
Next let $\gamma_1 := \gamma^*(f)$.
If $\gamma_1 > 0$, then part~(ii) with $F = \overline\partial f$, whose domain is contained in $\op{dom}(f)$, shows that $\overline\partial f$ is $\gamma_1^2$--tilted monotone.
Hence $\gamma^*(f)^2 \le \gamma^*_{\op{mono}}(f)$, and this is again trivial if $\gamma_1 = 0$.
Taking square roots gives $\gamma^*(f) \le (\gamma^*_{\op{mono}}(f))^{1/2}$.
\end{proof}

\subsection{Proof of Theorem~\ref{thm:mw-mono-alternative-characterizations}: Alternative First-order Characterization}
\label{apx:mw-mono-alternative-characterizations}

If an operator is $\gamma$--tilted monotone, then so is every operator
whose values are subsets of its values.  Since $f$ is lsc,
\eqref{eq:pc2-subdiff-inclusions} holds, and it gives the implications
(i) $\implies$ (ii) $\implies$ (iii).  If
(iii) holds, Theorem~\ref{thm:mwc-vs-mw-mono}\textup{(i)} shows that $f$ is
$\gamma$--tilted convex. Hence $\mathcal K$ is convex by
Theorem~\ref{thm:mwc-domain-is-convex},
$f|_{\operatorname{ri}(\mathcal K)}$ is locally Lipschitz by
Theorem~\ref{thm:gamma-MW-convex-is-locally-Lipschitz}, and $f$ is
radially continuous by Corollary~\ref{cor:mwc-radially-continuous}.
The restriction $\hat\partial f|_{\operatorname{ri}(\mathcal K)}$ is
$\gamma$--tilted monotone because $\hat\partial f$ is. Thus
condition~\textup{(iv)} holds.
For (iv) $\implies$ (v), let $\x\in\op{ri}(\mathcal K)$ be a point
where the relative gradient $\nabla f(\x)$ exists.  The argument that
gave \eqref{eq:pc2-gradient-is-regular} uses only
$\op{dom}(f)\subseteq\op{aff}(\mathcal K)$, so it shows that
$\nabla f(\x)\in\hat\partial f(\x)$.  Hence the inequality in (v) is a
special case of the $\gamma$--tilted monotonicity of
$\hat\partial f|_{\operatorname{ri}(\mathcal K)}$ in (iv), and the
other requirements in (v) are part of (iv).

It remains to prove (ii) $\implies$ (i), (iii) $\implies$ (ii),
(iv) $\implies$ (iii), and (v) $\implies$ (iv).

\subsubsection{Proof of (ii) \texorpdfstring{$\implies$}{implies} (i)}

\begin{proof}
Assume $\partial f$ is $\gamma$--tilted monotone.  By
\eqref{eq:pc2-subdiff-inclusions}, $\hat\partial f$ is $\gamma$--tilted monotone, so Theorem~\ref{thm:mwc-vs-mw-mono}\textup{(i)} shows that
$f$ is $\gamma$--tilted convex.  By \eqref{eq:pc2-subdiff-inclusions}
again, $f$ is $\gamma$--tilted convex with respect to $\hat\partial f$, and
Theorem~\ref{thm:mwc-domain-is-convex} shows that $\mathcal K$ is
convex.  Thus Lemma~\ref{lem:horizon-subdiff} applies to $f$ and gives,
for every $\x'\in\mathcal K$,
\begin{equation}
\label{eq:pc2-mono-normal}
\partial^\infty f(\x')
\subseteq
\hat N_{\mathcal K}(\x')
=\{\mathbf n:\langle\mathbf n,\y-\x'\rangle\leq0
\ \forall\y\in\mathcal K\}.
\end{equation}
The set $\hat N_{\mathcal K}(\x')$ is an intersection of closed
half-spaces, hence convex, so
$\op{conv}(\partial^\infty f(\x'))\subseteq\hat N_{\mathcal K}(\x')$.
By the definition of $\overline{\partial} f$, this gives
\begin{equation}
\label{eq:pc2-mono-clarke-bound}
\overline{\partial} f(\x')
=
\overline{\op{conv}(\partial f(\x')) + \op{conv}(\partial^\infty f(\x'))}
\subseteq
\overline{\op{conv}(\partial f(\x')) + \hat{N}_{\C{K}}(\x')},
\qquad \x'\in\mathcal K.
\end{equation}

Fix $\x \in \C{K}$ and define
\begin{align*}
V
&:=
\left\{ \vv \in \mathbb{R}^d
\mid
\langle \y - \x, \uu \rangle
\ge
\psi_\gamma(\langle \y - \x, \vv \rangle)
,\ \forall \y \in \C{K}, \uu \in \partial f(\y)
\right\} \\
&=
\left\{ \vv \in \mathbb{R}^d
\mid
\overline{\psi}_\gamma(\langle \y - \x, \uu \rangle)
\ge
\langle \y - \x, \vv \rangle
,\ \forall \y \in \C{K}, \uu \in \partial f(\y)
\right\} \\
&= \bigcap_{\y \in \C{K}, \uu \in \partial f(\y)}
\left\{ \vv \in \mathbb{R}^d
\mid
\overline{\psi}_\gamma(\langle \y - \x, \uu \rangle)
\ge
\langle \y - \x, \vv \rangle
\right\},
\end{align*}
where the second equality follows from Lemma~\ref{lem:psi_basics}:
$\psi_\gamma$ and $\overline{\psi}_\gamma$ are increasing and inverse
to each other.  Each set in the last intersection is a closed
half-space or $\mathbb R^d$, so $V$ is closed and convex.  The
$\gamma$--tilted monotonicity of $\partial f$, applied to the pairs
$(\x,\vv)$ with $\vv\in\partial f(\x)$ and $(\y,\uu)$, gives
$\partial f(\x)\subseteq V$, and hence
$\op{conv}(\partial f(\x))\subseteq V$.  Now let $\vv\in V$ and
$\mathbf n\in\hat N_{\mathcal K}(\x)$.  By \eqref{eq:pc2-mono-normal},
$\langle\y-\x,\mathbf n\rangle\leq0$ for every $\y\in\mathcal K$.
Since $\psi_\gamma$ is increasing,
\[
\psi_\gamma(\langle \y - \x, \vv+\mathbf n \rangle)
\leq
\psi_\gamma(\langle \y - \x, \vv \rangle)
\leq
\langle \y - \x, \uu \rangle,
\qquad
\y \in \C{K},\ \uu \in \partial f(\y),
\]
so $\vv+\mathbf n\in V$.  Consequently
$\op{conv}(\partial f(\x)) + \hat{N}_{\C{K}}(\x)\subseteq V$, and since
$V$ is closed,
\[
\overline{\op{conv}(\partial f(\x)) + \hat{N}_{\C{K}}(\x)} \subseteq V.
\]
By \eqref{eq:pc2-mono-clarke-bound} with $\x'=\x$, this gives
$\overline{\partial} f(\x)\subseteq V$.  Since $\x\in\mathcal K$ was
arbitrary, we have shown
\begin{equation}
\label{eq:pc2-mono-half-way}
\langle \y - \x, \uu \rangle
\ge
\psi_\gamma(\langle \y - \x, \vv \rangle)
,\qquad
\forall \x,\y \in \C{K},\ \vv \in \overline{\partial} f(\x),\ \uu \in \partial f(\y).
\end{equation}

Next, fix $\y \in \C{K}$ and define
\begin{align*}
U
&:=
\left\{ \uu \in \mathbb{R}^d
\mid
\langle \y - \x, \uu \rangle
\ge
\psi_\gamma(\langle \y - \x, \vv \rangle)
,\ \forall \x \in \C{K}, \vv \in \overline{\partial} f(\x)
\right\} \\
&= \bigcap_{\x \in \C{K}, \vv \in \overline{\partial} f(\x)}
\left\{ \uu \in \mathbb{R}^d
\mid
\langle \y - \x, \uu \rangle
\geq
\psi_\gamma(\langle \y - \x, \vv \rangle)
\right\}.
\end{align*}
Each set in the intersection is a closed half-space or
$\mathbb R^d$, so $U$ is closed and convex.  By
\eqref{eq:pc2-mono-half-way}, $\partial f(\y) \subseteq U$, and hence
$\op{conv}(\partial f(\y))\subseteq U$.  Now let $\uu\in U$ and
$\mathbf n\in\hat N_{\mathcal K}(\y)$.  By \eqref{eq:pc2-mono-normal}
with $\x'=\y$, we have $\langle\x-\y,\mathbf n\rangle\leq0$, that is,
$\langle\y-\x,\mathbf n\rangle\geq0$, for every $\x\in\mathcal K$.
Hence
$\langle\y-\x,\uu+\mathbf n\rangle\geq\langle\y-\x,\uu\rangle$ for
every $\x\in\mathcal K$, and $\uu+\mathbf n\in U$.  Since $U$ is
closed, it follows that
\[
\overline{\op{conv}(\partial f(\y)) + \hat{N}_{\C{K}}(\y)} \subseteq U,
\]
and \eqref{eq:pc2-mono-clarke-bound} with $\x'=\y$ gives
$\overline{\partial} f(\y)\subseteq U$.
Hence, for all $\y \in \C{K}$ and $\uu \in \overline{\partial} f(\y)$, we have
\[
\langle \y - \x, \uu \rangle
\ge
\psi_\gamma(\langle \y - \x, \vv \rangle)
,\qquad
\forall \x \in \C{K}, \vv \in \overline{\partial} f(\x).
\]
This is the $\gamma$--tilted monotonicity of $\overline{\partial} f$.
\end{proof}

\subsubsection{Proof of (iii) \texorpdfstring{$\implies$}{implies} (ii)}

\begin{proof}
Assume (iii).  Fix $\x,\y\in\mathcal K$, $\vv\in\partial f(\x)$, and
$\uu\in\partial f(\y)$.  By the definition of the limiting
subdifferential, there are sequences
\[
  \x_k\xrightarrow{f}\x,
  \quad \vv_k\in\hat\partial f(\x_k),
  \quad \vv_k\to\vv,
  \qquad
  \y_k\xrightarrow{f}\y,
  \quad \uu_k\in\hat\partial f(\y_k),
  \quad \uu_k\to\uu.
\]
Applying (iii) to the pairs $(\x_k,\vv_k)$ and $(\y_k,\uu_k)$ gives
\begin{equation}
\label{eq:pc2-mono-regular-k}
  \langle\uu_k,\y_k-\x_k\rangle
  \geq
  \psi_\gamma(\langle\vv_k,\y_k-\x_k\rangle).
\end{equation}
As $k\to\infty$, we have $\uu_k\to\uu$, $\vv_k\to\vv$, and
$\y_k-\x_k\to\y-\x$.  Since $\psi_\gamma$ is continuous, passing to the
limit in \eqref{eq:pc2-mono-regular-k} gives
\[
  \langle\uu,\y-\x\rangle
  \geq
  \psi_\gamma(\langle\vv,\y-\x\rangle),
\]
which proves that $\partial f$ is $\gamma$--tilted monotone.
\end{proof}

\subsubsection{Proof of (iv) \texorpdfstring{$\implies$}{implies} (iii)}

\begin{proof}
Assume (iv).  Put $A:=\op{aff}(\C{K})$ and $W:=A-A$, and work in the
Euclidean space $A$.  This does not change any of the inner products
below, because all differences of points of $\mathcal K$ lie in $W$.
We write $f|_{\op{ri}(\C{K})}$ for the restriction of $f$ to the
relatively open set $\op{ri}(\C{K})$; its subdifferentials are taken
in $A$ and consist of vectors of $W$.

Let $\x'\in\op{ri}(\C{K})$ and let $\mathbf a$ be a regular
subgradient of $f|_{\op{ri}(\C{K})}$ at $\x'$.  Since
$\op{ri}(\C{K})$ is open in $A$ and $\mathcal K\subseteq A$, all points
of $\mathcal K$ close to $\x'$ lie in $\op{ri}(\C{K})$.  So the
quotients in the definitions of the regular subdifferentials of
$f|_{\op{ri}(\C{K})}$ and of $f$ at $\x'$ agree near $\x'$, and
$\mathbf a\in\hat\partial f(\x')$.  Thus the monotonicity assumed in
(iv) also holds for the regular subdifferential of
$f|_{\op{ri}(\C{K})}$.  The proof of (iii) $\implies$ (ii) above uses
only the definitions of the regular and limiting subdifferentials and
the continuity of $\psi_\gamma$.  Applied to $f|_{\op{ri}(\C{K})}$ in
$A$, it shows that
\begin{equation}
\label{eq:pc2-relative-limiting-mono}
\langle\mathbf b,\y'-\x'\rangle
\geq
\psi_\gamma(\langle\mathbf a,\y'-\x'\rangle)
\end{equation}
for all $\x',\y'\in\op{ri}(\C{K})$,
$\mathbf a\in\partial(f|_{\op{ri}(\C{K})})(\x')$, and
$\mathbf b\in\partial(f|_{\op{ri}(\C{K})})(\y')$.

By condition~\textup{(iv)}, $f$ is locally Lipschitz on
$\op{ri}(\C{K})$ and radially continuous.  In particular, $f$ is
continuous on every line segment joining two points of $\C{K}$.

Fix $\x,\y\in\C{K}$, $\vv\in\hat\partial f(\x)$, and
$\uu\in\hat\partial f(\y)$.  If $\x=\y$, both sides of the required
inequality are zero.  So assume $\x\neq\y$ and
set $\dd:=\y-\x\neq 0$.  Choose $\z\in\op{ri}(\C{K})$ and, for
$\epsilon\in(0,1)$, define
\[
  \mathbf p_\epsilon(t)
  :=(1-\epsilon)(\x+t\dd)+\epsilon\z,
  \qquad t\in\mathbb R.
\]
For $t\in[0,1]$, the point $\x+t\dd$ lies in $\C{K}$, so Theorem~6.1
in~\cite{rockafellar15_convex_analy} gives
$\mathbf p_\epsilon(t)\in\op{ri}(\C{K})$.  Thus the segment
$\mathbf p_\epsilon([0,1])$ is contained in $\op{ri}(\C{K})$.  The set
$\{t\in\mathbb R:\mathbf p_\epsilon(t)\in\op{ri}(\C{K})\}$ is open,
because $\mathbf p_\epsilon$ is continuous and $\op{ri}(\C{K})$ is open
in $A$.  It contains $[0,1]$, so its connected component containing
$[0,1]$ is an open interval $J_\epsilon\supseteq[0,1]$ with
$\mathbf p_\epsilon(J_\epsilon)\subseteq\op{ri}(\C{K})$.  Define
\[
  h_\epsilon(t):=f(\mathbf p_\epsilon(t)),
  \qquad t\in J_\epsilon.
\]

We now verify the assumptions of Theorem~\ref{thm:chain-rule}.  The
function $f|_{\op{ri}(\C{K})}$ is proper and lsc on $\op{ri}(\C{K})$
(indeed continuous), while
$\mathbf p_\epsilon:\mathbb R\to A$ is affine, and hence continuously
differentiable, with
\[
  \nabla\mathbf p_\epsilon(s)a=(1-\epsilon)a\dd
  \qquad(a\in\mathbb R).
\]
On the open interval $J_\epsilon$, the function $h_\epsilon$ coincides
with $f|_{\op{ri}(\C{K})}\circ\mathbf p_\epsilon$, so the two functions
have the same subdifferentials there.
Moreover, the local Lipschitz property in condition~\textup{(iv)}
shows that $f|_{\op{ri}(\C{K})}$ is locally Lipschitz at
$\mathbf p_\epsilon(s)$ for every $s\in J_\epsilon$.
Thus the limiting subdifferential part of Theorem~\ref{thm:chain-rule} applies and gives the inclusion
\begin{equation}
\label{eq:pc2-chain-inclusion}
  \partial h_\epsilon(s)
  \subseteq
  \nabla\mathbf p_\epsilon(s)^\top
  \partial(f|_{\op{ri}(\C{K})})(\mathbf p_\epsilon(s))
  =
  \left\{
    \langle\mathbf a,(1-\epsilon)\dd\rangle
    \mathrel{\Big|}
    \mathbf a\in
    \partial(f|_{\op{ri}(\C{K})})(\mathbf p_\epsilon(s))
  \right\},
  \qquad s\in J_\epsilon.
\end{equation}
Also, since a composition of a locally Lipschitz function with an affine
map is locally Lipschitz, $h_\epsilon$ is locally Lipschitz on
$J_\epsilon$.

We claim that $\partial h_\epsilon$ is $\gamma$--tilted monotone on
$J_\epsilon$.  Fix $s,t\in J_\epsilon$,
$a\in\partial h_\epsilon(s)$, and $b\in\partial h_\epsilon(t)$.  By
\eqref{eq:pc2-chain-inclusion},
\[
  a=\langle \mathbf a_s,(1-\epsilon)\dd\rangle,
  \qquad
  b=\langle \mathbf a_t,(1-\epsilon)\dd\rangle
\]
for suitable
$\mathbf a_s\in\partial(f|_{\op{ri}(\C{K})})(\mathbf p_\epsilon(s))$
and
$\mathbf a_t\in\partial(f|_{\op{ri}(\C{K})})(\mathbf p_\epsilon(t))$.
Since
$\mathbf p_\epsilon(t)-\mathbf p_\epsilon(s)=(t-s)(1-\epsilon)\dd$,
the inequality \eqref{eq:pc2-relative-limiting-mono} with
$\x'=\mathbf p_\epsilon(s)$ and $\y'=\mathbf p_\epsilon(t)$ reads
\[
  \langle\mathbf a_t,(t-s)(1-\epsilon)\dd\rangle
  \geq
  \psi_\gamma\bigl(\langle\mathbf a_s,(t-s)(1-\epsilon)\dd\rangle\bigr),
\]
that is,
\begin{equation}
\label{eq:pc2-h-mono}
  (t-s)b
  \geq \psi_\gamma((t-s)a).
\end{equation}
This is precisely the required monotonicity inequality for
$\partial h_\epsilon$.

Because $h_\epsilon$ is locally Lipschitz on $J_\epsilon$,
Proposition~\ref{prop:subdiff-basics}-(ii) gives
$\overline{\nabla} h_\epsilon\subseteq\partial h_\epsilon
\subseteq\overline\partial h_\epsilon
=\operatorname{conv}(\overline{\nabla} h_\epsilon)$, and therefore
$\overline\partial h_\epsilon=\operatorname{conv}(\partial h_\epsilon)$
on $J_\epsilon$.
Fix $s,t\in J_\epsilon$.  For fixed $b\in\partial h_\epsilon(t)$, by
Lemma~\ref{lem:psi_basics} the set of $a\in\mathbb R$ satisfying
\eqref{eq:pc2-h-mono} is
$\{a:(t-s)a\leq\overline\psi_\gamma((t-s)b)\}$, a closed half-line or
$\mathbb R$.  It is convex and contains $\partial h_\epsilon(s)$, so it
contains $\overline\partial h_\epsilon(s)$.  Next, for fixed
$a\in\overline\partial h_\epsilon(s)$, the set of $b\in\mathbb R$
satisfying \eqref{eq:pc2-h-mono} is again a closed half-line or
$\mathbb R$.  It contains $\partial h_\epsilon(t)$, so it contains
$\overline\partial h_\epsilon(t)$.  Hence \eqref{eq:pc2-h-mono} holds
for all $a\in\overline\partial h_\epsilon(s)$ and
$b\in\overline\partial h_\epsilon(t)$, that is,
$\overline\partial h_\epsilon$ is also $\gamma$--tilted monotone on
$J_\epsilon$.

Fix $0<r<s<1$.  The function $h_\epsilon$ is continuous on
$[0,1]$ and locally Lipschitz on $J_\epsilon\supseteq[0,1]$.  Applying
Corollary~\ref{cor:Lipschitz-interior-mean-value}
on $[0,r]$ and $[s,1]$, respectively, gives points
$c_\epsilon\in(0,r)$ and $e_\epsilon\in(s,1)$ and numbers
$a_\epsilon\in\overline{\partial} h_\epsilon(c_\epsilon)$ and
$b_\epsilon\in\overline{\partial} h_\epsilon(e_\epsilon)$ such that
\[
  a_\epsilon
  =\frac{h_\epsilon(r)-h_\epsilon(0)}{r},
  \qquad
  b_\epsilon
  =\frac{h_\epsilon(1)-h_\epsilon(s)}{1-s}.
\]
Since $c_\epsilon<r<s<e_\epsilon$, the monotonicity of
$\overline\partial h_\epsilon$, applied to
$(c_\epsilon,a_\epsilon)$ and $(e_\epsilon,b_\epsilon)$, and the
positive homogeneity of $\psi_\gamma$ give
\[
  (e_\epsilon-c_\epsilon)b_\epsilon
  \geq\psi_\gamma\bigl((e_\epsilon-c_\epsilon)a_\epsilon\bigr)
  =(e_\epsilon-c_\epsilon)\psi_\gamma(a_\epsilon).
\]
Dividing by $e_\epsilon-c_\epsilon>0$ and substituting the values of
$a_\epsilon$ and $b_\epsilon$ yields
\begin{equation}
\label{eq:pc2-eps-secant}
  \frac{h_\epsilon(1)-h_\epsilon(s)}{1-s}
  \geq
  \psi_\gamma\!\left(
    \frac{h_\epsilon(r)-h_\epsilon(0)}{r}
  \right).
\end{equation}
For every fixed $t\in[0,1]$, the point $\mathbf p_\epsilon(t)$ lies on
the segment joining $\x+t\dd\in\C{K}$ to $\z\in\C{K}$, and it tends to
$\x+t\dd$ as $\epsilon\downarrow0$.  Continuity of $f$ on this segment
gives
\[
  h_\epsilon(t)\longrightarrow f(\x+t\dd)
  \qquad\text{as }\epsilon\downarrow0.
\]
Applying this with $t\in\{0,r,s,1\}$, letting $\epsilon\downarrow0$ in
\eqref{eq:pc2-eps-secant}, and using the continuity of $\psi_\gamma$, we
obtain
\begin{equation}
  \frac{f(\y)-f(\x+s\dd)}{1-s}
  \geq
  \psi_\gamma\!\left(
    \frac{f(\x+r\dd)-f(\x)}{r}
  \right)
  \qquad(0<r<s<1).
  \label{eq:boundary-mw-secant-slopes}
\end{equation}

Set
\[
  a:=\langle\vv,\dd\rangle,
  \qquad
  b:=\langle\uu,\dd\rangle.
\]
Since $\vv\in\hat\partial f(\x)$, the definition of the regular
subdifferential gives
\[
  \liminf_{\substack{\mathbf w\to\x\\
                     \mathbf w\in\C{K},\ \mathbf w\neq\x}}
  \frac{
    f(\mathbf w)-f(\x)-\langle\vv,\mathbf w-\x\rangle
  }{\|\mathbf w-\x\|}
  \geq0.
\]
The points $\mathbf w=\x+r\dd$ with $r\in(0,1]$ lie in
$\C{K}\setminus\{\x\}$.  Evaluating this lower limit along them as
$r\downarrow0$ yields
\begin{align*}
  0
  &\leq
  \liminf_{r\downarrow0}
  \frac{f(\x+r\dd)-f(\x)-r\langle\vv,\dd\rangle}
       {r\|\dd\|} \\
  &=\frac{1}{\|\dd\|}
  \left(
    \liminf_{r\downarrow0}
    \frac{f(\x+r\dd)-f(\x)}{r}-a
  \right).
\end{align*}
Consequently,
\begin{equation}
\label{eq:pc2-slope-at-x}
  a\leq\liminf_{r\downarrow0}
  \frac{f(\x+r\dd)-f(\x)}{r}.
\end{equation}
Similarly, $\uu\in\hat\partial f(\y)$ means that
\[
  \liminf_{\substack{\mathbf w\to\y\\
                     \mathbf w\in\C{K},\ \mathbf w\neq\y}}
  \frac{
    f(\mathbf w)-f(\y)-\langle\uu,\mathbf w-\y\rangle
  }{\|\mathbf w-\y\|}
  \geq0.
\]
This time, evaluating along
$\mathbf w=\x+s\dd=\y-(1-s)\dd$ as $s\uparrow1$ gives
\begin{align*}
  0
  &\leq
  \liminf_{s\uparrow1}
  \frac{f(\x+s\dd)-f(\y)+(1-s)\langle\uu,\dd\rangle}
       {(1-s)\|\dd\|} \\
  &=\frac{1}{\|\dd\|}
  \left(
    b-\limsup_{s\uparrow1}
    \frac{f(\y)-f(\x+s\dd)}{1-s}
  \right),
\end{align*}
so
\begin{equation}
\label{eq:pc2-slope-at-y}
  b\geq\limsup_{s\uparrow1}
  \frac{f(\y)-f(\x+s\dd)}{1-s}.
\end{equation}
Fix $r\in(0,1)$.  By \eqref{eq:boundary-mw-secant-slopes}, every
quotient in \eqref{eq:pc2-slope-at-y} with $s\in(r,1)$ is at least
$\psi_\gamma((f(\x+r\dd)-f(\x))/r)$.  Hence
\eqref{eq:pc2-slope-at-y} gives
\begin{equation}
\label{eq:pc2-b-lower}
  b\geq
  \psi_\gamma\!\left(
    \frac{f(\x+r\dd)-f(\x)}{r}
  \right),
  \qquad r\in(0,1).
\end{equation}
Let $L$ be the lower limit in \eqref{eq:pc2-slope-at-x}, so that
$L\geq a$, and choose $r_n\downarrow0$ such that
$(f(\x+r_n\dd)-f(\x))/r_n\to L$.  If $L=+\infty$, then the right-hand
side of \eqref{eq:pc2-b-lower} with $r=r_n$ tends to $+\infty$, which
is impossible.  So $L$ is finite.  Letting $n\to\infty$ in
\eqref{eq:pc2-b-lower} with $r=r_n$, and using the continuity and
monotonicity of $\psi_\gamma$, gives
\[
  \langle\uu,\y-\x\rangle
  =b
  \geq\psi_\gamma(L)
  \geq\psi_\gamma(a)
  =\psi_\gamma(\langle\vv,\y-\x\rangle).
\]
Thus $\hat\partial f$ is $\gamma$--tilted monotone.
\end{proof}

\subsubsection{Proof of (v) \texorpdfstring{$\implies$}{implies} (iv)}

\begin{proof}
Condition~\textup{(v)} includes all requirements of
condition~\textup{(iv)} except the monotonicity assertion, so we prove
only that assertion.  Put $A:=\op{aff}(\mathcal K)$ and
$W:=A-A$, and interpret gradients relative to $A$.  Fix
$\x,\y\in\op{ri}(\mathcal K)$,
$\vv\in\widehat{\partial}f(\x)$, and
$\uu\in\widehat{\partial}f(\y)$.  Let $\vv_A$ and $\uu_A$ be the
orthogonal projections of $\vv$ and $\uu$, respectively, onto $W$.
Every point of $\op{dom}(f)=\mathcal K$ differs from $\x$ and from $\y$
by a vector of $W$.  Hence the quotients in the definition of the
regular subdifferential do not change when $\vv$ and $\uu$ are replaced
by $\vv_A$ and $\uu_A$, which gives
\[
\vv_A\in\widehat{\partial}_A f(\x),
\qquad
\uu_A\in\widehat{\partial}_A f(\y),
\]
where the subdifferentials are taken relative to $A$.  For the same
reason, since $\y-\x\in W$,
\begin{equation}
\label{eq:pc2-mono-projection}
\langle\vv,\y-\x\rangle=\langle\vv_A,\y-\x\rangle,
\qquad
\langle\uu,\y-\x\rangle=\langle\uu_A,\y-\x\rangle.
\end{equation}

Let $\overline{\nabla}_A f$ denote the set of
Proposition~\ref{prop:subdiff-basics}-(ii) for $f|_A$ in the Euclidean
space $A$.  Let $\mathbf w\in\overline{\nabla}_A f(\x)$ and
$\mathbf z\in\overline{\nabla}_A f(\y)$.  Then there are sequences
$\x_k\to\x$ and $\y_k\to\y$ in $A$ such that the
relative gradients $\nabla_A f(\x_k)$ and $\nabla_A f(\y_k)$ exist and
converge to $\mathbf w$ and $\mathbf z$, respectively.  Since
$\op{ri}(\mathcal K)$ is open in $A$, the points $\x_k$ and $\y_k$ lie
in $\op{ri}(\mathcal K)$ for all large $k$.
Condition~\textup{(v)} gives
\[
\langle\nabla_A f(\y_k),\y_k-\x_k\rangle
\geq
\psi_\gamma\!\left(
\langle\nabla_A f(\x_k),\y_k-\x_k\rangle
\right).
\]
Letting $k\to\infty$ and using the continuity of $\psi_\gamma$ yields
\[
\langle\mathbf z,\y-\x\rangle
\geq
\psi_\gamma\bigl(\langle\mathbf w,\y-\x\rangle\bigr),
\]
and applying the increasing inverse $\overline{\psi}_\gamma$
(Lemma~\ref{lem:psi_basics}) gives
\begin{equation}
\label{eq:pc2-mono-gradient-limits}
\overline{\psi}_\gamma\bigl(\langle\mathbf z,\y-\x\rangle\bigr)
\geq
\langle\mathbf w,\y-\x\rangle,
\qquad
\mathbf w\in\overline{\nabla}_A f(\x),\
\mathbf z\in\overline{\nabla}_A f(\y).
\end{equation}

Since $f|_A$ is locally Lipschitz at $\x$ and at $\y$,
Proposition~\ref{prop:subdiff-basics}-(i) and~(ii), applied in the
Euclidean space $A$, give
\begin{equation}
\label{eq:pc2-mono-conv-hulls}
\vv_A
\in\widehat{\partial}_A f(\x)
\subseteq\op{conv}\bigl(\overline{\nabla}_A f(\x)\bigr),
\qquad
\uu_A
\in\widehat{\partial}_A f(\y)
\subseteq\op{conv}\bigl(\overline{\nabla}_A f(\y)\bigr).
\end{equation}
Fix $\mathbf z\in\overline{\nabla}_A f(\y)$.  The set of
$\mathbf w\in W$ satisfying \eqref{eq:pc2-mono-gradient-limits} is a
closed half-space of $W$ or all of $W$, hence convex.  It contains
$\overline{\nabla}_A f(\x)$, so by \eqref{eq:pc2-mono-conv-hulls} it
contains $\vv_A$.  Applying the increasing function $\psi_\gamma$
gives
\[
\langle\mathbf z,\y-\x\rangle
\geq
\psi_\gamma\bigl(\langle\vv_A,\y-\x\rangle\bigr),
\qquad
\mathbf z\in\overline{\nabla}_A f(\y).
\]
The set of $\mathbf z\in W$ satisfying this inequality is again a
closed half-space of $W$ or all of $W$.  It contains
$\overline{\nabla}_A f(\y)$, so by \eqref{eq:pc2-mono-conv-hulls} it
contains $\uu_A$.  Combining this with
\eqref{eq:pc2-mono-projection}, we obtain
\[
\langle\uu,\y-\x\rangle
=\langle\uu_A,\y-\x\rangle
\geq
\psi_\gamma\bigl(\langle\vv_A,\y-\x\rangle\bigr)
=\psi_\gamma\bigl(\langle\vv,\y-\x\rangle\bigr).
\]
Thus $\widehat{\partial} f|_{\operatorname{ri}(\mathcal K)}$ is
$\gamma$--tilted monotone, proving condition~\textup{(iv)}.
\end{proof}

\subsection{Proof of Theorem~\ref{thm:gamma-MW-monotone-on-lines}}
\label{apx:gamma-MW-monotone-on-lines}

\begin{proof}
\noindent\textbf{Forward direction.}
Assume that $\overline{\partial} f$ is $\gamma$--tilted monotone and put
$\mathcal K:=\operatorname{dom}(f)$.  Since
$\hat\partial f\subseteq\overline{\partial}f$ by
Proposition~\ref{prop:subdiff-basics}-(i), the operator $\hat\partial f$ is
$\gamma$--tilted monotone, and Theorem~\ref{thm:mwc-vs-mw-mono}-(i) shows that
$f$ is $\gamma$--tilted convex.
Hence $\mathcal K$ is convex by Theorem~\ref{thm:mwc-domain-is-convex}.
Fix a line
$\ell$ meeting $\mathcal K$, write
\[
  \ell=\{\mathbf b+t\mathbf d:t\in\mathbb R\},
  \qquad \|\mathbf d\|=1,
\]
and set
\[
  I:=\{t\in\mathbb R:\mathbf b+t\mathbf d\in\mathcal K\},
  \qquad
  h(t):=f(\mathbf b+t\mathbf d).
\]
If $I$ is a singleton, the only instance of~\eqref{eq:gamma-mw-mono} for
$\overline{\partial} h$ compares a point with itself, and both sides
vanish; so
$\overline{\partial} h$ is $\gamma$--tilted monotone.  Suppose, therefore, that $I$ is nondegenerate.  Then
$\ell\subseteq\operatorname{aff}(\mathcal K)$.  We work relative to
$\operatorname{aff}(\mathcal K)$; relative limiting subgradients embed as
ambient limiting subgradients because $f=+\infty$ outside this affine
hull.  Thus the relative limiting subdifferential of $f$ is also
$\gamma$--tilted monotone.

Choose $\mathbf a\in\operatorname{ri}(\mathcal K)$ and a sequence
$\epsilon_n\downarrow0$.  Define
\[
  \tau_n(t):=(1-\epsilon_n)(\mathbf b+t\mathbf d)
              +\epsilon_n\mathbf a,
  \qquad
  h_n(t):=f(\tau_n(t)).
\]
For $t\in I$, the point $\tau_n(t)$ belongs to
$\operatorname{ri}(\mathcal K)$ by~\eqref{eq:pc3-ri-segment}.  Let
$s,t\in\operatorname{int}(I)$.  The function $f$ is locally Lipschitz at
$\tau_n(s)$ and $\tau_n(t)$ by
Theorem~\ref{thm:gamma-MW-convex-is-locally-Lipschitz}, and the
derivative of $\tau_n$ is $(1-\epsilon_n)\mathbf d$.  Hence the second
part of Theorem~\ref{thm:chain-rule} shows that every
$p\in\partial h_n(s)$ and $q\in\partial h_n(t)$ have the form
\[
  p=(1-\epsilon_n)\langle\mathbf v,\mathbf d\rangle,
  \qquad
  q=(1-\epsilon_n)\langle\mathbf u,\mathbf d\rangle
\]
for some
$\mathbf v\in\partial f(\tau_n(s))$ and
$\mathbf u\in\partial f(\tau_n(t))$, where the subdifferentials of $f$
are relative to $\operatorname{aff}(\mathcal K)$.  Since
$\tau_n(t)-\tau_n(s)=(1-\epsilon_n)(t-s)\mathbf d$, the relative
$\gamma$--tilted monotonicity of $\partial f$ gives
\begin{equation}
\label{eq:pc3-hn-mono}
  (t-s)q
  =\langle\mathbf u,\tau_n(t)-\tau_n(s)\rangle
  \geq
  \psi_\gamma\!\left(
    \langle\mathbf v,\tau_n(t)-\tau_n(s)\rangle
  \right)
  =\psi_\gamma((t-s)p).
\end{equation}
Thus $\partial h_n|_{\operatorname{int}(I)}$ is $\gamma$--tilted monotone.

Each $h_n$ is a reparametrized line restriction of $f$, and it is
$\gamma$--tilted convex.  To see this, put $c_n:=1-\epsilon_n$ and
$\tilde h_n(r):=f(\tau_n(0)+r\mathbf d)$, so that
\begin{equation}
\label{eq:pc3-hn-rescale}
h_n(t)=\tilde h_n(c_nt)\qquad(t\in\mathbb R).
\end{equation}
The function $\tilde h_n$ is the restriction of $f$ to the line
$\tau_n(\mathbb R)$ in a unit-speed parametrization, and this line meets
$\mathcal K$; so Theorem~\ref{thm:gamma-MW-convex-on-lines} shows that
$\tilde h_n$ is $\gamma$--tilted convex.  The first part of
Theorem~\ref{thm:chain-rule}, applied to the inverse relation
$\tilde h_n(r)=h_n(r/c_n)$, gives
$\hat\partial h_n(t)\subseteq c_n\hat\partial\tilde h_n(c_nt)$.  Hence, if
$t,t'\in\operatorname{dom}(h_n)$ and $p\in\hat\partial h_n(t)$, then
$p=c_nw$ with $w\in\hat\partial\tilde h_n(c_nt)\subseteq
\overline{\partial}\tilde h_n(c_nt)$, and
\[
h_n(t')-h_n(t)=\tilde h_n(c_nt')-\tilde h_n(c_nt)
\geq\psi_\gamma(c_n(t'-t)w)=\psi_\gamma((t'-t)p).
\]
By Theorem~\ref{thm:mwc-alternative-characterizations}-(iii), $h_n$ is
$\gamma$--tilted convex.  Moreover, for $t\in I$ the value $h_n(t)$ equals
$g_{\epsilon_n}(t)$ in the proof of
Theorem~\ref{thm:gamma-MW-convex-on-lines} (with the same $\mathbf b$,
$\mathbf d$, $\mathbf a$), and $h(t)$ equals $g(t)$ there.  So
\eqref{eq:pc3-geps-pointwise}, which was derived from
Theorem~\ref{thm:gamma-MW-continuous-on-lines--strong-version}, gives
\[
  h_n(t)\longrightarrow h(t)
  \qquad(t\in I).
\]
Applying Theorem~\ref{thm:gamma-MW-convex-pointwise-limit} in one
dimension, with the convex set $I$ and the dense subset $I$, shows that
$h_n\to h$ locally uniformly on
$\operatorname{int}(I)$ and that $h$ is continuous there.  This local
uniform convergence gives a common bound for the $h_n$ on compact
subintervals, and hence a common local Lipschitz constant.  To see this, fix
$t_0\in\operatorname{int}(I)$ and $R>0$ with
$[t_0-3R,t_0+3R]\subseteq I\subseteq\operatorname{dom}(h_n)$.  Since
$[t_0-2R,t_0+2R]$ is a compact subset of $\operatorname{int}(I)$, after
discarding finitely many terms,
\begin{equation}
\label{eq:pc3-hn-bound}
\sup_{|t-t_0|\leq2R}|h_n(t)|\leq M:=1+\max_{|t-t_0|\leq2R}|h(t)|.
\end{equation}
Each $h_n$ is $\gamma$--tilted convex with respect to $\hat\partial h_n$,
whose domain is dense in $\operatorname{dom}(h_n)$ by
Proposition~\ref{prop:subdiff-basics}-(i).  So
Theorem~\ref{thm:approximate-gamma-supports-local-Lipschitz} and
\eqref{eq:pc3-hn-bound} give
\begin{equation}
\label{eq:pc3-hn-Lip}
|h_n(t)-h_n(t')|\leq\frac{2M}{\gamma^2R}|t-t'|
\qquad(t,t'\in(t_0-R,t_0+R)).
\end{equation}

Fix $s,t\in\operatorname{int}(I)$,
$p\in\partial h(s)$, and $q\in\partial h(t)$.  By \eqref{eq:pc3-hn-Lip}
with $t_0=s$ and with $t_0=t$, the functions $h_n$ are equi-Lipschitz on
small open intervals around $s$ and around $t$, and they converge
pointwise there to $h$.  Applying the last part of
Theorem~\ref{thm:lipschitz-limit} on these intervals gives
\[
  s_n\to s,
  \qquad
  p_n\in\partial h_n(s_n),
  \qquad
  p_n\to p,
\]
and
\[
  t_n\to t,
  \qquad
  q_n\in\partial h_n(t_n),
  \qquad
  q_n\to q.
\]
After discarding finitely many terms, $s_n,t_n\in\operatorname{int}(I)$,
and hence \eqref{eq:pc3-hn-mono} gives
\[
  (t_n-s_n)q_n
  \geq
  \psi_\gamma((t_n-s_n)p_n).
\]
Passing to the limit, using the continuity of $\psi_\gamma$, shows that
\begin{equation}
\label{eq:pc3-h-mono}
  (t-s)q\geq\psi_\gamma((t-s)p).
\end{equation}
Thus $\partial h|_{\operatorname{int}(I)}$ is $\gamma$--tilted monotone.

Finally, $h$ is $\gamma$--tilted convex by
Theorem~\ref{thm:gamma-MW-convex-on-lines}, and we check condition~(iv) of
Theorem~\ref{thm:mw-mono-alternative-characterizations} for $h$.  The
domain $I$ is convex and
$\operatorname{ri}(\operatorname{dom}(h))=\operatorname{int}(I)$.  The
restriction $h|_I$ is continuous by
Theorem~\ref{thm:gamma-MW-continuous-on-lines}, so $h$ is radially
continuous.  The function $h$ is locally Lipschitz on
$\operatorname{int}(I)$ by
Theorem~\ref{thm:gamma-MW-convex-is-locally-Lipschitz}.  Since
$\hat\partial h\subseteq\partial h$, \eqref{eq:pc3-h-mono} shows that
$\hat\partial h|_{\operatorname{int}(I)}$ is $\gamma$--tilted monotone.
Theorem~\ref{thm:mw-mono-alternative-characterizations} now shows that
$\overline{\partial} h$ is $\gamma$--tilted monotone.

Now let $J\subseteq\mathbb R$ be a nonempty closed interval such that
$J\cap I\neq\varnothing$, and restrict $h$ to $J$, with the value
$+\infty$ outside $J$.  The function $h|_J$ is proper and lsc, because $J$
is closed, and its domain $J\cap I$ is convex.  If $J\cap I$ is a
singleton, $\overline{\partial}(h|_J)$ is $\gamma$--tilted monotone, because the
only instance of the inequality compares a point with itself.
Otherwise,
\[
  \operatorname{ri}(J\cap I)\subseteq\operatorname{ri}(I),
\]
and $h|_J$ agrees locally with $h$ at every point of
$\operatorname{ri}(J\cap I)$.  Thus
$\hat\partial(h|_J)=\hat\partial h$ on $\operatorname{ri}(J\cap I)$, and
\eqref{eq:pc3-h-mono} shows that the regular subdifferential of
$h|_J$ is $\gamma$--tilted monotone on the relative interior of its domain.
For $t\in\operatorname{ri}(J\cap I)$, $p\in\hat\partial(h|_J)(t)$, and
$s\in J\cap I$, we have $p\in\hat\partial h(t)\subseteq\overline{\partial}h(t)$,
so the $\gamma$--tilted convexity of $h$ gives
$h(s)-h(t)\geq\psi_\gamma((s-t)p)$.
So the support characterization in
Theorem~\ref{thm:mwc-alternative-characterizations}-(iv) shows that $h|_J$ is
$\gamma$--tilted convex.  Moreover, $h|_J$ is continuous on $J\cap I$ because
$h$ is continuous on $I$, and it is locally Lipschitz on
$\operatorname{ri}(J\cap I)$ because $h$ is.  Hence
Theorem~\ref{thm:mw-mono-alternative-characterizations}-(iv) shows that
$\overline{\partial}(h|_J)$ is $\gamma$--tilted monotone.  This also proves the
additional assertion and completes the forward direction.

\medskip
\noindent\textbf{Reverse direction.}
Assume that $\overline{\partial}(f|_\ell)$ is $\gamma$--tilted monotone for
every line $\ell$ meeting $\operatorname{dom}(f)$.  Fix
$\x,\y\in\operatorname{dom}(f)$,
$\vv\in\hat\partial f(\x)$, and
$\uu\in\hat\partial f(\y)$.  If $\x=\y$, both sides of the required
inequality vanish.  Otherwise, put
\[
  \mathbf d:=\frac{\y-\x}{\|\y-\x\|},
  \qquad
  q(t):=f(\x+t\mathbf d),
\]
which is $f|_\ell$ for the line $\ell$ through $\x$ and $\y$, in a
unit-speed parametrization.
The first part of Theorem~\ref{thm:chain-rule}, applied with
$F(t):=\x+t\mathbf d$ at $t=0$ and at $t=\|\y-\x\|$, gives
\[
  \langle\vv,\mathbf d\rangle\in\hat\partial q(0),
  \qquad
  \langle\uu,\mathbf d\rangle
  \in\hat\partial q(\|\y-\x\|).
\]
These sets are contained in $\overline{\partial}q(0)$ and
$\overline{\partial}q(\|\y-\x\|)$ by
Proposition~\ref{prop:subdiff-basics}-(i).  Since
$\overline{\partial} q$ is $\gamma$--tilted monotone, applying
\eqref{eq:gamma-mw-mono} to the points $0$ and $\|\y-\x\|$ and
multiplying out $\|\y-\x\|\mathbf d=\y-\x$ gives
\[
  \langle\uu,\y-\x\rangle
  \geq
  \psi_\gamma\!\left(
    \langle\vv,\y-\x\rangle
  \right).
\]
Thus $\hat\partial f$ is $\gamma$--tilted monotone, and
Theorem~\ref{thm:mw-mono-alternative-characterizations} shows that
$\overline{\partial} f$ is $\gamma$--tilted monotone.
\end{proof}

\subsection{Proof of Theorem~\ref{thm:gamma-MW-monotone-pointwise-limit}}
\label{apx:gamma-MW-monotone-pointwise-limit}

\begin{proof}
Since $\hat\partial f_n\subseteq\overline{\partial} f_n$,
the operator $\hat\partial f_n$ is $\gamma$--tilted monotone, and
Theorem~\ref{thm:mwc-vs-mw-mono}-(i) shows that each $f_n$ is
$\gamma$--tilted convex.
Theorem~\ref{thm:gamma-MW-convex-pointwise-limit} therefore gives a
continuous function $f$ satisfying~(i), and it shows that
$\bar f:=\op{lsc}(f)$ is $\gamma$--tilted convex.

For the monotonicity in~(ii), put
\[
  A:=\operatorname{aff}(\mathcal K),
  \qquad
  V:=A-A,
\]
and work in the Euclidean space $A$, with the balls $\mathbb B_V$ of $A$
as in the proof of Theorem~\ref{thm:gamma-MW-convex-pointwise-limit}.
Choose $\mathbf d_0\in\mathcal D$.
After discarding finitely many terms, $\mathbf d_0\in\operatorname{dom}(f_n)$,
so $f_n|_A$ is proper and lsc.  For every line $\ell\subseteq A$ meeting
its effective domain, the forward direction of
Theorem~\ref{thm:gamma-MW-monotone-on-lines} shows that
$\overline{\partial}(f_n|_\ell)$ is $\gamma$--tilted monotone.  The reverse
direction, now applied in the Euclidean space $A$, shows that the relative
Clarke subdifferential $\overline{\partial}(f_n|_A)$ is
$\gamma$--tilted monotone.  Hence, by
Theorem~\ref{thm:mw-mono-alternative-characterizations}, the relative
limiting subdifferential $\partial(f_n|_A)$ is also
$\gamma$--tilted monotone.
By Theorem~\ref{thm:mwc-vs-mw-mono}, these restrictions, which we continue
to denote by $f_n$, are also $\gamma$--tilted convex.

On $O:=\operatorname{ri}(\mathcal K)$, the convergence $f_n\to f$ is
locally uniform and locally equi-Lipschitz.  To see the latter, fix
$\x_0\in O$ and choose $R>0$ such that
\[
  \overline{\mathbb B_V(\x_0,3R)}\subseteq O.
\]
The function $f$ is continuous on this compact ball, and $f_n\to f$
uniformly on it.  Hence, after discarding finitely many terms,
\[
|f_n|\leq M:=1+\max_{\overline{\mathbb B_V(\x_0,3R)}}|f|
\qquad\text{on }\overline{\mathbb B_V(\x_0,3R)}.
\]
In particular, this ball is contained in
$\operatorname{dom}(f_n)$, so $\operatorname{aff}(\operatorname{dom}(f_n))=A$,
and $M$ bounds $|f_n|$ on
$\overline{\mathbb B_V(\x_0,2R)}$.
Since each $f_n$ is $\gamma$--tilted convex with respect to $\hat\partial f_n$,
whose domain is dense in $\operatorname{dom}(f_n)$ by
Proposition~\ref{prop:subdiff-basics}-(i),
Theorem~\ref{thm:approximate-gamma-supports-local-Lipschitz} gives the
common Lipschitz constant $2M/(\gamma^2R)$ on $\mathbb B_V(\x_0,R)$.

Fix $\x,\y\in O$, $\vv\in\partial f(\x)$, and
$\uu\in\partial f(\y)$, where these subdifferentials are relative to
$A$.  Applying the last part of Theorem~\ref{thm:lipschitz-limit} on
balls $\mathbb B_V(\x,R)$ and $\mathbb B_V(\y,R')$ of the kind just
constructed gives
\[
  \x_n\to\x,
  \qquad
  \vv_n\in\partial f_n(\x_n),
  \qquad
  \vv_n\to\vv,
\]
and
\[
  \y_n\to\y,
  \qquad
  \uu_n\in\partial f_n(\y_n),
  \qquad
  \uu_n\to\uu.
\]
After discarding finitely many terms, $\x_n,\y_n\in O$, so the relative
$\gamma$--tilted monotonicity of $\partial f_n$ gives
\[
  \langle\uu_n,\y_n-\x_n\rangle
  \geq
  \psi_\gamma\!\left(
    \langle\vv_n,\y_n-\x_n\rangle
  \right).
\]
Passing to the limit and using the continuity of $\psi_\gamma$ yields
\begin{equation}
\label{eq:pc3-f-mono}
  \langle\uu,\y-\x\rangle
  \geq
  \psi_\gamma\!\left(
    \langle\vv,\y-\x\rangle
  \right).
\end{equation}
Thus the relative subdifferential $\partial f|_O$ is
$\gamma$--tilted monotone.

The functions $f$ and $\bar f$ agree on $O$.  If
$\x\in O$ and $\vv\in\hat\partial\bar f(\x)$, then the orthogonal
projection $P_V\vv$ belongs to the relative regular subdifferential of
$f$ at $\x$, as in \eqref{eq:pc3-proj}, and hence to the relative
subdifferential $\partial f(\x)$ by
Proposition~\ref{prop:subdiff-basics}-(i).  Let $\x,\y\in O$,
$\vv\in\hat\partial\bar f(\x)$, and $\uu\in\hat\partial\bar f(\y)$.  Since
$\y-\x\in V$, \eqref{eq:pc3-f-mono} applied to $P_V\vv$ and $P_V\uu$ gives
\[
\langle\uu,\y-\x\rangle=\langle P_V\uu,\y-\x\rangle
\geq\psi_\gamma(\langle P_V\vv,\y-\x\rangle)
=\psi_\gamma(\langle\vv,\y-\x\rangle).
\]
So $\hat\partial\bar f|_O$ is $\gamma$--tilted monotone.  Moreover,
\[
  \operatorname{ri}(\operatorname{dom}(\bar f))=O,
\]
because $O\subseteq\operatorname{dom}(\bar f)
\subseteq\overline{\mathcal K}$ and $\operatorname{dom}(\bar f)$ is
convex by Theorem~\ref{thm:mwc-domain-is-convex}.  The
$\gamma$--tilted convex function $\bar f$ is radially continuous by
Corollary~\ref{cor:mwc-radially-continuous} and locally Lipschitz on $O$
by Theorem~\ref{thm:gamma-MW-convex-is-locally-Lipschitz}.  Hence
condition~(iv) of
Theorem~\ref{thm:mw-mono-alternative-characterizations} holds for $\bar f$,
and that theorem proves that
$\overline{\partial}\bar f$ is $\gamma$--tilted monotone.

Finally, under the additional pointwise-convergence assumption,
Theorem~\ref{thm:gamma-MW-convex-pointwise-limit} gives
$\op{lsc}(g)=\bar f$.  Hence
$\overline{\partial}(\op{lsc}(g))=\overline{\partial}\bar f$ is
$\gamma$--tilted monotone.
\end{proof}

\subsection{Proof of Theorem~\ref{thm:gamma-MW-Jensen-basics}}
\label{apx:gamma-MW-Jensen-basics}

\begin{proof}
Throughout, for $\y\ne\z$ in $\mathcal K$ and $\lambda\in(0,1)$ we put
$\x:=\lambda\y+(1-\lambda)\z$ and use the slopes $s_\pm$
of~\eqref{eq:secant-slopes}.  Since $\norm{\x-\z}$ and $\norm{\y-\x}$ are
positive, $s_->0$ exactly when $f(\x)>f(\z)$, and $s_+\ge0$ exactly
when $f(\y)\ge f(\x)$.

\medskip
\noindent\textbf{(i): Quasiconvexity and convexity of the effective domain.}
Suppose that $f$ is $\gamma$--tilted secant convex.  The definition
requires $\x\in\mathcal K$ for all $\y\ne\z$ in $\mathcal K$ and
$\lambda\in(0,1)$, so $\mathcal K$ is convex.  If $f(\x)>f(\z)$, then
$s_->0$, so $\psi_\gamma(s_-)\ge0$ is defined, and~\eqref{eq:secant-form}
gives $s_+\ge0$, that is, $f(\y)\ge f(\x)$.  Hence
$f(\x)\le\max\{f(\y),f(\z)\}$.  This inequality is immediate when
$\y=\z$, when $\lambda\in\{0,1\}$, and when $\y$ or $\z$ lies in
$\mathcal X\setminus\mathcal K$, so $f$ is quasiconvex.

\medskip
\noindent\textbf{(ii): Convex functions.}
Let $f$ be convex.  Then $\mathcal K$ is convex, and the slopes of the
secants of $f$ increase, so $s_+\ge s_-$.  By
Lemma~\ref{lem:psi_basics}(2), $s_-\ge\psi_\gamma(s_-)$ whenever
$\psi_\gamma(s_-)$ is defined.  Hence~\eqref{eq:secant-form} holds, and
$f$ is $\gamma$--tilted secant convex.

\medskip
\noindent\textbf{(iii): Suprema.}
Here and in (v) we use the Jensen form of
Proposition~\ref{prop:secant-jensen}, whose right side is nondecreasing
in the values.  For every $\gamma\in[0,1]$ and $\lambda\in[0,1]$, the map
$(a,b)\mapsto M_{\gamma,\lambda}(a,b)$ is nondecreasing in each
argument and satisfies
\begin{equation}
M_{\gamma,\lambda}(a,b)\leq\max\{a,b\}.
\label{eq:gamma-MW-Jensen-mean-monotone-and-bounded}
\end{equation}
For $\gamma=0$, these facts follow immediately from
\eqref{eq:zero-MW-Jensen-mean}.  Now let $\gamma\in(0,1]$, and put
$D_1:=\gamma\lambda+(1-\lambda)$ and $D_2:=\lambda+\gamma(1-\lambda)$.
Since $\gamma\le1$, we have $D_1\ge\gamma\lambda+\gamma(1-\lambda)=\gamma$
and likewise $D_2\ge\gamma$, so
\begin{equation}
D_1\ge\gamma>0,
\qquad
D_2\ge\gamma>0.
\label{eq:pc1-J-denominators}
\end{equation}
By~\eqref{eq:gamma-MW-Jensen-mean}, with
$\omega:=\gamma\lambda/D_1$ and $\nu:=\lambda/D_2$,
\begin{equation}
M_{\gamma,\lambda}(a,b)
=
\begin{cases}
\omega a+(1-\omega)b,& a\leq b,\\
\nu a+(1-\nu)b,& a>b,
\end{cases}
\qquad
1-\omega=\frac{1-\lambda}{D_1},
\quad
1-\nu=\frac{\gamma(1-\lambda)}{D_2},
\label{eq:pc1-J-weights}
\end{equation}
and $\omega,\nu\in[0,1]$.  Each case of~\eqref{eq:pc1-J-weights} is a
weighted arithmetic mean of $a$ and $b$ with nonnegative coefficients,
so it is at most $\max\{a,b\}$.  Both cases give $M_{\gamma,\lambda}(a,a)=a$.

For monotonicity in $a$, fix $b$ and let $a_1<a_2$.  If $a_1$ and $a_2$
fall in the same case of~\eqref{eq:pc1-J-weights}, then
$M_{\gamma,\lambda}(a_1,b)\le M_{\gamma,\lambda}(a_2,b)$ because
$\omega,\nu\ge0$.  If $a_1\le b<a_2$, then
$M_{\gamma,\lambda}(a_1,b)\le\max\{a_1,b\}=b$, while
$M_{\gamma,\lambda}(a_2,b)=\nu a_2+(1-\nu)b\ge b$.  For
monotonicity in $b$, fix $a$ and let $b_1<b_2$.  If $b_1$ and $b_2$ fall
in the same case, the claim follows from $1-\omega,1-\nu\ge0$.  If
$b_1<a\le b_2$, then $M_{\gamma,\lambda}(a,b_1)\le\max\{a,b_1\}=a$,
while $M_{\gamma,\lambda}(a,b_2)=\omega a+(1-\omega)b_2\ge a$.  Hence
$M_{\gamma,\lambda}$ is nondecreasing in each variable.

Let
$\{f_i:\mathcal X\to\mathbb R\cup\{+\infty\}\}_{i\in I}$ be a
non-empty family of $\gamma$--tilted secant convex functions and
$g:=\sup_{i\in I}f_i$.  Since $I\ne\emptyset$ and each $f_i>-\infty$,
$g$ takes values in $\mathbb R\cup\{+\infty\}$.
Let $\y,\z\in\operatorname{dom}(g)$ and $\lambda\in[0,1]$.  Since
$-\infty<f_i\leq g$, the values $f_i(\y)$ and $f_i(\z)$ are finite for
every $i\in I$.  The $\gamma$--tilted secant convexity of $f_i$
and the monotonicity in~\eqref{eq:gamma-MW-Jensen-mean-monotone-and-bounded},
applied with $f_i(\y)\le g(\y)$ and $f_i(\z)\le g(\z)$, give
\begin{align*}
f_i(\lambda\y+(1-\lambda)\z)
&\leq
M_{\gamma,\lambda}\bigl(f_i(\y),f_i(\z)\bigr)
\leq
M_{\gamma,\lambda}\bigl(g(\y),g(\z)\bigr)
\qquad(i\in I).
\end{align*}
Taking the supremum over $i$ proves that $g$ is $\gamma$--tilted secant convex.

\medskip
\noindent\textbf{(iv): Convex restrictions.}
Now let $f$ be $\gamma$--tilted secant convex and let
$\mathcal C\subseteq\mathcal X$ be convex.  If
$\y,\z\in\operatorname{dom}(f|_{\mathcal C})=\mathcal C\cap\mathcal K$,
then their convex combinations belong to $\mathcal C$, and the defining
condition for $f$ is exactly the defining condition for
$f|_{\mathcal C}$.  Thus $f|_{\mathcal C}$ is $\gamma$--tilted secant convex.

\medskip
\noindent\textbf{(v): The lsc envelope.}
Assume that $\gamma\in(0,1]$ and that $f$ is $\gamma$--tilted secant convex.
By (i), $\mathcal K$ is convex.  If $\mathcal K=\emptyset$, then
$f\equiv+\infty$, $\op{lsc}(f)\equiv+\infty$, and both claims are
immediate.  So assume $\mathcal K\ne\emptyset$, and put
$A:=\operatorname{aff}(\mathcal K)$ and $k:=\dim A$.  We use two more
properties of $M_{\gamma,\lambda}$.  First, by~\eqref{eq:pc1-J-weights},
$M_{\gamma,\lambda}(a,b)-a$ equals $(1-\omega)(b-a)$ or $(1-\nu)(b-a)$.
By~\eqref{eq:pc1-J-weights} and~\eqref{eq:pc1-J-denominators},
$1-\omega\le(1-\lambda)/\gamma$ and $1-\nu\le1-\lambda\le(1-\lambda)/\gamma$.
Hence
\begin{equation}
\bigl|M_{\gamma,\lambda}(a,b)-a\bigr|
\le\frac{1-\lambda}{\gamma}\,|b-a|
\qquad\forall a,b\in\mathbb R.
\label{eq:pc1-J-near-a}
\end{equation}
Second, for fixed $\lambda$, the map $(a,b)\mapsto M_{\gamma,\lambda}(a,b)$
is continuous, because both cases in~\eqref{eq:pc1-J-weights} are
continuous and they agree when $a=b$.

We first show a local upper bound: for every
$\w\in\operatorname{ri}(\mathcal K)$ there are $r>0$ and $C\in\mathbb R$
such that
\begin{equation}
\mathbb B(\w,r)\cap A\subseteq\mathcal K
\qquad\text{and}\qquad
f\le C\ \text{ on }\mathbb B(\w,r)\cap A.
\label{eq:pc1-J-upper}
\end{equation}
If $k=0$, then $A=\{\w\}$, and $r:=1$ and $C:=f(\w)$ work.
Let $k\ge1$.  Choose $\delta>0$ with
$\mathbb B(\w,\delta)\cap A\subseteq\mathcal K$, choose an orthonormal
basis $\mathbf e_1,\ldots,\mathbf e_k$ of $A-A$, and put
$r:=\delta/(2\sqrt k)$.  The $2^k$ points
$\w+r\sum_i\sigma_i\mathbf e_i$ with $\sigma_i\in\{-1,1\}$ lie in $A$ at
distance $r\sqrt k=\delta/2$ from $\w$, so they lie in $\mathcal K$.
Their convex hull is the cube
$\{\w+\sum_i s_i\mathbf e_i:|s_i|\le r\}$, which contains
$\mathbb B(\w,r)\cap A$.  Quasiconvexity from (i) extends to finite
convex combinations by induction on the number of points: if
$\mu_m<1$, then
$\sum_{j=1}^m\mu_j\mathbf p_j=(1-\mu_m)\mathbf q+\mu_m\mathbf p_m$ with
$\mathbf q:=\sum_{j<m}\mu_j\mathbf p_j/(1-\mu_m)$.  Hence $f\le C$
on the cube, where $C$ is the largest value of $f$ at its vertices.  This
proves~\eqref{eq:pc1-J-upper}, since
$\mathbb B(\w,r)\cap A\subseteq\mathbb B(\w,\delta)\cap A\subseteq\mathcal K$.

To show that $f$ is lsc on $\operatorname{ri}(\mathcal K)$, fix
$\x\in\operatorname{ri}(\mathcal K)$, and take $r$ and $C$
from~\eqref{eq:pc1-J-upper} with $\w=\x$.  Let
$\uu\in\mathbb B(\x,r)\cap A$.  The reflected point $\vv:=2\x-\uu$ lies
in $A$ and satisfies $\|\vv-\x\|=\|\uu-\x\|<r$, so
$f(\uu)\le C$ and $f(\vv)\le C$.  Since $\x=\frac12\uu+\frac12\vv$, the
inequality \eqref{eq:gamma-MW-Jensen-convex} and monotonicity of $M_{\gamma,1/2}$ in its second
argument give
\[
f(\x)
\leq
M_{\gamma,1/2}\bigl(f(\uu),f(\vv)\bigr)
\leq
M_{\gamma,1/2}\bigl(f(\uu),C\bigr)
=\frac{\gamma f(\uu)+C}{\gamma+1},
\]
where the equality is the case $a\le b$ of~\eqref{eq:gamma-MW-Jensen-mean}
with $\lambda=1/2$.  Solving for $f(\uu)$ gives
\begin{equation}
L\le f(\uu)\le C
\qquad\forall\uu\in\mathbb B(\x,r)\cap A,
\qquad
L:=\frac{(1+\gamma)f(\x)-C}{\gamma}.
\label{eq:pc1-J-lower}
\end{equation}

Let $\uu_n\in\mathcal K$ tend to $\x$.  Since $\mathcal K\subseteq A$,
after discarding finitely many terms we have
$\uu_n\in\mathbb B(\x,r)\cap A$.  Put
$\tau_n:=\sqrt{\|\uu_n-\x\|}+1/n$.  For all sufficiently large $n$,
$\tau_n\in(0,1)$; moreover $\tau_n\to0$ and
$\|\uu_n-\x\|/\tau_n\le\sqrt{\|\uu_n-\x\|}\to0$.  Define
\[
\vv_n
:=
\frac{\x-(1-\tau_n)\uu_n}{\tau_n}
=\x+\frac{1-\tau_n}{\tau_n}(\x-\uu_n).
\]
Then $\vv_n\in A$ and $\|\vv_n-\x\|\le\|\uu_n-\x\|/\tau_n\to0$.  So for
all sufficiently large $n$, $\vv_n\in\mathbb B(\x,r)\cap A$, and
by~\eqref{eq:pc1-J-lower} both $f(\uu_n)$ and $f(\vv_n)$ lie in $[L,C]$.
Since $\x=(1-\tau_n)\uu_n+\tau_n\vv_n$, inequality \eqref{eq:gamma-MW-Jensen-convex}
and~\eqref{eq:pc1-J-near-a} with $\lambda=1-\tau_n$ give
\[
f(\x)
\leq
M_{\gamma,1-\tau_n}\bigl(f(\uu_n),f(\vv_n)\bigr)
\leq
f(\uu_n)+\frac{\tau_n}{\gamma}\bigl|f(\vv_n)-f(\uu_n)\bigr|
\leq
f(\uu_n)+\frac{\tau_n}{\gamma}(C-L).
\]
Since $\tau_n\to0$, taking the lower limit gives
\[
f(\x)\leq\liminf_{n\to\infty}f(\uu_n).
\]
Now choose $\w_n\in\mathcal X$ with $\w_n\to\x$ and
$f(\w_n)\to\op{lsc}(f)(\x)$.  Since
$\op{lsc}(f)(\x)\le f(\x)<+\infty$, eventually $\w_n\in\mathcal K$, and
the last display, applied to $\uu_n:=\w_n$, gives $f(\x)\le\op{lsc}(f)(\x)$.  This proves
\begin{equation}
\label{eq:gamma-MW-Jensen-basics-lsc-on-ri}
f=\op{lsc}(f)
\qquad\text{on }\operatorname{ri}(\mathcal K).
\end{equation}

Set $\bar f:=\op{lsc}(f)$.  We next show that $\bar f$ never takes the
value $-\infty$.  Outside $\overline{\mathcal K}$, the function $\bar f$
equals $+\infty$ by definition.  Fix $\x\in\operatorname{ri}(\mathcal K)$
and $\y\in\overline{\mathcal K}$.  Choose $\delta>0$ with
$\mathbb B(\x,\delta)\cap A\subseteq\mathcal K$, and choose
$\theta\in(0,1)$ so small that
$\theta\|\x-\y\|/(1-\theta)<\delta/2$.  Then
\[
\z:=\frac{\x-\theta\y}{1-\theta}
=\x+\frac{\theta}{1-\theta}(\x-\y)
\in A,
\qquad
\|\z-\x\|<\delta/2,
\]
so $\mathbb B(\z,\delta/2)\cap A\subseteq\mathbb B(\x,\delta)\cap A\subseteq\mathcal K$
and $\z\in\operatorname{ri}(\mathcal K)$.  Take $r_{\z}$ and $C$
from~\eqref{eq:pc1-J-upper} with $\w=\z$.  Choose
$\y_n\in\mathcal X$ with $\y_n\to\y$ and $f(\y_n)\to\bar f(\y)$.  If
$\bar f(\y)=+\infty$, there is nothing to prove.  Otherwise
$\y_n\in\mathcal K$ for all large $n$.  The points
\[
\z_n:=\frac{\x-\theta\y_n}{1-\theta}
\]
lie in $A$ and tend to $\z$, so for large $n$ they lie in
$\mathbb B(\z,r_{\z})\cap A$ and satisfy $f(\z_n)\le C$.  Since
$\x=\theta\y_n+(1-\theta)\z_n$, inequality \eqref{eq:gamma-MW-Jensen-convex} and monotonicity
give
\[
f(\x)
\leq
M_{\gamma,\theta}\bigl(f(\y_n),f(\z_n)\bigr)
\leq
M_{\gamma,\theta}\bigl(f(\y_n),C\bigr).
\]
If $f(\y_n)\le C$, the case $a\le b$ of~\eqref{eq:gamma-MW-Jensen-mean}
turns this into
$f(\x)\le\bigl(\gamma\theta f(\y_n)+(1-\theta)C\bigr)/(\gamma\theta+1-\theta)$.
Solving for $f(\y_n)$ shows that, for all large $n$,
\[
f(\y_n)
\ge
\min\left\{C,\
\frac{(\gamma\theta+1-\theta)f(\x)-(1-\theta)C}{\gamma\theta}\right\}.
\]
Letting $n\to\infty$ gives $\bar f(\y)>-\infty$.
Thus $\bar f:\mathbb R^d\to\mathbb R\cup\{+\infty\}$.

For the $\gamma$--tilted secant convexity of $\bar f$, fix
$\y,\z\in\operatorname{dom}(\bar f)$ and $\lambda\in[0,1]$.  By the
definition of $\bar f$ as a lower limit, there are sequences in
$\mathcal X$, which lie in $\mathcal K$ for large $n$ because
$\bar f(\y),\bar f(\z)$ are finite, such that
\[
\y_n\to\y,
\qquad
f(\y_n)\to\bar f(\y),
\qquad
\z_n\to\z,
\qquad
f(\z_n)\to\bar f(\z).
\]
The points $\lambda\y_n+(1-\lambda)\z_n$ lie in $\mathcal K$ and tend to
$\lambda\y+(1-\lambda)\z$.  The definition of $\bar f$, inequality
\eqref{eq:gamma-MW-Jensen-convex} for $f$, and the continuity of $M_{\gamma,\lambda}$ give
\begin{align*}
\bar f(\lambda\y+(1-\lambda)\z)
&\leq
\liminf_{n\to\infty}
f(\lambda\y_n+(1-\lambda)\z_n)\\
&\leq
\lim_{n\to\infty}
M_{\gamma,\lambda}\bigl(f(\y_n),f(\z_n)\bigr)
=
M_{\gamma,\lambda}\bigl(\bar f(\y),\bar f(\z)\bigr).
\end{align*}
Thus $\bar f$ is $\gamma$--tilted secant convex, and
\eqref{eq:gamma-MW-Jensen-basics-lsc-on-ri} proves the remaining
claim about the relative interior.

\medskip
\noindent\textbf{(vi): Restrictions to lines.}
Finally, suppose that $f$ is $\gamma$--tilted secant convex and let
$\ell$ be a line intersecting $\mathcal X$.  The set
$\ell\cap\mathcal X$ is convex, so $f|_\ell$ is $\gamma$--tilted secant convex by (iv).  Conversely,
suppose all the stated line restrictions are $\gamma$--tilted secant convex.  Any two distinct points $\y,\z\in\mathcal K$ lie on a line
$\ell$, and the condition for $f|_\ell$ at $\y$ and $\z$ is the condition
for $f$ at $\y$ and $\z$, so $f$ is $\gamma$--tilted secant convex.
\end{proof}

\subsection{Proof of Theorem~\ref{thm:gamma-MW-Jensen-equivalent-def}: Characterizations of \texorpdfstring{$\gamma$}{gamma}--tilted secant convexity}
\label{apx:gamma-MW-Jensen-equivalent-def}

\subsubsection{A Scalar Characterization of the Finite-dimensional Mean}
\label{apx:mean-characterization}

We collect the sublevel, balance, threshold, max, and variational
descriptions of the mean $M^{(m)}_{\gamma,\boldsymbol\lambda}$, together with its binary
specialization and its limit as $\gamma\downarrow0$.

\begin{lemma}[Equivalent formulations of the finite-dimensional mean]
\label{lem:finite-dimensional-gamma-MW-Jensen-mean-characterization}
Let $m\geq2$, $\boldsymbol\lambda\in\Delta_m$,
$\a\in\mathbb R^m$, and $\gamma\in(0,1]$.  Set
\[
\mu:=M^{(m)}_{\gamma,\boldsymbol\lambda}(\a).
\]
For $r\in\mathbb R$, write
$r_+:=\max\{r,0\}$ and $r_-:=\max\{-r,0\}$.  Then the following
equivalent characterizations hold.
\begin{itemize}
\item[(i)] For every $c\in\mathbb R$,
\begin{equation}
\label{eq:finite-dimensional-gamma-MW-Jensen-mean-characterization}
c\leq\mu
\quad\Longleftrightarrow\quad
\sum_{i=1}^m
\lambda_i\overline{\psi}_{\sqrt\gamma}(a_i-c)
\geq0.
\end{equation}

\item[(ii)] The number $\mu$ is the unique $t\in\mathbb R$ satisfying
\[
\sum_{i=1}^m
\lambda_i\overline{\psi}_{\sqrt\gamma}(a_i-t)
=0.
\]

\item[(iii)] The number $\mu$ is the unique $t\in\mathbb R$ such that,
upon setting
\[
I_-(t):=\{i:a_i<t\},
\qquad
I_+(t):=\{i:a_i\geq t\},
\]
one has
\[
t
=
\frac{
\gamma\sum_{i\in I_-(t)}\lambda_i a_i
+\sum_{i\in I_+(t)}\lambda_i a_i
}{
\gamma\sum_{i\in I_-(t)}\lambda_i
+\sum_{i\in I_+(t)}\lambda_i
}.
\]

\item[(iv)] One has
\[
\mu
=
\max_{\w\in[\gamma,1]^m}
\frac{
\sum_{i=1}^m\lambda_i w_i a_i
}{
\sum_{i=1}^m\lambda_i w_i
}.
\]

\item[(v)] One has
\[
\mu
=
\underset{t\in\mathbb R}{\operatorname{argmin}}
\sum_{i=1}^m\lambda_i
\left((a_i-t)_+^2+\gamma(a_i-t)_-^2\right),
\]
where the minimizer is unique.
\end{itemize}
The sets in~\textup{(iii)} can be identified by sorting the
coordinates of $\a$ and selecting the interval containing the
corresponding value on the right-hand side.

In the binary case, with $m=2$ and
$\boldsymbol\lambda=(\lambda,1-\lambda)$,
\[
M_{\gamma,\lambda}(a,b)
=
\max\left\{
\frac{\gamma\lambda a+(1-\lambda)b}
     {\gamma\lambda+(1-\lambda)},
\frac{\lambda a+\gamma(1-\lambda)b}
     {\lambda+\gamma(1-\lambda)}
\right\}.
\]
Finally,
\[
\lim_{\eta\downarrow0}
M^{(m)}_{\eta,\boldsymbol\lambda}(\a)
= \max\{a_i:\lambda_i>0\}
\le \max_i a_i.
\]
Thus this limit is $\max_i a_i$ when every $\lambda_i$ is positive;
in particular, for $m=2$ and $\lambda\in(0,1)$, it is
$\max\{a,b\}$.
\end{lemma}

\begin{proof}
Since $\gamma>0$ and $\sum_{i=1}^m\lambda_i=1$, every
$\w\in[\gamma,1]^m$ satisfies
\begin{equation}
\label{eq:pc2-positive-denominator}
\sum_{i=1}^m\lambda_i w_i
\geq
\gamma\sum_{i=1}^m\lambda_i
=\gamma
>0.
\end{equation}
For $S\subseteq\{1,\ldots,m\}$, let $R_S$ denote the corresponding
ratio in~\eqref{eq:gamma-MW-Jensen-subset-max}, and for
$c\in\mathbb R$ put
\[
G_S(c):=
\sum_{i\in S}\lambda_i(a_i-c)
+\gamma\sum_{i\notin S}\lambda_i(a_i-c).
\]
The denominator of $R_S$ is the sum in
\eqref{eq:pc2-positive-denominator} with $w_i=1$ for $i\in S$ and
$w_i=\gamma$ otherwise, so it is positive.  Multiplying the inequality
$c\leq R_S$ by this denominator shows that $c\leq R_S$ if and only if
$G_S(c)\geq0$.  Since $\mu=\max_S R_S$, we get
\begin{equation}
\label{eq:pc2-mu-via-GS}
c\leq\mu
\quad\Longleftrightarrow\quad
\max_{S\subseteq\{1,\ldots,m\}}G_S(c)\geq0.
\end{equation}
The term of index $i$ in $G_S(c)$ is $\lambda_i(a_i-c)$ if $i\in S$
and $\gamma\lambda_i(a_i-c)$ if $i\notin S$.  Since $\lambda_i\geq0$,
the choice $S=\{i:a_i\geq c\}$ maximizes every term at once, and the
maximal term is $\lambda_i\max\{a_i-c,\gamma(a_i-c)\}$.  By the
definition of $\overline{\psi}_{\sqrt\gamma}$,
\begin{equation}
\label{eq:pc2-max-psibar}
\max\{r,\gamma r\}
=
\sqrt\gamma\,\overline{\psi}_{\sqrt\gamma}(r)
=
\begin{cases}
\gamma r,&r<0,\\
r,&r\geq0,
\end{cases}
\qquad r\in\mathbb R.
\end{equation}
Consequently,
\[
\max_{S\subseteq\{1,\ldots,m\}}G_S(c)
=
\sqrt\gamma\sum_{i=1}^m
\lambda_i\overline{\psi}_{\sqrt\gamma}(a_i-c).
\]
Since $\sqrt\gamma>0$, substituting this into
\eqref{eq:pc2-mu-via-GS} proves
\eqref{eq:finite-dimensional-gamma-MW-Jensen-mean-characterization}.

Let
\[
Q(t):=
\sum_{i=1}^m
\lambda_i\overline{\psi}_{\sqrt\gamma}(a_i-t).
\]
Each function $t\mapsto\overline{\psi}_{\sqrt\gamma}(a_i-t)$ is
continuous and strictly decreasing, with limits $+\infty$ and
$-\infty$ as $t$ tends to $-\infty$ and $+\infty$, respectively.  The
weights are nonnegative and at least one of them is positive.  Hence
$Q$ is continuous and strictly decreasing, with the same limits, and it
has a unique root $t_0$.  By
\eqref{eq:finite-dimensional-gamma-MW-Jensen-mean-characterization},
\[
Q(c)\geq0
\quad\Longleftrightarrow\quad
c\leq\mu.
\]
Since $Q(t_0)=0$, this gives $t_0\leq\mu$.  For every $t>t_0$ we have
$Q(t)<0$, so this gives $t>\mu$; hence $\mu\leq t_0$.  Thus $\mu=t_0$
is the unique root of $Q$.  This proves the equivalence of
the sublevel characterization~\textup{(i)} and the balance
characterization~\textup{(ii)}.

By \eqref{eq:pc2-max-psibar}, for every $t\in\mathbb R$,
\[
\sqrt\gamma\,Q(t)
=
\gamma\sum_{i\in I_-(t)}\lambda_i(a_i-t)
+\sum_{i\in I_+(t)}\lambda_i(a_i-t).
\]
Rearranging, $Q(t)=0$ holds if and only if
\[
t\left(
\gamma\sum_{i\in I_-(t)}\lambda_i
+\sum_{i\in I_+(t)}\lambda_i
\right)
=
\gamma\sum_{i\in I_-(t)}\lambda_i a_i
+\sum_{i\in I_+(t)}\lambda_i a_i.
\]
The factor multiplying $t$ is positive by
\eqref{eq:pc2-positive-denominator}.  So $Q(t)=0$ if and only if $t$
satisfies the identity in~\textup{(iii)}.  Since $\mu$ is the unique
root of $Q$, it is the unique $t$ satisfying~\textup{(iii)}.

For the box-max representation, set
$S_*:=\{i:a_i\geq\mu\}$.  The balance equation $Q(\mu)=0$, multiplied
by $\sqrt\gamma$ and rewritten with \eqref{eq:pc2-max-psibar}, gives
\begin{equation}
\label{eq:pc2-balance-at-mu}
\sum_{i\in S_*}\lambda_i(a_i-\mu)
+\gamma\sum_{i\notin S_*}\lambda_i(a_i-\mu)
=0.
\end{equation}
For $\w\in[\gamma,1]^m$, let
$N_\w:=\sum_{i=1}^m\lambda_i w_i a_i$ and
$D_\w:=\sum_{i=1}^m\lambda_i w_i$ denote the numerator and
denominator of the ratio in~\textup{(iv)}, so that
$N_\w-\mu D_\w=\sum_{i=1}^m\lambda_i w_i(a_i-\mu)$.  Choose $w_i=1$
for $i\in S_*$ and $w_i=\gamma$ otherwise.  Then
\eqref{eq:pc2-balance-at-mu} states that $N_\w-\mu D_\w=0$, so the
corresponding ratio equals $\mu$.

Now let $\w\in[\gamma,1]^m$ be arbitrary.  For $i\in S_*$, we have
$a_i-\mu\geq0$ and $w_i\leq1$, so
$\lambda_i w_i(a_i-\mu)\leq\lambda_i(a_i-\mu)$.  For $i\notin S_*$, we
have $a_i-\mu<0$ and $w_i\geq\gamma$, so
$\lambda_i w_i(a_i-\mu)\leq\gamma\lambda_i(a_i-\mu)$.  Summing and
using \eqref{eq:pc2-balance-at-mu} gives
\[
N_\w-\mu D_\w
\leq
\sum_{i\in S_*}\lambda_i(a_i-\mu)
+\gamma\sum_{i\notin S_*}\lambda_i(a_i-\mu)
=0.
\]
Since $D_\w>0$ by \eqref{eq:pc2-positive-denominator}, every ratio is
at most $\mu$.  Thus its maximum is $\mu$, proving~\textup{(iv)}.

Denote by $\Phi(t)$ the objective in condition~\textup{(v)}.  For
every $a\in\mathbb R$, the functions $t\mapsto(a-t)_+^2$ and
$t\mapsto(a-t)_-^2$ are continuously differentiable, with derivatives
$-2(a-t)_+$ and $2(a-t)_-$.  Hence $\Phi$ is continuously
differentiable, and \eqref{eq:pc2-max-psibar} gives
\[
\Phi'(t)
=
-2\left(
\sum_{i:a_i\geq t}\lambda_i(a_i-t)
+\gamma\sum_{i:a_i<t}\lambda_i(a_i-t)
\right)
=-2\sqrt\gamma\,Q(t).
\]
Since $Q$ is strictly decreasing with unique root $\mu$, we have
$\Phi'<0$ on $(-\infty,\mu)$ and $\Phi'>0$ on $(\mu,\infty)$.  So
$\Phi$ is strictly decreasing on $(-\infty,\mu]$ and strictly
increasing on $[\mu,\infty)$, and $\mu$ is its unique minimizer.
This proves~\textup{(v)}.

For $m=2$, write
\[
R_1:=\frac{\gamma\lambda a+(1-\lambda)b}{D_1},
\quad
D_1:=\gamma\lambda+(1-\lambda),
\qquad
R_2:=\frac{\lambda a+\gamma(1-\lambda)b}{D_2},
\quad
D_2:=\lambda+\gamma(1-\lambda)
\]
for the two ratios in the displayed single-max formula.  Both
denominators are positive by \eqref{eq:pc2-positive-denominator}.
Since the weights of $a$ and $b$ in each ratio sum to one,
\[
R_1=a+\beta_1(b-a),
\quad
\beta_1:=\frac{1-\lambda}{D_1},
\qquad
R_2=a+\beta_2(b-a),
\quad
\beta_2:=\frac{\gamma(1-\lambda)}{D_2}.
\]
A direct computation gives $D_2-\gamma D_1=\lambda(1-\gamma^2)$, and
therefore
\[
R_1-R_2
=(\beta_1-\beta_2)(b-a)
=\frac{(1-\lambda)(D_2-\gamma D_1)}{D_1D_2}(b-a)
=\frac{\lambda(1-\lambda)(1-\gamma^2)}{D_1D_2}(b-a).
\]
The coefficient of $b-a$ is nonnegative.  Hence $R_1\geq R_2$ when
$a\leq b$, and $R_2\geq R_1$ when $a>b$.  By
\eqref{eq:gamma-MW-Jensen-mean}, $M_{\gamma,\lambda}(a,b)$ equals $R_1$
when $a\leq b$ and $R_2$ when $a>b$, so in both cases it equals
$\max\{R_1,R_2\}$.
The subset-max~\eqref{eq:gamma-MW-Jensen-subset-max} for $m=2$ has four ratios: $S=\{1\}$ gives $R_2$, $S=\{2\}$ gives $R_1$, and $S=\emptyset$ and $S=\{1,2\}$ both give the arithmetic mean $a+(1-\lambda)(b-a)$.
Since $D_1\le1$ and $D_2\ge\gamma$, we have $\beta_1\ge1-\lambda\ge\beta_2$, so the arithmetic mean lies between $R_1$ and $R_2$.
Hence $M^{(2)}_{\gamma,(\lambda,1-\lambda)}(a,b)=\max\{R_1,R_2\}=M_{\gamma,\lambda}(a,b)$.

For the limit as $\eta\downarrow0$, let
$P:=\{i:\lambda_i>0\}$, which is non-empty, and
$a^*:=\max\{a_i:i\in P\}$.  For $S\subseteq\{1,\ldots,m\}$ and
$\eta\in(0,1]$, let $R_S(\eta)$ be the ratio
in~\eqref{eq:gamma-MW-Jensen-subset-max} with $\eta$ in place of
$\gamma$.  If $\sum_{i\in S}\lambda_i>0$, then, as $\eta\downarrow0$,
\[
R_S(\eta)
\longrightarrow
\frac{\sum_{i\in S}\lambda_i a_i}{\sum_{i\in S}\lambda_i},
\]
which is a weighted average of the $a_i$ with $i\in S\cap P$ and hence
at most $a^*$.  If $\sum_{i\in S}\lambda_i=0$, then the factor $\eta$
cancels and $R_S(\eta)=\sum_{i\notin S}\lambda_i a_i$, a weighted
average of the $a_i$ with $i\in P$ that does not depend on $\eta$ and
is at most $a^*$.  Thus every ratio has a limit bounded above by
$a^*$.  Choose $j\in P$ with $a_j=a^*$.  Then
$R_{\{j\}}(\eta)\to a_j=a^*$.  The maximum of finitely many convergent
functions converges to the maximum of their limits, so
$M^{(m)}_{\eta,\boldsymbol\lambda}(\a)\to a^*$.  Finally,
$a^*\leq\max_i a_i$, with equality when $P=\{1,\ldots,m\}$.
\end{proof}

\subsubsection{Equivalence of (i) and (ii)}

\begin{proof}
Fix
$\y,\z\in\mathcal K$ and $\lambda\in(0,1)$, and set
$\x:=\lambda\y+(1-\lambda)\z$.  If $\y=\z$, then $\x=\y$, and
\eqref{eq:gamma-MW-Jensen-equivalent-def} holds with equality, so let
$\y\ne\z$.  We first show that $\x\in\mathcal K$
under either condition.  Under condition~\textup{(i)}, this is part of
Definition~\ref{def:gamma-MW-Jensen-convex}.  Under
condition~\textup{(ii)}, suppose that $f(\x)=+\infty$.  Then
$f(\y)-f(\x)=f(\z)-f(\x)=-\infty$, both transformed differences
in~\eqref{eq:gamma-MW-Jensen-equivalent-def} equal $-\infty$, and the
left-hand side of~\eqref{eq:gamma-MW-Jensen-equivalent-def} is
$-\infty$.  This contradicts the asserted inequality, so again
$\x\in\mathcal K$.

Therefore $a:=f(\y)$, $b:=f(\z)$, and $m:=f(\x)$ are real, and the
slopes~\eqref{eq:secant-slopes} satisfy~\eqref{eq:secant-values}.  Put
$\rho:=\sqrt\gamma>0$ and $c:=\lambda(1-\lambda)\norm{\y-\z}>0$.  By
\eqref{eq:secant-values}, $\lambda(a-m)=cs_+$ and $(1-\lambda)(b-m)=-cs_-$.
Positive homogeneity of $\overline\psi_\rho$ and
Lemma~\ref{lem:psi_basics}(3) give
\[
\lambda\,\overline\psi_\rho(a-m)+(1-\lambda)\,\overline\psi_\rho(b-m)
=c\bigl[\overline\psi_\rho(s_+)+\overline\psi_\rho(-s_-)\bigr]
=c\bigl[\overline\psi_\rho(s_+)-\psi_\rho(s_-)\bigr].
\]
So~\eqref{eq:gamma-MW-Jensen-equivalent-def} holds if and only if
$\overline\psi_\rho(s_+)\ge\psi_\rho(s_-)$.  Since $\psi_\rho$ is
increasing and inverts $\overline\psi_\rho$, this is equivalent to
$s_+\ge\psi_\rho(\psi_\rho(s_-))=\psi_\gamma(s_-)$ by
Lemma~\ref{lem:psi_basics}(1), (4), and~(6), which
is~\eqref{eq:secant-form}.
\end{proof}

\subsubsection{Equivalence of (iii) and (iv)}

\begin{proof}
Assume condition~\textup{(iii)}.  Let $m\geq2$, let
$\y_1,\ldots,\y_m\in\mathcal K$, and let
$\lambda_1,\ldots,\lambda_m>0$ satisfy $\sum_{i=1}^m\lambda_i=1$.  Put
$\x:=\sum_{i=1}^m\lambda_i\y_i$.  Every $f(\y_i)$ is real, and every
ratio in~\eqref{eq:gamma-MW-Jensen-subset-max} has a positive
denominator, so the right-hand side of
\eqref{eq:gamma-MW-Jensen-finite-point} is a real number.  Hence
$f(\x)<+\infty$, and properness implies that $\x\in\mathcal K$.
Lemma~\ref{lem:finite-dimensional-gamma-MW-Jensen-mean-characterization}--\textup{(i)},
applied with
\[
a_i=f(\y_i),
\qquad
c=f\left(\sum_{i=1}^m\lambda_i\y_i\right),
\]
shows that \eqref{eq:gamma-MW-Jensen-finite-point} is equivalent to
\eqref{eq:gamma-MW-Jensen-finite-point-psi}.  Thus
condition~\textup{(iii)} implies condition~\textup{(iv)}.

Conversely, assume condition~\textup{(iv)}, fix the data in
condition~\textup{(iii)}, and put $\x:=\sum_{i=1}^m\lambda_i\y_i$ and
$P:=\{i:\lambda_i>0\}$.  A zero-weight index contributes nothing to the
numerator and to the denominator of any ratio
in~\eqref{eq:gamma-MW-Jensen-subset-max}.  So each ratio for
$S\subseteq\{1,\ldots,m\}$ equals the ratio for $S\cap P$ formed with
the indices in $P$ only, and zero-weight coordinates do not affect
$M^{(m)}_{\gamma,\boldsymbol\lambda}$.  If $P=\{j\}$, then
$\lambda_j=1$, the barycenter is $\y_j$, and every ratio
in~\eqref{eq:gamma-MW-Jensen-subset-max} equals $f(\y_j)$.  So
$M^{(m)}_{\gamma,\boldsymbol\lambda}
(f(\y_1),\ldots,f(\y_m))=f(\y_j)=f(\x)$, and the desired inequality is
an equality.  If $P$ has at least two elements, we apply
condition~\textup{(iv)} to the points $\y_i$, $i\in P$, with the
positive weights $\lambda_i$, $i\in P$, whose barycenter is $\x$.  It
first implies that $\x\in\mathcal K$: otherwise $f(\x)=+\infty$, all
the transformed differences would equal $-\infty$, and
\eqref{eq:gamma-MW-Jensen-finite-point-psi} would fail.
Lemma~\ref{lem:finite-dimensional-gamma-MW-Jensen-mean-characterization}--\textup{(i)},
applied to the indices in $P$ with $c=f(\x)$, then gives
$f(\x)\leq M^{(|P|)}_{\gamma,(\lambda_i)_{i\in P}}((f(\y_i))_{i\in P})$.
By the observation on zero-weight indices, the right-hand side equals
$M^{(m)}_{\gamma,\boldsymbol\lambda}(f(\y_1),\ldots,f(\y_m))$, which
is the desired inequality.
\end{proof}

\subsubsection{Equivalence of (ii) and (iv)}

\begin{proof}
The implication (iv) $\implies$ (ii) follows by taking $m=2$.
Conversely, assume condition~\textup{(ii)} holds, and recall that
$\mathcal K=\operatorname{dom}(f)$.  By the definition of
$\overline{\psi}_{\sqrt{\gamma}}$, the function
$\overline{\psi}_{\sqrt{\gamma}}$ is strictly increasing,
$\overline{\psi}_{\sqrt{\gamma}}(0)=0$, and
$\overline{\psi}_{\sqrt{\gamma}}(u)$ has the same sign as $u$.

Fix $\uu,\vv\in\mathcal K$ and $t\in(0,1)$, and set
\[
\w:=(1-t)\uu+t\vv.
\]
Condition~\textup{(ii)}, applied with $\y=\uu$, $\z=\vv$, and
$\lambda=1-t$, entails that $\w\in\mathcal K$, since otherwise
properness would give $f(\w)=+\infty$, so both transformed
differences in~\textup{(ii)} would equal $-\infty$, contradicting
the asserted inequality.
Since $\uu$, $\vv$, and $t$ were arbitrary, $\mathcal K$ is convex.
Condition~\textup{(ii)} also gives
\begin{equation}
(1-t)\overline{\psi}_{\sqrt{\gamma}}\bigl(f(\uu)-f(\w)\bigr)
+t\overline{\psi}_{\sqrt{\gamma}}\bigl(f(\vv)-f(\w)\bigr)\geq0.
\label{eq:gamma-MW-Jensen-proof-two-point}
\end{equation}
Since the same inequality is automatic for $t\in\{0,1\}$,
\eqref{eq:gamma-MW-Jensen-proof-two-point} holds for every
$\uu,\vv\in\mathcal K$ and every $t\in[0,1]$.

The inequality \eqref{eq:gamma-MW-Jensen-proof-two-point} has a strict
consequence.  Let $c\in\mathbb R$, and
suppose that
\[
f(\uu)\leq c,
\qquad
f(\vv)\leq c,
\]
at least one of these inequalities is strict, and $0<t<1$.  We claim
that
\begin{equation}
f((1-t)\uu+t\vv)<c.
\label{eq:gamma-MW-Jensen-proof-strict-two-point}
\end{equation}
Suppose to the contrary that $f(\w)\geq c$, where
$\w=(1-t)\uu+t\vv$.  Then
\[
f(\uu)-f(\w)\leq0,
\qquad
f(\vv)-f(\w)\leq0.
\]
At least one of these two inequalities is strict: if, for example,
$f(\uu)<c$, then $f(\uu)<c\leq f(\w)$ and hence
$f(\uu)-f(\w)<0$; the argument is identical if $f(\vv)<c$.
Since $\overline{\psi}_{\sqrt{\gamma}}$ is strictly increasing and vanishes
at zero, it follows that
\[
\overline{\psi}_{\sqrt{\gamma}}(f(\uu)-f(\w))\leq0,
\qquad
\overline{\psi}_{\sqrt{\gamma}}(f(\vv)-f(\w))\leq0,
\]
with at least one strict inequality.  The weights $1-t$ and $t$ are
both positive, so the left-hand side of
\eqref{eq:gamma-MW-Jensen-proof-two-point} is strictly negative.  This
contradicts \eqref{eq:gamma-MW-Jensen-proof-two-point}, and proves
\eqref{eq:gamma-MW-Jensen-proof-strict-two-point}.

This strict conclusion extends to finitely many points: if
$\vv_1,\ldots,\vv_N\in\mathcal K$ and
\begin{equation}
\begin{gathered}
\mu_r>0,\qquad \sum_{r=1}^N\mu_r=1,\qquad
f(\vv_r)\leq c\quad(1\leq r\leq N),\\
f(\vv_{r_0})<c\quad\text{for at least one }r_0,
\end{gathered}
\label{eq:gamma-MW-Jensen-proof-strict-finite-assumptions}
\end{equation}
then
\begin{equation}
f\left(\sum_{r=1}^N\mu_r\vv_r\right)<c.
\label{eq:gamma-MW-Jensen-proof-strict-finite}
\end{equation}
To prove this implication, relabel the points so that $r_0=1$, and
define
\[
M_k:=\sum_{r=1}^k\mu_r,
\qquad
\z_k:=\frac{1}{M_k}\sum_{r=1}^k\mu_r\vv_r
\qquad(1\leq k\leq N).
\]
Each $\z_k$ belongs to $\mathcal K$ because it is a convex
combination of points in the convex set $\mathcal K$.
Since $M_1=\mu_1$, we have $\z_1=\vv_1$, and hence
$f(\z_1)<c$.  Suppose inductively that $f(\z_k)<c$ for some $k<N$.
The identities
\[
M_{k+1}=M_k+\mu_{k+1}
\]
and
\[
\z_{k+1}
=\frac{M_k}{M_{k+1}}\z_k
+\frac{\mu_{k+1}}{M_{k+1}}\vv_{k+1}
\]
show that $\z_{k+1}$ is a convex combination of $\z_k$ and
$\vv_{k+1}$: both coefficients are strictly positive, and their sum
is $(M_k+\mu_{k+1})/M_{k+1}=1$.  Since
$f(\z_k)<c$ and $f(\vv_{k+1})\leq c$,
\eqref{eq:gamma-MW-Jensen-proof-strict-two-point} implies that
$f(\z_{k+1})<c$.  Induction gives $f(\z_N)<c$.  Finally,
$M_N=1$, so $\z_N=\sum_{r=1}^N\mu_r\vv_r$.  This proves
\eqref{eq:gamma-MW-Jensen-proof-strict-finite}.

For the finite-point inequality in condition~\textup{(iv)}, let
$m\geq2$, let $\y_1,\ldots,\y_m\in\mathcal K$, and let
$\lambda_1,\ldots,\lambda_m>0$ satisfy
$\sum_{i=1}^m\lambda_i=1$.  Put
\[
\x:=\sum_{i=1}^m\lambda_i\y_i,
\qquad c:=f(\x).
\]
The convexity of $\mathcal K$ proved above implies that
$\x\in\mathcal K$, so $c$ is finite.

Suppose, toward a contradiction, that
\begin{equation}
\sum_{i=1}^m\lambda_i
\overline{\psi}_{\sqrt{\gamma}}(f(\y_i)-c)<0.
\label{eq:gamma-MW-Jensen-proof-failure}
\end{equation}
Define
\[
a_i:=\overline{\psi}_{\sqrt{\gamma}}(f(\y_i)-c),
\qquad
L:=\{i:a_i<0\},
\qquad
H:=\{j:a_j\geq0\},
\]
and
\[
A:=\sum_{i\in L}\lambda_i(-a_i),
\qquad
B:=\sum_{j\in H}\lambda_j a_j.
\]
Since the sets $L$ and $H$ partition $\{1,\ldots,m\}$, the
left-hand side of \eqref{eq:gamma-MW-Jensen-proof-failure} equals
\[
\sum_{i\in L}\lambda_i a_i
+\sum_{j\in H}\lambda_j a_j
=-A+B.
\]
Consequently, \eqref{eq:gamma-MW-Jensen-proof-failure} implies that
$-A+B<0$, or equivalently,
\begin{equation}
A>B\geq0.
\label{eq:gamma-MW-Jensen-proof-AB}
\end{equation}
Here $B\geq0$ because $a_j\geq0$ for every $j\in H$ and all the
weights are positive.  The inequality $A>B\geq0$ gives $A>0$; since
the sum defining $A$ would equal zero if $L$ were empty, $L$ is nonempty.  If $i\in L$, then $a_i<0$.  Since
$\overline{\psi}_{\sqrt{\gamma}}$ has the same sign as its argument,
$a_i<0$ implies that $f(\y_i)-c<0$, and hence $f(\y_i)<c$.

If $H$ were empty, then every $\y_i$ would satisfy $f(\y_i)<c$.
The positive weights $\lambda_i$ sum to one and have barycenter
$\x$, so \eqref{eq:gamma-MW-Jensen-proof-strict-finite} would imply
that $f(\x)<c$, contradicting the definition $c=f(\x)$.
Therefore $H$ is nonempty.

For $i\in L$ and $j\in H$, define
\[
p_{ij}:=\frac{\lambda_i\lambda_j a_j}{A},
\qquad
q_{ij}:=\frac{\lambda_i(-a_i)\lambda_j}{A},
\qquad
s_{ij}:=p_{ij}+q_{ij},
\]
and set
\[
\z_{ij}:=\frac{p_{ij}\y_i+q_{ij}\y_j}{s_{ij}}.
\]
These quantities are well defined: since $A>0$, $\lambda_i>0$,
$\lambda_j>0$, $a_j\geq0$, and $-a_i>0$, we have
$p_{ij}\geq0$ and $q_{ij}>0$, and hence $s_{ij}>0$.  Thus
\[
\frac{p_{ij}}{s_{ij}}\geq0,
\qquad
\frac{q_{ij}}{s_{ij}}>0,
\qquad
\frac{p_{ij}}{s_{ij}}+\frac{q_{ij}}{s_{ij}}=1.
\]
It follows from the convexity of $\mathcal K$ that
$\z_{ij}\in\mathcal K$.  Direct substitution also
gives
\begin{equation}
p_{ij}a_i+q_{ij}a_j
=\frac{\lambda_i\lambda_j}{A}
\bigl(a_j a_i+(-a_i)a_j\bigr)
=0.
\label{eq:gamma-MW-Jensen-proof-balance}
\end{equation}

We claim that $f(\z_{ij})\leq c$.  Suppose instead that
$f(\z_{ij})>c$.  For $h\in\{i,j\}$, this strict inequality implies
that
\[
f(\y_h)-f(\z_{ij})<f(\y_h)-c.
\]
Strict monotonicity of $\overline{\psi}_{\sqrt{\gamma}}$ and the definition of
$a_h$ therefore give
\[
\overline{\psi}_{\sqrt{\gamma}}(f(\y_i)-f(\z_{ij}))<a_i,
\qquad
\overline{\psi}_{\sqrt{\gamma}}(f(\y_j)-f(\z_{ij}))<a_j.
\]
Multiplication of the first inequality by $p_{ij}/s_{ij}\geq0$ gives
a weak inequality, which is strict when $p_{ij}>0$.  Multiplication
of the second inequality by $q_{ij}/s_{ij}>0$ gives a strict
inequality.  Adding the resulting inequalities and using
\eqref{eq:gamma-MW-Jensen-proof-balance} therefore yields
\[
\frac{p_{ij}}{s_{ij}}
\overline{\psi}_{\sqrt{\gamma}}(f(\y_i)-f(\z_{ij}))
+\frac{q_{ij}}{s_{ij}}
\overline{\psi}_{\sqrt{\gamma}}(f(\y_j)-f(\z_{ij}))
<\frac{p_{ij}a_i+q_{ij}a_j}{s_{ij}}
=0.
\]
On the other hand,
\[
\z_{ij}
=\frac{p_{ij}}{s_{ij}}\y_i
+\frac{q_{ij}}{s_{ij}}\y_j.
\]
Applying \eqref{eq:gamma-MW-Jensen-proof-two-point} to the endpoints
$\y_i,\y_j$ with $t=q_{ij}/s_{ij}$ shows that the left-hand side of
the preceding strict inequality is nonnegative, a contradiction.
Hence
\begin{equation}
\label{eq:pc2-zij-bound}
f(\z_{ij})\leq c,
\qquad i\in L,\ j\in H.
\end{equation}

For $i\in L$, set
\[
r_i:=\left(1-\frac BA\right)\lambda_i>0.
\]
The asserted inequality $r_i>0$ holds because $\lambda_i>0$ and the strict
inequality \eqref{eq:gamma-MW-Jensen-proof-AB} implies $0\leq B/A<1$.
For each $i\in L$, the definition of $p_{ij}$ gives
\[
\sum_{j\in H}p_{ij}
=\frac{\lambda_i}{A}\sum_{j\in H}\lambda_j a_j
=\frac BA\lambda_i.
\]
Similarly, for each $j\in H$, the definition of $q_{ij}$ gives
\[
\sum_{i\in L}q_{ij}
=\frac{\lambda_j}{A}\sum_{i\in L}\lambda_i(-a_i)
=\lambda_j.
\]
Using these two identities and $s_{ij}=p_{ij}+q_{ij}$, we obtain
\begin{align}
\sum_{i\in L}\sum_{j\in H}s_{ij}+\sum_{i\in L}r_i
&=\frac BA\sum_{i\in L}\lambda_i
  +\sum_{j\in H}\lambda_j
  +\left(1-\frac BA\right)\sum_{i\in L}\lambda_i \notag\\
&=\sum_{i\in L}\lambda_i+\sum_{j\in H}\lambda_j \notag\\
&=\sum_{i=1}^m\lambda_i=1.
\label{eq:pc2-new-weights}
\end{align}
Furthermore, $s_{ij}\z_{ij}=p_{ij}\y_i+q_{ij}\y_j$.  Hence, by the
same two identities and the definition of $r_i$,
\begin{align}
\sum_{i\in L}\sum_{j\in H}s_{ij}\z_{ij}
  +\sum_{i\in L}r_i\y_i
&=\sum_{i\in L}\sum_{j\in H}
  (p_{ij}\y_i+q_{ij}\y_j)+\sum_{i\in L}r_i\y_i \notag\\
&=\sum_{i\in L}\left(\frac BA\lambda_i+r_i\right)\y_i
  +\sum_{j\in H}\lambda_j\y_j \notag\\
&=\sum_{i=1}^m\lambda_i\y_i=\x.
\label{eq:pc2-new-barycenter}
\end{align}
By \eqref{eq:pc2-new-weights} and \eqref{eq:pc2-new-barycenter},
$\x$ is a convex combination of the finitely many points
$\z_{ij}$, for $i\in L$ and $j\in H$, and $\y_i$, for $i\in L$,
with respective coefficients $s_{ij}>0$ and $r_i>0$.  By
\eqref{eq:pc2-zij-bound}, every $\z_{ij}$ has value at most $c$.  Every
$\y_i$ in this combination has $i\in L$ and hence value strictly
less than $c$; since $L$ is nonempty, at least
one point in the combination has value strictly less than $c$.
Therefore the implication
\eqref{eq:gamma-MW-Jensen-proof-strict-finite-assumptions}
$\Rightarrow$ \eqref{eq:gamma-MW-Jensen-proof-strict-finite} gives
$f(\x)<c$, which again contradicts $c=f(\x)$.

Hence \eqref{eq:gamma-MW-Jensen-proof-failure} is impossible, and since the
points and weights were arbitrary, for every finite
convex combination,
\[
\sum_{i=1}^m\lambda_i
\overline{\psi}_{\sqrt{\gamma}}\bigl(f(\y_i)-f(\x)\bigr)\geq0,
\qquad
\x=\sum_{i=1}^m\lambda_i\y_i.
\]
Thus condition~\textup{(iv)} holds, which together with the preceding two
subsections proves the equivalence of conditions~\textup{(i)}--\textup{(iv)};
none of these arguments uses lower semicontinuity.
\end{proof}

\subsubsection{(v) \texorpdfstring{$\implies$}{implies} (iv)}

\begin{proof}
Assume~\textup{(v)}.  Let
$\y_1,\ldots,\y_m\in\mathcal K$ and
$\lambda_1,\ldots,\lambda_m\in[0,1]$ satisfy
\[
\sum_{i=1}^m\lambda_i=1,
\qquad
\x:=\sum_{i=1}^m\lambda_i\y_i.
\]
Since $\mathcal K$ is convex, $\x\in\mathcal K$, so $f(\x)$ is real.
First suppose that $\x\in\operatorname{ri}(\mathcal K)$ and choose
$\v\in F(\x)$, which is possible because
$\operatorname{ri}(\mathcal K)\subseteq\operatorname{dom}(F)$.  For
every $i$, the $\sqrt{\gamma}$--tilted convexity of $f$ with respect to $F$
gives
\[
f(\y_i)-f(\x)
\geq
\psi_{\sqrt{\gamma}}(\langle\y_i-\x,\v\rangle).
\]
By Lemma~\ref{lem:psi_basics}, $\overline{\psi}_{\sqrt{\gamma}}$ is
increasing and inverse to $\psi_{\sqrt{\gamma}}$, so
\[
\langle\y_i-\x,\v\rangle
\leq
\overline{\psi}_{\sqrt{\gamma}}\bigl(f(\y_i)-f(\x)\bigr).
\]
Multiplying by $\lambda_i\geq0$, summing over $i$, and using
$\sum_{i=1}^m\lambda_i(\y_i-\x)=\x-\x=\BF{0}$, we obtain
\begin{equation}
\label{eq:pc2-ri-finite-point}
0
=
\left\langle
\sum_{i=1}^m\lambda_i(\y_i-\x),\v
\right\rangle
\leq
\sum_{i=1}^m
\lambda_i
\overline{\psi}_{\sqrt{\gamma}}\bigl(f(\y_i)-f(\x)\bigr).
\end{equation}

To remove the assumption that $\x$ belongs to the relative
interior, choose $\z\in\operatorname{ri}(\mathcal K)$ and, for
$t\in(0,1]$, set
\[
\y_{i,t}:=(1-t)\y_i+t\z,
\qquad
\x_t:=\sum_{i=1}^m\lambda_i\y_{i,t}
=(1-t)\x+t\z.
\]
By Theorem~6.1 in~\cite{rockafellar15_convex_analy}, all these points
belong to $\operatorname{ri}(\mathcal K)$.  So
\eqref{eq:pc2-ri-finite-point}, applied to the points $\y_{i,t}$ with
barycenter $\x_t$, gives
\begin{equation}
\label{eq:pc2-ri-finite-point-t}
\sum_{i=1}^m
\lambda_i
\overline{\psi}_{\sqrt{\gamma}}
\bigl(f(\y_{i,t})-f(\x_t)\bigr)
\geq0.
\end{equation}
The hypotheses of
Theorem~\ref{thm:gamma-MW-continuous-on-lines--strong-version} are
satisfied by $f$: it is proper and lsc, its domain $\mathcal K$ is
convex, exact $\sqrt{\gamma}$--tilted convexity implies approximate
$\sqrt{\gamma}$--tilted convexity, and
$\mathcal K\cap\operatorname{dom}(F)$ is dense in $\mathcal K$ because
it contains $\operatorname{ri}(\mathcal K)$.  For each $i$, the points
$\y_{i,t}$ lie on the line through $\y_i$ and $\z$, which meets
$\op{ri}(\mathcal K)$ at $\z$; the same holds for $\x_t$, $\x$, and
$\z$.  (If $\y_i=\z$, then $\y_{i,t}=\y_i$ for all $t$, and
likewise for $\x$.)  Hence the theorem gives
\[
f(\y_{i,t})\longrightarrow f(\y_i),
\qquad
f(\x_t)\longrightarrow f(\x)
\qquad(t\downarrow0).
\]
Letting $t\downarrow0$ in \eqref{eq:pc2-ri-finite-point-t} and using
the continuity of $\overline{\psi}_{\sqrt{\gamma}}$ gives the
finite-point inequality \eqref{eq:gamma-MW-Jensen-finite-point-psi},
and hence condition~\textup{(iv)}.
\end{proof}

\subsubsection{(iv) \texorpdfstring{$\implies$}{implies} (v)}

\begin{proof}
Assume condition~\textup{(iv)}.  Let $\y,\z\in\mathcal K$ and
$\lambda\in(0,1)$, and set
$\x:=\lambda\y+(1-\lambda)\z$.  If $\x\notin\mathcal K$, then
properness gives $f(\x)=+\infty$, so both transformed differences in
condition~\textup{(iv)} with $m=2$ equal $-\infty$, contradicting its
asserted inequality.  So $\x\in\mathcal K$ for every
$\lambda\in(0,1)$, and for $\lambda\in\{0,1\}$ the point $\x$ is $\z$
or $\y$.  Hence $\mathcal K$ is convex.  Set
$k:=\operatorname{dim}(\mathcal K)$ and
\[
V:=\operatorname{span}(\mathcal K-\mathcal K),
\]
a $k$--dimensional subspace.  For every $\x\in\mathcal K$, the affine
hull of $\mathcal K$ is $\x+V$.

Define a multi-valued operator
$F:\mathbb R^d\rightrightarrows\mathbb R^d$ as follows.  For
$\x\notin\operatorname{ri}(\mathcal K)$, set $F(\x):=\emptyset$.  For
$\x\in\operatorname{ri}(\mathcal K)$, define the closed halfspaces in
the $k$--dimensional space $V$ by
\[
H_{\y}
:=
\left\{
\v\in V:
\langle\y-\x,\v\rangle
\leq
\overline{\psi}_{\sqrt{\gamma}}\bigl(f(\y)-f(\x)\bigr)
\right\},
\qquad \y\in\mathcal K,
\]
where $H_{\x}=V$, and set
\[
F(\x):=\bigcap_{\y\in\mathcal K}H_{\y}.
\]
We show that this intersection is nonempty.  If $k=0$,
then $\mathcal K=\{\x\}$, $V=\{\BF{0}\}$, and $F(\x)=H_{\x}=\{\BF{0}\}$.
Assume $k\geq1$.

Consider any finite family $H_{\y_1},\ldots,H_{\y_m}$.
By Farkas' lemma in the space $V$, its intersection is nonempty
provided that
\begin{equation}
\label{eq:pc2-farkas-condition}
\theta_i\ge0,\quad
\sum_{i=1}^m\theta_i(\y_i-\x)=0
\quad\Longrightarrow\quad
\sum_{i=1}^m\theta_i\overline\psi_{\sqrt\gamma}
       \bigl(f(\y_i)-f(\x)\bigr)\ge0.
\end{equation}
If $A:=\sum_i\theta_i=0$, every coefficient vanishes and the
implication is immediate.  Otherwise discard zero coefficients
and put $\lambda_i:=\theta_i/A$ on the remaining indices.
Then $\x=\sum_i\lambda_i\y_i$, and condition~\textup{(iv)} proves
the required inequality after multiplication by $A$.  If only one
coefficient is positive, its point is $\x$ and the inequality follows
from $\overline\psi_{\sqrt\gamma}(0)=0$.
Thus every finite subfamily has nonempty intersection.

Choose an orthonormal basis $\mathbf e_1,\ldots,\mathbf e_k$ of $V$.
Since $\x\in\operatorname{ri}(\mathcal K)$, there is $\varepsilon>0$
such that $\x\pm\varepsilon\mathbf e_j\in\mathcal K$ for every $j$.
The finite intersection
\[
P:=\bigcap_{j=1}^k
\bigl(H_{\x+\varepsilon\mathbf e_j}\cap
      H_{\x-\varepsilon\mathbf e_j}\bigr)
\]
is nonempty by the preceding paragraph and is compact: these
inequalities bound every coordinate $\langle\v,\mathbf e_j\rangle$
above and below by finite numbers.  The closed subsets
$P\cap H_{\y}$, $\y\in\mathcal K$, have the finite-intersection
property, again by the preceding paragraph.  Compactness of $P$
therefore gives
\[
\bigcap_{\y\in\mathcal K}H_{\y}\ne\emptyset.
\]
Hence
$\operatorname{ri}(\mathcal K)\subseteq\operatorname{dom}(F)$.  Let
$\x\in\operatorname{ri}(\mathcal K)$, $\v\in F(\x)$, and
$\y\in\mathcal K$.  By the definition of $H_{\y}$,
$\langle\y-\x,\v\rangle\leq
\overline{\psi}_{\sqrt{\gamma}}(f(\y)-f(\x))$.  Since
$\psi_{\sqrt{\gamma}}$ is increasing and
$\psi_{\sqrt{\gamma}}\circ\overline{\psi}_{\sqrt{\gamma}}$ is the
identity (Lemma~\ref{lem:psi_basics}), applying
$\psi_{\sqrt{\gamma}}$ gives
\[
f(\y)-f(\x)
\geq
\psi_{\sqrt{\gamma}}(\langle\y-\x,\v\rangle).
\]
This holds for every $\x\in\operatorname{ri}(\mathcal K)$,
$\v\in F(\x)$, and $\y\in\mathcal K$, and $F(\x)=\emptyset$ for all
other $\x$.  Thus $f$ is $\sqrt{\gamma}$--tilted convex with respect to
$F$, and condition~\textup{(v)} holds.
\end{proof}

\subsection{Proof of Theorem~\ref{thm:gamma-MW-Jensen-vs-gamma-MW-convex}}
\label{apx:gamma-MW-Jensen-vs-gamma-MW-convex}

\subsubsection{(i)}

\begin{proof}
Suppose that $f$ is $\gamma$--tilted secant convex and put
$\mathcal K:=\operatorname{dom}(f)$.  The set $\mathcal K$ is convex
by Theorem~\ref{thm:gamma-MW-Jensen-basics}(i).
Fix $\x,\y\in\mathcal K$ and $\v\in\widehat\partial f(\x)$, and
write $\dd:=\y-\x$ and $\Delta:=f(\y)-f(\x)$.
The case $\dd=0$ is immediate.  For $0<t<1$, put
$\x_t:=\x+t\dd=t\y+(1-t)\x$.  Definition~\ref{def:gamma-MW-Jensen-convex},
applied to the points $\y$ and $\x$ with weight $t$ on $\y$, gives
$\x_t\in\mathcal K$ and
\[
\frac{f(\y)-f(\x_t)}{(1-t)\norm{\dd}}
\ge
\psi_\gamma\Bigl(\frac{f(\x_t)-f(\x)}{t\norm{\dd}}\Bigr).
\]
Applying the increasing inverse $\overline\psi_\gamma$ of $\psi_\gamma$
(Lemma~\ref{lem:psi_basics}(1) and~(6)) and multiplying by $\norm{\dd}$,
which commutes with $\overline\psi_\gamma$ by positive homogeneity, gives
\begin{equation}\label{eq:sec-to-tilted-t}
\frac{f(\x_t)-f(\x)}{t}
\le
\overline\psi_\gamma\Bigl(\frac{f(\y)-f(\x_t)}{1-t}\Bigr).
\end{equation}
Regular subdifferentiability gives
$f(\x_t)\ge f(\x)+t\langle\v,\dd\rangle+o(t)$ as $t\downarrow0$.  Hence
$\liminf_{t\downarrow0}f(\x_t)\ge f(\x)$ and
\[
\langle\v,\dd\rangle
\le\liminf_{t\downarrow0}
\frac{f(\x_t)-f(\x)}{t}.
\]
Since $\overline\psi_\gamma$ is continuous and increasing, the right side
of~\eqref{eq:sec-to-tilted-t} has limit superior at most
$\overline\psi_\gamma(\Delta)$ as $t\downarrow0$.  Therefore
$\langle\v,\dd\rangle\le\overline\psi_\gamma(\Delta)$.
Since $\psi_\gamma$ is the increasing inverse of
$\overline\psi_\gamma$, this is
\[
f(\y)-f(\x)\ge
\psi_\gamma(\langle\v,\y-\x\rangle).
\]
Thus $f$ is $\gamma$--tilted convex with respect to
$\widehat\partial f$.  The equivalence of (iii) and (i) in
Theorem~\ref{thm:mwc-alternative-characterizations} proves the assertion.
\end{proof}

\subsubsection{(ii)}

\begin{proof}
Suppose that $\overline\partial f$ is $\gamma$--tilted monotone.
Put $\mathcal K:=\operatorname{dom}(f)$; it is convex by
Corollary~\ref{cor:mono-structural-properties}.
Fix distinct $\y_0,\y_1\in\mathcal K$, put
$\ell:=\|\y_1-\y_0\|$ and $\mathbf e:=(\y_1-\y_0)/\ell$, and define
\[
g(t):=
\begin{cases}
f(\y_0+t\mathbf e),&0\le t\le\ell,\\
+\infty,&\text{otherwise}.
\end{cases}
\]
Convexity of $\mathcal K$ gives $\operatorname{dom}(g)=[0,\ell]$,
and $g$ is proper and lsc.  The closed-interval assertion of
Theorem~\ref{thm:gamma-MW-monotone-on-lines}, in the unit-speed
coordinate $t$, shows that $\overline\partial g$ is
$\gamma$--tilted monotone.  Corollary~\ref{cor:mono-structural-properties}(ii)
gives continuity on $[0,\ell]$ and local Lipschitz continuity on
$(0,\ell)$.
Fix $t\in(0,\ell)$ and set
\[
A:=\frac{g(t)-g(0)}t,\qquad
B:=\frac{g(\ell)-g(t)}{\ell-t}.
\]
Applying Corollary~\ref{cor:Lipschitz-interior-mean-value} to $g$ on
$[0,t]$ and $[t,\ell]$ gives points $u\in(0,t)$ and $v\in(t,\ell)$
with $A\in\overline\partial g(u)$ and
$B\in\overline\partial g(v)$.
Tilted monotonicity and positive homogeneity imply
\[
(v-u)B\ge\psi_\gamma\bigl((v-u)A\bigr)
=(v-u)\psi_\gamma(A),\qquad
B\ge\psi_\gamma(A).
\]
The point $\x:=\y_0+t\mathbf e=\lambda\y_1+(1-\lambda)\y_0$, where
$\lambda:=t/\ell$, lies in $\mathcal K$, and $A$ and $B$ are the slopes
$s_-$ and $s_+$ of~\eqref{eq:secant-slopes} for $\y=\y_1$ and $\z=\y_0$.
So $B\ge\psi_\gamma(A)$ is~\eqref{eq:secant-form}.  Since $\y_0$, $\y_1$,
and $t$ were arbitrary, $f$ is $\gamma$--tilted secant convex.
\end{proof}

\subsection{Proof of Theorem~\ref{thm:gamma-MW-Jensen-pointwise-limit}}
\label{apx:gamma-MW-Jensen-pointwise-limit}

\begin{proof}
Set
\[
\mathcal K_0:=\operatorname{ri}(\mathcal K),
\qquad
\mathcal D_0:=\mathcal D\cap\mathcal K_0.
\]
Then $\mathcal D_0$ is dense in $\mathcal K_0$, because $\mathcal K_0$ is
open relative to $\operatorname{aff}(\mathcal K)$ and
$\mathcal D\subseteq\mathcal K\subseteq\operatorname{aff}(\mathcal K)$ is
dense in $\mathcal K$.  Choose a point of $\mathcal D_0$; the values of
$f_n$ there are finite for large $n$.  So, after discarding
finitely many terms, the restrictions
\[
h_n:=f_n|_{\mathcal K_0}
\]
are proper.  By Theorem~\ref{thm:gamma-MW-Jensen-basics}\textup{(iv)}, each $h_n$
is $\gamma$--tilted secant convex.  By
Theorem~\ref{thm:gamma-MW-Jensen-basics}\textup{(v)}, so is its lsc envelope
\[
\bar h_n:=\op{lsc}(h_n):\mathbb R^d\to\mathbb R\cup\{+\infty\},
\]
and
\begin{equation}
\label{eq:gamma-MW-Jensen-limit-envelope-agreement}
h_n=\bar h_n
\qquad\text{on }\operatorname{ri}(\operatorname{dom}(h_n)).
\end{equation}
Thus $\bar h_n$ is lsc, never equal to $-\infty$, and finite on the
nonempty set $\operatorname{ri}(\operatorname{dom}(h_n))$; hence it is
proper.
Theorem~\ref{thm:gamma-MW-Jensen-vs-gamma-MW-convex}\textup{(i)}
shows that it is $\gamma$--tilted convex.

For every
compact set $C\subseteq\mathcal K_0$, one can choose finitely many
points $\dd_1,\ldots,\dd_m\in\mathcal D_0$ such that
\begin{equation}
\label{eq:pc3-hull}
C\subseteq
\operatorname{ri}\bigl(\operatorname{conv}\{\dd_1,\ldots,\dd_m\}\bigr)
\subseteq\mathcal K_0,
\qquad
\operatorname{aff}\{\dd_1,\ldots,\dd_m\}
=\operatorname{aff}(\mathcal K).
\end{equation}
To see this, let $V:=\operatorname{span}(\mathcal K-\mathcal K)$ and use
the balls $\mathbb B_V$ of the proof of
Theorem~\ref{thm:gamma-MW-convex-pointwise-limit}.  For
$\x_0\in C$, choose $R>0$ with
$\overline{\mathbb B_V(\x_0,5R)}\subseteq\mathcal K_0$.  The construction
leading to~\eqref{eq:pc3-3R-in-P} gives points $\y_1,\ldots,\y_k\in\mathcal D$
with $\|\y_i-\x_0\|<4R+R/4$, hence $\y_i\in\mathcal D_0$, whose convex
hull contains $\overline{\mathbb B_V(\x_0,3R)}$.  Finitely many balls
$\mathbb B_V(\x_0,R)$ with $\x_0\in C$ cover $C$.  Let
$\dd_1,\ldots,\dd_m$ be all the points $\y_i$ obtained for these
finitely many centers, and put
$Q:=\operatorname{conv}\{\dd_1,\ldots,\dd_m\}$.  Then $Q\subseteq\mathcal K_0$
because $\mathcal K_0$ is convex.  The set $Q$ contains a ball of
$\operatorname{aff}(\mathcal K)$, so
$\operatorname{aff}(Q)=\operatorname{aff}(\mathcal K)$, and each covering
ball is open in $\operatorname{aff}(\mathcal K)$ and contained in $Q$,
hence contained in $\operatorname{ri}(Q)$.  This proves~\eqref{eq:pc3-hull}.

Since $f_n(\dd_i)$ converges to a finite value for each $i$, all these
points belong to $\operatorname{dom}(h_n)$ for sufficiently large
$n$.  That domain is convex by
Theorem~\ref{thm:gamma-MW-Jensen-basics}\textup{(i)}, so
$Q\subseteq\operatorname{dom}(h_n)\subseteq\operatorname{aff}(\mathcal K)$.
As $\operatorname{aff}(Q)=\operatorname{aff}(\mathcal K)$, the two convex
sets $Q$ and $\operatorname{dom}(h_n)$ have the same affine hull, so
$\operatorname{ri}(Q)\subseteq\operatorname{ri}(\operatorname{dom}(h_n))$.
By \eqref{eq:pc3-hull}, $C\subseteq\operatorname{ri}(\operatorname{dom}(h_n))$,
and \eqref{eq:gamma-MW-Jensen-limit-envelope-agreement} then gives
\begin{equation}
\label{eq:gamma-MW-Jensen-limit-eventual-agreement}
f_n=h_n=\bar h_n
\qquad\text{on }C
\end{equation}
for all sufficiently large $n$.

Taking $C$ to be a singleton in~
\eqref{eq:gamma-MW-Jensen-limit-eventual-agreement} shows that
$\bar h_n$ converges to finite values on the dense set $\mathcal D_0$.
Theorem~\ref{thm:gamma-MW-convex-pointwise-limit}, applied to
$\bar h_n$ with parameter $\gamma$ and convex set $\mathcal K_0$,
therefore gives a continuous function
$f:\operatorname{ri}(\mathcal K_0)=\mathcal K_0\to\mathbb R$ such that
$\bar h_n\to f$ uniformly on
compact subsets of $\mathcal K_0$.  Applying
\eqref{eq:gamma-MW-Jensen-limit-eventual-agreement} once more, now to
an arbitrary compact $C\subseteq\mathcal K_0$, proves that
$f_n\to f$ uniformly on $C$.  This proves~\textup{(i)}.

Let $\y\ne\z$ in $\mathcal K_0$ and $\lambda\in(0,1)$, and put
$\x:=\lambda\y+(1-\lambda)\z\in\mathcal K_0$.  By~(i), $f_n$ converges to
the finite values of $f$ at $\x$, $\y$, and $\z$, so the
slopes~\eqref{eq:secant-slopes} of $f_n$ converge to those of $f$.  Since
$\gamma>0$, $\psi_\gamma$ is continuous, and passing to the limit
in~\eqref{eq:secant-form} for $f_n$ gives~\eqref{eq:secant-form} for $f$.
Hence $f$ is $\gamma$--tilted secant convex,
proving~\textup{(ii)}.

Now suppose that $f_n\to g$ pointwise on $\mathcal K$.  The function
$g$ cannot take the value $-\infty$.  To see this, observe that $g=f$ on
$\operatorname{ri}(\mathcal K)$, since both are pointwise limits of $f_n$
there, and suppose that
$g(\y)=-\infty$ for some $\y\in\mathcal K$.  Choose
$\x\in\operatorname{ri}(\mathcal K)$; then $\x\ne\y$ because $g(\x)=f(\x)$
is finite.  As $\theta\downarrow0$, the point
$(\x-\theta\y)/(1-\theta)$ of $\operatorname{aff}(\mathcal K)$ tends to
$\x$, so one may choose $\theta\in(0,1)$ such that
\[
\z:=\frac{\x-\theta\y}{1-\theta}
\in\operatorname{ri}(\mathcal K).
\]
Then $\x=\theta\y+(1-\theta)\z$ and $\y\ne\z$.  For large $n$, $f_n(\y)$
is finite because $f_n(\y)\to-\infty$ and $f_n$ is proper, while
$f_n(\x)\to f(\x)$ and $f_n(\z)\to f(\z)$.  The slope $s_-$ of $f_n$ over
$[\z,\x]$ therefore converges to a finite limit, so $\psi_\gamma(s_-)$
stays bounded, while the slope $s_+=(f_n(\y)-f_n(\x))/\norm{\y-\x}$
tends to $-\infty$.  This contradicts~\eqref{eq:secant-form} for $f_n$.

Finally, let $\y\ne\z$ in $\operatorname{dom}(g)$ and $\lambda\in(0,1)$,
and put $\x:=\lambda\y+(1-\lambda)\z\in\mathcal K$.  Since $g(\y)$ and
$g(\z)$ are finite, $\y,\z\in\operatorname{dom}(f_n)$ for large $n$, and
Theorem~\ref{thm:gamma-MW-Jensen-basics}\textup{(i)} gives
$f_n(\x)\le\max\{f_n(\y),f_n(\z)\}$.  Passing to the limit,
$g(\x)\le\max\{g(\y),g(\z)\}<+\infty$, and $g(\x)>-\infty$ by the
previous paragraph, so $\x\in\operatorname{dom}(g)$.  The slopes of $f_n$
then converge to those of $g$, and the continuity of $\psi_\gamma$
gives~\eqref{eq:secant-form} for $g$.  Hence $g$ is
$\gamma$--tilted secant convex.
\end{proof}

\subsection{Proof of Theorem~\ref{thm:H-convex-equivalent-def}}
\label{apx:H-convex-equivalent-def}

\begin{proof}
First let $h(\y)=c+\psi_\gamma(\langle\y-\w,\z\rangle)$ be a
member of $\mathcal H_\gamma$.
Superadditivity of $\psi_\gamma$ gives, for every
$\x,\y\in\mathbb R^d$,
\[
h(\y)-h(\x)\ge\psi_\gamma(\langle\y-\x,\z\rangle).
\]
Thus $h$ is $\gamma$--tilted convex with respect to the constant
operator $F(\x)=\{\z\}$, so
Theorem~\ref{thm:gamma-MW-Jensen-equivalent-def}(v), applied with
parameter $\gamma^2$, shows that $h$ is $\gamma^2$--tilted secant convex.
By Theorem~\ref{thm:gamma-MW-Jensen-basics}(iii), every proper
nonempty pointwise supremum of members of $\mathcal H_\gamma$ is
$\gamma^2$--tilted secant convex, which proves the forward implication.

Conversely, let $f$ be proper, lsc, and $\gamma^2$--tilted secant convex,
and put $\mathcal K:=\operatorname{dom}(f)$.
Theorem~\ref{thm:gamma-MW-Jensen-equivalent-def} gives an operator
$F$ whose domain contains $\operatorname{ri}(\mathcal K)$ and with
respect to which $f$ is $\gamma$--tilted convex.  Define
\[
\mathcal V:=
\{h\in\mathcal H_\gamma:h\le f\text{ on }\mathbb R^d\},
\qquad q:=\sup_{h\in\mathcal V}h.
\]
For each $\x\in\operatorname{ri}(\mathcal K)$ and $\z\in F(\x)$,
the function
$\y\mapsto f(\x)+\psi_\gamma(\langle\y-\x,\z\rangle)$
belongs to $\mathcal V$.  Since $\mathcal K$ is nonempty and convex,
its relative interior is nonempty.  Hence $\mathcal V\ne\emptyset$,
and
\begin{equation}
\label{eq:H-equivalence-agreement-relative-interior}
q\le f\quad\text{on }\mathbb R^d,\qquad
q=f\quad\text{on }\operatorname{ri}(\mathcal K).
\end{equation}
A fixed member of $\mathcal V$ bounds $q$ below everywhere, and
$q$ is finite on $\operatorname{ri}(\mathcal K)$, so $q$ is proper;
it is lsc as a supremum of continuous functions.  By the forward
implication, $q$ is $\gamma^2$--tilted secant convex.

We claim that
\begin{equation}
\label{eq:H-equivalence-domain-inclusion}
\mathcal K\subseteq\operatorname{dom}(q)\subseteq\overline{\mathcal K}.
\end{equation}
The first inclusion follows from $q\le f$ and $q>-\infty$.
For the second, fix $\y\notin\overline{\mathcal K}$.
Strict separation gives an affine function $a$ such that
$a\le0$ on $\overline{\mathcal K}$ and $a(\y)>0$.
Choose $h_0=c+\psi_\gamma\circ L\in\mathcal V$, with $L$ affine,
and for $r>0$ set $h_r:=c+\psi_\gamma\circ(L+ra)$.
Every such function belongs to $\mathcal H_\gamma$: an affine
function with nonzero linear part can be written as
$\langle\,\cdot-\w,\z\rangle$, and a constant argument can be
absorbed into the outer constant.
On $\mathcal K$, monotonicity of $\psi_\gamma$ gives
$h_r\le h_0\le f$; outside $\mathcal K$ the inequality
$h_r\le f$ is automatic.  Thus $h_r\in\mathcal V$.
Because $a(\y)>0$ and $\gamma>0$,
$h_r(\y)\to+\infty$ as $r\to\infty$, so
$q(\y)=+\infty$, proving the second inclusion.

Both $\mathcal K$ and $\operatorname{dom}(q)$ are convex, so
the inclusions in~\eqref{eq:H-equivalence-domain-inclusion} show
that their closures, and hence their relative interiors, coincide.
Corollary~\ref{cor:gamma-MW-Jensen-structural-properties}(iii),
applied to $f$ and $q$, gives
\[
q=\op{lsc}\bigl(q|_{\operatorname{ri}(\operatorname{dom}(q))}\bigr)
=\op{lsc}\bigl(f|_{\operatorname{ri}(\mathcal K)}\bigr)=f.
\]
Thus $f$ is $\mathcal H_\gamma$--convex.
\end{proof}

\subsection{Proofs of the separating examples}
\subsubsection{Proof of Example~\ref{ex:sep-1}}
\label{apx:example-1}

\begin{proof}
The function $f$ is continuous on $[-1,1]$ and equals $+\infty$ outside this closed interval, so it is proper and lsc.
For $-1\le x<y\le1$ we write
\[
A_{x,y}:=\frac{f(y)-f(x)}{y-x}
\]
for the secant slopes of $f$.

The following three one-dimensional reductions are also used in the proofs of Example~\ref{ex:sep-2} and Lemma~\ref{lem:relations-integral-test}.
Let $g:\mathbb R\to\overline{\mathbb R}$ be proper and lsc with $\op{dom}(g)$ an interval, let $\eta\in(0,1]$, and write $A_{x,y}$ for the secant slopes of $g$.
Fix $a<b$ in $\op{dom}(g)$, $u\in\overline\partial g(a)$, and $w\in\overline\partial g(b)$.
By positive homogeneity and Lemma~\ref{lem:psi_basics}(3), $\psi_\eta(-sr)=-s\,\overline\psi_\eta(r)$ for $s>0$.

For monotonicity, inequality \eqref{eq:gamma-mw-mono} with $(\x,\x^*,\y,\y^*)=(a,u,b,w)$ reads $(b-a)w\ge\psi_\eta((b-a)u)=(b-a)\psi_\eta(u)$, that is, $w\ge\psi_\eta(u)$.
With $(\x,\x^*,\y,\y^*)=(b,w,a,u)$ it reads $(a-b)u\ge\psi_\eta((a-b)w)=-(b-a)\overline\psi_\eta(w)$, that is, $u\le\overline\psi_\eta(w)$.
Applying the increasing function $\psi_\eta$ and using Lemma~\ref{lem:psi_basics}(6) turns the second condition into $w\ge\psi_\eta(u)$ as well.
For $a=b$ both sides of \eqref{eq:gamma-mw-mono} vanish.
Hence
\begin{equation}
\label{eq:pb-mono-1d}
\overline\partial g\text{ is }\eta\text{--tilted monotone}
\iff
w\ge\psi_\eta(u)
\quad
\forall a<b\text{ in }\op{dom}(g),\ u\in\overline\partial g(a),\ w\in\overline\partial g(b).
\end{equation}

For $\eta$--tilted convexity, inequality \eqref{eq:gamma-mwc} with $(\x,\y,\v)=(a,b,u)$ reads $g(b)-g(a)\ge(b-a)\psi_\eta(u)$.
With $(\x,\y,\v)=(b,a,w)$ it reads $g(a)-g(b)\ge-(b-a)\overline\psi_\eta(w)$.
Dividing both by $b-a>0$, we obtain
\begin{equation}
\label{eq:pb-mwc-1d}
g\text{ is }\eta\text{--tilted convex}
\iff
\psi_\eta(u)\le A_{a,b}\le\overline\psi_\eta(w)
\quad
\forall a<b\text{ in }\op{dom}(g),\ u\in\overline\partial g(a),\ w\in\overline\partial g(b).
\end{equation}

For $\eta$--tilted secant convexity, assume in addition that $g$ is nondecreasing on $\op{dom}(g)$, and let $a<x<b$ in $\op{dom}(g)$.
With $\z=a$ and $\y=b$ in Definition~\ref{def:gamma-MW-Jensen-convex}, the slopes~\eqref{eq:secant-slopes} are $s_-=A_{a,x}\ge0$ and $s_+=A_{x,b}$, so~\eqref{eq:secant-form} reads $A_{x,b}\ge\eta A_{a,x}$.
With $\z=b$ and $\y=a$, they are $s_-=-A_{x,b}\le0$ and $s_+=-A_{a,x}$, so~\eqref{eq:secant-form} reads $-A_{a,x}\ge-A_{x,b}/\eta$, which is the same inequality.
Since $\op{dom}(g)$ is an interval, every point between two points of the domain lies in it.
Therefore
\begin{equation}
\label{eq:pb-M-1d}
g\text{ is }\eta\text{--tilted secant convex}
\iff
A_{x,b}\ge\eta A_{a,x}
\quad
\forall a<x<b\text{ in }\op{dom}(g).
\end{equation}

We return to $f$.
The Clarke subdifferential of $f$ is
\begin{equation}
\label{eq:pb-ex1-subdiff}
\overline{\partial}f(x)=
\begin{cases}
(-\infty,1], & x=-1,\\
\{1\}, & -1<x<0,\\
[\gamma,1], & x=0,\\
\{\gamma\}, & 0<x<1,\\
[\gamma,+\infty), & x=1.
\end{cases}
\end{equation}
On $(-1,0)$ and $(0,1)$ the function is differentiable on a neighborhood of each point, so Proposition~\ref{prop:subdiff-basics}(iv) applies.
Near $0$, $f$ is Lipschitz with gradients $1$ on the left and $\gamma$ on the right, so Proposition~\ref{prop:subdiff-basics}(ii) gives $\overline\partial f(0)=\op{conv}\{1,\gamma\}=[\gamma,1]$.
At $x=\pm1$, write $f=\delta_{[-1,1]}+f_0$, where $f_0(x)=x$ for $x\le0$ and $f_0(x)=\gamma x$ for $x\ge0$.
The function $f_0$ is continuously differentiable near $\pm1$, with $f_0'(-1)=1$ and $f_0'(1)=\gamma$.
By Proposition~\ref{prop:subdiff-basics}(v), $\overline\partial f(\pm1)=f_0'(\pm1)+N_{[-1,1]}(\pm1)$, where the normal cone $N_{[-1,1]}(\pm1)$ is $(-\infty,0]$ at $-1$ and $[0,+\infty)$ at $1$.

By \eqref{eq:pb-ex1-subdiff}, every point $a<1$ satisfies $\overline\partial f(a)\subseteq(-\infty,1]$, and every point $b>-1$ satisfies $\overline\partial f(b)\subseteq[\gamma,+\infty)$.
So for $a<b$, $u\in\overline\partial f(a)$, and $w\in\overline\partial f(b)$, we have
\begin{equation}
\label{eq:pb-ex1-pq}
u\le1,
\qquad
w\ge\gamma.
\end{equation}
Since $\psi_\gamma$ is increasing, \eqref{eq:pb-ex1-pq} gives $w\ge\gamma=\psi_\gamma(1)\ge\psi_\gamma(u)$, so $\overline{\partial}f$ is $\gamma$--tilted monotone by \eqref{eq:pb-mono-1d}.
For $\eta>\gamma$, take $a=-1/2$, $u=1$, $b=1/2$, and $w=\gamma$.
Then $\psi_\eta(u)=\eta>\gamma=w$, so \eqref{eq:pb-mono-1d} fails and $\overline\partial f$ is not $\eta$--tilted monotone.  Hence
\[
\gamma^*_{\op{mono}}(f)=\gamma.
\]

Since $f$ is piecewise linear with slopes $1$ and $\gamma$, every
secant slope $A_{x,y}$ is a weighted average of these slopes, so
\begin{equation}
\label{eq:pb-ex1-secant}
\gamma\le A_{x,y}\le1
\qquad
\forall\,-1\le x<y\le1.
\end{equation}
For $a<b$, $u\in\overline\partial f(a)$, and $w\in\overline\partial f(b)$, the bounds \eqref{eq:pb-ex1-pq} and the monotonicity of $\psi_\gamma$ and $\overline\psi_\gamma$ give
\[
\psi_\gamma(u)\le\psi_\gamma(1)=\gamma,
\qquad
\overline\psi_\gamma(w)\ge\overline\psi_\gamma(\gamma)=1.
\]
Combined with \eqref{eq:pb-ex1-secant}, this gives $\psi_\gamma(u)\le A_{a,b}\le\overline\psi_\gamma(w)$, so $f$ is $\gamma$--tilted convex by \eqref{eq:pb-mwc-1d}.
On the other hand, take $a=0$, $u=1\in\overline{\partial}f(0)$, and $b>0$.
Then $A_{0,b}=\gamma$, and \eqref{eq:pb-mwc-1d} with parameter $\eta$ would require $\gamma\ge\psi_\eta(1)=\eta$.
So $f$ is not $\eta$--tilted convex for $\eta>\gamma$.  Consequently,
\[
\gamma^*(f)=\gamma.
\]

Finally, $f$ is nondecreasing, so \eqref{eq:pb-M-1d} applies.
For $-1\le a<x<b\le1$, \eqref{eq:pb-ex1-secant} gives $A_{x,b}\ge\gamma\ge\gamma A_{a,x}$, so $f$ is $\gamma$--tilted secant convex.
Choosing $a<0$, $x=0$, and $b>0$ gives
$A_{a,0}=1$ and $A_{0,b}=\gamma$.
Then \eqref{eq:pb-M-1d} with parameter $\eta$ would require $\gamma\ge\eta$, so no larger parameter is possible.
Therefore
\[
\gamma^*_{\op{sec}}(f)=\gamma.
\]
Since $\gamma^*(f)=\gamma<\sqrt\gamma$, we have $f\notin\mathcal F^{\sqrt\gamma}$.
Since $\gamma^*_{\op{mono}}(f)=\gamma$, we have $f\in\mathcal F^\gamma_{\op{mono}}$, and the remaining inclusions are parts~(ii) and~(i) of Theorem~\ref{thm:gamma-MW-Jensen-vs-gamma-MW-convex}.
\end{proof}

\subsubsection{Proof of Example~\ref{ex:sep-2}}
\label{apx:example-2}

\begin{proof}
Note that $1+c=1/\gamma$.
The formula $x/(1+cx)$ defines a continuously differentiable function on the open set $(-1/c,+\infty)\supseteq[0,1]$, so $f$ is proper and lsc.
On $[0,1]$, the quotient rule gives
\[
f'(x)=\frac{(1+cx)-cx}{(1+cx)^2}=\frac{1}{(1+cx)^2}.
\]
For $0\le x<y\le1$,
\[
f(y)-f(x)=\frac{y(1+cx)-x(1+cy)}{(1+cx)(1+cy)}=\frac{y-x}{(1+cx)(1+cy)},
\]
and hence
\begin{equation}
A_{x,y}:=\frac{f(y)-f(x)}{y-x}
=\frac{1}{(1+cx)(1+cy)}.
\label{eq:relations-example-secant}
\end{equation}
By Proposition~\ref{prop:subdiff-basics}(v), the Clarke subdifferential of the proper extension is $f'(x)$ plus the normal cone of $[0,1]$ at $x$:
\begin{equation}
\label{eq:pb-ex2-subdiff}
\overline{\partial}f(x)
=
\begin{cases}
(-\infty,1], & x=0,\\
\{f'(x)\}, & x\in(0,1),\\
[\gamma^2,+\infty), & x=1.
\end{cases}
\end{equation}
Here $f'(0)=1$ and $f'(1)=(1+c)^{-2}=\gamma^2$.
Since $f'$ is positive on $[0,1]$, \eqref{eq:pb-ex2-subdiff} shows that, for $a<b$ in $[0,1]$,
\begin{equation}
\label{eq:pb-ex2-extreme}
\max\overline\partial f(a)=f'(a)>0,
\qquad
\min\overline\partial f(b)=f'(b)>0.
\end{equation}

To compute $\gamma^*(f)$, let $\eta\in(0,1]$ and $0\le x<y\le1$.
Since $\psi_\eta$ and $\overline\psi_\eta$ are increasing,
\eqref{eq:pb-ex2-extreme} shows that the two inequalities in \eqref{eq:pb-mwc-1d} hold for all $u\in\overline\partial f(x)$ and $w\in\overline\partial f(y)$ if and only if they hold for $u=f'(x)$ and $w=f'(y)$.
Because $f'(x),f'(y)>0$, these two inequalities are $A_{x,y}\ge\eta f'(x)$ and $A_{x,y}\le f'(y)/\eta$.
By \eqref{eq:relations-example-secant}, they are equivalent to
\[
\eta
\le
\frac{A_{x,y}}{f'(x)}
=
\frac{1+cx}{1+cy}
\qquad\text{and}\qquad
\eta
\le
\frac{f'(y)}{A_{x,y}}
=
\frac{1+cx}{1+cy},
\]
respectively.
Consequently, $f$ is $\eta$--tilted convex if and only if
\[
\eta
\le
\inf_{0\le x<y\le1}\frac{1+cx}{1+cy}
=\frac{1}{1+c}
=\gamma.
\]
The infimum is attained at $x=0$ and $y=1$, because the ratio increases in $x$ and decreases in $y$.
Thus $\gamma^*(f)=\gamma$.

To compute $\gamma^*_{\op{mono}}(f)$, let $x<y$ in $[0,1]$.
Since $\psi_\eta$ is increasing, \eqref{eq:pb-ex2-extreme} shows that the condition in \eqref{eq:pb-mono-1d} holds for all $u\in\overline\partial f(x)$ and $w\in\overline\partial f(y)$ if and only if $f'(y)\ge\psi_\eta(f'(x))=\eta f'(x)$.
This is equivalent to
\[
\eta
\le
\frac{f'(y)}{f'(x)}
=
\left(\frac{1+cx}{1+cy}\right)^2.
\]
It follows that
\[
\gamma^*_{\op{mono}}(f)
=
\inf_{0\le x<y\le1}
\left(\frac{1+cx}{1+cy}\right)^2
=\frac{1}{(1+c)^2}
=\gamma^2.
\]

For $\gamma^*_{\op{sec}}(f)$, we use \eqref{eq:pb-M-1d}, which applies because $f$ is increasing on $[0,1]$.
For $0\le a<x<b\le1$, \eqref{eq:relations-example-secant} gives
\[
\frac{A_{x,b}}{A_{a,x}}
=
\frac{(1+ca)(1+cx)}{(1+cx)(1+cb)}
=
\frac{1+ca}{1+cb}.
\]
So the condition $A_{x,b}\ge\eta A_{a,x}$ in \eqref{eq:pb-M-1d} is equivalent to
\[
\eta\le\frac{1+ca}{1+cb}.
\]
Therefore
\[
\gamma^*_{\op{sec}}(f)
=
\inf_{0\le a<b\le1}\frac{1+ca}{1+cb}
=\frac{1}{1+c}
=\gamma.
\]
Combining the three computations proves the claimed equalities.  Since
$\gamma^*_{\op{mono}}(f)=\gamma^2<\gamma=
\gamma^*_{\op{sec}}(f)$, we also obtain
$f\in\mathcal F_{\op{sec}}^{\gamma}$ and
$f\notin\mathcal F_{\op{mono}}^\gamma$.
The remaining inclusions are Theorem~\ref{thm:gamma-MW-Jensen-vs-gamma-MW-convex}(i) and Theorem~\ref{thm:mwc-vs-mw-mono}(ii) with $F=\overline\partial f$.
\end{proof}

\subsubsection{Proof of Example~\ref{ex:sep-3}}
\label{apx:example-3}

The following lemma is tailored to the two-piece, piecewise-linear
derivative used in the next example.

\begin{lemma}[Two-piece affine integral test]
\label{lem:relations-integral-test}
Let $I=[\ell,r]$ with $\ell<r$, fix $m\in(\ell,r)$, and let
$q:I\to(0,\infty)$ be continuous and nonincreasing.  Suppose that $q$ is
affine on each of $[\ell,m]$ and $[m,r]$.  For $c\in\mathbb R$, define
$f:\mathbb R\to\overline{\mathbb R}$ by
\[
f(x):=
\begin{cases}
c+\displaystyle\int_\ell^x q(t)\,dt, & x\in I,\\[1mm]
+\infty, & x\notin I.
\end{cases}
\]
For $\ell\le x<y\le r$, write
\[
A_{x,y}:=\frac{f(y)-f(x)}{y-x}.
\]
Then
\begin{align}
\gamma^*(f)
&=
\min\left\{
\inf_{\ell\le x<r}\frac{A_{x,r}}{q(x)},
\inf_{\ell<y\le r}\frac{q(y)}{A_{\ell,y}}
\right\},
\label{eq:integral-test-mwc}\\
\gamma^*_{\op{mono}}(f)
&=
\frac{q(r)}{q(\ell)},
\label{eq:integral-test-mono}\\
\gamma^*_{\op{sec}}(f)
&=
\inf_{\ell<x<r}
\frac{A_{x,r}}{A_{\ell,x}}.
\label{eq:integral-test-J}
\end{align}
The secants in these formulas are explicit:
\begin{align*}
A_{\ell,x}
&=\frac{q(\ell)+q(x)}2,
&&\ell<x\le m,\\
A_{x,r}
&=\frac{q(x)+q(r)}2,
&&m\le x<r,\\
A_{x,r}
&=\frac{(m-x)(q(x)+q(m))+(r-m)(q(m)+q(r))}
{2(r-x)},
&&\ell\le x\le m,\\
A_{\ell,y}
&=\frac{(m-\ell)(q(\ell)+q(m))+(y-m)(q(m)+q(y))}
{2(y-\ell)},
&&m\le y\le r.
\end{align*}
\end{lemma}

\begin{proof}
Extend $q$ to $\mathbb R$ by $q(\ell)$ on $(-\infty,\ell)$ and by $q(r)$ on $(r,+\infty)$.
The extension is continuous, so $x\mapsto c+\int_\ell^xq(t)\,dt$ is continuously differentiable on $\mathbb R$ and agrees with $f$ on $I$.
Hence $f$ is proper and lsc.
By Proposition~\ref{prop:subdiff-basics}(v), its Clarke subdifferential is $q(x)$ plus the normal cone of $I$ at $x$:
\[
\overline{\partial}f(x)=
\begin{cases}
(-\infty,q(\ell)], & x=\ell,\\
\{q(x)\}, & \ell<x<r,\\
[q(r),+\infty), & x=r.
\end{cases}
\]
In particular, for $x<y$ in $I$,
\begin{equation}
\label{eq:pb-IT-extreme}
\max\overline\partial f(x)=q(x)>0,
\qquad
\min\overline\partial f(y)=q(y)>0.
\end{equation}
The secant $A_{x,y}$ is the average of $q$ over $[x,y]$.
Since $q$ is nonincreasing, this average does not increase when the right end point moves to the right, and does not decrease when the left end point moves to the left.
Therefore
\begin{equation}
\label{eq:pb-IT-secant-mono}
A_{x,r}\le A_{x,y}\le A_{\ell,y}
\qquad
\forall\,\ell\le x<y\le r.
\end{equation}
Since $q(r)\le q(t)\le q(\ell)$ on $I$, all the ratios in \eqref{eq:integral-test-mwc}--\eqref{eq:integral-test-J} lie in $[q(r)/q(\ell),1]$, so the infima are positive.

Fix $\eta\in(0,1]$ and $x<y$ in $I$.
Since $\psi_\eta$ and $\overline\psi_\eta$ are increasing, \eqref{eq:pb-IT-extreme} shows that the two inequalities in \eqref{eq:pb-mwc-1d} hold for all subgradients at $x$ and $y$ if and only if they hold for $u=q(x)$ and $w=q(y)$.
Since $q$ is positive, these two inequalities are
\[
A_{x,y}\ge\eta q(x),
\qquad
A_{x,y}\le\eta^{-1}q(y).
\]
By \eqref{eq:pb-IT-secant-mono}, the first inequality holds for every $y\in(x,r]$ if and only if it holds for $y=r$.
Likewise, the second inequality holds for every $x\in[\ell,y)$ if and only if it holds for $x=\ell$.
Thus $f$ is $\eta$--tilted convex if and only if
\[
\eta\le\frac{A_{x,r}}{q(x)}\quad\forall x\in[\ell,r)
\qquad\text{and}\qquad
\eta\le\frac{q(y)}{A_{\ell,y}}\quad\forall y\in(\ell,r].
\]
The largest such $\eta$ is the right-hand side of \eqref{eq:integral-test-mwc}, which proves \eqref{eq:integral-test-mwc}.

For $\eta$--tilted monotonicity, \eqref{eq:pb-IT-extreme} and the monotonicity of $\psi_\eta$ show that the condition in \eqref{eq:pb-mono-1d} for $x<y$ reduces to
\[
q(y)\ge\psi_\eta(q(x))=\eta q(x).
\]
Since $q$ is nonincreasing, $q(y)/q(x)\ge q(r)/q(\ell)$ for all $x<y$, with equality at $x=\ell$ and $y=r$.
So the minimum of $q(y)/q(x)$ is
$q(r)/q(\ell)$, which proves \eqref{eq:integral-test-mono}.

Next, $f$ is strictly increasing because $q>0$, so by \eqref{eq:pb-M-1d}, $f$ is $\eta$--tilted secant convex if and only if
\[
A_{x,b}\ge\eta A_{a,x}
\qquad\forall\,\ell\le a<x<b\le r.
\]
By \eqref{eq:pb-IT-secant-mono},
\[
A_{x,b}\ge A_{x,r},
\qquad
A_{a,x}\le A_{\ell,x}.
\]
Hence $A_{x,b}/A_{a,x}\ge A_{x,r}/A_{\ell,x}$, with equality when
$a=\ell$ and $b=r$.  Taking the largest admissible $\eta$ therefore
proves \eqref{eq:integral-test-J}.

Finally, on an interval where $q$ is affine, its integral is the length
of the interval times the average of its endpoint values, which gives the first two secant formulas directly.
For the last two, we split the interval $[x,r]$ or $[\ell,y]$ at $m$ and apply the same rule on each piece.
For example, for $\ell\le x\le m$,
\[
f(r)-f(x)
=\int_x^mq(t)\,dt+\int_m^rq(t)\,dt
=\frac{(m-x)(q(x)+q(m))}2+\frac{(r-m)(q(m)+q(r))}2,
\]
and dividing by $r-x$ gives the third formula.
The fourth formula follows in the same way.
\end{proof}

\begin{proof}[Proof of Example~\ref{ex:sep-3}]
Apply Lemma~\ref{lem:relations-integral-test} with
$\ell=0$, $m=s$, $r=1$, and $c=0$.  The function $q$ is continuous, positive, nonincreasing, and affine on $[0,s]$ and on $[s,1]$.
Put
\[
k:=\frac{2(1-s)}{s(2-s)},
\]
so that $q(x)=1-kx$ on $[0,s]$.
Since $1-a=\frac{2(1-s)}{2-s}$, we have $k=(1-a)/s$, and
\begin{equation}
\label{eq:pb-ex3-values}
q(0)=1,
\qquad
q(s)=1-ks=a=q(1),
\qquad
\frac{2a}{1+a}=\frac{2s/(2-s)}{2/(2-s)}=s.
\end{equation}
By the first secant formula of Lemma~\ref{lem:relations-integral-test} and \eqref{eq:pb-ex3-values},
\begin{equation}
\label{eq:pb-ex3-f}
f(s)=s\,\frac{1+a}{2}=\frac{s}{2-s}=a,
\qquad
f(y)=a+a(y-s)\quad(s\le y\le1),
\qquad
f(1)=a(2-s)=s.
\end{equation}
The monotonicity formula \eqref{eq:integral-test-mono} gives
\[
\gamma^*_{\op{mono}}(f)
=\frac{q(1)}{q(0)}
=a
=\frac{\gamma}{2-\gamma}.
\]

We next compute $\gamma^*(f)$ from \eqref{eq:integral-test-mwc}, starting with its second infimum.  If $0<y\le s$, then $A_{0,y}=(1+q(y))/2$.
Since the map $u\mapsto 2u/(1+u)$ is increasing on $(0,\infty)$ and $q(y)\ge q(s)=a$,
\eqref{eq:pb-ex3-values} gives
\[
\frac{q(y)}{A_{0,y}}
=\frac{2q(y)}{1+q(y)}
\ge\frac{2a}{1+a}
=s.
\]
If $s\le y\le1$, then $q(y)=a$ and $A_{0,y}=f(y)/y=a(1+y-s)/y$ by \eqref{eq:pb-ex3-f}.
Hence
\[
\frac{q(y)}{A_{0,y}}
=\frac{y}{1+y-s}
\ge s,
\]
where the inequality is equivalent to $y(1-s)\ge s(1-s)$, that is, to $y\ge s$.
Equality holds at $y=s$, so the second infimum in
\eqref{eq:integral-test-mwc} equals $s$.

For the first infimum, $A_{x,1}=a=q(x)$ and hence $A_{x,1}/q(x)=1$ when $s\le x<1$.  For
$0\le x\le s$, we have $f(x)=x-kx^2/2$ and $f(1)=s$, so
\[
A_{x,1}-s q(x)
=\frac{s-x+kx^2/2-(s-skx)(1-x)}{1-x}
=\frac{x\bigl[s-1+sk+kx\left(\frac12-s\right)\bigr]}{1-x}.
\]
Using $sk=1-a$, this becomes
\begin{equation}
\label{eq:pb-ex3-bracket}
A_{x,1}-s q(x)
=\frac{x}{1-x}
\left[s-a+kx\left(\frac12-s\right)\right].
\end{equation}
Here $s-a=\frac{s(1-s)}{2-s}>0$.
If $s\le1/2$, both terms in the bracket of \eqref{eq:pb-ex3-bracket} are nonnegative and the first is positive.
If $s>1/2$, the bracket is decreasing in $x$, and its value at $x=s$ is
\[
s-a+(1-a)\left(\frac12-s\right)
=\frac{s(1-s)}{2-s}+\frac{(1-s)(1-2s)}{2-s}
=\frac{(1-s)^2}{2-s}>0.
\]
In both cases the bracket is positive on $[0,s]$, so \eqref{eq:pb-ex3-bracket} gives $A_{x,1}\ge s q(x)$ for every $x\in[0,s]$, with equality at
$x=0$.
Together with the case $s\le x<1$, this shows that the first infimum in \eqref{eq:integral-test-mwc} equals $s$, and
formula~\eqref{eq:integral-test-mwc} yields
\[
\gamma^*(f)=s=\gamma.
\]

It remains to compute $\gamma^*_{\op{sec}}(f)$ from \eqref{eq:integral-test-J}.  For $s\le x<1$, \eqref{eq:pb-ex3-f} gives $A_{x,1}=a$ and $A_{0,x}=a(1+x-s)/x$.
So the ratio in \eqref{eq:integral-test-J} is
\[
\frac{A_{x,1}}{A_{0,x}}
=\frac{x}{1+x-s}
\ge s,
\]
as in the computation of the second infimum above.
For $0<x\le s$, write $x=sz$ with $z\in(0,1]$, and set
\[
h:=\frac{1-s}{2-s},
\qquad\text{so that}\qquad
1-a=2h,
\quad
kx=(1-a)z=2hz.
\]
Then $A_{0,x}=(1+q(x))/2=1-kx/2=1-hz$.
Also $f(1)-f(x)=s-sz+kx^2/2=s(1-z+hz^2)$ and $1-x=1-sz$.
Hence
\begin{equation}
\label{eq:pb-ex3-ratio}
\frac{A_{x,1}}{A_{0,x}}
=
s\frac{1-z+h z^2}{(1-sz)(1-hz)}
=
s\left(
1-h(1-s)\,\phi(z)
\right),
\qquad
\phi(z):=\frac{z(1-z)}{(1-sz)(1-hz)}.
\end{equation}
The second equality in \eqref{eq:pb-ex3-ratio} holds because
\[
(1-sz)(1-hz)-h(1-s)z(1-z)
=1-(s+2h-hs)z+hz^2
=1-z+hz^2,
\]
where $s+2h-hs=s+h(2-s)=s+(1-s)=1$.

To maximize $\phi$ on $(0,1)$, put $P:=(1-s)(1-h)$.
Polynomial division and partial fractions give
\[
\frac1{\phi(z)}
=\frac{(1-sz)(1-hz)}{z(1-z)}
=-sh+\frac1z+\frac{P}{1-z},
\qquad z\in(0,1).
\]
By the Cauchy--Schwarz inequality,
\[
\left(\frac1z+\frac{P}{1-z}\right)\bigl(z+(1-z)\bigr)
\ge\bigl(1+\sqrt P\bigr)^2,
\]
with equality if and only if $(1-z)/z=\sqrt P$, that is, at
\[
z_*=
\frac{1}{1+\sqrt{(1-s)(1-h)}}\in(0,1).
\]
Hence the unique maximum of $\phi$ on $(0,1)$ is attained at $z_*$, and
\begin{equation}
\label{eq:pb-ex3-phimax}
\max_{0<z<1}\phi(z)
=\frac1{(1+\sqrt P)^2-sh}
=\frac{1}{\bigl(\sqrt{1-s}+\sqrt{1-h}\bigr)^2}.
\end{equation}
The last equality uses $(1+\sqrt P)^2-sh=2-s-h+2\sqrt P=\bigl(\sqrt{1-s}+\sqrt{1-h}\bigr)^2$.
At $z=1$ we have $\phi(1)=0$, so \eqref{eq:pb-ex3-phimax} is also the maximum over $(0,1]$.
Since $1-h=(2-s)^{-1}$ and $h(1-s)=(1-s)^2/(2-s)$,
\[
h(1-s)\max_{0<z\le1}\phi(z)
=\frac{(1-s)^2/(2-s)}{\bigl(\sqrt{1-s}+(2-s)^{-1/2}\bigr)^2}
=\left(\frac{1-s}{1+\sqrt{(1-s)(2-s)}}\right)^2.
\]
By \eqref{eq:pb-ex3-ratio}, the infimum of the ratio over $0<x\le s$ is attained at $x=sz_*$ and equals
\[
s\left[
1-
\left(
\frac{1-s}{1+\sqrt{(1-s)(2-s)}}
\right)^2
\right]
<s.
\]
The ratio is at least $s$ on $[s,1)$, so this value is the infimum in \eqref{eq:integral-test-J}, and
\[
\gamma^*_{\op{sec}}(f)
=s\left[
1-
\left(
\frac{1-s}{1+\sqrt{(1-s)(2-s)}}
\right)^2
\right]
<s.
\]
Substituting $s=\gamma$ proves the asserted formula and the strict
inequality $\gamma^*_{\op{sec}}(f)<\gamma$.  Since $\gamma^*_{\op{sec}}(f)<\gamma$, we have $f\notin\mathcal F_{\op{sec}}^{\gamma}$.
Since $\gamma^*(f)=\gamma$, we have $f\in\mathcal F^\gamma$, and Theorem~\ref{thm:mwc-vs-mw-mono}(ii) with $F=\overline\partial f$ and Theorem~\ref{thm:gamma-MW-Jensen-vs-gamma-MW-convex}(ii) with parameter $\gamma^2$ give $\mathcal F^{\gamma}\subseteq\mathcal F^{\gamma^2}_{\op{mono}}\subseteq\mathcal F_{\op{sec}}^{\gamma^2}$.
\end{proof}

\subsection{Proof of Example~\ref{ex:quadratic}}
\label{apx:quadratic}

\begin{proof}
Write $h=h_c$. It is a polynomial, so it is continuously differentiable on $\R$, and $h'(x)=-(2x+c)<0$ on $[0,1]$. For $x,y\in[0,1]$,
\begin{equation}\label{eq:quad-ratio}
h(y)-h(x)=-(y-x)(x+y+c),\qquad h'(x)(y-x)=-(2x+c)(y-x).
\end{equation}
Let $\gamma\in(0,1]$. The proper extension $\widetilde h$ is proper, lsc, and radially continuous, because $h$ is continuous on the closed interval $[0,1]$. By Corollary~\ref{cor:mwc-smooth-test}, $h\in\mathcal F^\gamma$ if, for all $x,y\in[0,1]$,
\begin{equation}\label{eq:m6-quad-grad}
h(y)-h(x)\ge\psi_\gamma\bigl(h'(x)(y-x)\bigr).
\end{equation}
Conversely, $h'(x)\in\overline\partial\widetilde h(x)$ by Proposition~\ref{prop:subdiff-basics}(v), so if $\widetilde h$ is $\gamma$--tilted convex, then \eqref{eq:gamma-mwc} with $\v=h'(x)$ is \eqref{eq:m6-quad-grad}. Thus $h\in\mathcal F^\gamma$ exactly when \eqref{eq:m6-quad-grad} holds for all $x,y\in[0,1]$. If $y=x$, both sides of \eqref{eq:m6-quad-grad} vanish.

If $y<x$, then $h'(x)(y-x)=(2x+c)(x-y)>0$, so the right side of \eqref{eq:m6-quad-grad} is $\gamma(2x+c)(x-y)$, and by \eqref{eq:quad-ratio} the left side is $(x-y)(x+y+c)$. Dividing by $(x-y)(2x+c)>0$ shows that \eqref{eq:m6-quad-grad} is equivalent to
\begin{equation}\label{eq:quad-backward}
\gamma\le\frac{c+x+y}{c+2x}.
\end{equation}
The function $x\mapsto(c+x)/(c+2x)$ has derivative $-c/(c+2x)^2<0$, so its minimum over $[0,1]$ is $(c+1)/(c+2)$, attained at $x=1$. Since $y\ge0$, this gives
\begin{equation}\label{eq:m6-quad-backward-bound}
\frac{c+x+y}{c+2x}\ge\frac{c+x}{c+2x}\ge\frac{c+1}{c+2},
\end{equation}
with equality throughout at $(x,y)=(1,0)$.

If $y>x$, then $h'(x)(y-x)=-(2x+c)(y-x)<0$, so the right side of \eqref{eq:m6-quad-grad} is $-(2x+c)(y-x)/\gamma$, and by \eqref{eq:quad-ratio} the left side is $-(y-x)(x+y+c)$. Dividing by $-(y-x)<0$ reverses the inequality and gives $x+y+c\le(2x+c)/\gamma$, so \eqref{eq:m6-quad-grad} is equivalent to
\begin{equation}\label{eq:quad-forward}
\gamma\le\frac{c+2x}{c+x+y}.
\end{equation}
The function $x\mapsto(c+2x)/(c+x+1)$ has derivative $(c+2)/(c+x+1)^2>0$, so its minimum over $[0,1]$ is $c/(c+1)$, attained at $x=0$. Since $y\le1$, this gives
\begin{equation}\label{eq:m6-quad-forward-bound}
\frac{c+2x}{c+x+y}\ge\frac{c+2x}{c+x+1}\ge\frac c{c+1},
\end{equation}
with equality throughout at $(x,y)=(0,1)$.

Since $(c+1)^2-c(c+2)=1>0$, we have $c/(c+1)<(c+1)/(c+2)$. If $\gamma\le c/(c+1)$, then \eqref{eq:m6-quad-backward-bound} and \eqref{eq:m6-quad-forward-bound} show that \eqref{eq:quad-backward} and \eqref{eq:quad-forward} hold for all pairs, so $h\in\mathcal F^\gamma$. If $\gamma>c/(c+1)$, then \eqref{eq:quad-forward} fails at $(x,y)=(0,1)$, where its right side equals $c/(c+1)$, so $h\notin\mathcal F^\gamma$. Hence $\gamma^*(h)=c/(c+1)$.

For $\gamma^*_{\op{sec}}$, let $\gamma\in(0,1]$, let $0\le s<u\le1$, and let $x\in(s,u)$. By \eqref{eq:quad-ratio}, the slope of $h$ over $[s,x]$ is $-(x+s+c)$, and its slope over $[x,u]$ is $-(u+x+c)$. In Definition~\ref{def:gamma-MW-Jensen-convex} the point $\x$ lies between $\z$ and $\y$, and the two orders of $s$ and $u$ give two conditions. If $\z=s$ and $\y=u$, then $s_-=-(x+s+c)<0$ and $s_+=-(u+x+c)$, and \eqref{eq:secant-form} reads $-(u+x+c)\ge-(x+s+c)/\gamma$. If $\z=u$ and $\y=s$, then $s_-=u+x+c>0$ and $s_+=x+s+c$, and \eqref{eq:secant-form} reads $x+s+c\ge\gamma(u+x+c)$. Both conditions are equivalent to
\begin{equation}\label{eq:quad-jensen}
\gamma\le\frac{x+s+c}{u+x+c}.
\end{equation}
Since $s,x\ge0$ and $u\le1$, $(x+s+c)(c+1)-c(u+x+c)=x+s+cs+c(1-u)\ge0$, so the right side of \eqref{eq:quad-jensen} is at least $c/(c+1)$. Hence \eqref{eq:quad-jensen} holds for all such $s,u,x$ when $\gamma\le c/(c+1)$. At $s=0$ and $u=1$ the right side equals $(x+c)/(x+c+1)$, which tends to $c/(c+1)$ as $x\downarrow0$, so for $\gamma>c/(c+1)$, \eqref{eq:quad-jensen} fails for $x$ close to $0$. Therefore $\widetilde h$ is $\gamma$--tilted secant convex exactly when $\gamma\le c/(c+1)$, and $\gamma^*_{\op{sec}}(h)=c/(c+1)$.

For the monotonicity parameter, let $\kappa=c/(c+2)$, let $x,y\in(0,1)$, and put $u=h'(x)(y-x)$ and $v=h'(y)(y-x)$. Since $h'(t)=-(2t+c)$ is negative with $c\le|h'(t)|\le c+2$ on $[0,1]$, either $u=v=0$ or $u$ and $v$ are nonzero with the same sign, and
\begin{equation}\label{eq:quad-mono-uv}
\kappa|u|\le\kappa(c+2)|y-x|=c|y-x|\le|v|\le(c+2)|y-x|=\frac{c}{\kappa}|y-x|\le\frac{|u|}{\kappa}.
\end{equation}
If $u\ge0$, then $v\ge0$, and \eqref{eq:quad-mono-uv} gives $v\ge\kappa u=\psi_\kappa(u)$. If $u<0$, then $v<0$, and \eqref{eq:quad-mono-uv} gives $v=-|v|\ge-|u|/\kappa=u/\kappa=\psi_\kappa(u)$. Corollary~\ref{cor:mono-smooth-test} therefore gives $\widetilde h\in\mathcal F^{\kappa}_{\op{mono}}$. For $\eta\le c/(c+2)$, Lemma~\ref{lem:psi_basics} gives $\psi_\eta\le\psi_{c/(c+2)}$, so $\overline\partial\widetilde h$ is also $\eta$--tilted monotone. Conversely, $h'(0)=-c\in\overline\partial\widetilde h(0)$ and $h'(1)=-(c+2)\in\overline\partial\widetilde h(1)$ by Proposition~\ref{prop:subdiff-basics}(v). The monotonicity inequality \eqref{eq:gamma-mw-mono} with parameter $\eta$ at $x=0$ and $y=1$ reads $-(c+2)\ge\psi_\eta(-c)=-c/\eta$, which is equivalent to $\eta\le c/(c+2)$. Hence $\overline\partial\widetilde h$ is $\eta$--tilted monotone exactly when $\eta\le c/(c+2)$, and $\gamma^*_{\op{mono}}(h)=c/(c+2)$. Finally, for $\gamma\in(0,1)$,
\[
\frac c{c+1}\ge\gamma\iff c(1-\gamma)\ge\gamma\iff c\ge\frac\gamma{1-\gamma},\qquad
\frac c{c+2}\ge\gamma\iff c(1-\gamma)\ge2\gamma\iff c\ge\frac{2\gamma}{1-\gamma}.\qedhere
\]
\end{proof}

\subsection{Proof of Theorem~\ref{thm:decomposition}}
\label{apx:decomposition}

\begin{proof}[Proof of Theorem~\ref{thm:decomposition}]
Let $h_c(\sigma)=-\sigma^2-c\sigma$ on $[0,1]$, so that $p_i(\x)=\frac\tau2h_c(x_i)$. The identity \eqref{eq:decomposition} holds because
\[
q(\x)+\sum_ip_i(\x)=f(\x)+\frac\tau2\sum_ix_i^2+\frac{\tau c}2\sum_ix_i-\frac\tau2\sum_i(x_i^2+cx_i)=f(\x).
\]
The Hessian of $q$ is $\nabla^2f+\tau I\succeq0$ on $[0,1]^n$. For $\x,\y$ in the cube, $\varphi(t)=q(\x+t(\y-\x))$ has $\varphi''(t)=(\y-\x)^\top\nabla^2q(\x+t(\y-\x))(\y-\x)\ge0$ on $[0,1]$, so $\varphi$ is convex on $[0,1]$. Hence $q(\x+t(\y-\x))\le(1-t)q(\x)+tq(\y)$ for $t\in[0,1]$, and $q$ is convex on the cube.

If $\tau=0$, every $p_i$ vanishes on the cube, and its proper extension is the indicator $\delta_{[0,1]^n}$, which is convex, proper, and lsc. By Theorem~\ref{thm:mwc_one}, its Clarke subdifferential is $1$--tilted monotone, and by \eqref{eq:psi-endpoint-chain} it is $\gamma$--tilted monotone, so $p_i\in\mathcal F^\gamma_{\op{mono}}$. If $\tau>0$, then $p_i(\x)=\frac\tau2h_c(x_i)$. Example~\ref{ex:quadratic} gives $h_c\in\mathcal F^\gamma_{\op{mono}}$, and the proper extension of $h_c$ is proper, lsc, and radially continuous, so Corollary~\ref{cor:mono-smooth-test} gives
\begin{equation}\label{eq:dec-hc}
h_c'(t)(t-s)\ge\psi_\gamma\bigl(h_c'(s)(t-s)\bigr)\qquad(s,t\in(0,1)).
\end{equation}
Let $\x,\y\in(0,1)^n$, and put $s=x_i$ and $t=y_i$. Since $\nabla p_i(\z)=\frac\tau2h_c'(z_i)\BF e_i$,
\[
\ip{\nabla p_i(\y)}{\y-\x}=\frac\tau2h_c'(t)(t-s),\qquad\ip{\nabla p_i(\x)}{\y-\x}=\frac\tau2h_c'(s)(t-s).
\]
Since $\psi_\gamma(\lambda z)=\lambda\psi_\gamma(z)$ for $\lambda>0$, multiplying \eqref{eq:dec-hc} by $\tau/2$ gives $\ip{\nabla p_i(\y)}{\y-\x}\ge\psi_\gamma\bigl(\ip{\nabla p_i(\x)}{\y-\x}\bigr)$. The function $p_i$ is continuous on the cube, so its proper extension is proper, lsc, and radially continuous, and Corollary~\ref{cor:mono-smooth-test} gives $p_i\in\mathcal F^\gamma_{\op{mono}}$. In both cases $p_i\in\mathcal F^\gamma_{\op{mono}}$, and Theorem~\ref{thm:gamma-MW-Jensen-vs-gamma-MW-convex} gives $p_i\in\mathcal F^\gamma_{\op{sec}}\subseteq\mathcal F^\gamma$.

On the cube, $0\le x_i\le1$, so $\norm{\nabla p_i(\x)}=\frac\tau2(2x_i+c)\le\frac\tau2(c+2)$ and $|p_i(\x)|=\frac\tau2(x_i^2+cx_i)\le\frac\tau2(1+c)$. Differentiating \eqref{eq:decomposition} gives $\nabla q(\x)=\nabla f(\x)+\frac\tau2(2\x+c\one)$, and $\norm{2\x+c\one}\le2\norm\x+c\norm\one\le(c+2)\sqrt n$ because $\norm\x\le\sqrt n$ and $\norm\one=\sqrt n$. Hence $\norm{\nabla q(\x)}\le L+\frac\tau2(c+2)\sqrt n$. Finally, every term $x_i^2+cx_i$ lies in $[0,1+c]$, so $|q(\x)|\le|f(\x)|+\frac\tau2\sum_i(x_i^2+cx_i)\le B+\frac{\tau n}2(c+1)$.
\end{proof}

\subsection{Proof of Example~\ref{ex:mono-max}}
\label{apx:mono-max}

\begin{proof}
\noindent\textbf{Concave functions with one kink.}
Let $a>0$, $b:=\gamma a$, $t_0\in(L,U)$, and $c\in\R$, and let
\[
\varphi(t):=\min\bigl\{a(t-t_0),\ b(t-t_0)\bigr\}+c ,
\]
so that $\varphi$ has slope $a$ on $(-\infty,t_0]$ and slope $b<a$ on $[t_0,\infty)$. Let $\widetilde\varphi$ be the proper extension of $\varphi|_{\mathcal K}$. Its Clarke subdifferential is
\begin{equation}\label{eq:mm-subdiff}
\overline\partial\widetilde\varphi(t)=
\begin{cases}
(-\infty,a], & t=L,\\
\{a\}, & L<t<t_0,\\
[b,a], & t=t_0,\\
\{b\}, & t_0<t<U,\\
[b,\infty), & t=U.
\end{cases}
\end{equation}
At $t\in(L,U)\setminus\{t_0\}$, $\widetilde\varphi$ is affine near $t$, and Proposition~\ref{prop:subdiff-basics}(iv) applies. At $t_0$, $\widetilde\varphi$ is Lipschitz near $t_0$, the limits of derivatives at nearby points are $a$ and $b$, and Proposition~\ref{prop:subdiff-basics}(ii) gives $\op{conv}\{a,b\}=[b,a]$. Near $L$, $\widetilde\varphi$ coincides with the proper extension to $\mathcal K$ of the affine function $t\mapsto a(t-t_0)+c$, and Proposition~\ref{prop:subdiff-basics}(v) with $N_{\mathcal K}(L)=(-\infty,0]$ gives $(-\infty,a]$. The case $t=U$ is symmetric, with $N_{\mathcal K}(U)=[0,\infty)$. By \eqref{eq:mm-subdiff},
\begin{equation}\label{eq:mm-bounds}
t^*\le a\ \text{ if }t<U,\qquad t^*\ge b\ \text{ if }t>L,\qquad\text{for } t\in\mathcal K,\ t^*\in\overline\partial\widetilde\varphi(t).
\end{equation}

Let $x,y\in\mathcal K$, $x^*\in\overline\partial\widetilde\varphi(x)$, $y^*\in\overline\partial\widetilde\varphi(y)$, and $d:=y-x$. If $d=0$, both sides of \eqref{eq:gamma-mw-mono} vanish. If $d>0$, then $x<U$ and $y>L$, so \eqref{eq:mm-bounds} gives $x^*\le a$ and $y^*\ge b$. Since $\psi_\gamma$ is nondecreasing and $ad>0$,
\begin{equation}\label{eq:mm-pos}
y^*d\ \ge\ bd=\gamma ad=\psi_\gamma(ad)\ \ge\ \psi_\gamma(x^*d).
\end{equation}
If $d<0$, then $x>L$ and $y<U$, so $x^*\ge b$ and $y^*\le a$. Since $bd<0$, we have $\psi_\gamma(bd)=bd/\gamma=ad$, and
\begin{equation}\label{eq:mm-neg}
y^*d\ \ge\ ad=\psi_\gamma(bd)\ \ge\ \psi_\gamma(x^*d).
\end{equation}
By \eqref{eq:mm-pos} and \eqref{eq:mm-neg}, $\overline\partial\widetilde\varphi$ is $\gamma$--tilted monotone. The function $\widetilde\varphi$ is proper and lsc, since $\varphi$ is continuous and $\mathcal K$ is closed, so $\varphi|_{\mathcal K}\in\mathcal F^\gamma_{\op{mono}}$.

The function $f_2$ is $\varphi$ with $a=1$, $t_0=0$, and $c=0$, and $f_1$ is $\varphi$ with $a=\sigma$, $t_0=2$, and $c=2\sigma-\delta$, since both affine pieces of $f_1$ equal $2\sigma-\delta$ at $t=2$. Both kinks lie in $(L,U)$ because $L<0$ and $U>2$, so $f_1,f_2\in\mathcal F^\gamma_{\op{mono}}$.

\medskip
\noindent\textbf{The maximum.}
For $t\in[L,0]$ we have $f_2(t)=t$, $f_1(t)=\sigma t-\delta$, and $|t|\le|L|$, so
\begin{equation}\label{eq:mm-left}
f_1(t)-f_2(t)=(1-\sigma)|t|-\delta\ \le\ (1-\sigma)|L|-\delta=-\frac\delta2<0 .
\end{equation}
For $t\in[2,U]$ we have $f_2(t)=\gamma t$, $f_1(t)=\gamma \sigma(t-2)+2\sigma-\delta$, and $2\sigma-2\gamma-\delta=\delta$, so
\begin{equation}\label{eq:mm-right}
f_1(t)-f_2(t)=\delta-\gamma(1-\sigma)(t-2)\ \ge\ \delta-\gamma(1-\sigma)(U-2)=\frac\delta2>0 .
\end{equation}
By \eqref{eq:mm-left}, $h(t)=t$ on $[L,0]$, and by \eqref{eq:mm-right}, $h(t)=\gamma \sigma(t-2)+2\sigma-\delta$ on $[2,U]$. Let $\widetilde h$ be the proper extension of $h$, and take $x:=L/2\in(L,0)$ and $y:=(2+U)/2\in(2,U)$. The function $\widetilde h$ is affine near $x$ with slope $1$ and affine near $y$ with slope $\gamma \sigma$, so Proposition~\ref{prop:subdiff-basics}(iv) gives $1\in\overline\partial\widetilde h(x)$ and $\gamma \sigma\in\overline\partial\widetilde h(y)$. Suppose that $\overline\partial\widetilde h$ is $\gamma'$--tilted monotone for some $\gamma'\in[0,1]$. Since $d:=y-x>0$, inequality \eqref{eq:gamma-mw-mono} with $x^*=1$ and $y^*=\gamma \sigma$ reads
\begin{equation}\label{eq:mm-violation}
\gamma \sigma\,d\ \ge\ \psi_{\gamma'}(d)=\gamma' d ,
\end{equation}
so $\gamma'\le\gamma\sigma$ and
$\gamma^*_{\op{mono}}(h)\le\gamma\sigma<\gamma$.

For the reverse bound, the middle branches cross at $t=1$, because
$(\sigma t-\delta)-\gamma t=\delta(t-1)$ on $[0,2]$.
Together with~\eqref{eq:mm-left} and~\eqref{eq:mm-right}, this gives
\[
h(t)=
\begin{cases}
t,&L\le t\le0,\\
\gamma t,&0\le t\le1,\\
\sigma t-\delta,&1\le t\le2,\\
\gamma\sigma(t-2)+2\sigma-\delta,&2\le t\le U.
\end{cases}
\]
All interior slopes lie in $[\gamma\sigma,1]$;
at a kink the Clarke subdifferential is the interval between the
adjacent slopes, and at $L,U$ it includes the respective normal ray.
Consequently, for $a<b$ in $\mathcal K$, every
$u\in\overline\partial\widetilde h(a)$ and
$w\in\overline\partial\widetilde h(b)$ satisfy
\[
u\le1,\qquad w\ge\gamma\sigma,
\qquad
w\ge\gamma\sigma=\psi_{\gamma\sigma}(1)
\ge\psi_{\gamma\sigma}(u).
\]
By~\eqref{eq:pb-mono-1d}, this proves
$h\in\mathcal F_{\op{mono}}^{\gamma\sigma}$ and hence
$\gamma^*_{\op{mono}}(h)=\gamma\sigma$.

Since $f_1,f_2\in\mathcal F_{\op{mono}}^\gamma$,
Theorem~\ref{thm:gamma-MW-Jensen-vs-gamma-MW-convex} gives
$f_1,f_2\in\mathcal F_{\op{sec}}^\gamma$. The class $\mathcal F_{\op{sec}}^\gamma$ is
closed under proper suprema by Theorem~\ref{thm:gamma-MW-Jensen-basics}(iii), and
Theorem~\ref{thm:gamma-MW-Jensen-vs-gamma-MW-convex} again gives
$h\in\mathcal F_{\op{sec}}^\gamma\subseteq\mathcal F^\gamma$.
On the other hand, $1\in\overline\partial\widetilde h(0)$ and
$h(t)-h(0)=\gamma t$ for $0<t<1$.
The support inequality at $0$ with support vector $1$ and parameter
$\eta>0$ would require $\gamma t\ge\psi_\eta(t)=\eta t$,
so $\gamma^*(h)\le\gamma$.
Together with the inclusions and the lower bounds established above, this
gives $\gamma^*_{\op{sec}}(h)=\gamma^*(h)=\gamma$.
\end{proof}

\section{Proofs for Section~\ref{sec:examples}}
\label{apx:examples}

\subsection{Proof of Lemma~\ref{lem:slope-ratio}}
\label{apx:slope-ratio}

\begin{proof}[Proof of Lemma~\ref{lem:slope-ratio}]
Let $\kappa=A/B\in(0,1]$, and let $I$ be the open interval on which $H$ is continuously differentiable. The set $U=\{\x\in\R^d:\ip\a\x\in I\}$ is open and contains $\mathcal K$, and $f(\x)=H(\ip\a\x)$ is continuously differentiable on $U$ with $\nabla f(\x)=H'(\ip\a\x)\a$. Proposition~\ref{prop:subdiff-basics}(v) therefore gives $\overline\partial\widetilde f(\x)=\nabla f(\x)+N_{\mathcal K}(\x)$ for $\x\in\mathcal K$, where $\widetilde f$ is the extension of $f$ by $+\infty$. Fix $\x,\y\in\mathcal K$, let $\dd=\y-\x$, and write arbitrary elements $\x^*\in\overline\partial\widetilde f(\x)$ and $\y^*\in\overline\partial\widetilde f(\y)$ as $\x^*=\nabla f(\x)+\BF n_x$ and $\y^*=\nabla f(\y)+\BF n_y$ with $\BF n_x\in N_{\mathcal K}(\x)$ and $\BF n_y\in N_{\mathcal K}(\y)$. The definition of the normal cone, applied at $\x$ with the point $\y$ and at $\y$ with the point $\x$, gives
\begin{equation}\label{eq:slope-normals}
\ip{\BF n_x}{\dd}\le0,\qquad\ip{\BF n_y}{\dd}\ge0.
\end{equation}
Let $s=\ip\a\x$ and $t=\ip\a\y$, so that $s,t\in J$ and $\ip\a\dd=t-s$. Let
\begin{equation}\label{eq:m6-slope-uv}
u=\ip{\nabla f(\x)}{\dd}=H'(s)(t-s),\qquad v=\ip{\nabla f(\y)}{\dd}=H'(t)(t-s).
\end{equation}
By \eqref{eq:m6-slope-uv}, $|u|=|H'(s)|\,|t-s|$ and $|v|=|H'(t)|\,|t-s|$. The slope bounds $A\le|H'(s)|,|H'(t)|\le B$ and the identity $A=\kappa B$ then give
\begin{equation}\label{eq:slope-ratio-uv}
\kappa|u|\le\kappa B|t-s|=A|t-s|\le|v|\le B|t-s|=\frac A\kappa|t-s|\le\frac{|u|}\kappa.
\end{equation}
Since $H'(s)$ and $H'(t)$ are nonzero with the same sign, \eqref{eq:m6-slope-uv} shows that either $u=v=0$ (when $t=s$) or $u$ and $v$ are nonzero with the same sign. Together with \eqref{eq:slope-ratio-uv}, this gives
\begin{equation}\label{eq:m6-slope-psi}
v\ge\psi_\kappa(u).
\end{equation}
If $u\ge0$, then $v\ge0$, and \eqref{eq:slope-ratio-uv} gives $v=|v|\ge\kappa|u|=\kappa u=\psi_\kappa(u)$. If $u<0$, then $v<0$, and \eqref{eq:slope-ratio-uv} gives $v=-|v|\ge-|u|/\kappa=u/\kappa=\psi_\kappa(u)$. By \eqref{eq:slope-normals}, $\ip{\y^*}{\dd}=v+\ip{\BF n_y}{\dd}\ge v$ and $\ip{\x^*}{\dd}=u+\ip{\BF n_x}{\dd}\le u$. Since $\psi_\kappa$ is nondecreasing (Lemma~\ref{lem:psi_basics}), these two bounds and \eqref{eq:m6-slope-psi} give
\[
\ip{\y^*}{\dd}\ge v\ge\psi_\kappa(u)\ge\psi_\kappa\bigl(\ip{\x^*}{\dd}\bigr).
\]
This is \eqref{eq:gamma-mw-mono} with parameter $\kappa$, and $\op{dom}(\overline\partial\widetilde f)\subseteq\mathcal K$, so $\overline\partial\widetilde f$ is $\kappa$--tilted monotone. The extension $\widetilde f$ is proper, and it is lsc because $f$ is continuous and $\mathcal K$ is closed, so $\widetilde f\in\mathcal F^\kappa_{\op{mono}}$.
\end{proof}

\subsection{Proof of Example~\ref{ex:poly-coordinate}}

\begin{proof}[Proof of Example~\ref{ex:poly-coordinate}]
Apply Lemma~\ref{lem:slope-ratio} with $\mathcal K=[0,1]^n$, the vector $\a=\BF e_i$, and $H(s)=-s^k-as$, which is continuously differentiable on $\R$. Then $J=\{x_i:\x\in[0,1]^n\}=[0,1]$. For $s\in[0,1]$ we have $0\le ks^{k-1}\le k$, so
\begin{equation}\label{eq:m6-poly-slope}
-(a+k)\le H'(s)=-(ks^{k-1}+a)\le-a<0.
\end{equation}
By \eqref{eq:m6-poly-slope}, $H'$ is negative on $J$ with $a\le|H'|\le a+k$, and Lemma~\ref{lem:slope-ratio} with $A=a$ and $B=a+k$ gives $f\in\mathcal F^{a/(a+k)}_{\op{mono}}$. For $\gamma\in(0,1)$,
\begin{equation}\label{eq:m6-poly-threshold}
\frac a{a+k}\ge\gamma\iff a\ge\gamma(a+k)\iff a(1-\gamma)\ge k\gamma\iff a\ge\frac{k\gamma}{1-\gamma}.
\end{equation}
To see that the monotonicity parameter is exact, take $\x=\BF0$, $\y=\BF e_i$, and the ordinary gradients, which belong to the constrained subdifferentials because the zero normal is admissible. For any admissible parameter $\eta>0$, the monotonicity inequality gives $-a/\eta\le-(a+k)$, hence $\eta\le a/(a+k)$.

Suppose that $a\ge k\gamma/(1-\gamma)$, so that $\gamma\le a/(a+k)$ by \eqref{eq:m6-poly-threshold}. Lemma~\ref{lem:psi_basics} gives $\psi_\gamma(z)\le\psi_{a/(a+k)}(z)$ for every real $z$, so each instance of \eqref{eq:gamma-mw-mono} with parameter $a/(a+k)$ implies the same instance with parameter $\gamma$. Hence $f\in\mathcal F^\gamma_{\op{mono}}$, and Corollary~\ref{cor:relations} places $f$ in the two other classes.
\end{proof}

\subsection{Proof of Example~\ref{ex:radial} and Examples~\ref{ex:bowl-cosine} and~\ref{ex:cauchy}}

\begin{proof}[Proof of Example~\ref{ex:radial}]
Taking $r=s$ in the definition of $\delta$ gives $\delta\le1$, so $\delta\in(0,1]$ and $\psi_\delta$ is defined. Let $g(X)=\varphi(\norm X)$ on $\R^m$. For $X\ne\BF0$ the chain rule gives $\nabla g(X)=\varphi'(\norm X)X/\norm X$. At $\BF0$, the mean value theorem gives $|g(X)-g(\BF0)|=|\varphi(\norm X)-\varphi(0)|\le\norm X\max_{[0,\norm X]}|\varphi'|$, and the maximum tends to $|\varphi'(0)|=0$ as $X\to\BF0$, so $g$ is differentiable at $\BF0$. Hence
\begin{equation}\label{eq:m6-radial-grad}
\nabla g(X)=\varphi'(\norm X)\frac X{\norm X}\quad(X\ne\BF0),\qquad\nabla g(\BF0)=\BF0.
\end{equation}
By \eqref{eq:m6-radial-grad}, $\norm{\nabla g(X)}=|\varphi'(\norm X)|\to0$ as $X\to\BF0$, so $\nabla g$ is continuous. Hence $f(\x)=g(A\x-\BF b)$ is continuously differentiable on $\R^d$ with $\nabla f(\x)=A^\top\nabla g(A\x-\BF b)$, and Proposition~\ref{prop:subdiff-basics}(v) gives $\overline\partial\widetilde f(\x)=\nabla f(\x)+N_{\mathcal K}(\x)$ for $\x\in\mathcal K$. As in the proof of Lemma~\ref{lem:slope-ratio}, write $\x^*=\nabla f(\x)+\BF n_x$ and $\y^*=\nabla f(\y)+\BF n_y$ with normal vectors satisfying \eqref{eq:slope-normals}, where $\dd=\y-\x$. Then $\ip{\y^*}{\dd}\ge\ip{\nabla f(\y)}{\dd}$ and $\ip{\x^*}{\dd}\le\ip{\nabla f(\x)}{\dd}$. Since $\psi_\delta$ is nondecreasing, it suffices to show
\begin{equation}\label{eq:radial-mono}
\ip{\nabla f(\y)}{\y-\x}\ge\psi_\delta\bigl(\ip{\nabla f(\x)}{\y-\x}\bigr)\qquad(\x,\y\in\mathcal K).
\end{equation}
Let $X=A\x-\BF b$, $Y=A\y-\BF b$, $p=\norm X$, and $q=\norm Y$, so that $p,q\le R$ and $A(\y-\x)=Y-X$. Then $\ip{\nabla f(\x)}{\y-\x}=\ip{\nabla g(X)}{Y-X}$ and $\ip{\nabla f(\y)}{\y-\x}=\ip{\nabla g(Y)}{Y-X}$. If $p=0$, then by \eqref{eq:m6-radial-grad} the right side of \eqref{eq:radial-mono} is $\psi_\delta(0)=0$ and the left side is $\varphi'(q)q\ge0$. If $q=0$, the left side is $0$ and the right side is $\psi_\delta(-\varphi'(p)p)\le0$. Otherwise $p,q\in(0,R]$, and we let $\xi=\ip{X/p}{Y-X}$ and $\zeta=\ip{Y/q}{Y-X}$. By \eqref{eq:m6-radial-grad}, \eqref{eq:radial-mono} reads
\begin{equation}\label{eq:m6-radial-target}
\varphi'(q)\zeta\ge\psi_\delta\bigl(\varphi'(p)\xi\bigr).
\end{equation}
The Cauchy--Schwarz inequality gives $\ip XY\le pq$, so $\xi=\ip XY/p-p\le q-p$ and $\zeta=q-\ip XY/q\ge q-p$, that is,
\begin{equation}\label{eq:radial-order}
\xi\le q-p\le\zeta.
\end{equation}
Since $p,q\in(0,R]$, $\varphi'(p),\varphi'(q)>0$. By \eqref{eq:radial-order}, $\xi\le\zeta$, so exactly one of the following three cases occurs.

If $\xi\ge0$, then $q\ge p$ by \eqref{eq:radial-order}, and the definition of $\delta$ with $(r,s)=(p,q)$ gives $\varphi'(q)\ge\delta\varphi'(p)$. Since $\zeta\ge\xi\ge0$,
\[
\varphi'(q)\zeta\ge\varphi'(q)\xi\ge\delta\varphi'(p)\xi=\psi_\delta(\varphi'(p)\xi).
\]
If $\xi<0\le\zeta$, the left side of \eqref{eq:m6-radial-target} is nonnegative and the right side is negative. If $\zeta<0$, then $q<p$ by \eqref{eq:radial-order}, and the definition of $\delta$ with $(r,s)=(q,p)$ gives $\varphi'(p)\ge\delta\varphi'(q)$, that is, $\varphi'(q)\le\varphi'(p)/\delta$. Multiplying this by $\zeta<0$ and then using $\zeta\ge\xi$ gives
\[
\varphi'(q)\zeta\ge\frac{\varphi'(p)}\delta\zeta\ge\frac{\varphi'(p)}\delta\xi=\psi_\delta(\varphi'(p)\xi),
\]
where the last equality holds because $\xi<0$. This proves \eqref{eq:m6-radial-target}, hence \eqref{eq:radial-mono}, so $\overline\partial\widetilde f$ is $\delta$--tilted monotone. The extension $\widetilde f$ is proper, and it is lsc because $f$ is continuous and $\mathcal K$ is closed, so $\widetilde f\in\mathcal F^\delta_{\op{mono}}$. The last sentence of the example is the case $A=I$, $\BF b=\BF0$, $\mathcal K=\R^d$, and $R=\infty$, together with Corollary~\ref{cor:relations}.
\end{proof}

\begin{proof}[Proof of Example~\ref{ex:bowl-cosine}]
Here $f(\x)=\varphi(\norm\x)$ with $\varphi(r)=\frac12r^2-b\cos(\omega r)$. With $g(u)=u+\kappa\sin u$ and $\kappa=b\omega^2$,
\begin{equation}\label{eq:m6-bowl-phiprime}
\varphi'(r)=r+b\omega\sin(\omega r)=\frac{g(\omega r)}\omega,\qquad\varphi''(r)=g'(\omega r)=1+\kappa\cos(\omega r),
\end{equation}
so $\varphi'(0)=0$. We use Example~\ref{ex:radial} with $A=I$, $\BF b=\BF0$, $\mathcal K=\R^d$, and $R=\infty$. Once we show that $g>0$ on $(0,\infty)$, \eqref{eq:m6-bowl-phiprime} gives $\varphi'>0$ on $(0,\infty)$, and the substitution $u=\omega r$, $v=\omega s$ in the definition of $\delta$ gives
\begin{equation}\label{eq:bowl-delta}
\delta=\inf\{g(v)/g(u):0<u\le v\}.
\end{equation}

If $\kappa\le1$, then $g'(u)=1+\kappa\cos u\ge1-\kappa\ge0$, so $g$ is nondecreasing with $g(0)=0$. By \eqref{eq:m6-bowl-phiprime}, $\varphi'\ge0$ and $\varphi''\ge0$ on $[0,\infty)$, so $\varphi$ is nondecreasing and convex there. For $\x,\y\in\R^d$ and $\lambda\in[0,1]$, the triangle inequality and these two properties give $f(\lambda\x+(1-\lambda)\y)\le\varphi(\lambda\norm\x+(1-\lambda)\norm\y)\le\lambda f(\x)+(1-\lambda)f(\y)$, so $f$ is convex.

Let $\kappa>1$, $s=s_\kappa=\arccos(-1/\kappa)$, and $s'=2\pi-s$. Since $-1/\kappa\in(-1,0)$, we have $s\in(\pi/2,\pi)$ and $s'\in(\pi,3\pi/2)$. On $[0,2\pi]$, $\cos u>-1/\kappa$ exactly on $[0,s)\cup(s',2\pi]$, so $g'=1+\kappa\cos u$ is positive on $[0,s)$, negative on $(s,s')$, and positive on $(s',2\pi]$. Since $s\in(0,\pi)$, $\sin s=\sqrt{1-1/\kappa^2}$, and $\sin s'=-\sin s$, so $\kappa\sin s=\sqrt{\kappa^2-1}=-\kappa\sin s'$ and
\begin{equation}\label{eq:m6-bowl-gs}
g(s)=s+\sqrt{\kappa^2-1},\qquad g(s')=2\pi-s-\sqrt{\kappa^2-1}.
\end{equation}
Let $G(\kappa)=2\pi-\arccos(-1/\kappa)-\sqrt{\kappa^2-1}$ for $\kappa\ge1$, so that $G(\kappa)=g(s')$ by \eqref{eq:m6-bowl-gs}. For $\kappa>1$, the derivative of $\arccos(-1/\kappa)$ is $-1/(\kappa\sqrt{\kappa^2-1})$, so
\[
G'(\kappa)=\frac1{\kappa\sqrt{\kappa^2-1}}-\frac\kappa{\sqrt{\kappa^2-1}}=\frac{1-\kappa^2}{\kappa\sqrt{\kappa^2-1}}=-\frac{\sqrt{\kappa^2-1}}\kappa<0.
\]
Also $G$ is continuous on $[1,\infty)$, $G(1)=2\pi-\pi=\pi$, and $G(\kappa)\le2\pi-\pi/2-\sqrt{\kappa^2-1}\to-\infty$ as $\kappa\to\infty$. Hence $G$ vanishes at exactly one point of $(1,\infty)$, which is the number $\kappa_0$ of Example~\ref{ex:bowl-cosine}, and
\begin{equation}\label{eq:m6-bowl-gsprime-pos}
g(s')=G(\kappa)>0\qquad(1<\kappa<\kappa_0).
\end{equation}
From now on let $1<\kappa<\kappa_0$. Let $a_k=s+2\pi k$ and $b_k=s'+2\pi k$ for $k\ge0$, and $b_{-1}=0$. Since $g(u+2\pi)=g(u)+2\pi$ and $g'$ has period $2\pi$, the sign pattern of $g'$ on $[0,2\pi]$ shows that, for every $k\ge0$, $g$ is increasing on $[b_{k-1},a_k]$ and decreasing on $[a_k,b_k]$, with
\begin{equation}\label{eq:bowl-extrema}
g(a_k)=g(s)+2\pi k,\qquad g(b_k)=g(s')+2\pi k.
\end{equation}
Since $g$ is decreasing on $[s,s']$, we have $g(s')<g(s)$. The function $g$ is positive on $(0,a_0]$, because it increases there from $g(0)=0$. On $[a_k,b_{k+1}]$ it is at least $\min\{g(b_k),g(b_{k+1})\}=g(b_k)$, which is positive by \eqref{eq:bowl-extrema} and \eqref{eq:m6-bowl-gsprime-pos}. Hence $g>0$ on $(0,\infty)$, and \eqref{eq:bowl-delta} holds.

The monotonicity pattern also gives, for every $j\ge0$,
\begin{equation}\label{eq:m6-bowl-unimodal}
\min_{[p,p']}g=\min\{g(p),g(p')\}\quad\text{for }b_{j-1}\le p\le p'\le b_j,\qquad\max_{[b_{j-1},b_j]}g=g(a_j).
\end{equation}
Let $0<u\le v$, and let $k\ge0$ be such that $u\in(b_{k-1},b_k]$. The second part of \eqref{eq:m6-bowl-unimodal} with $j=k$ gives $g(u)\le g(a_k)$. Next, $g(v)\ge\min\{g(u),g(b_k)\}$: if $v\le b_k$, this is the first part of \eqref{eq:m6-bowl-unimodal} on $[u,b_k]$, and if $v>b_k$, then $v\in[b_j,b_{j+1}]$ for some $j\ge k$, and the first part of \eqref{eq:m6-bowl-unimodal} on $[b_j,b_{j+1}]$ together with \eqref{eq:bowl-extrema} gives $g(v)\ge\min\{g(b_j),g(b_{j+1})\}=g(b_j)\ge g(b_k)$. Since $0<g(u)\le g(a_k)$ and $g(b_k)>0$,
\[
\frac{g(v)}{g(u)}\ge\min\Bigl\{1,\frac{g(b_k)}{g(u)}\Bigr\}\ge\min\Bigl\{1,\frac{g(b_k)}{g(a_k)}\Bigr\}=\frac{g(s')+2\pi k}{g(s)+2\pi k}\ge\frac{g(s')}{g(s)}=\gamma_\kappa.
\]
The equality uses \eqref{eq:bowl-extrema} and $g(s')<g(s)$, and the last inequality holds because $t\mapsto(g(s')+t)/(g(s)+t)$ has derivative $(g(s)-g(s'))/(g(s)+t)^2>0$ for $t\ge0$. The identity $g(s')/g(s)=\gamma_\kappa$ is \eqref{eq:m6-bowl-gs}. At $(u,v)=(s,s')$ the ratio equals $\gamma_\kappa$, so \eqref{eq:bowl-delta} gives $\delta=\gamma_\kappa>0$. Example~\ref{ex:radial} then gives $f\in\mathcal F^{\gamma_\kappa}_{\op{mono}}$, and Corollary~\ref{cor:relations} places $f$ in the two other classes.

To prove exactness, fix a unit vector $\BF e$, put $r_-=s_\kappa/\omega$ and $r_+=(2\pi-s_\kappa)/\omega$, and take $\x=r_-\BF e$, $\y=r_+\BF e$. Their gradients project onto $\y-\x$ as the positive numbers $\varphi'(r_-)(r_+-r_-)$ and $\varphi'(r_+)(r_+-r_-)$. Monotonicity at parameter $\eta$ therefore requires
\[
\eta\,\varphi'(r_-)(r_+-r_-)
\le\varphi'(r_+)(r_+-r_-),
\]
so $\eta\le\varphi'(r_+)/\varphi'(r_-)=\gamma_\kappa$. Together with membership at $\gamma_\kappa$, this proves $\gamma^*_{\op{mono}}(f)=\gamma_\kappa$.

For nonconvexity, fix a unit vector $\BF e$. Since $\cos$ is even, $f(t\BF e)=\frac12t^2-b\cos(\omega t)$ for $t\in\R$, whose second derivative $1+\kappa\cos(\omega t)$ equals $1-\kappa<0$ at $t=\pi/\omega$. For $\kappa=2$, $s_2=\arccos(-1/2)=2\pi/3$ and $\sqrt{\kappa^2-1}=\sqrt3$. Then $G(2)=4\pi/3-\sqrt3>0$, so $2<\kappa_0$ because $G$ is decreasing with $G(\kappa_0)=0$, and
\[
\gamma_2=\frac{4\pi/3-\sqrt3}{2\pi/3+\sqrt3}=\frac{4\pi-3\sqrt3}{2\pi+3\sqrt3}.\qedhere
\]
\end{proof}

\begin{proof}[Proof of Example~\ref{ex:cauchy}]
Both losses have the form $\varphi(|r(\x)|)$, which is the case $m=1$, $A=\a^\top$, and $\BF b=\beta$ of Example~\ref{ex:radial}, because $\norm{A\x-\BF b}=|r(\x)|\le R$ on $\mathcal K$. In both cases $\varphi$ is continuously differentiable on $[0,\infty)$, $\varphi'(0)=0$, $\varphi'>0$ on $(0,\infty)$, and for some $\sigma_0>0$, $\varphi'$ is increasing on $[0,\sigma_0]$ and decreasing on $[\sigma_0,\infty)$. For such $\varphi$ we first show that
\begin{equation}\label{eq:m6-cauchy-delta}
\delta=\frac{\varphi'(R)}{\varphi'(\sigma_0)}\qquad\text{if }R\ge\sigma_0.
\end{equation}
Let $0<r\le s\le R$. If $s\le\sigma_0$, then $\varphi'(s)\ge\varphi'(r)$, so $\varphi'(s)/\varphi'(r)\ge1\ge\varphi'(R)/\varphi'(\sigma_0)$, where the last step holds because $\varphi'(\sigma_0)$ is the maximum of $\varphi'$. If $s>\sigma_0$, then $s\in(\sigma_0,R]$, so $\varphi'(s)\ge\varphi'(R)$, and $\varphi'(r)\le\varphi'(\sigma_0)$. In both cases $\varphi'(s)/\varphi'(r)\ge\varphi'(R)/\varphi'(\sigma_0)$, with equality at $(r,s)=(\sigma_0,R)$, which proves \eqref{eq:m6-cauchy-delta}. If instead $R\le\sigma_0$, then $\varphi'\ge0$ is nondecreasing on $[0,R]$, so $\varphi$ is nondecreasing and convex on $[0,R]$. The function $|r|$ is convex on $\mathcal K$ with values in $[0,R]$, so for $\x,\y\in\mathcal K$ and $\lambda\in[0,1]$,
\[
\varphi\bigl(|r(\lambda\x+(1-\lambda)\y)|\bigr)\le\varphi\bigl(\lambda|r(\x)|+(1-\lambda)|r(\y)|\bigr)\le\lambda\varphi(|r(\x)|)+(1-\lambda)\varphi(|r(\y)|),
\]
and the loss is convex on $\mathcal K$.

For the Cauchy loss, $\varphi(s)=\log(1+s^2/\sigma^2)$, and
\[
\varphi'(s)=\frac{2s}{\sigma^2+s^2},\qquad\varphi''(s)=\frac{2(\sigma^2-s^2)}{(\sigma^2+s^2)^2},
\]
so the above holds with $\sigma_0=\sigma$ and $\varphi'(\sigma)=1/\sigma$. The condition $R\ge\sigma$ is $q\ge1$, and then \eqref{eq:m6-cauchy-delta} gives
\[
\delta=\frac{\varphi'(R)}{\varphi'(\sigma)}=\frac{2R\sigma}{\sigma^2+R^2}=\frac{2\sqrt q}{1+q},
\]
where the last step divides numerator and denominator by $\sigma^2$ and uses $R/\sigma=\sqrt q$. If $q\le1$, then $R\le\sigma=\sigma_0$, and the loss is convex.

For the Geman--McClure loss, $\varphi(s)=s^2/(\sigma^2+s^2)$, and
\[
\varphi'(s)=\frac{2\sigma^2s}{(\sigma^2+s^2)^2},\qquad\varphi''(s)=\frac{2\sigma^2(\sigma^2-3s^2)}{(\sigma^2+s^2)^3},
\]
so the above holds with $\sigma_0=\sigma/\sqrt3$. Since $\sigma^2+\sigma_0^2=4\sigma^2/3$, the maximum is $\varphi'(\sigma/\sqrt3)=(2\sigma^3/\sqrt3)\cdot9/(16\sigma^4)=3\sqrt3/(8\sigma)$. The condition $R\ge\sigma/\sqrt3$ is $q\ge1/3$, and then \eqref{eq:m6-cauchy-delta} gives
\[
\delta=\frac{\varphi'(R)}{\varphi'(\sigma/\sqrt3)}=\frac{2\sigma^2R}{(\sigma^2+R^2)^2}\cdot\frac{8\sigma}{3\sqrt3}=\frac{16\sqrt q}{3\sqrt3\,(1+q)^2},
\]
where the last step divides numerator and denominator by $\sigma^4$. If $q\le1/3$, then $R\le\sigma_0$, and the loss is convex.

For the second claim in (b), take $g=r^2$, which is convex and continuously differentiable because $r$ is affine, $J=[0,R^2]$, and $\phi(u)=u/(\sigma^2+u)$ in Example~\ref{ex:transform}. Then $\phi\circ g$ is the Geman--McClure loss, and $\phi'(u)=\sigma^2/(\sigma^2+u)^2>0$ on the open interval $(-\sigma^2,\infty)\supseteq J$. For $u\ne v$ in $J$, let $\Lambda(u,v)=(\phi(v)-\phi(u))/(\phi'(u)(v-u))$. Dividing \eqref{eq:relative-secants} by $\phi'(u)(v-u)$, which is positive for $v>u$ and negative for $v<u$, shows that \eqref{eq:relative-secants} holds exactly when $\Lambda(u,v)\ge a$ for $v>u$ and $\Lambda(u,v)\le b$ for $v<u$ (the case $v=u$ is trivial). Here
\[
\phi(v)-\phi(u)=\frac{\sigma^2(v-u)}{(\sigma^2+u)(\sigma^2+v)},\qquad\text{so}\qquad\Lambda(u,v)=\frac{\sigma^2+u}{\sigma^2+v}.
\]
For $v>u$ in $J$, $\Lambda(u,v)\ge\sigma^2/(\sigma^2+R^2)=1/(1+q)$, and for $v<u$ in $J$, $\Lambda(u,v)\le(\sigma^2+R^2)/\sigma^2=1+q$. Example~\ref{ex:transform} with $a=1/(1+q)$ and $b=1+q$ gives the parameter $\min\{a,1/b\}=1/(1+q)$. For $q\ge1/3$, comparison of the two parameters gives
\[
\frac1{1+q}\ge\frac{16\sqrt q}{3\sqrt3(1+q)^2}
\iff 3\sqrt3(1+q)\ge16\sqrt q
\iff q\ge\frac{101+16\sqrt{37}}{27}.
\]
For the last equivalence, set $t=\sqrt q$: the roots of
$3\sqrt3t^2-16t+3\sqrt3$ are $(8\pm\sqrt{37})/(3\sqrt3)$, and the smaller root is below $1/\sqrt3$. Finally, as functions of $r$, the second derivatives of the two losses are $2(\sigma^2-r^2)/(\sigma^2+r^2)^2$ and $2\sigma^2(\sigma^2-3r^2)/(\sigma^2+r^2)^3$, which are negative exactly when $|r|>\sigma$ and when $|r|>\sigma/\sqrt3$, respectively.
\end{proof}

\subsection{Proofs of Example~\ref{ex:fractional} and Example~\ref{ex:ratios}}

\begin{proof}[Proof of Example~\ref{ex:fractional}]
We first prove $\gamma$--tilted secant convexity. Since $Z$ is continuous and positive on the compact set $\mathcal K$, $\gamma\in(0,1]$, and for all $\y,\z\in\mathcal K$,
\begin{equation}\label{eq:m6-frac-rho}
\frac{Z(\z)}{Z(\y)}\ge\frac{\min_{\mathcal K}Z}{\max_{\mathcal K}Z}=\gamma.
\end{equation}
Let $\y\ne\z$ in $\mathcal K$ and $\lambda\in(0,1)$, and let $\x=\lambda\y+(1-\lambda)\z$, which lies in the convex set $\mathcal K$, $w_y=\lambda Z(\y)>0$, and $w_z=(1-\lambda)Z(\z)>0$. Convexity of $W$ and $W=fZ$ give $W(\x)\le\lambda W(\y)+(1-\lambda)W(\z)=w_yf(\y)+w_zf(\z)$. If $Z$ is affine, then $Z(\x)=w_y+w_z$, and dividing by $Z(\x)$ gives \eqref{eq:frac-mean} below. If $Z$ is concave and $W\ge0$, then $Z(\x)\ge w_y+w_z>0$ and $W(\x)\ge0$, so $f(\x)=W(\x)/Z(\x)\le W(\x)/(w_y+w_z)$. In both cases
\begin{equation}\label{eq:frac-mean}
f(\x)\le\frac{w_yf(\y)+w_zf(\z)}{w_y+w_z}.
\end{equation}
Put $a=f(\y)$, $b=f(\z)$, and $m=f(\x)$. Multiplying \eqref{eq:frac-mean} by $w_y+w_z$ gives $\lambda Z(\y)(a-m)\ge(1-\lambda)Z(\z)(m-b)$, and by \eqref{eq:secant-values} this is $Z(\y)s_+\ge Z(\z)s_-$. By \eqref{eq:m6-frac-rho}, applied to both orders of $\y$ and $\z$, the ratio $r:=Z(\z)/Z(\y)$ lies in $[\gamma,1/\gamma]$. If $s_-\ge0$, then $s_+\ge rs_-\ge\gamma s_-=\psi_\gamma(s_-)$. If $s_-<0$, then $s_+\ge rs_-\ge s_-/\gamma=\psi_\gamma(s_-)$. Hence the extension of $f$ by $+\infty$ is $\gamma$--tilted secant convex. It is proper and lsc, because $f$ is continuous on the compact set $\mathcal K$, so Theorem~\ref{thm:gamma-MW-Jensen-vs-gamma-MW-convex}(i) makes it $\gamma$--tilted convex. The inclusion $\mathcal F^\gamma\subseteq\mathcal F^{\gamma^2}_{\op{mono}}$ of Corollary~\ref{cor:relations} makes its Clarke subdifferential $\gamma^2$--tilted monotone.

For \eqref{eq:fractional-identity}, differentiating $W=fZ$ gives $\nabla f(\x)=(\nabla W(\x)-f(\x)\nabla Z(\x))/Z(\x)$. For $\x,\y\in\mathcal K$,
\begin{equation}\label{eq:frac-start}
Z(\y)\bigl(f(\y)-f(\x)\bigr)=W(\y)-f(\x)Z(\y).
\end{equation}
Convexity of $W$ gives
\begin{equation}\label{eq:m6-frac-N}
W(\y)\ge W(\x)+\ip{\nabla W(\x)}{\y-\x}.
\end{equation}
If $Z$ is affine, $Z(\y)=Z(\x)+\ip{\nabla Z(\x)}{\y-\x}$. If $Z$ is concave and $W\ge0$, then $Z(\y)\le Z(\x)+\ip{\nabla Z(\x)}{\y-\x}$ and $f(\x)\ge0$. In both cases
\begin{equation}\label{eq:m6-frac-Q}
-f(\x)Z(\y)\ge-f(\x)\bigl(Z(\x)+\ip{\nabla Z(\x)}{\y-\x}\bigr).
\end{equation}
Inserting \eqref{eq:m6-frac-N} and \eqref{eq:m6-frac-Q} in \eqref{eq:frac-start} and using $W(\x)=f(\x)Z(\x)$ gives
\[
Z(\y)\bigl(f(\y)-f(\x)\bigr)\ge\ip{\nabla W(\x)-f(\x)\nabla Z(\x)}{\y-\x}=Z(\x)\ip{\nabla f(\x)}{\y-\x},
\]
and dividing by $Z(\y)>0$ gives \eqref{eq:fractional-identity}. If $W$ and $Z$ are affine, \eqref{eq:m6-frac-N} and \eqref{eq:m6-frac-Q} are equalities, so \eqref{eq:fractional-identity} is an equality.
\end{proof}

\begin{proof}[Proof of the claims after Example~\ref{ex:fractional} and of Example~\ref{ex:ratios}]
For $f(t)=t/(1+ct)$ on $[0,1]$ with $c=(1-\gamma)/\gamma$, take $W(t)=t$ and $Z(t)=1+ct$. Both are affine, $Z$ is positive on $(-1/c,\infty)\supseteq[0,1]$, and $Z([0,1])=[1,1+c]=[1,1/\gamma]$, so $\min_{[0,1]}Z/\max_{[0,1]}Z=\gamma$ and Example~\ref{ex:fractional} gives $f\in\mathcal F^\gamma$. At $x=0$ and $y=1$, $s=f'(0)=1>0$, and \eqref{eq:fractional-identity} holds with equality, so $f(1)-f(0)=(Z(0)/Z(1))s=\gamma s$. For $\theta\in(\gamma,1]$ this gives $f(1)-f(0)<\theta s=\psi_\theta(s)$. Since $f'(0)\in\overline\partial\widetilde f(0)$ by Proposition~\ref{prop:subdiff-basics}(v), \eqref{eq:gamma-mwc} with parameter $\theta$ fails, so $f\notin\mathcal F^\theta$ and $\gamma^*(f)=\gamma$. Example~\ref{ex:sep-2} gives $\gamma^*_{\op{sec}}(f)=\gamma$ and $\gamma^*_{\op{mono}}(f)=\gamma^2$.

In Example~\ref{ex:ratios}(a), $W(\x)=\sqrt{\x^\top\Sigma\x}=\norm{\Sigma^{1/2}\x}$ is convex as a norm of a linear map. It is continuously differentiable on the open set $\R^n\setminus\{\BF0\}$, because $\Sigma^{1/2}\x\ne\BF0$ for $\x\ne\BF0$, and this set contains the probability simplex $\Delta_n$. The function $Z(\x)=\ip\mu\x$ is affine, and for $\x\in\Delta_n$ the value $\ip\mu\x=\sum_i\mu_ix_i$ is a convex combination of the $\mu_i$, so $\min_i\mu_i\le Z(\x)\le\max_i\mu_i$, with equality at vertices. Hence $Z>0$ on $\Delta_n$, $\gamma=\min_i\mu_i/\max_i\mu_i$, and the first case of Example~\ref{ex:fractional} applies. In Example~\ref{ex:ratios}(b), $W(\x)=\ip{\BF c}\x+P_0$ is affine, hence convex, and it is positive on $\{\x\ge\BF0\}$ because $\BF c\ge\BF0$ and $P_0>0$. The denominator $R$ is concave and continuously differentiable on an open convex neighborhood of $\mathcal K$, and it is positive on $\mathcal K$, so the second case of Example~\ref{ex:fractional} applies.
\end{proof}

\subsection{Proofs of Example~\ref{ex:sup-H} and Example~\ref{ex:distorted-convex}}

\begin{proof}[Proof of Example~\ref{ex:sup-H}]
Each $h_i$ is continuous, so $f$ is lsc. It is proper, because $f\ge h_i>-\infty$ everywhere and $f$ is finite at some point. By definition $f$ is $\mathcal H_\gamma$--convex, Theorem~\ref{thm:H-convex-equivalent-def} makes it $\gamma^2$--tilted secant convex, and Theorem~\ref{thm:gamma-MW-Jensen-vs-gamma-MW-convex}(i) makes it $\gamma^2$--tilted convex.
\end{proof}

\begin{proof}[Proof of the monotonicity claim in Example~\ref{ex:distorted-convex}]
Let $f=\psi_\gamma\circ\phi$. Since $\phi:\R^d\to\R$ is finite and convex, it is locally Lipschitz, and its Clarke subdifferential equals its convex subdifferential $\partial\phi$. The function $f$ is finite and continuous, hence proper and lsc. Since $\overline\partial\psi_\gamma(u)\subseteq[\gamma,1/\gamma]$ for every $u$, the Clarke chain rule \citep[Theorem~2.3.9]{clarke90_optim_nonsm_analy} gives
\[
\overline\partial f(\x)
\subseteq\operatorname{co}\{s\g:s\in[\gamma,1/\gamma],\ \g\in\partial\phi(\x)\}
=[\gamma,1/\gamma]\,\partial\phi(\x).
\]
For the equality, a convex combination $\sum_i\lambda_i s_i\g_i$ equals $s\g$, where $s=\sum_i\lambda_i s_i\in[\gamma,1/\gamma]$ and
$\g=\sum_i(\lambda_i s_i/s)\g_i\in\partial\phi(\x)$ by convexity of the subdifferential.

Fix $\x,\y$, put $\dd=\y-\x$, and write arbitrary
$\x^*\in\overline\partial f(\x)$ and $\y^*\in\overline\partial f(\y)$ as $\x^*=s\g$ and $\y^*=t\BF h$, with $s,t\in[\gamma,1/\gamma]$, $\g\in\partial\phi(\x)$, and $\BF h\in\partial\phi(\y)$. Monotonicity of $\partial\phi$ and $t/s\in[\gamma^2,1/\gamma^2]$ give
\begin{align*}
\ip{\y^*}{\dd}
&=t\ip{\BF h}{\dd}
\ge t\ip{\g}{\dd}
=\frac ts\ip{\x^*}{\dd}\\
&\ge\min\left\{\gamma^2\ip{\x^*}{\dd},
\frac1{\gamma^2}\ip{\x^*}{\dd}\right\}
=\psi_{\gamma^2}(\ip{\x^*}{\dd}).
\end{align*}
Thus $\overline\partial f$ is $\gamma^2$--tilted monotone.
\end{proof}

\subsection{Proof of the exactness claim in Example~\ref{ex:two-slope}}

\begin{proof}[Proof of the exactness claim in Example~\ref{ex:two-slope}]
If $a=b$, then $b/a=1$ is the largest possible parameter, so let $0<b<a$ and $\theta\in(b/a,1]$. The function $f$ is differentiable at $x=1$ with $f'(1)=b$, so $\overline\partial f(1)=\{b\}$. For $y=-T$ with $T>0$, $f(y)-f(x)=-aT-b$, and $b(y-x)=-b(T+1)<0$, so $\psi_\theta(b(y-x))=-b(T+1)/\theta$. The inequality \eqref{eq:gamma-mwc} with parameter $\theta$ requires $-aT-b\ge-b(T+1)/\theta$, that is,
\begin{equation}\label{eq:m6-two-slope}
\theta(aT+b)\le b(T+1),\quad\text{equivalently}\quad T(\theta a-b)\le b(1-\theta).
\end{equation}
Since $\theta a-b>0$, \eqref{eq:m6-two-slope} fails for $T>b(1-\theta)/(\theta a-b)$, so $f$ is not $\theta$--tilted convex and $\gamma^*(f)\le b/a$. Example~\ref{ex:two-slope} shows $\gamma^*_{\op{mono}}(f)\ge b/a$, and Corollary~\ref{cor:relations} gives $\gamma^*_{\op{mono}}(f)\le\gamma^*_{\op{sec}}(f)\le\gamma^*(f)$, so all three parameters equal $b/a$.
\end{proof}

\subsection{Proof of Example~\ref{ex:transform}}

\begin{proof}[Proof of Example~\ref{ex:transform}]
Let $U\supseteq\mathcal K$ be an open set on which $g$ is continuously differentiable, and $I\supseteq J$ an open interval on which $\phi$ is continuously differentiable. Then $\phi\circ g$ is continuously differentiable on the open set $U\cap g^{-1}(I)\supseteq\mathcal K$, with gradient $\phi'(g(\x))\nabla g(\x)$. The parameter $\theta=\min\{a,1/b\}$ lies in $(0,1]$ and satisfies $\theta\le a$ and $b\le1/\theta$. Fix $\x,\y\in\mathcal K$, and let $u=g(\x)$, $v=g(\y)$, and $s=\ip{\nabla g(\x)}{\y-\x}$. Convexity of $g$ gives
\begin{equation}\label{eq:transform-convex}
v-u\ge s.
\end{equation}
The proper extension of $(\phi\circ g)|_{\mathcal K}$ is proper, lsc, and radially continuous, because $\phi\circ g$ is continuous on the closed set $\mathcal K$. The vector $\y-\x$ lies in the linear subspace parallel to the affine hull of $\mathcal K$, so the gradient of $\phi\circ g$ relative to that hull has the same inner product with $\y-\x$ as $\phi'(u)\nabla g(\x)$. By Corollary~\ref{cor:mwc-smooth-test}, it therefore suffices to show $\phi(v)-\phi(u)\ge\psi_\theta(\phi'(u)s)$. Recall $\phi'(u)>0$. If $s\ge0$, then $v-u\ge s\ge0$ by \eqref{eq:transform-convex}. The first line of \eqref{eq:relative-secants}, then \eqref{eq:transform-convex} with $a\phi'(u)>0$, then $\theta\le a$ with $\phi'(u)s\ge0$, give
\[
\phi(v)-\phi(u)\ge a\phi'(u)(v-u)\ge a\phi'(u)s\ge\theta\phi'(u)s=\psi_\theta(\phi'(u)s).
\]
If $s<0$ and $v\ge u$, then $\phi(v)-\phi(u)\ge0>\psi_\theta(\phi'(u)s)$ because $\phi$ is increasing. If $s<0$ and $v<u$, the second line of \eqref{eq:relative-secants}, then \eqref{eq:transform-convex} with $b\phi'(u)>0$, then $b\le1/\theta$ with $\phi'(u)s<0$, give
\[
\phi(v)-\phi(u)\ge b\phi'(u)(v-u)\ge b\phi'(u)s\ge\phi'(u)s/\theta=\psi_\theta(\phi'(u)s).
\]
Corollary~\ref{cor:mwc-smooth-test} therefore gives $(\phi\circ g)|_{\mathcal K}\in\mathcal F^\theta$.
\end{proof}

\section{Proofs for Section~\ref{sec:linearization}}
\label{apx:precision}

Throughout this appendix, $\mathcal K$ is compact and convex with diameter at most $D$, a feasible point $\z_1\in\mathcal K$ is known, $f:\mathcal K\to\R$ is nonpositive and $G$--Lipschitz with $f\in\mathcal F^\gamma$, and every vector $q_f(\z)$ returned by the oracle is a support vector with $\norm{q_f(\z)}\le G$. We use two facts about $\psi_\gamma$ repeatedly: $\psi_\gamma(\sigma)\ge0$ for $\sigma\ge0$, and $\psi_\gamma(\sigma)=\sigma/\gamma$ for $\sigma<0$.

\subsection{Proof of Lemma~\ref{lem:violation-oracle}}
\label{apx:violation}

Fix $\x\in\mathcal K$, a vector $\v$ with $\norm\v\le G$, and $h>0$. Nonpositivity of $f$ is not used in this subsection. Define
\[
\ell(\y)=\ip{\v}{\y-\x},\qquad W(\y)=\ell(\y)-\gamma f(\y)\qquad(\y\in\mathcal K).
\]

\begin{proof}[Proof of Lemma~\ref{lem:violation-oracle}]
If $D=0$, then $\mathcal K=\{\x\}$ and $\y_\v=\x$ attains $\Phi_\x(\v)$. If $G=0$, then $\v=\BF0$ and $f$ is constant, so the maximand in \eqref{eq:violation} is the constant $(1-\gamma)f(\x)$, and again $\y_\v=\x$ attains $\Phi_\x(\v)$. In both cases one oracle call suffices. Assume $D,G>0$. For $\y\in\mathcal K$ we have $-\ip{\v}{\x-\y}=\ip{\v}{\y-\x}=\ell(\y)$, so
\begin{equation}\label{eq:po-violation-W}
f(\x)-\gamma f(\y)-\ip{\v}{\x-\y}=f(\x)+W(\y)\qquad\text{and}\qquad\Phi_\x(\v)=f(\x)+\max_{\mathcal K}W,
\end{equation}
where the maximum exists because $W$ is continuous and $\mathcal K$ is compact. Since $|f(\y)-f(\x)|\le G\norm{\y-\x}\le GD$ for $\y\in\mathcal K$,
\begin{equation}\label{eq:f-range}
f(\mathcal K)\subseteq[f(\x)-DG,\,f(\x)+DG].
\end{equation}

The procedure uses the grid
\[
A=\{f(\x)-DG+2jDG/J:\ j=0,\dots,J\},\qquad J=\max\{1,\lceil2\gamma DG/h\rceil\}.
\]
It contains both end points of the interval in \eqref{eq:f-range}. Since $2\gamma DG/h>0$, we have $J=\lceil2\gamma DG/h\rceil$, so $2\gamma DG/h\le J<2\gamma DG/h+1$. The first inequality shows that consecutive points are $2DG/J\le h/\gamma$ apart, and the second gives $|A|=J+1\le2+2\gamma DG/h$. For each level $a\in A$, let $N=\max\{1,\lceil(2DG/h)^2\rceil\}$ and $\eta=D/(G\sqrt N)$, start from the known feasible point $\z_1$, and for $j=1,\dots,N$ query the oracle at $\z_j$ and set
\begin{equation}\label{eq:level-updates}
b_j=\begin{cases}
q_f(\z_j),& f(\z_j)>a+h/\gamma,\\
-\v,& f(\z_j)\le a+h/\gamma,
\end{cases}
\qquad \z_{j+1}=\Pi_{\mathcal K}(\z_j-\eta b_j).
\end{equation}
We call $\z_j$ \emph{recorded} if the second case occurs, so every recorded point satisfies $f(\z_j)\le a+h/\gamma$. If some point is recorded, let $\y_a$ be a recorded point with the largest value of $\ell$. After all levels are processed, the procedure returns, among the points $\y_a$, a point $\y_\v$ with the largest value of $W$, which is known from the stored oracle values.

The analysis rests on the standard projected-subgradient potential. For $\uu\in\mathcal K$, projection onto $\mathcal K$ does not increase distances and fixes $\uu$, so
\[
\norm{\z_{j+1}-\uu}^2\le\norm{\z_j-\eta b_j-\uu}^2=\norm{\z_j-\uu}^2-2\eta\ip{b_j}{\z_j-\uu}+\eta^2\norm{b_j}^2.
\]
Since $\norm{q_f(\z_j)}\le G$ and $\norm\v\le G$, we have $\norm{b_j}\le G$. Rearranging and dividing by $2\eta$ gives
\begin{equation}\label{eq:one-step}
\ip{b_j}{\z_j-\uu}\le\frac{\norm{\z_j-\uu}^2-\norm{\z_{j+1}-\uu}^2}{2\eta}+\frac{\eta G^2}2.
\end{equation}
Summing \eqref{eq:one-step} over $j=1,\dots,N$, the squared distances telescope, and $\norm{\z_1-\uu}^2\le D^2$ gives
\begin{equation}\label{eq:po-telescoped}
\sum_{j=1}^N\ip{b_j}{\z_j-\uu}\le\frac{\norm{\z_1-\uu}^2-\norm{\z_{N+1}-\uu}^2}{2\eta}+\frac{N\eta G^2}2\le\frac{D^2}{2\eta}+\frac{N\eta G^2}2.
\end{equation}
Dividing \eqref{eq:po-telescoped} by $N$ and substituting $\eta=D/(G\sqrt N)$ makes each of the two resulting terms equal to $DG/(2\sqrt N)$, and since $\sqrt N\ge2DG/h$, this gives
\begin{equation}\label{eq:level-potential}
\frac1N\sum_{j=1}^N\ip{b_j}{\z_j-\uu}\le\frac{D^2}{2\eta N}+\frac{\eta G^2}2=\frac{DG}{\sqrt N}\le\frac h2.
\end{equation}

Fix a level $a$ with $S_a=\{\y\in\mathcal K:f(\y)\le a\}\ne\emptyset$. The set $S_a$ is a closed subset of the compact set $\mathcal K$, hence compact, so $\ell$ attains its maximum $p_a$ on $S_a$ at some $\uu$. We claim that some recorded point has $\ell\ge p_a-h$, so that $\y_a$ exists and
\begin{equation}\label{eq:level-guarantee}
f(\y_a)\le a+h/\gamma,\qquad \ell(\y_a)\ge p_a-h.
\end{equation}
Suppose instead that every recorded point has $\ell<p_a-h$ (this includes the case that no point is recorded). We show that $\ip{b_j}{\z_j-\uu}>h$ for every $j$; averaging over $j$ then contradicts \eqref{eq:level-potential}. If $\z_j$ is not recorded, then $f(\z_j)>a+h/\gamma$ and $f(\uu)\le a$, so
\begin{equation}\label{eq:po-gap}
\Delta=f(\z_j)-f(\uu)>(a+h/\gamma)-a=h/\gamma.
\end{equation}
With $\sigma=\ip{q_f(\z_j)}{\z_j-\uu}$, inequality \eqref{eq:gamma-mwc} at $\z_j$ with comparator $\uu$ gives $\psi_\gamma(-\sigma)\le f(\uu)-f(\z_j)=-\Delta<0$. Since $\psi_\gamma$ is negative only on negative arguments, $-\sigma<0$, so $\psi_\gamma(-\sigma)=-\sigma/\gamma$. Then $-\sigma/\gamma\le-\Delta$ and \eqref{eq:po-gap} give $\ip{b_j}{\z_j-\uu}=\sigma\ge\gamma\Delta>h$. If $\z_j$ is recorded, then $b_j=-\v$ and $\ip{b_j}{\z_j-\uu}=\ip{\v}{\uu-\z_j}=\ell(\uu)-\ell(\z_j)=p_a-\ell(\z_j)>h$. This proves the claim. The recorded point with $\ell\ge p_a-h$ satisfies $f\le a+h/\gamma$, and $\y_a$ has the largest $\ell$ among recorded points, which gives \eqref{eq:level-guarantee}.

Now let $\y^*$ maximize $W$ on $\mathcal K$, and let $a$ be the smallest point of $A$ with $a\ge f(\y^*)$. It exists because the largest point of $A$ is $f(\x)+DG\ge f(\y^*)$ by \eqref{eq:f-range}. If $a$ is the smallest point $f(\x)-DG$ of $A$, then $a\le f(\y^*)$ by \eqref{eq:f-range}. Otherwise the preceding point of $A$ is below $f(\y^*)$ and at most $h/\gamma$ below $a$. In both cases
\begin{equation}\label{eq:po-best-level}
f(\y^*)\le a\le f(\y^*)+h/\gamma.
\end{equation}
Since $\y^*\in S_a$, \eqref{eq:level-guarantee} applies, and $p_a\ge\ell(\y^*)$. Combining \eqref{eq:level-guarantee} with \eqref{eq:po-best-level} and $p_a\ge\ell(\y^*)$ gives
\[
f(\y_a)\le f(\y^*)+\frac{2h}\gamma,\qquad\ell(\y_a)\ge\ell(\y^*)-h,
\]
and therefore
\[
W(\y_a)=\ell(\y_a)-\gamma f(\y_a)\ge\ell(\y^*)-h-\gamma f(\y^*)-2h=W(\y^*)-3h.
\]
The returned point satisfies $W(\y_\v)\ge W(\y_a)$. By \eqref{eq:po-violation-W}, $V_\v=f(\x)+W(\y_\v)\ge f(\x)+W(\y^*)-3h=\Phi_\x(\v)-3h\ge\Phi_\x(\v)-4h$, and $V_\v\le\Phi_\x(\v)$ because $\y_\v\in\mathcal K$. This proves \eqref{eq:violation-accuracy}.

There are at most $2+2\gamma DG/h$ levels, each with $N\le1+(2DG/h)^2\le2+4(DG/h)^2$ iterations, and each iteration uses one oracle call and one projection. Together with the query at $\x$, this gives at most $1+2(2+2\gamma DG/h)(2+4(DG/h)^2)$ calls, which is \eqref{eq:candidate-complexity}.
\end{proof}

\subsection{Proof of Theorem~\ref{thm:query}(i)}
\label{apx:query-proof}

Parts (ii) and (iii) are proved in Appendices~\ref{apx:defective} and~\ref{apx:candidate-precision}. In this subsection $f$ is nonpositive and the oracle is exact.

\begin{proof}[Proof of Theorem~\ref{thm:query}(i)]
Write $q=q_f(\x)$. If $q=\BF0$, Algorithm~\ref{alg:query} returns $\BF0$, and \eqref{eq:gamma-mwc} gives $f(\y)-f(\x)\ge\psi_\gamma(0)=0$ for every $\y$, and since $f(\y)\le0$ and $\gamma\le1$,
\[
f(\x)-\gamma f(\y)=\bigl(f(\x)-f(\y)\bigr)+(1-\gamma)f(\y)\le0=\ip{\BF0}{\x-\y},
\]
which is \eqref{eq:approx-linearization} with $\varepsilon=0$. If $\gamma=1$, the algorithm returns $q$, and since $\psi_1(\sigma)=\sigma$, \eqref{eq:gamma-mwc} reads $f(\x)-f(\y)\le\ip{q}{\x-\y}$, which is again \eqref{eq:approx-linearization} with $\varepsilon=0$. Assume from now on that $q\ne\BF0$ and $\gamma<1$, so that every vector of $[\gamma q,q]$ is $sq$ for some $s\in[\gamma,1]$.

Let
\begin{equation}\label{eq:feasible-scalars}
T^*=\{s\in[\gamma,1]:\Phi_\x(sq)\le0\}.
\end{equation}
The vector $sq$ lies in \eqref{eq:g-feasible} exactly when $s\in T^*$, so Theorem~\ref{thm:linearization} gives $T^*\ne\emptyset$. We also use that $\Phi_\x$ is $D$--Lipschitz. For vectors $\v,\v'$, the two maximands in \eqref{eq:violation} differ at each $\y$ by $\ip{\v'-\v}{\x-\y}$, whose absolute value is at most $D\norm{\v-\v'}$, and maxima of two functions that differ by at most a constant differ by at most that constant:
\begin{equation}\label{eq:phi-lipschitz}
|\Phi_\x(\v)-\Phi_\x(\v')|\le D\norm{\v-\v'}.
\end{equation}

The algorithm maintains an interval $[\underline t,\overline t]\subseteq[\gamma,1]$ with
\begin{equation}\label{eq:bisection-invariant}
[\underline t,\overline t]\cap T^*\ne\emptyset,
\end{equation}
which holds initially with $[\underline t,\overline t]=[\gamma,1]$. Consider a candidate $\v=t\,q$ with $t=(\underline t+\overline t)/2$. Since $\norm\v\le\norm q\le G$, Lemma~\ref{lem:violation-oracle} applies with $h=\varepsilon/16$. If the candidate is accepted, that is, $V_\v\le\varepsilon/2$, then \eqref{eq:violation-accuracy} gives
\[
\Phi_\x(\v)\le V_\v+4h\le\frac\varepsilon2+\frac\varepsilon4<\varepsilon,
\]
which by \eqref{eq:violation} is \eqref{eq:approx-linearization} for $\widehat g=\v$, and $\v\in[\gamma q,q]$.

If the candidate is rejected, let $d=\ip{q}{\x-\y_\v}$. Since $\ip{\v}{\x-\y_\v}=t\,d$, the witness satisfies
\begin{equation}\label{eq:rejected-witness}
f(\x)-\gamma f(\y_\v)-t\,d=V_\v>\varepsilon/2,
\end{equation}
while every $s\in T^*$ satisfies, by the definition of $\Phi_\x$ with $\y=\y_\v$,
\begin{equation}\label{eq:feasible-at-witness}
f(\x)-\gamma f(\y_\v)-s\,d\le\Phi_\x(sq)\le0.
\end{equation}
Subtracting \eqref{eq:feasible-at-witness} from \eqref{eq:rejected-witness} gives
\begin{equation}\label{eq:po-side}
(s-t)\,d>\varepsilon/2>0\qquad(s\in T^*).
\end{equation}
By \eqref{eq:po-side}, if $d>0$, every $s\in T^*$ satisfies $s>t$, so $[t,\overline t]\cap T^*=[\underline t,\overline t]\cap T^*$, and setting $\underline t=t$ keeps \eqref{eq:bisection-invariant}. If $d<0$, every $s\in T^*$ satisfies $s<t$, and setting $\overline t=t$ keeps \eqref{eq:bisection-invariant}. The case $d=0$ cannot occur, since then $(s-t)d=0$ for the element $s$ provided by \eqref{eq:bisection-invariant}.

Each rejection halves $\overline t-\underline t$, so after $j$ rejections
\begin{equation}\label{eq:po-bisection-length}
\overline t-\underline t=(1-\gamma)2^{-j}\le2^{-j}.
\end{equation}
A test is performed after $j$ rejections only if $DG(\overline t-\underline t)>\varepsilon/4$, which by \eqref{eq:po-bisection-length} requires $DG\,2^{-j}>\varepsilon/4$, that is, $j<\log_2(4DG/\varepsilon)$. So the loop performs at most $\max\{0,\lceil\log_2(4DG/\varepsilon)\rceil\}=O(\log(2+DG/\varepsilon))$ tests. If it ends without acceptance, then $DG(\overline t-\underline t)\le\varepsilon/4$. Take $s\in[\underline t,\overline t]\cap T^*$, which exists by \eqref{eq:bisection-invariant}, and $\widehat g=\frac{\underline t+\overline t}2q$. Then $|\frac{\underline t+\overline t}2-s|\le\frac{\overline t-\underline t}2$ and $\norm q\le G$, so $\norm{\widehat g-sq}\le\frac{\overline t-\underline t}2G$. By \eqref{eq:phi-lipschitz} and $\Phi_\x(sq)\le0$,
\[
\Phi_\x(\widehat g)\le\Phi_\x(sq)+DG\,\frac{\overline t-\underline t}2\le0+\frac\varepsilon8<\varepsilon,
\]
which is \eqref{eq:approx-linearization}; also $\widehat g\in[\gamma q,q]$ because $\frac{\underline t+\overline t}2\in[\gamma,1]$.

By Lemma~\ref{lem:violation-oracle} with $h=\varepsilon/16$, each test uses $O\bigl((1+\gamma DG/\varepsilon)(1+(DG/\varepsilon)^2)\bigr)$ calls. Multiplying by the number of tests and adding the query at $\x$ gives \eqref{eq:query-complexity}. The parameter $\gamma\le1$ enters only through the factor $1+\gamma DG/\varepsilon$, so no inverse power of $\gamma$ appears.
\end{proof}

\subsection{Proof of Theorem~\ref{thm:query}(ii): inexact oracle}
\label{apx:defective}

In this subsection $f:\mathcal K\to\R$ is nonpositive and $G$--Lipschitz with $f\in\mathcal F^\gamma$. The oracle returns $\widehat f(\z)$ with $|\widehat f(\z)-f(\z)|\le\zeta$ and $\a_\z$ with $\norm{\a_\z}\le G$ satisfying \eqref{eq:noisy-oracle}. The first lemma replaces Theorem~\ref{thm:linearization} when the base vector has a support defect.

\begin{lemma}[Comparison from a defective base vector]\label{lem:defective-field}
Let $f:\mathcal K\to\R$ be continuous and nonpositive on a convex set $\mathcal K$ with $f\in\mathcal F^\gamma$, let $\x\in\mathcal K$, and suppose a vector $q$ satisfies $f(\y)-f(\x)\ge\psi_\gamma(\ip{q}{\y-\x})-\xi$ for all $\y\in\mathcal K$, with $\xi\ge0$. Then some $\v\in[\gamma q,q]$ satisfies
\[
f(\x)-\gamma f(\y)-\ip{\v}{\x-\y}\le\xi\qquad(\y\in\mathcal K).
\]
\end{lemma}
\begin{proof}
We follow the proof of Theorem~\ref{thm:linearization} with the defect carried along. Write $r=-f\ge0$, $p=-q$, $s_\y=\ip{p}{\y-\x}$, $c_\y=\gamma r(\y)-r(\x)$, and $c'_\y=c_\y-\xi$. Since $\ip{q}{\y-\x}=-s_\y$, the hypothesis reads $r(\y)-r(\x)\le-\psi_\gamma(-s_\y)+\xi$. Evaluating $\psi_\gamma$ on the two signs as in the derivation of \eqref{eq:reward-support} gives
\begin{equation}\label{eq:defective-support}
r(\y)-r(\x)\le\begin{cases}s_\y/\gamma+\xi,&s_\y\ge0,\\ \gamma s_\y+\xi,&s_\y<0.\end{cases}
\end{equation}
By \eqref{eq:scalar-constraint}, for a real $t$, $f(\x)-\gamma f(\y)-\ip{t\,q}{\x-\y}=c_\y-t\,s_\y$. So $t\,q$ satisfies the claim for $\y$ exactly when $c'_\y\le t\,s_\y$.

If $s_\y=0$, \eqref{eq:defective-support} gives $r(\y)\le r(\x)+\xi$, so
\begin{equation}\label{eq:po-def-zero}
c_\y\le\gamma r(\x)+\gamma\xi-r(\x)=-(1-\gamma)r(\x)+\gamma\xi\le\xi,
\end{equation}
that is, $c'_\y\le0$, and the constraint holds for every $t$. In particular, if $q=\BF0$, then $s_\y=0$ for every $\y$ and $\v=\BF0$ proves the lemma, so assume $q\ne\BF0$. If $s_\y>0$, \eqref{eq:defective-support} gives
\begin{equation}\label{eq:po-def-lower}
c_\y\le\gamma r(\x)+s_\y+\gamma\xi-r(\x)=s_\y-(1-\gamma)r(\x)+\gamma\xi\le s_\y+\xi,
\end{equation}
so $c'_\y\le s_\y$, and the constraint $t\ge c'_\y/s_\y$ is a lower bound that is at most $1$. If $s_\y<0$, \eqref{eq:defective-support} and $r(\y)\ge0$ give
\begin{equation}\label{eq:po-def-upper}
c_\y=\bigl(r(\y)-r(\x)\bigr)-(1-\gamma)r(\y)\le\gamma s_\y+\xi,
\end{equation}
so $c'_\y\le\gamma s_\y$, and dividing by $s_\y<0$ shows that the constraint $t\le c'_\y/s_\y$ is an upper bound that is at least $\gamma$.

Let $\y_+,\y_-\in\mathcal K$ with $s_+=s_{\y_+}>0>s_-=s_{\y_-}$, let $\lambda=-s_-/(s_+-s_-)$, and let $\z=\lambda\y_++(1-\lambda)\y_-$, so $s_\z=0$ as in the proof of Theorem~\ref{thm:linearization}. Then $r(\z)\le r(\x)+\xi$ by \eqref{eq:defective-support} with $s_\z=0$, and $r(\z)\ge\gamma(\lambda r(\y_+)+(1-\lambda)r(\y_-))$ by Lemma~\ref{lem:interpolation}. Hence
\begin{equation}\label{eq:po-def-pairing}
\lambda c_{\y_+}+(1-\lambda)c_{\y_-}=\gamma\bigl(\lambda r(\y_+)+(1-\lambda)r(\y_-)\bigr)-r(\x)\le r(\z)-r(\x)\le\xi,
\end{equation}
and subtracting $\xi=\lambda\xi+(1-\lambda)\xi$ gives $\lambda c'_{\y_+}+(1-\lambda)c'_{\y_-}\le0$. This is \eqref{eq:pairing} with $c'$ in place of $c$, and the computation that leads from \eqref{eq:pairing} to \eqref{eq:po-ratio-compare} gives $c'_{\y_+}/s_+\le c'_{\y_-}/s_-$. So every lower bound is at most every upper bound. As in \eqref{eq:po-tstar}, let $t^*=\max\{\gamma,\sup_{s_\y>0}c'_\y/s_\y\}$. By \eqref{eq:po-def-lower}, $t^*\in[\gamma,1]$. By \eqref{eq:po-def-upper} and the comparison of bounds, $t^*$ is at most every upper bound, and by its definition it is at least every lower bound. So $t^*$ satisfies every constraint, and the vector $\v=t^*q\in[\gamma q,q]$ proves the lemma.
\end{proof}

\begin{proof}[Proof of Theorem~\ref{thm:query}(ii)]
Query $\widehat f(\x)$ and $q=\a_\x$, let $h=\varepsilon/16$, and let $\ell$, $W$, and $\Phi_\x$ be as in Appendix~\ref{apx:violation}. Three cases are handled directly. If $\gamma=1$, return $q$: with $\psi_1(\sigma)=\sigma$, \eqref{eq:noisy-oracle} at $\z=\x$ gives $f(\x)-f(\y)\le\ip{q}{\x-\y}+\xi$. If $q=\BF0$, return $\BF0$: Lemma~\ref{lem:defective-field} gives $f(\x)-\gamma f(\y)\le\xi$. If $\zeta>DG$, return $q$: for $\y\in\mathcal K$, the Lipschitz bound and $f(\y)\le0$ give $f(\x)-\gamma f(\y)=(f(\x)-f(\y))+(1-\gamma)f(\y)\le GD$, and $-\ip{q}{\x-\y}\le\norm q\norm{\x-\y}\le GD$, so $f(\x)-\gamma f(\y)-\ip{q}{\x-\y}\le2DG<2\zeta\le(2+4\gamma)\zeta$. In each case \eqref{eq:noisy-comparison} holds. From now on $\gamma<1$, $q\ne\BF0$, and $\zeta\le DG$.

Each candidate $\v=tq$ with $t\in[\gamma,1]$ is evaluated by the procedure of Appendix~\ref{apx:violation} with two changes. The grid $A$ covers $[\widehat f(\x)-DG-\zeta,\ \widehat f(\x)+DG+\zeta]$ with gaps at most $h/\gamma$ and at most $2+2\gamma(DG+\zeta)/h\le2+4\gamma DG/h$ points; by \eqref{eq:f-range} and $|\widehat f(\x)-f(\x)|\le\zeta$, this interval contains $f(\mathcal K)$. The update \eqref{eq:level-updates} is replaced by
\begin{equation}\label{eq:noisy-level-update}
b_j=\begin{cases}
\a_{\z_j},&\widehat f(\z_j)>a+\xi+\zeta+h/\gamma,\\
-\v,&\widehat f(\z_j)\le a+\xi+\zeta+h/\gamma,
\end{cases}
\end{equation}
with the same $N$ and $\eta$. Points in the second case are recorded, and they satisfy $f(\z_j)\le\widehat f(\z_j)+\zeta\le a+\xi+2\zeta+h/\gamma$. Let $\y_a$ be a recorded point with the largest $\ell$.

Fix a level $a$ with $S_a\ne\emptyset$, and let $\uu\in S_a$ maximize $\ell$ on $S_a$, with $p_a=\ell(\uu)$. Inequality \eqref{eq:level-potential} uses only $\norm{b_j}\le G$, so it holds here. We claim that some recorded point has $\ell\ge p_a-h$. Otherwise every $j$ has $\ip{b_j}{\z_j-\uu}>h$, contradicting \eqref{eq:level-potential}. For a recorded $\z_j$ this is the same computation as in Appendix~\ref{apx:violation}. For an unrecorded $\z_j$, the value error and $f(\uu)\le a$ give
\begin{equation}\label{eq:po-noisy-gap}
f(\z_j)\ge\widehat f(\z_j)-\zeta>a+\xi+h/\gamma\ge f(\uu)+\xi+h/\gamma.
\end{equation}
With $\sigma=\ip{\a_{\z_j}}{\z_j-\uu}$, \eqref{eq:noisy-oracle} at $\z_j$ with $\y=\uu$ and \eqref{eq:po-noisy-gap} give $\psi_\gamma(-\sigma)\le f(\uu)-f(\z_j)+\xi<-h/\gamma<0$. Hence $-\sigma<0$, so $\psi_\gamma(-\sigma)=-\sigma/\gamma$, and $-\sigma/\gamma<-h/\gamma$ gives $\ip{b_j}{\z_j-\uu}=\sigma>h$. The recorded point with $\ell\ge p_a-h$ satisfies the bound on $f$ for recorded points, and $\y_a$ has the largest $\ell$ among recorded points. Hence
\begin{equation}\label{eq:noisy-level-guarantee}
f(\y_a)\le a+\xi+2\zeta+h/\gamma,\qquad \ell(\y_a)\ge p_a-h.
\end{equation}

Let $\y^*$ maximize $W$, and let $a$ be the smallest grid point with $a\ge f(\y^*)$. The grid covers an interval containing $f(\mathcal K)$ with gaps at most $h/\gamma$, so the argument for \eqref{eq:po-best-level} gives $a\le f(\y^*)+h/\gamma$. Since $\y^*\in S_a$ and $p_a\ge\ell(\y^*)$, \eqref{eq:noisy-level-guarantee} gives $f(\y_a)\le f(\y^*)+\xi+2\zeta+2h/\gamma$ and $\ell(\y_a)\ge\ell(\y^*)-h$. Therefore $W(\y_a)=\ell(\y_a)-\gamma f(\y_a)\ge\ell(\y^*)-h-\gamma f(\y^*)-\gamma\xi-2\gamma\zeta-2h$, that is,
\begin{equation}\label{eq:noisy-best-level}
W(\y_a)\ge W(\y^*)-3h-\gamma\xi-2\gamma\zeta.
\end{equation}
For each produced point the report $R(\y_a)=\widehat f(\x)-\gamma\widehat f(\y_a)-\ip{\v}{\x-\y_a}$ is computable. By \eqref{eq:po-violation-W}, the true value is $f(\x)-\gamma f(\y_a)-\ip{\v}{\x-\y_a}=f(\x)+W(\y_a)$, and the value errors give
\begin{equation}\label{eq:po-report-error}
\bigl|R(\y_a)-f(\x)-W(\y_a)\bigr|\le|\widehat f(\x)-f(\x)|+\gamma|\widehat f(\y_a)-f(\y_a)|\le(1+\gamma)\zeta.
\end{equation}
The procedure returns the produced point $\y_\v$ with the largest report and sets $V_\v=R(\y_\v)$. Using $V_\v\ge R(\y_a)$, \eqref{eq:po-report-error}, \eqref{eq:noisy-best-level}, and $\Phi_\x(\v)=f(\x)+W(\y^*)$ for the first inequality below, and \eqref{eq:po-report-error} at $\y_\v$ for the second,
\begin{equation}\label{eq:noisy-violation-estimate}
\Phi_\x(\v)-3h-\gamma\xi-(1+3\gamma)\zeta\le V_\v,\qquad
f(\x)+W(\y_\v)\ge V_\v-(1+\gamma)\zeta.
\end{equation}

The outer bisection uses $T^*_\xi=\{s\in[\gamma,1]:\Phi_\x(sq)\le\xi\}$, which is nonempty by Lemma~\ref{lem:defective-field}, and maintains $[a,b]\cap T^*_\xi\ne\emptyset$ from $[a,b]=[\gamma,1]$. A candidate $\v=tq$ with $t=(a+b)/2$ is accepted if $V_\v\le\xi+(1+\gamma)\zeta+\varepsilon/2$. Then the first inequality of \eqref{eq:noisy-violation-estimate} and $3h=3\varepsilon/16$ give
\[
\Phi_\x(\v)\le\xi+(1+\gamma)\zeta+\frac\varepsilon2+\frac{3\varepsilon}{16}+\gamma\xi+(1+3\gamma)\zeta
=\frac{11\varepsilon}{16}+(1+\gamma)\xi+(2+4\gamma)\zeta,
\]
which implies \eqref{eq:noisy-comparison} by \eqref{eq:violation}. If the candidate is rejected, then $V_\v>\xi+(1+\gamma)\zeta+\varepsilon/2$. With $d=\ip{q}{\x-\y_\v}$, the identity $f(\x)+W(\y_\v)=f(\x)-\gamma f(\y_\v)-t\,d$ and the second inequality of \eqref{eq:noisy-violation-estimate} give
\begin{equation}\label{eq:po-noisy-reject}
f(\x)-\gamma f(\y_\v)-t\,d\ge V_\v-(1+\gamma)\zeta>\xi+\varepsilon/2,
\end{equation}
while every $s\in T^*_\xi$ satisfies $f(\x)-\gamma f(\y_\v)-s\,d\le\Phi_\x(sq)\le\xi$. Subtracting this from \eqref{eq:po-noisy-reject} gives $(s-t)\,d>\varepsilon/2$, which is \eqref{eq:po-side} with $T^*_\xi$ in place of $T^*$, and the bisection step is justified exactly as in Appendix~\ref{apx:query-proof}. If the loop ends with $DG(b-a)\le\varepsilon/4$, then for $s\in[a,b]\cap T^*_\xi$, \eqref{eq:phi-lipschitz} gives $\Phi_\x(\frac{a+b}2q)\le\Phi_\x(sq)+DG(b-a)/2\le\xi+\varepsilon/8$, which implies \eqref{eq:noisy-comparison}. Every output lies in $[\gamma q,q]$. Each test uses at most $2+4\gamma DG/h$ levels with $N$ iterations each, and the count of tests in Appendix~\ref{apx:query-proof} gives $O(\log(2+DG/\varepsilon))$ tests, which gives \eqref{eq:query-complexity}.
\end{proof}

\subsection{Proof of Theorem~\ref{thm:query}(iii): bit model}
\label{apx:candidate-precision}

\begin{proof}[Proof of Theorem~\ref{thm:query}(iii)]
First suppose that $\gamma$, $D$, and $G$ are positive rationals and that a bound $|f|\le B$ on $\mathcal K$ is known. Let $\xi_0,\zeta_0\ge0$ be the defects of the supplied oracle in the sense of Theorem~\ref{thm:query}(ii); for an exact oracle, $\xi_0=\zeta_0=0$. The rational implementation runs the algorithm of part (ii) with tolerance $\varepsilon/2$ and with the rounding described next.

Values are rounded to rationals within $\varepsilon/48$, so the rounded values have value error at most
\begin{equation}\label{eq:rounded-value-error}
\zeta=\zeta_0+\varepsilon/48.
\end{equation}
To round vectors, let $\nu=\gamma\varepsilon/(16D)$. Approximate $\a_\z$ by a rational $\a$ with $\norm{\a-\a_\z}\le\nu/4$, and compute a rational $r_\a$ with $\norm\a\le r_\a\le\norm\a+\nu/2$. If $r_\a\le G$, keep $\a$. Otherwise replace $\a$ by $\a'=(G/r_\a)\a$, which has $\norm{\a'}=G\norm\a/r_\a\le G$ and
\[
\norm{\a'-\a}=\norm\a\Bigl(1-\frac G{r_\a}\Bigr)\le r_\a-G\le\norm\a+\frac\nu2-G\le\norm{\a_\z}+\frac\nu4+\frac\nu2-G\le\frac{3\nu}4,
\]
using $\norm\a\le r_\a$ and $\norm{\a_\z}\le G$, and the triangle inequality gives $\norm{\a'-\a_\z}\le\norm{\a'-\a}+\norm{\a-\a_\z}\le3\nu/4+\nu/4=\nu$. In both cases the stored vector, again called $\a$, has $\norm\a\le G$ and $\norm{\a-\a_\z}\le\nu$. For $\y\in\mathcal K$, $|\ip{\a-\a_\z}{\y-\z}|\le\nu D$, and $\psi_\gamma$ is $(1/\gamma)$--Lipschitz because it is piecewise linear with slopes $\gamma$ and $1/\gamma$. So
\begin{equation}\label{eq:po-rounded-psi}
\psi_\gamma(\ip{\a_\z}{\y-\z})\ge\psi_\gamma(\ip{\a}{\y-\z})-\nu D/\gamma,
\end{equation}
and \eqref{eq:noisy-oracle} for $\a_\z$ with defect $\xi_0$, combined with \eqref{eq:po-rounded-psi}, yields \eqref{eq:noisy-oracle} for $\a$ with defect
\begin{equation}\label{eq:rounded-defect}
\xi=\xi_0+D\nu/\gamma=\xi_0+\varepsilon/16.
\end{equation}

Projections are computed approximately: $\z_{j+1}$ is a rational point of $\mathcal K$ within distance $\kappa$ of $P_j=\Pi_{\mathcal K}(\z_j-\eta b_j)$. For $\uu\in\mathcal K$, $\norm{P_j-\uu}\le D$ because both points lie in $\mathcal K$, so
\begin{equation}\label{eq:po-approx-proj}
\norm{\z_{j+1}-\uu}^2\le(\norm{P_j-\uu}+\kappa)^2\le\norm{P_j-\uu}^2+2D\kappa+\kappa^2.
\end{equation}
In the derivation of \eqref{eq:one-step}, the bound $\norm{P_j-\uu}^2\le\norm{\z_j-\eta b_j-\uu}^2$ still holds, so \eqref{eq:po-approx-proj} adds $(2D\kappa+\kappa^2)/(2\eta)$ to the right side of \eqref{eq:one-step}. Averaging over $j$ as in \eqref{eq:po-telescoped} and \eqref{eq:level-potential} gives
\begin{equation}\label{eq:level-projection-precision}
\frac1N\sum_{j=1}^N\ip{b_j}{\z_j-\uu}\le\frac{D^2}{2\eta N}+\frac{\eta G^2}2+\frac{2D\kappa+\kappa^2}{2\eta}.
\end{equation}
Choose $N=\lceil\max\{1,(16DG/h)^2\}\rceil$, a rational $\eta$ with $D/(2G\sqrt N)\le\eta\le D/(G\sqrt N)$, and $\kappa\le\min\{D,\eta h/(64D)\}$. Since $\sqrt N\ge16DG/h$, the three terms on the right of \eqref{eq:level-projection-precision} satisfy
\begin{equation}\label{eq:po-three-terms}
\frac{D^2}{2\eta N}\le\frac{DG}{\sqrt N}\le\frac h{16},\qquad\frac{\eta G^2}2\le\frac{DG}{2\sqrt N}\le\frac h{32},\qquad\frac{2D\kappa+\kappa^2}{2\eta}\le\frac{3D\kappa}{2\eta}\le\frac{3h}{128},
\end{equation}
where the first bound uses $\eta\ge D/(2G\sqrt N)$, the second uses $\eta\le D/(G\sqrt N)$, and the third uses $\kappa^2\le D\kappa$ and $\kappa\le\eta h/(64D)$. By \eqref{eq:po-three-terms}, the right side of \eqref{eq:level-projection-precision} is at most $(8+4+3)h/128=15h/128<h/2$, so \eqref{eq:level-potential} holds with approximate projections and the argument for \eqref{eq:noisy-level-guarantee} applies unchanged. The number of iterations changes by an absolute factor and has no inverse power of $\gamma$.

By \eqref{eq:rounded-value-error} and \eqref{eq:rounded-defect}, part (ii) applies with tolerance $\varepsilon/2$, and \eqref{eq:noisy-comparison} gives the comparison error
\[
\frac\varepsilon2+(1+\gamma)\Bigl(\xi_0+\frac\varepsilon{16}\Bigr)+(2+4\gamma)\Bigl(\zeta_0+\frac\varepsilon{48}\Bigr)
\le\frac{3\varepsilon}4+(1+\gamma)\xi_0+(2+4\gamma)\zeta_0,
\]
using $1+\gamma\le2$ and $2+4\gamma\le6$, which give $(1+\gamma)\varepsilon/16+(2+4\gamma)\varepsilon/48\le\varepsilon/8+\varepsilon/8$. The right side is at most $\varepsilon+(1+\gamma)\xi_0+(2+4\gamma)\zeta_0$, so \eqref{eq:noisy-comparison} holds with the defects $\xi_0,\zeta_0$ of the supplied oracle. The output is a rational multiple $t\,\a$ of the stored base vector with $t\in[\gamma,1]$, so it lies in $[\gamma\a,\a]$ and has norm at most $G$. All grid points, thresholds, step sizes, and reports are obtained from the input by polynomially many arithmetic operations, every oracle input is a stored rational point, and the accuracies $\varepsilon/48$, $\nu$, and $\kappa$ have bit length polynomial in $\log(2+1/\varepsilon)+\log(1/\gamma)+\log(2+D+D^{-1}+G+G^{-1}+B)+\log d$. With $1/\varepsilon$, $D$, and $G$ polynomially bounded, the number of calls \eqref{eq:query-complexity} is polynomial, so the running time is polynomial.

Finally, let $\gamma$ be given by certified rational approximations. If the known bound is $B=0$, then $f\equiv0$ and $\BF0$ is exact. Otherwise choose a rational $\gamma'\in[\gamma/2,\gamma]$ with $\gamma-\gamma'\le\varepsilon/(4B)$. By Lemma~\ref{lem:psi_basics}(7) when $\gamma'<\gamma$, and trivially when $\gamma'=\gamma$,
\begin{equation}\label{eq:po-psi-monotone}
\psi_{\gamma'}(z)\le\psi_\gamma(z)\qquad(z\in\R).
\end{equation}
By \eqref{eq:po-psi-monotone}, inequality \eqref{eq:gamma-mwc} for $\gamma$ implies \eqref{eq:gamma-mwc} for $\gamma'$, so $f\in\mathcal F^{\gamma'}$, and \eqref{eq:noisy-oracle} also holds with $\gamma'$. Running the rational construction at $\gamma'$ with tolerance $3\varepsilon/4$ in place of $\varepsilon$, the bound above gives comparison error at most $\frac34\cdot\frac{3\varepsilon}4+(1+\gamma')\xi_0+(2+4\gamma')\zeta_0$, so
\begin{equation}\label{eq:po-gamma-prime}
f(\x)-\gamma'f(\y)\le\ip{\widehat g}{\x-\y}+\frac{3\varepsilon}4+(1+\gamma')\xi_0+(2+4\gamma')\zeta_0\qquad(\y\in\mathcal K).
\end{equation}
Since $-f(\y)\le B$ and $\gamma-\gamma'\le\varepsilon/(4B)$,
\[
f(\x)-\gamma f(\y)=f(\x)-\gamma'f(\y)+(\gamma-\gamma')(-f(\y))\le f(\x)-\gamma'f(\y)+\frac\varepsilon4.
\]
Adding this to \eqref{eq:po-gamma-prime} and using $\gamma'\le\gamma$ gives \eqref{eq:noisy-comparison} at the original $\gamma$. For $B>0$ the segment guarantee is $[\gamma'\a,\a]$, while the norm bound and comparison hold at the original parameter; the separate $B=0$ branch returns zero and needs no segment assertion.
\end{proof}

\subsection{Proof of Corollary~\ref{cor:ombq}}
\label{apx:linear-regret}

For an interval $I=[s,e]$ and comparators $\uu_s,\dots,\uu_e$, let $P_I(\uu)=\sum_{t=s+1}^e\norm{\uu_t-\uu_{t-1}}$.

\begin{lemma}[Projected online gradient descent]\label{lem:ogd}
Let $\norm{\a_t}\le G$ and $\x_{t+1}=\Pi_{\mathcal K}(\x_t-\eta\a_t)$ with $\eta>0$. For every interval $I=[s,e]$ and $\uu_s,\dots,\uu_e\in\mathcal K$,
\begin{equation}\label{eq:ogd-potential}
\sum_{t\in I}\ip{\a_t}{\x_t-\uu_t}\le\frac{D^2+2DP_I(\uu)}{2\eta}+\frac{\eta G^2|I|}2.
\end{equation}
\end{lemma}
\begin{proof}
The derivation of \eqref{eq:one-step} gives, for each $t$,
\begin{equation}\label{eq:ogd-one-step}
\ip{\a_t}{\x_t-\uu_t}\le\frac{\norm{\x_t-\uu_t}^2-\norm{\x_{t+1}-\uu_t}^2}{2\eta}+\frac{\eta G^2}2.
\end{equation}
Summing the numerators over $t\in I$ and regrouping,
\[
\sum_{t=s}^e\bigl(\norm{\x_t-\uu_t}^2-\norm{\x_{t+1}-\uu_t}^2\bigr)
=\norm{\x_s-\uu_s}^2-\norm{\x_{e+1}-\uu_e}^2+\sum_{t=s+1}^{e}\bigl(\norm{\x_t-\uu_t}^2-\norm{\x_t-\uu_{t-1}}^2\bigr).
\]
The first term is at most $D^2$ and the second is nonpositive. For each $t$ in the last sum, with $\BF m=\x_t-\uu_t$ and $\BF m'=\x_t-\uu_{t-1}$,
\[
\norm{\BF m}^2-\norm{\BF m'}^2=(\norm{\BF m}-\norm{\BF m'})(\norm{\BF m}+\norm{\BF m'})\le\norm{\BF m-\BF m'}\cdot2D=2D\norm{\uu_t-\uu_{t-1}}.
\]
So the numerators sum to at most $D^2+2DP_I(\uu)$, and summing \eqref{eq:ogd-one-step} over $t\in I$ gives \eqref{eq:ogd-potential}.
\end{proof}

\begin{proof}[Proof of Corollary~\ref{cor:ombq}]
In round $t$, Theorem~\ref{thm:query}(ii) applied to $f_t$ at $\x_t$, with tolerance $\varepsilon_t$, defect $\xi_t$, and value error $\zeta_t$, gives for every $\uu_t\in\mathcal K$
\begin{equation}\label{eq:round-comparison}
f_t(\x_t)-\gamma f_t(\uu_t)\le\ip{\widehat g_t}{\x_t-\uu_t}+\mathrm{err}_t,
\end{equation}
where an exact oracle is the case $\xi_t=\zeta_t=0$. The returned vector satisfies $\norm{\widehat g_t}\le G$ by Theorem~\ref{thm:query}, including its irrational-parameter implementation, and Algorithm~\ref{alg:OMBQ} feeds $\a_t=\widehat g_t$ to $\mathcal A$, which played $\x_t$. Summing \eqref{eq:round-comparison} over $t\in I$ and applying \eqref{eq:linear-bound} gives
\[
\Reg_\gamma(I;\uu_{s:e})\le\sum_{t\in I}\ip{\widehat g_t}{\x_t-\uu_t}+\mathrm{Err}(I)\le\mathcal B_{\mathcal A}(I;\uu_{s:e})+\mathrm{Err}(I),
\]
which is \eqref{eq:ombq-transfer}. For static regret, the function $\sum_{t\in I}f_t$ is continuous on the compact set $\mathcal K$ and has a minimizer $\uu$, and
\begin{equation}\label{eq:static-as-fixed}
\SReg_\gamma(I)=\sum_{t\in I}f_t(\x_t)-\gamma\sum_{t\in I}f_t(\uu)=\Reg_\gamma(I;\uu,\dots,\uu).
\end{equation}

A fixed comparator has $P_{[1,T]}(\uu)=0$, so Lemma~\ref{lem:ogd} with $I=[1,T]$ and $\eta=D/(G\sqrt T)$ gives
\begin{equation}\label{eq:po-ogd-static}
\mathcal B_{\mathcal A}=\frac{D^2}{2\eta}+\frac{\eta G^2T}2=\frac{DG\sqrt T}2+\frac{DG\sqrt T}2=DG\sqrt T.
\end{equation}
Then \eqref{eq:static-as-fixed}, \eqref{eq:ombq-transfer}, and \eqref{eq:po-ogd-static} give $\SReg_\gamma(T)\le DG\sqrt T+\mathrm{Err}([1,T])$.

If $\mathrm{err}_t\le c/\sqrt t$, then $\mathrm{Err}(I)\le c\sum_{t=s}^et^{-1/2}$. For $t\ge1$, $\sqrt t-\sqrt{t-1}=1/(\sqrt t+\sqrt{t-1})\ge1/(2\sqrt t)$ because $\sqrt{t-1}\le\sqrt t$, so $t^{-1/2}\le2(\sqrt t-\sqrt{t-1})$, and summing over $t=s,\dots,e$ telescopes:
\[
\mathrm{Err}(I)\le c\sum_{t=s}^et^{-1/2}\le2c\bigl(\sqrt e-\sqrt{s-1}\bigr).
\]
Since $(\sqrt{s-1}+\sqrt{e-s+1})^2=e+2\sqrt{(s-1)(e-s+1)}\ge e$, we have $\sqrt e-\sqrt{s-1}\le\sqrt{e-s+1}=\sqrt{|I|}$, so $\mathrm{Err}(I)\le2c\sqrt{|I|}$, which is \eqref{eq:interval-error-condition}. With an exact oracle and $\varepsilon_t=c/\sqrt t$, $\mathrm{err}_t=\varepsilon_t$. Finally, round $t$ calls the procedure of Theorem~\ref{thm:query} once, with the number of calls in \eqref{eq:query-complexity} for $\varepsilon=\varepsilon_t$, and performs one update of $\mathcal A$. With inverse-polynomial tolerances, Theorem~\ref{thm:query}(iii) and the rational implementation of projected online gradient descent below give polynomial time per round.
\end{proof}

\paragraph{Rational implementation of the linear learners.}
For static regret, choose one rational step size
\[
\frac{D}{G\sqrt T}\le\eta\le\frac{5D}{4G\sqrt T}.
\]
Certified approximation of $\sqrt T$ supplies such a step size with polynomial encoding length, and the exact-projection potential is then at most $(9/8)DG\sqrt T$. If the update in round $t$ is a feasible rational point within distance $\kappa_t$ of the exact projection, the squared-distance estimate in Appendix~\ref{apx:candidate-precision} adds at most
\[
\frac{2D\kappa_t+\kappa_t^2}{2\eta}
\le\frac{3D\kappa_t}{2\eta}
\]
to the one-round bound whenever $\kappa_t\le D$. Taking
$\kappa_t\le\min\{D,\eta G/(4T^2)\}$ makes the total addition on every interval at most $3DG/(8T)$. Thus the implemented static linear regret is at most $(3/2)DG\sqrt T$, and comparison errors at most $DG/(2\sqrt T)$ per round give the stated $2DG\sqrt T$ bound. The constant-step potential for moving comparators admits the same rational approximation and projection-error accounting. For a strongly adaptive learner, rational convex combinations and the linear experts' errors can likewise be charged to its explicit linear-regret bound.

\section{Proofs for Section~\ref{sec:hardness}}
\label{apx:hardness}

\subsection{Proof of Theorem~\ref{thm:online-gap-hardness-criterion}}
\label{apx:hardness-proof}

\begin{proof}[Proof of Theorem~\ref{thm:online-gap-hardness-criterion}]
Fix $\alpha\in(\beta,1]$ and apply
Definition~\ref{def:online-gap-property} to a SAT formula of size $n$.
Write $N$ for the resulting encoding length, which does not depend on the horizon, and
$\Delta=\alpha-\tau>0$.  We may assume $P(N)\ge1$.
For a horizon $T$ chosen below, draw indices $J_1,\ldots,J_T$
independently and uniformly from $I$, independently of the learner's
random tape $\omega$.  In round $t$, obtain the learner's rational
action $\x_t$, then reveal $f_{J_t}$ and all the prescribed feedback.
Accept the SAT formula if $Q(\x_t)>\tau$ for some $t\le T$.
The score is computed exactly at rational points, so this test is
exact.  On an unsatisfiable formula, item~(iv) of
Definition~\ref{def:online-gap-property} gives $Q(\x_t)\le\tau$ for
every $t$, so an unsatisfiable formula is never accepted.

If the guarantee is stated for deterministic loss sequences, condition
on the complete index sequence.  The learner's tape remains independent
of it, so the assumed regret bound applies; averaging over the index
sequence gives the corresponding expected or high-probability bound
for this simulation.  If the guarantee is given directly for the
independent-sampling model, this conditioning step is unnecessary.

Suppose the formula is satisfiable, and fix $\x^*$ with $Q(\x^*)=1$,
which exists by item~(iii) of Definition~\ref{def:online-gap-property}.
Put
\[
Z_t=f_{J_t}(\x_t)-\alpha f_{J_t}(\x^*),\qquad
\mathcal A_t=\sigma(\omega,J_1,\ldots,J_t).
\]
The comparator $\x^*$ is fixed, so
$\sum_t f_{J_t}(\x^*)\ge\min_{\x\in\mathcal K}\sum_t f_{J_t}(\x)$.
Multiplying this by $-\alpha<0$ and adding $\sum_t f_{J_t}(\x_t)$
gives, by \eqref{eq:static},
\begin{equation}
\label{eq:hardness-Z-regret}
\sum_{t=1}^T Z_t\le\SReg_\alpha(T).
\end{equation}
The action $\x_t$ is a function of $\omega$ and of the feedback in
rounds $1,\ldots,t-1$, so it is $\mathcal A_{t-1}$-measurable, while
$J_t$ is uniform on $I$ and independent of $\mathcal A_{t-1}$.
Hence, applying \eqref{eq:abstract-hardness-average-loss} at $\x_t$
and at $\x^*$,
\begin{equation}
\label{eq:hardness-conditional-mean}
\mu_t:=\mathbb E[Z_t\mid\mathcal A_{t-1}]
=\frac1{|I|}\sum_{i\in I}f_i(\x_t)-\frac{\alpha}{|I|}\sum_{i\in I}f_i(\x^*)
=-\frac{Q(\x_t)}{\Lambda}+\frac{\alpha Q(\x^*)}{\Lambda}
=\frac{\alpha-Q(\x_t)}{\Lambda}.
\end{equation}
Summing \eqref{eq:hardness-conditional-mean} over $t$ gives
\begin{equation}
\label{eq:ph-sum-mu}
\sum_{t=1}^T\mu_t
=\frac1\Lambda\Bigl(\alpha T-\sum_{t=1}^T Q(\x_t)\Bigr).
\end{equation}

\paragraph{Expected regret.}
By the tower property, $\mathbb E Z_t=\mathbb E\mu_t$ for every $t$.
Taking expectations in~\eqref{eq:hardness-Z-regret} and using
\eqref{eq:ph-sum-mu} and \eqref{eq:hardness-expected-regret-assumption},
\[
\frac1\Lambda\Bigl(\alpha T-\mathbb E\sum_{t=1}^T Q(\x_t)\Bigr)
=\mathbb E\sum_{t=1}^T Z_t
\le\mathbb E[\SReg_\alpha(T)]
\le P(N)T^{1-\rho}.
\]
Multiplying by $\Lambda/T$ and rearranging yields
\begin{equation}
\label{eq:ph-expected-average}
\mathbb E\!\left[\frac1T\sum_{t=1}^T Q(\x_t)\right]
\ge\alpha-\Lambda P(N)T^{-\rho}.
\end{equation}
Choose a polynomially bounded integer horizon satisfying
\begin{equation}
\label{eq:hardness-expectation-horizon}
T\ge\left(\frac{2\Lambda P(N)}{\Delta}\right)^{1/\rho}.
\end{equation}
Such a horizon exists because $N$, $\Lambda$, and $\Delta^{-1}$ are
polynomially bounded in $n$, and $\rho$ is fixed.  Any integer
polynomial in $n$ that exceeds the right-hand side can be used, so no
irrational power has to be computed.
By \eqref{eq:hardness-expectation-horizon},
$\Lambda P(N)T^{-\rho}\le\Delta/2$.  Let
$X=\max_{t\le T}Q(\x_t)\in[0,1]$.  Since $X$ is at least the average
score, \eqref{eq:ph-expected-average} gives
\begin{equation}
\label{eq:ph-EX-lower}
\mathbb EX\ge\alpha-\frac\Delta2=\frac{\alpha+\tau}2.
\end{equation}
On the other hand, $X\le\tau$ on the event $\{X\le\tau\}$ and $X\le1$
on its complement, so
\begin{equation}
\label{eq:ph-EX-upper}
\mathbb EX\le\tau\,\mathbb P(X\le\tau)+\mathbb P(X>\tau)
=\tau+(1-\tau)\mathbb P(X>\tau).
\end{equation}
Since $\tau<\alpha\le1$, we have $1-\tau>0$.  Combining
\eqref{eq:ph-EX-lower} with \eqref{eq:ph-EX-upper} and dividing by
$1-\tau$ gives
\begin{equation}
\label{eq:hardness-success-probability}
\mathbb P(X>\tau)\ge
\frac{\alpha-\tau}{2(1-\tau)}=:q_0.
\end{equation}
Since $1-\tau\le1$, we have $q_0\ge\Delta/2$, so $q_0$ is
inverse-polynomial.  Repeat the simulation independently $k$ times
with $k\ge(\log4)/q_0$, and accept if any run accepts.  All runs fail
with probability at most $(1-q_0)^k\le e^{-q_0k}\le1/4$, so this
procedure accepts a satisfiable formula with probability at least
$3/4$.  Since $1/q_0\le2/\Delta$, an integer polynomial upper bound for
$k$ can again be used.  The argument uses only $Q\in[0,1]$; it does not
need $\SReg_\alpha(T)\ge0$.

\paragraph{High-probability regret.}
Let $D_t=\mu_t-Z_t$.  Since $Z_t$ is $\mathcal A_t$-measurable and
$\mu_t=\mathbb E[Z_t\mid\mathcal A_{t-1}]$, the sequence $(D_t)$ is a
martingale difference sequence with respect to $(\mathcal A_t)$.
Each loss takes values in $[-1,0]$ and $\alpha\in(0,1]$, so
$Z_t\in[-1,\alpha]\subseteq[-1,1]$; hence $\mu_t\in[-1,1]$ and
$|D_t|\le2$.  The Azuma--Hoeffding inequality with increments bounded
by $2$ gives $\mathbb P\{\sum_t D_t\ge\lambda\}\le\exp(-\lambda^2/(8T))$
for $\lambda>0$.  The choice $\lambda=2\sqrt{2T\log9}$ has
$\lambda^2=8T\log9$, so
\begin{equation}
\label{eq:ph-azuma}
\mathbb P\!\left\{\sum_{t=1}^T D_t
\ge2\sqrt{2T\log9}\right\}\le\frac19.
\end{equation}
Let $\mathcal E_1$ be the event in
\eqref{eq:hardness-high-prob-regret-assumption}, and let $\mathcal E_2$
be the complement of the event in \eqref{eq:ph-azuma}.  By the union
bound, $\mathbb P(\mathcal E_1\cap\mathcal E_2)\ge1-1/9-1/9=7/9$.
On $\mathcal E_1\cap\mathcal E_2$, the identity
$\sum_t\mu_t=\sum_t Z_t+\sum_t D_t$ and \eqref{eq:hardness-Z-regret}
give
\[
\sum_{t=1}^T\mu_t
\le\SReg_\alpha(T)+\sum_{t=1}^T D_t
< P(N)T^{1-\rho}+2\sqrt{2T\log9}.
\]
Substituting \eqref{eq:ph-sum-mu}, multiplying by $\Lambda/T$, and
rearranging shows that on $\mathcal E_1\cap\mathcal E_2$
\begin{equation}
\label{eq:ph-hp-average}
\frac1T\sum_{t=1}^T Q(\x_t)
\ge\alpha-\Lambda P(N)T^{-\rho}
-2\Lambda\sqrt{\frac{2\log9}{T}}.
\end{equation}
Choose an integer polynomial horizon satisfying
\begin{equation}
\label{eq:hardness-probability-horizon}
T\ge\left(\frac{4\Lambda P(N)}{\Delta}\right)^{1/\rho},
\qquad
T\ge2\log9\left(\frac{8\Lambda}{\Delta}\right)^2.
\end{equation}
The first condition gives $\Lambda P(N)T^{-\rho}\le\Delta/4$, and the
second gives $\sqrt{2\log9/T}\le\Delta/(8\Lambda)$, hence
$2\Lambda\sqrt{2\log9/T}\le\Delta/4$.  By \eqref{eq:ph-hp-average},
the average score on $\mathcal E_1\cap\mathcal E_2$ is at least
$\alpha-\Delta/2=(\alpha+\tau)/2>\tau$.  The largest score is at least
the average, so some $Q(\x_t)$ exceeds $\tau$ and the rule accepts.
One run therefore accepts with probability at least $7/9>3/4$; put
$k=1$.

\paragraph{Finite sampling and running time.}
Let $M=|I|$.  If $M=1$, the unique index is deterministic.  Otherwise,
let $\ell=\lceil\log_2 M\rceil$, so that $2^{\ell-1}<M\le2^\ell$.
Sample uniformly from the integers in $[0,2^\ell-1]$ using $\ell$
random bits, and retain the sample if it is below $M$.  A retained
sample is uniform on $\{0,\ldots,M-1\}$, and each attempt succeeds with
probability $M/2^\ell>1/2$.  Cap the number of attempts for each of
the at most $kT$ draws at $c=\lceil\log_2(16kT)\rceil$, and reject the
SAT formula if all $c$ attempts for some draw fail.  A single draw
fails with probability at most $2^{-c}\le1/(16kT)$, so by the union
bound some draw fails with probability at most $1/16$.  Run on the same
random bits, the capped sampler and uncapped independent sampling
produce the same indices unless some draw fails, so the acceptance
probabilities differ by at most $1/16$.  Thus the bounded-time
algorithm accepts a satisfiable formula with probability at least
$3/4-1/16>1/2$, and it never accepts an unsatisfiable formula because
acceptance still requires an iterate with $Q(\x_t)>\tau$.

The simulation, feedback, and exact score tests all take polynomial
time.  This proves $\mathrm{SAT}\in\RP$, hence $\NP\subseteq\RP$;
the reverse inclusion always holds.
\end{proof}

\subsection{Proof of Theorem~\ref{thm:direct-hard-profile-criterion}}
\label{apx:hard-profile-criterion}

The proof reduces from Label Cover, using the gap theorem below and a rounding lemma for the projected masses.

A Label Cover instance
$\Gamma=(V_L,V_R,E,\Sigma_L,\Sigma_R,(\pi_e)_{e\in E})$ has a bipartite
graph with no isolated vertices and projection maps
$\pi_e:\Sigma_L\to\Sigma_R$.  A labeling assigns $a_u\in\Sigma_L$
to each $u\in V_L$ and $b_v\in\Sigma_R$ to each $v\in V_R$.
An edge $e=(u,v)$ is satisfied when $\pi_e(a_u)=b_v$.
The value $\operatorname{val}(\Gamma)$ is the largest fraction of
satisfied edges, with parallel edges counted separately.

\begin{theorem}[Fixed-gap Label Cover
{\citep{moshkovitz2010two,moshkovitz2015projection}}]
\label{thm:label-cover}
\label{thm:hardness-label-cover-gap}
For every fixed rational $\delta\in(0,1)$ there is a deterministic
polynomial-time reduction from SAT to Label Cover that has value
$1$ on satisfiable formulas and value at most $\delta$ on
unsatisfiable formulas.  The alphabets and the two degrees of the
biregular graph are bounded by constants depending only on $\delta$.
\end{theorem}

The statement follows by specializing the soundness parameter in
Moshkovitz~\citep[Theorem~2.1]{moshkovitz2015projection} to a fixed constant;
it relies on a proved Label Cover theorem, not on the Projection Games
Conjecture.

Write $k_L=|\Sigma_L|$, $k_R=|\Sigma_R|$, and let
\begin{equation}
\label{eq:hardness-feasible-set}
\mathcal K_\Gamma
=\prod_{u\in V_L}\left\{\x_u\in\mathbb R_+^{\Sigma_L}:
\sum_{a\in\Sigma_L}x_{u,a}\le1\right\}.
\end{equation}
For a right vertex of degree $\ell_v$, set $w_v=\ell_v/|E|$ and
\begin{equation}
\label{eq:hardness-projected-mass}
z_{v,b}(\x)
=\frac1{\ell_v}\sum_{e=(u,v)\in E}
\sum_{a:\pi_e(a)=b}x_{u,a}.
\end{equation}
For every feasible point, $z_{v,b}\ge0$ and
$\sum_b z_{v,b}\le1$; also $\sum_v w_v=1$.

\begin{lemma}[Pointwise rounding and the profile gap]
\label{lem:gap}
\label{lem:hardness-label-cover-rounding}
Given a rational $\x\in\mathcal K_\Gamma$, one can compute in polynomial
time a labeling of value $L(\x)$ such that
\begin{equation}
\label{eq:hardness-rounding-dominates-masses}
L(\x)\ge\sum_v w_v\max_b z_{v,b}(\x).
\end{equation}
If $H(0)=0$, $H(1)=1$, $0\le H(s)\le s$, and
$H(s)\le as+bs^2$ with $a\ge0$ and $b>0$, then
\begin{equation}
\label{eq:hard-profile-score}
Q_\Gamma^H(\x):=\sum_v w_v\sum_b H(z_{v,b}(\x))\in[0,1]
\end{equation}
and
\begin{equation}
\label{eq:hard-profile-rounding-bound}
Q_\Gamma^H(\x)\le a+bL(\x)
\le a+b\operatorname{val}(\Gamma).
\end{equation}
If $\Gamma$ is fully satisfiable, a one-hot point has $Q_\Gamma^H=1$.
\end{lemma}

\begin{proof}
We first verify the facts stated before the lemma.  Each $a\in\Sigma_L$
has exactly one image $\pi_e(a)$, so summing
\eqref{eq:hardness-projected-mass} over $b\in\Sigma_R$ gives
\begin{equation}
\label{eq:ph-mass-sum}
\sum_{b\in\Sigma_R}z_{v,b}(\x)
=\frac1{\ell_v}\sum_{e=(u,v)\in E}\sum_{a\in\Sigma_L}x_{u,a}\le1,
\end{equation}
because each of the $\ell_v$ inner sums is at most one.  Every
coordinate is nonnegative, so $z_{v,b}\ge0$, and therefore
$z_{v,b}\in[0,1]$.  Every edge has exactly one right endpoint, so
$\sum_v \ell_v=|E|$ and $\sum_v w_v=1$.

For each $v$, choose $b_v$ maximizing $z_{v,b}(\x)$ over $b$.  For
$u\in V_L$ and $a\in\Sigma_L$, let $N_u(a)$ be the number of edges
$e=(u,v)$ at $u$ with $\pi_e(a)=b_v$, and choose $a_u$ maximizing
$N_u(a)$.  Under the labeling $(a_u)_u,(b_v)_v$, an edge $e=(u,v)$ is
satisfied exactly when $\pi_e(a_u)=b_v$, so exactly $N_u(a_u)$ edges at
$u$ are satisfied and
\begin{equation}
\label{eq:ph-rounding-value}
L(\x)=\frac1{|E|}\sum_{u\in V_L}N_u(a_u).
\end{equation}
Since $x_{u,a}\ge0$, $\sum_a x_{u,a}\le1$, and
$0\le N_u(a)\le N_u(a_u)$,
\begin{equation}
\label{eq:ph-rounding-block}
\sum_{a\in\Sigma_L}x_{u,a}N_u(a)
\le N_u(a_u)\sum_{a\in\Sigma_L}x_{u,a}
\le N_u(a_u).
\end{equation}
Writing $N_u(a)$ as a sum over the edges at $u$, exchanging the order
of summation, and grouping the edges by their right endpoint gives, by
\eqref{eq:hardness-projected-mass},
\begin{equation}
\label{eq:ph-rounding-exchange}
\sum_{u\in V_L}\sum_{a\in\Sigma_L}x_{u,a}N_u(a)
=\sum_{e=(u,v)\in E}\ \sum_{a:\pi_e(a)=b_v}x_{u,a}
=\sum_{v\in V_R}\ell_v z_{v,b_v}(\x).
\end{equation}
Summing \eqref{eq:ph-rounding-block} over $u$ and using
\eqref{eq:ph-rounding-value} and \eqref{eq:ph-rounding-exchange}, we get
\[
|E|L(\x)\ge\sum_{v\in V_R}\ell_v z_{v,b_v}(\x).
\]
Dividing by $|E|$ and using $w_v=\ell_v/|E|$ and
$z_{v,b_v}=\max_b z_{v,b}$ proves
\eqref{eq:hardness-rounding-dominates-masses}.  All the choices use
exact rational comparisons and edge counts, so the labeling is computed
in polynomial time.

Now let $H$ satisfy the hypotheses of the second part.  Since
$z_{v,b}(\x)\in[0,1]$, the bounds $0\le H(s)\le s$, together with
\eqref{eq:ph-mass-sum} and $\sum_v w_v=1$, give
\[
0\le Q_\Gamma^H(\x)\le\sum_v w_v\sum_b z_{v,b}(\x)\le\sum_v w_v=1.
\]
For each $v$, by \eqref{eq:ph-mass-sum},
\begin{equation}
\label{eq:ph-square-mass}
\sum_b z_{v,b}^2
\le \Bigl(\max_b z_{v,b}\Bigr)\sum_bz_{v,b}
\le\max_b z_{v,b}.
\end{equation}
Applying $H(s)\le as+bs^2$ to every term of \eqref{eq:hard-profile-score}
and then using \eqref{eq:ph-mass-sum} with $a\ge0$, \eqref{eq:ph-square-mass}
with $b>0$, and finally \eqref{eq:hardness-rounding-dominates-masses}, we get
\[
Q_\Gamma^H(\x)
\le a\sum_vw_v\sum_b z_{v,b}
+b\sum_vw_v\sum_b z_{v,b}^2
\le a\sum_v w_v+b\sum_v w_v\max_b z_{v,b}
\le a+bL(\x).
\]
The value $\operatorname{val}(\Gamma)$ is the maximum over all labelings, so
$L(\x)\le\operatorname{val}(\Gamma)$.  This proves
\eqref{eq:hard-profile-rounding-bound}.

For a satisfying labeling $(a_u)_u,(b_v)_v$, take $x_{u,a_u}=1$ and every
other coordinate zero; this point lies in $\mathcal K_\Gamma$.  Every edge
$e=(u,v)$ has $\pi_e(a_u)=b_v$, so \eqref{eq:hardness-projected-mass}
gives $z_{v,b_v}=1$ and $z_{v,b}=0$ for $b\ne b_v$.  The endpoint values
$H(1)=1$ and $H(0)=0$ then give $Q_\Gamma^H=\sum_v w_v=1$.
\end{proof}

\begin{proof}[Proof of Theorem~\ref{thm:direct-hard-profile-criterion}]
Fix $\alpha\in(\beta,1]$, and choose a rational
$\theta\in(\beta,\alpha)$.  Take the profile $H_\theta$ from the
hypothesis, and write $a=a_\theta$ and $b=b_\theta$.  Since $a<\theta$
and $b>0$, every rational $\delta$ with
$0<\delta<\min\{1,(\theta-a)/b\}$ satisfies
\begin{equation}
\label{eq:hard-profile-threshold}
\tau:=a+b\delta<\theta<\alpha.
\end{equation}
Fix such a $\delta$ and use Theorem~\ref{thm:hardness-label-cover-gap}
with it.  Enumerate $V_L=[p]$ and $\Sigma_L=[q]$, so that
$\mathcal K_\Gamma=\mathcal K_{p,q}$, and let $\ell$ be the common right degree.
For $v\in V_R$ with incident edges $e_j=(u_j,v)$, $j=1,\ldots,\ell$, and
for $b\in\Sigma_R$, put $R_j=\{a:\pi_{e_j}(a)=b\}$.  Then
\eqref{eq:hardness-projected-mass} reads
\[
z_{v,b}(\x)=\frac1\ell\sum_{j=1}^\ell\sum_{a\in R_j}x_{u_j,a},
\]
which is the local form \eqref{eq:hard-profile-local-map}.

By the lifting hypothesis, all losses $-H_\theta(z_{v,b})/S$ belong
to $\mathcal L(\mathcal K_\Gamma)$, where
$S=S_{\theta,p,q,\ell}$ is common to the family.  Index the family by
$I=E\times\Sigma_R$ and, for $e=(u,v)$, define
\begin{equation}
\label{eq:hardness-individual-loss}
f_{e,b}(\x)=-\frac{H_\theta(z_{v,b}(\x))}{S}.
\end{equation}
The same $(v,b)$ loss thus occurs once for each of the $\ell_v$ edges at
$v$.  Grouping the edges by their right endpoint and using
$w_v=\ell_v/|E|$, a uniform sample has mean
\begin{equation}
\label{eq:hard-profile-average-loss}
\frac1{|E|k_R}\sum_{e,b}f_{e,b}(\x)
=-\frac1{|E|k_RS}\sum_v \ell_v\sum_b H_\theta(z_{v,b}(\x))
=-\frac{Q_\Gamma^{H_\theta}(\x)}{Sk_R}.
\end{equation}
Take $\Lambda=Sk_R$ and $Q=Q_\Gamma^{H_\theta}$, so
\eqref{eq:hard-profile-average-loss} is
\eqref{eq:abstract-hardness-average-loss}.
The profile $H_\theta$ maps into $[0,1]$, so $H_\theta\ge0$; by
\eqref{eq:hard-profile-scalar-bounds} it satisfies the hypotheses of
the second part of Lemma~\ref{lem:gap} with $a\ge0$ and $b>0$, and that
lemma gives $Q:\mathcal K_\Gamma\to[0,1]$.  On a satisfiable formula,
$\operatorname{val}(\Gamma)=1$, and the lemma gives a point with $Q=1$.  On
an unsatisfiable formula, $\operatorname{val}(\Gamma)\le\delta$, so
\eqref{eq:hard-profile-rounding-bound} and
\eqref{eq:hard-profile-threshold} give $Q(\x)\le a+b\delta=\tau$ for
every $\x\in\mathcal K_\Gamma$.  Also $\tau\ge0$ is rational and
$\tau<\alpha$.

For fixed $\theta$ and $\delta$, the alphabets and degrees are
constants, so the graph, domain, loss family, and rational data have
polynomial encoding length and $S$, hence $\Lambda$, is polynomially
bounded.  By \eqref{eq:hard-profile-threshold}, the gap $\alpha-\tau$
is at least the fixed positive number $\alpha-\theta$.  The exact score
consists of polynomially many rational profile evaluations, and the
other required computations are polynomial by hypothesis, which
verifies every condition in Definition~\ref{def:online-gap-property}.
\end{proof}

\subsection{Proofs of Lemmas~\ref{lem:hardness-derivative-ratio} and~\ref{lem:hardness-tight-monotone-profile}}
\label{apx:hardness-profile}

\begin{proof}[Proof of Lemma~\ref{lem:hardness-derivative-ratio}]
Put $\kappa=A/B\in(0,1]$.  The function $-H$ is continuously
differentiable near $[0,1]$, and its derivative $-H'$ is negative with
absolute value in $[A,B]$ on $[0,1]$.  Lemma~\ref{lem:slope-ratio} with
$\mathcal K=[0,1]$ and $\a=1$ therefore gives
$\Psi\in\mathcal F^\kappa_{\op{mono}}$, so
$\gamma^*_{\op{mono}}(\Psi)\ge\kappa$.  By
Proposition~\ref{prop:subdiff-basics}(v) and $H'(0)=A$, $H'(1)=B$,
\[
\overline\partial\Psi(s)=-H'(s)+N_{[0,1]}(s)
=\begin{cases}
(-\infty,-A],&s=0,\\
\{-H'(s)\},&0<s<1,\\
[-B,+\infty),&s=1.
\end{cases}
\]

Conversely, suppose $\overline\partial\Psi$ is $\eta$--tilted monotone for
some $\eta\in(0,1]$.  Take $s=0$, $t=1$, $u=-A\in\overline\partial\Psi(0)$,
and $v=-B\in\overline\partial\Psi(1)$.  Then \eqref{eq:gamma-mw-mono}
reads $-B\ge\psi_\eta(-A)=-A/\eta$, hence $\eta\le A/B=\kappa$.  Thus
$\gamma^*_{\op{mono}}(\Psi)\le\kappa$, which proves exactness.
\end{proof}

\begin{proof}[Proof of Lemma~\ref{lem:hardness-tight-monotone-profile}]
Differentiate~\eqref{eq:hardness-polynomial-profile} and use
$(m+1)/(m+\kappa)=B$ from \eqref{eq:hardness-polynomial-constants} to
obtain
\begin{align}
\label{eq:hardness-profile-derivatives}
H'(s)&=B\bigl[1-(1-\kappa)(1-s)^m\bigr],\\
H''(s)&=m(1-\kappa)B(1-s)^{m-1}.
\label{eq:ph-second-derivative}
\end{align}
For $s\in[0,1]$, we have $0\le(1-s)^m\le1$, with value $1$ at $s=0$
and value $0$ at $s=1$.  Hence \eqref{eq:hardness-profile-derivatives}
and $A=\kappa B$ give
\begin{equation}
\label{eq:ph-poly-slope-range}
A\le H'(s)\le B\quad(0\le s\le1),\qquad H'(0)=A,\qquad H'(1)=B.
\end{equation}
Similarly, $0\le(1-s)^{m-1}\le1$ on $[0,1]$, so
\eqref{eq:ph-second-derivative} and $C=m(1-\kappa)B/2$ give
$0\le H''\le2C$ there.
Substitution in \eqref{eq:hardness-polynomial-profile} gives $H(0)=0$
and $H(1)=\bigl((m+1)-(1-\kappa)\bigr)/(m+\kappa)=1$.  By
\eqref{eq:ph-poly-slope-range}, $H'\ge A>0$, so $H$ is increasing and
$H(s)\ge H(0)=0$.  Since $H''\ge0$, $H$ is convex on $[0,1]$, and
\[
H(s)=H\bigl((1-s)\cdot0+s\cdot1\bigr)\le(1-s)H(0)+sH(1)=s.
\]
Taylor's formula with integral remainder at $0$, together with
$H(0)=0$, $H'(0)=A$, and $H''\le2C$, gives
\[
H(s)=As+\int_0^s(s-t)H''(t)\,dt
\le As+2C\int_0^s(s-t)\,dt=As+Cs^2.
\]
This proves \eqref{eq:hardness-polynomial-bounds}.  The polynomial $H$
is $C^1$ on $\mathbb R$, and \eqref{eq:ph-poly-slope-range} holds with
$A>0$, so Lemma~\ref{lem:hardness-derivative-ratio} gives
$\gamma^*_{\op{mono}}(\Psi)=A/B=\kappa$, which is
\eqref{eq:hardness-exact-mono-parameter}.  Next,
\[
A=\frac{\kappa(m+1)}{m+\kappa}
=\kappa+\frac{\kappa(1-\kappa)}{m+\kappa},
\]
and the last term is positive and decreases to $0$ as $m\to\infty$, so
$A\downarrow\kappa$.  Finally, $m-1$ is even, so $(1-s)^{m-1}\ge0$ for
every real $s$, and \eqref{eq:ph-second-derivative} gives $H''(s)\ge0$
on $\mathbb R$.  Thus the ambient polynomial is convex.
\end{proof}

\subsection{Proofs of Theorem~\ref{thm:hardness} and Corollary~\ref{cor:endpoints}}
\label{apx:hardness-application}

\begin{proof}[Proof of Theorem~\ref{thm:hardness}]
We first verify the profile criterion for the class of nonpositive,
$1$--Lipschitz losses whose proper extensions lie in
$\mathcal F_{\op{mono}}^\gamma$.  Given rational
$\theta\in(\gamma,1)$, choose rational $\kappa\in(\gamma,\theta)$
and a positive odd integer $m$ satisfying
\eqref{eq:hardness-degree-choice}.  By
\eqref{eq:hardness-polynomial-constants},
$A=\kappa(m+1)/(m+\kappa)$.  Since $m+\kappa>0$ and $\theta>\kappa$,
\[
A<\theta
\iff \kappa m+\kappa<\theta m+\theta\kappa
\iff \kappa(1-\theta)<m(\theta-\kappa)
\iff m>\frac{\kappa(1-\theta)}{\theta-\kappa}.
\]
Hence \eqref{eq:hardness-degree-choice} gives $A<\theta$.
Write $H=H_{\kappa,m}$.  By Lemma~\ref{lem:hardness-tight-monotone-profile},
$H$ maps $[0,1]$ into $[0,1]$, is nondecreasing and $B$--Lipschitz
there, and satisfies \eqref{eq:hard-profile-scalar-bounds} with
$a_\theta=A$ and $b_\theta=C$.  Also $0\le A<\theta$, $C>0$, and both
are rational.

\paragraph{Geometry of the local map.}
Fix $p,q,\ell$ and a map $z=\langle\a,\cdot\rangle$ of
\eqref{eq:hard-profile-local-map}.  The coordinate of $\a$ at $(u,a)$ is
the number of indices $j$ with $u_j=u$ and $a\in R_j$, divided by $\ell$,
so it belongs to $[0,1]$.  Also
\begin{equation}
\label{eq:ph-a-norm}
\sum_i a_i=\frac1\ell\sum_{j=1}^\ell|R_j|\le q,
\qquad
\|\a\|_2^2\le\Bigl(\max_i a_i\Bigr)\sum_i a_i\le q.
\end{equation}
For $\x\in\mathcal K_{p,q}$, all coordinates are nonnegative and every
block sums to at most one, so
\begin{equation}
\label{eq:ph-z-range}
0\le z(\x)\le\frac1\ell\sum_{j=1}^\ell\sum_{a=1}^q x_{u_j,a}\le1.
\end{equation}
Set $S=1+q(B+2C)$ and let $f=-H\circ z/S$ on $\mathcal K_{p,q}$.

\paragraph{Monotonicity.}
The finite-valued ambient formula is the polynomial
$-H(\langle\a,\x\rangle)/S$, and its gradient is $-H'(z(\x))\a/S$.
Proposition~\ref{prop:subdiff-basics}(v) gives, for the proper extension
$\widetilde f=-H\circ z/S+\delta_{\mathcal K_{p,q}}$,
\begin{equation}
\label{eq:hardness-lifted-subdifferential}
\overline\partial\widetilde f(\x)
=-\frac{H'(z(\x))}{S}\a+N_{\mathcal K_{p,q}}(\x).
\end{equation}
By Lemma~\ref{lem:hardness-tight-monotone-profile}, $A\le H'\le B$ on
$[0,1]$, and $z(\mathcal K_{p,q})\subseteq[0,1]$ by \eqref{eq:ph-z-range}.
So the derivative of the polynomial $-H/S$ is negative with absolute
value in $[A/S,B/S]$ on $z(\mathcal K_{p,q})$, and
Lemma~\ref{lem:slope-ratio} gives
$\widetilde f\in\mathcal F^{A/B}_{\op{mono}}=\mathcal F^\kappa_{\op{mono}}$.
Since $\gamma<\kappa$, Lemma~\ref{lem:psi_basics}(7) shows that each
instance of \eqref{eq:gamma-mw-mono} with parameter $\kappa$ implies the
same instance with parameter $\gamma$, so
$\widetilde f\in\mathcal F_{\op{mono}}^\gamma$.
A lifted loss can have a parameter larger than $\kappa$, for example
when $z$ is constant; the reduction uses only membership in
$\mathcal F_{\op{mono}}^\gamma$.

\paragraph{Normalization and individual minimization.}
By \eqref{eq:hardness-polynomial-bounds}, $0\le H\le1$ on $[0,1]$,
which together with \eqref{eq:ph-z-range} and $S\ge1$ gives $f\in[-1,0]$.
Lemma~\ref{lem:hardness-tight-monotone-profile} gives $H'\le B$ and
$H''\le2C$ on $[0,1]$.  Together with \eqref{eq:ph-a-norm},
$\|\a\a^\top\|_{\op{op}}=\|\a\|_2^2$, $\sqrt q\le q$, and
$S\ge\max\{Bq,2Cq\}$, this gives, on the feasible set,
\begin{align*}
\|\nabla f(\x)\|_2
&=\frac{H'(z(\x))}{S}\|\a\|_2
\le\frac{B\sqrt q}{S}\le1,\\
\nabla^2 f(\x)
&=-\frac{H''(z(\x))}{S}\a\a^\top,
&\|\nabla^2f(\x)\|_{\op{op}}
&\le\frac{2Cq}{S}\le1.
\end{align*}
The feasible set is convex, so it contains the segment between any two
of its points, and the mean value inequality along that segment turns
these bounds into $1$--Lipschitz continuity of both $f$ and its
gradient on the feasible set.
The gradient $-H'(z(\x))\a/S$ is coordinatewise nonpositive on the set,
because $H'\ge A>0$ and $\a\ge0$.
For odd $m$, $H''\ge0$ on all of $\mathbb R$, so the ambient
polynomial is concave, and its Hessian $-H''(z(\x))\a\a^\top/S$ has rank
at most one everywhere.
These Lipschitz and Hessian bounds are stated on the feasible set only.
Finally, $H'>0$ on $[0,1]$, so $H$ is strictly increasing there, and by
\eqref{eq:ph-z-range} any maximizer of $z$ over $\mathcal K_{p,q}$ is an
exact minimizer of $f$.  The linear function $z$ is a sum of one term
$\langle\a_u,\x_u\rangle$ per block, and the blocks are constrained
separately.  Each block maximizes its term by putting mass one on a
coordinate of largest positive coefficient, or by putting mass zero
if all coefficients are zero.

\paragraph{Rational implementations.}
For fixed $\theta$, the rational constants $\kappa,A,B,C$ and the
degree $m+1$ are fixed, and the coefficients of $\a$ have polynomial
bit length.  Evaluating the displayed polynomial in its factored
form and evaluating its derivatives require a fixed number of
rational arithmetic operations, with polynomial output bit length.
The resulting gradient is a valid generalized gradient of the
proper extension by \eqref{eq:hardness-lifted-subdifferential},
because zero belongs to every normal cone.
The normalization $S$ is rational and polynomially bounded.

Linear optimization and Euclidean projection over the product of
simplices are also exact polynomial-time rational operations.
For projection onto a block, first clip negative coordinates to
zero.  If the clipped vector has sum at most one, it is the projection,
since it is the projection onto the nonnegative orthant and lies in
the block.  Otherwise the constraint $\sum_a x_{u,a}\le1$ is active at
the projection, since an inactive constraint would make the projection
equal to the clipped vector.  In that case, sort the input coordinates
and find the active set on which subtraction of a common threshold
makes the sum one; the threshold is a rational average.
Blockwise projection and the exact linear minimizers therefore
have polynomial bit complexity.  Every coordinate of a feasible point
lies in $[0,1]$, so two feasible points differ by at most one in each
of the $pq$ coordinates, and the diameter is at most $\sqrt{pq}$.

All hypotheses of Theorem~\ref{thm:direct-hard-profile-criterion}
are thus verified, so the class of nonpositive $1$--Lipschitz losses in $\mathcal F^\gamma_{\op{mono}}$ has the $\gamma$--online gap
property, and Theorem~\ref{thm:online-gap-hardness-criterion} proves
its hardness.  The inclusions in~\eqref{eq:hardness-three-class-inclusion}
prove the two other conclusions.  For each fixed target $\alpha$,
choose one rational $\theta\in(\gamma,\alpha)$ and use the
corresponding fixed profile and fixed Label Cover soundness.
The right degree and alphabet size are then constants.  A loss
$-H(z_{v,b})/S$ involves the $\ell$ edges at $v$ and at most $k_L$ labels
per edge, so it depends on at most $\ell k_L$ coordinates, and its
polynomial degree $m+1$ is also constant.  This establishes all the
claimed restrictions on the common hard family.

\paragraph{Preserving the single-quadratic result for
$\mathcal F^\gamma$.}
For the last assertion, take the class of nonpositive, $1$--Lipschitz
losses whose proper extensions lie in $\mathcal F^\gamma$, and use
quadratic profiles
\[
h_\theta(s)=\theta s+(1-\theta)s^2.
\]
The strict inequality $a_\theta<\theta$ in the profile criterion rules
out $a_\theta=\theta$, so at a criterion index $\theta\in(\gamma,1)$
we choose any rational $\eta\in(\gamma,\theta)$ and use $h_\eta$, with
$a_\theta=\eta$ and $b_\theta=1-\eta$.  Then $h_\eta(0)=0$,
$h_\eta(1)=1$, and
\begin{equation}
\label{eq:ph-quad-slope}
h_\eta'(s)=\eta+2(1-\eta)s\in[\eta,2-\eta]\qquad(0\le s\le1),
\end{equation}
so $h_\eta$ is increasing and $2$--Lipschitz on $[0,1]$ with values in
$[0,1]$.  Since $(1-\eta)s^2\le(1-\eta)s$ on $[0,1]$, we get
$h_\eta(s)\le s$, and $h_\eta(s)\le\eta s+(1-\eta)s^2$ holds with
equality.  Also $0\le a_\theta=\eta<\theta$ and $b_\theta=1-\eta>0$.

Writing $c=\eta/(1-\eta)$, we have
$-h_\eta(s)=(1-\eta)(-s^2-cs)$, which is $(1-\eta)$ times the function
$h_c$ of Example~\ref{ex:quadratic}.  That example gives
$\gamma^*(h_c)=c/(c+1)=\eta$.  Multiplying a function by a positive
constant multiplies its Clarke subdifferential by the same constant,
and $\psi_\eta$ is positively homogeneous, so
$\gamma^*(-h_\eta)=\eta$.  Since zero lies in every normal cone,
$-h_\eta'(s)$ belongs to the Clarke subdifferential of the proper
extension at $s$, and \eqref{eq:gamma-mwc} gives, for $s,t\in[0,1]$,
\begin{equation}
\label{eq:ph-quad-scalar}
-h_\eta(t)+h_\eta(s)
\ge\psi_\eta\bigl(-h_\eta'(s)(t-s)\bigr).
\end{equation}
Let $S=2q$, $f=-h_\eta\circ z/S$, and
$\widetilde f=f+\delta_{\mathcal K_{p,q}}$.  As in
\eqref{eq:hardness-lifted-subdifferential}, every
$\vv\in\overline\partial\widetilde f(\x)$ has the form
$\vv=-h_\eta'(z(\x))\a/S+\mathbf n_x$ with
$\langle\mathbf n_x,\y-\x\rangle\le0$ for $\y\in\mathcal K_{p,q}$.  Put
$s=z(\x)$ and $t=z(\y)$, so $t-s=\langle\a,\y-\x\rangle$.  By
\eqref{eq:ph-quad-scalar}, positive homogeneity of $\psi_\eta$, the sign
of $\langle\mathbf n_x,\y-\x\rangle$ with monotonicity of $\psi_\eta$,
and Lemma~\ref{lem:psi_basics}(7) with $\gamma<\eta$,
\[
f(\y)-f(\x)
=\frac{-h_\eta(t)+h_\eta(s)}{S}
\ge\psi_\eta\!\left(-\frac{h_\eta'(s)}{S}\langle\a,\y-\x\rangle\right)
\ge\psi_\eta(\langle\y-\x,\vv\rangle)
\ge\psi_\gamma(\langle\y-\x,\vv\rangle).
\]
Thus the lifted losses are $\eta$--tilted convex, hence $\gamma$--tilted convex.
Since $0\le h_\eta\le1$ on $[0,1]$ and $S\ge1$, they have values in
$[-1,0]$.  By \eqref{eq:ph-quad-slope}, $h_\eta''=2(1-\eta)$, and
\eqref{eq:ph-a-norm},
\[
\|\nabla(-h_\eta\circ z/S)\|_2
\le\frac{2\sqrt q}{2q}\le1,\qquad
\|\nabla^2(-h_\eta\circ z/S)\|_{\op{op}}
\le\frac{2(1-\eta)q}{2q}\le1.
\]
They are coordinatewise nonincreasing on the domain, because
$h_\eta'>0$ there and $\a\ge0$, and they are concave quadratics with
Hessian $-2(1-\eta)\a\a^\top/S$ of rank at most one.  Since $h_\eta$ is
increasing on $[0,1]$, they are minimized by maximizing $z$, and all
implementations are rational.
Theorem~\ref{thm:direct-hard-profile-criterion} with $\beta=\gamma$ and
Theorem~\ref{thm:online-gap-hardness-criterion} then give the hardness
conclusion for $\mathcal F^\gamma$ with single concave quadratic losses.
Finally, $h_\theta$ is a polynomial, and $h_\theta'(s)=\theta+2(1-\theta)s$
satisfies $\theta\le h_\theta'\le2-\theta$ on $[0,1]$ with
$h_\theta'(0)=\theta>0$ and $h_\theta'(1)=2-\theta$.
Lemma~\ref{lem:hardness-derivative-ratio} with
$A=\theta$ and $B=2-\theta$ therefore proves
\eqref{eq:quadratic-mono-obstruction}.
\end{proof}

\begin{proof}[Endpoint assertion in Corollary~\ref{cor:endpoints}]
For a proposed $\alpha>0$, choose $0<\gamma<\min\{\alpha,1\}$
and use Theorem~\ref{thm:hardness}.
Every constructed loss is of the form $f=-H\circ z/S$ with $S>0$ and
$H'>0$ on $[0,1]$, which contains the range of $z$ by
\eqref{eq:ph-z-range}; here $H$ is $H_{\kappa,m}$ or $h_\eta$.  Let
$\x,\y$ be feasible with $f(\y)<f(\x)$.  Then $H(z(\y))>H(z(\x))$, and
$H$ is strictly increasing on $[0,1]$, so $z(\y)>z(\x)$.  Hence
\[
\langle\nabla f(\x),\y-\x\rangle
=-\frac{H'(z(\x))}{S}\bigl(z(\y)-z(\x)\bigr)<0.
\]
Every element of the Clarke subdifferential of the proper extension
$\widetilde f$ at $\x$ is $\nabla f(\x)+\mathbf n_x$ with
$\langle\mathbf n_x,\y-\x\rangle\le0$, so its inner product with
$\y-\x$ is also negative.  Thus, if $\langle\y-\x,\vv\rangle\ge0$ for an element
$\vv\in\overline\partial\widetilde f(\x)$, then $f(\y)\ge f(\x)$.  This
is pseudoconvexity on the feasible set.  For $c\in\mathbb R$, the
sublevel set $\{\x\in\mathcal K_{p,q}:f(\x)\le c\}$ equals
$\{\x\in\mathcal K_{p,q}:H(z(\x))\ge-cS\}$.  Since $H$ is continuous and
strictly increasing on $[0,1]$ with $H(0)=0$ and $H(1)=1$, this set is
the whole feasible set if $-cS\le0$, empty if $-cS>1$, and
$\{\x\in\mathcal K_{p,q}:z(\x)\ge H^{-1}(-cS)\}$ otherwise.  Each case is
the intersection of the feasible set with a halfspace or is empty, so
the losses are quasiconvex as well.  This covers the quadratic
specialization, which uses $H=h_\eta$.  An algorithm for either
endpoint class would thus contradict the hardness theorem unless
$\NP=\RP$.
\end{proof}

\clearpage
\bibliography{references}

\end{document}